\documentclass[11pt, opre]{informs3} %
\usepackage[margin=1in]{geometry}

\OneAndAHalfSpacedXI 
\usepackage{setspace}
\usepackage{dsfont}
\usepackage{multirow}
\usepackage{makecell}
\usepackage{chngcntr}
\usepackage{natbib}
\usepackage{setspace}
 \bibpunct[, ]{(}{)}{,}{a}{}{,}%
 \def\bibfont{\small}%

\TheoremsNumberedThrough  
\ECRepeatTheorems

\EquationsNumberedThrough    
\usepackage{longtable}
\usepackage{tabularx}
\usepackage{booktabs}
\usepackage{etex}
\usepackage{multirow}
\usepackage{bbold}
\usepackage{amsmath, bbm}
\usepackage{mathtools}
\usepackage{array}
\usepackage{hyperref}
\usepackage[caption=false]{subfig}
\usepackage{graphicx}
\usepackage{soul}

\usepackage{algorithm}
\usepackage{algorithmicx}
\usepackage{capt-of}
\usepackage[noend]{algpseudocode}
\usepackage{amsbsy, amsfonts, amsgen, amsmath, amsopn, amssymb, amstext,bm,amsxtra, bezier, color, enumerate, graphicx, latexsym, verbatim,pictexwd, supertabular, url, dsfont,leftidx, mathrsfs, appendix,setspace}
\usepackage{pbox}

\usepackage[dvipsnames]{xcolor}
\usepackage{tcolorbox}

\usepackage{setspace}

\usepackage{enumitem}

\usepackage{algorithm,algpseudocode,float}
\usepackage{lipsum}
\usepackage{multibib}
\newcites{ec}{Appendix References}

\makeatletter
\newenvironment{breakablealgorithm}
  {%
   \begin{center}
     \refstepcounter{algorithm}%
     \hrule height.8pt depth0pt \kern2pt
     \begingroup
     \let\oldcaption\caption
     \renewcommand{\caption}[2][\relax]{%
       {\raggedright\textbf{\ALG@name~\thealgorithm} ##2\par}%
       \ifx\relax##1\relax
         \addcontentsline{loa}{algorithm}{\protect\numberline{\thealgorithm}##2}%
       \else
         \addcontentsline{loa}{algorithm}{\protect\numberline{\thealgorithm}##1}%
       \fi
       \kern2pt\hrule\kern2pt
     }%
  }{%
     \let\caption\oldcaption
     \endgroup
     \kern2pt\hrule\relax
   \end{center}
  }
\makeatother

\usepackage[normalem]{ulem}

\begin{document}

\RUNTITLE{Identifying All $\varepsilon$-Best Arms In Linear Bandits}

\TITLE{Identifying All $\varepsilon$-Best Arms in (Misspecified) Linear Bandits}

\ARTICLEAUTHORS{%
\AUTHOR{Zhekai Li}
\AFF{Department of Civil and Environmental Engineering, Massachusetts Institute of Technology, \EMAIL{zhekaili@mit.edu}}
\AUTHOR{Tianyi Ma}
\AFF{School of Operations Research and Information Engineering, Cornell University, \EMAIL{tm693@cornell.edu}}
\AUTHOR{Cheng Hua}
\AFF{Antai College of Economics and Management, Shanghai Jiao Tong University, \EMAIL{cheng.hua@sjtu.edu.cn}}
\AUTHOR{Ruihao Zhu}
\AFF{SC Johnson College of Business, Cornell University, \EMAIL{ruihao.zhu@cornell.edu}}

} 

\ABSTRACT{%
Motivated by the need to efficiently identify multiple candidates in high trial-and-error cost tasks such as drug discovery, we propose a near-optimal algorithm to identify all $\varepsilon$-best arms (\emph{i.e.}, those at most $\varepsilon$ worse than the optimum). Specifically, we introduce LinFACT, an algorithm designed to optimize the identification of all $\varepsilon$-best arms in linear bandits. We establish a novel information-theoretic lower bound on the sample complexity of this problem and demonstrate that LinFACT achieves instance optimality by matching this lower bound up to a logarithmic factor. A key ingredient of our proof is to integrate the lower bound directly into the scaling process for upper bound derivation, determining the termination round and thus the sample complexity. We also extend our analysis to settings with model misspecification and generalized linear models. Numerical experiments, including synthetic and real drug discovery data, demonstrate that LinFACT identifies more promising candidates with reduced sample complexity, offering significant computational efficiency and accelerating early-stage exploratory experiments.

\vspace{0.2cm}

}

\KEYWORDS{ranking and selection; sequential decision making; simulation; adaptive experiment; model misspecification}

\maketitle

%


\section{Introduction}\label{sec_Introduction} 
This paper addresses the problem of identifying the best set of options from a finite pool of candidates. A decision-maker sequentially selects candidates for evaluation, observing independent noisy rewards that reflect their quality. The goal is to strategically allocate measurement efforts to identify the desired candidates.
This problem belongs to the class of pure exploration problems, which fall under the bandit framework but differ from traditional multi-armed bandits (MABs) that balance exploration and exploitation to minimize cumulative regret. Instead, pure exploration focuses on efficient information gathering to confidently apply the chosen best or best set of options. This approach is particularly relevant in applications such as drug discovery and product testing, where identifying the most promising candidates is followed by utilizing them under high-cost conditions, such as clinical trials or large-scale manufacturing tests. 

Conventional pure exploration focuses on identifying the optimal candidate, often referred to as the best arm in the bandit setting. However, in many real-world scenarios, candidates with rewards falling slightly below the optimum may later demonstrate advantageous traits, such as fewer side effects, simpler manufacturing processes, or lower resistance during implementation. Motivated by this insight, this paper aims to identify all $\varepsilon$-best candidates (\emph{i.e.}, those whose performance is at most $\varepsilon$ worse than the optimum). This approach is especially valuable when exploring a range of nearly optimal options is necessary. Promoting multiple promising candidates not only mitigates risk but also increases the chances that at least one will prove successful. 

This setting captures many applications in real-world scenarios. For example:
\begin{itemize}
    \item \textit{Drug Discovery}: In drug discovery, pharmaceutical companies aim to identify as many promising drug candidates as possible during preclinical stages. These candidates, known as preclinical candidate compounds (PCCs), are optimized compounds prepared for preclinical testing to assess efficacy, safety, and pharmacokinetics before advancing to clinical trials. Given the inherent risks and high failure rates in subsequent drug development \citep{das2021acceleratedanti}, starting with a larger pool of potential candidates increases the chances of identifying at least one successful, marketable drug.
    
    \item \textit{Assortment Design}: In e-commerce \citep{boyd2003revenue, elmaghraby2003dynamic, feng2022robust}, recommender systems \citep{huang2007analyzing, peukert2023editor}, and streaming services \citep{doi:10.1287/mnsc.2022.4307, doi:10.1287/mnsc.2017.2875}, expanding the consideration set (\emph{e.g.}, products, movies, or songs) can improve user satisfaction and increase revenues. Offering a diverse range of recommendations helps cater to varying tastes and preferences. 
    
    \item \textit{Automatic Machine Learning}: In automatic machine learning (AutoML) \citep{thornton2013auto}, the goal is to automate the process of selecting algorithms and tuning hyperparameters by providing multiple promising choices for predictive tasks. 
    Due to the randomness of limited data, the best out-of-sample model may not always be optimal. Therefore, providing users with a diverse set of models that yield good results is critical to assisting them in selecting the best algorithm and hyperparameters.
\end{itemize}

\subsection{Main Contributions}\label{subsec_Main Contributions} 
This paper focuses on identifying all $\varepsilon$-best arms in the linear bandit setting and presents contributions in both the algorithmic and theoretical dimensions.

\begin{itemize}
    \item \textit{$\delta$-Probably Approximately Correct (PAC) Algorithm:} On the algorithmic front, we introduce LinFACT (\textbf{Lin}ear \textbf{F}ast \textbf{A}rm \textbf{C}lassification with \textbf{T}hreshold estimation), a $\delta$-PAC (see Section \ref{subsec_PAC} for a formal definition) phase-based algorithm for identifying all $\varepsilon$-best arms in linear bandit problems. LinFACT demonstrates superior effectiveness compared to existing pure exploration algorithms.
    
    \item \textit{Matching Bound of Sample Complexity:} We make two key technical contributions. First, we derive an information-theoretic lower bound on the problem complexity for identifying all $\varepsilon$-best arms in linear bandits. To the best of our knowledge, this is the first such result in the literature. Second, we establish two distinct upper bounds on the expected sample complexity of LinFACT, illustrating the differences between various optimal design criteria used in its implementation. Notably, we demonstrate that LinFACT achieves instance optimality when using the $\mathcal{XY}$-optimal design criterion, matching the lower bound up to a logarithmic factor. The $\mathcal{XY}$-optimal design focuses on contrasting pairs of arms rather than evaluating each arm individually. Our analysis leverages the lower bound directly in defining the classification termination round and in scaling the upper bound.
   
    \item \textit{Accounting for Misspecified Models and GLMs:} We extend our framework beyond linear models to handle misspecified linear bandits and generalized linear models (GLMs). For both cases, we provide theoretical upper bounds on the expected sample complexity. Furthermore, we analyze how prior knowledge of model misspecification impacts the algorithmic upper bounds and performance, and how the incorporation of GLMs influences the sample complexity.  
    
    \item \textit{Numerical Studies with Real-World Datasets:} We conduct extensive numerical experiments to demonstrate that our LinFACT algorithm outperforms existing methods in terms of sample complexity, computational efficiency, and reliable identification of all $\varepsilon$-best arms. In experiments with synthetic data, LinFACT outperforms other baselines in both adaptive and static settings. Using a real-world drug discovery dataset \citep{free1964mathematical}, we further show that LinFACT achieves superior performance compared to previous algorithms. Notably, LinFACT is computationally efficient with time complexity of $O(Kd^2)$, which is lower than $O(nKd^2)$ of Lazy TTS ($n$ is a non-negligible number) \citep{rivera2024optimal}, $O(Kd^3)$ of top $m$ algorithms \citep{reda2021top}, and $O(K^2\log K)$ of KGCB \citep{negoescu2011knowledge}. 
\end{itemize}

\subsection{Related Literature}\label{subsec_Related Literature} 
\vspace{2mm}
\noindent\textbf{Pure Exploration.}\label{subsubsec_Pure Exploration Problem} The multi-armed bandits (MABs) model has been a critical paradigm for addressing the exploration-exploitation trade-off since its introduction by \citet{thompson1933likelihood} in the context of medical trials. While much of the research focuses on minimizing cumulative regret \citep{bubeck2012regret, lattimore2020bandit}, our work focuses on the pure exploration setting \citep{koenig1985procedure}, where the goal is to select a subset of arms and evaluation is based on the final outcome. This distinction highlights the context-specific benefits of each approach: MABs are suited for tasks where the goal is to optimize rewards in real-time, balancing exploration and exploitation, whereas pure exploration is focused on identifying a set of satisfactory arms, without the immediate concern for reward maximization. In pure exploration, the algorithm prioritizes information gathering over reward collection, transforming the objective from reward-centric to information-centric. The focus is on efficiently acquiring sufficient information about all arms for confident identification. 

The origins of pure exploration problems date back to the 1950s in the context of stochastic simulation, specifically within ordinal optimization \citep{shin2018tractable, shen2021ranking} or the Ranking and Selection (R\&S) problem, first addressed by \citet{bechhofer1954single}. Various methodologies have since been developed to solve the canonical R\&S problem, including elimination-type algorithms \citep{kim2001fully, bubeck2013multiple, fan2016indifference}, Optimal Computing Budget Allocation (OCBA) \citep{chen2000simulation}, knowledge-gradient algorithms \citep{frazier2008knowledge, frazier2009knowledge, ryzhov2012knowledge}, UCB-type algorithms \citep{kaufmann2013information}, and the unified gap-based exploration (UGapE) algorithm \citep{gabillon2012best}.  Comprehensive reviews of the R\&S problem can be found in \citet{kim2006selecting, hong2021review}, with the most recent overview provided by \citet{li2024surprising}. 

The general framework of pure exploration encompasses various exploration tasks \citep{qin2025dual}, including best arm identification (BAI) \citep{mannor2004sample, even2006action, doi:10.1287/opre.2019.1911, komiyama2023rate, simchi-levi2024experimentation}, top $m$ identification \citep{kalyanakrishnan2010efficient, kalyanakrishnan2012pac}, threshold bandits \citep{locatelli2016optimal, abernethy2016threshold}, and satisficing bandits \citep{feng2025satisficing}. In applications such as drug discovery, pharmacologists aim to identify a set of highly potent drug candidates from potentially millions of compounds, with only the selected candidates advancing to more extensive testing. Given the uncertainty of final outcomes and the high cost of trial-and-error, identifying multiple promising candidates simultaneously is crucial. To minimize the cost of early-stage exploration, adaptive, sequential experimental designs are necessary, as they require fewer experiments compared to fixed designs. 

\vspace{2mm}
\noindent\textbf{All $\varepsilon$-Best Arms Identification.} Conventional objectives, such as identifying the top $m$ best arms or all arms above a certain threshold, often face significant challenges. In the top $m$ task, selecting a small $m$ may exclude promising candidates, while choosing a large $m$ may include ineffective options, requiring an impractically large number of experiments. Similarly, setting a threshold too high can exclude viable candidates. Both approaches depend on prior knowledge of the problem to achieve good performance, which may not be available in real-world applications. 

In contrast, identifying all $\varepsilon$-best arms (those within $\varepsilon$ of the best arm) overcomes these limitations. This approach promotes broader exploration while providing a robust guarantee: no significantly suboptimal arms will be selected, thereby improving the reliability of downstream tasks \citep{mason2020finding}. The all $\varepsilon$-best arms identification problem generalizes both the top $m$ and threshold bandit problems. It reduces to the top $m$ problem if the number of $\varepsilon$-best arms is known in advance, and to a threshold bandit problem if the value of the best arm is known. 

\citet{mason2020finding} introduced the problem complexity for identifying all $\varepsilon$-best arms and derived a lower bound in the low-confidence regime. However, their lower bound involves a summation that may be unnecessary, indicating room for improvement. Building on Mason’s work, \citet{al2022complexity} derived tighter lower bounds by fully characterizing the alternative bandit instances that an optimal sampling strategy must distinguish and eliminate. They also proposed the asymptotically optimal Track-and-Stop algorithm. However, both \citet{mason2020finding} and \citet{al2022complexity} consider stochastic bandits without structures. In contrast, we study this problem in the linear bandit setting \citep{abe1999associative}, which leverages structural relationships among arms. This presents new challenges, but also allows it to handle more complex scenarios. As a result, our work establishes the first information-theoretic lower bound for identifying all $\varepsilon$-best arms in linear bandits, applicable to any $\delta$-PAC algorithm. 

An extended literature review of misspecified linear bandits and generalized linear bandit models can be found in Section \ref{sec_Literature Review for Misspecified Linear Bandits and Generalized Linear Bandits} of the online appendix.

\section{Problem Formulation}\label{sec_Problem Formulation} 
\noindent\textbf{Notations.} In this paper, we denote the set of positive integers up to $N$ by $[N] = \{1, \ldots, N\}$. Vectors and matrices are represented using boldface notation. The inner product of two vectors is denoted by $\langle \cdot, \cdot \rangle$. We define the weighted matrix norm $\Vert \boldsymbol{x} \Vert_{\boldsymbol{A}}$ as $\sqrt{\boldsymbol{x}^\top \boldsymbol{A} \boldsymbol{x}}$, where $\boldsymbol{A}$ is a positive semi-definite matrix that weights and scales the norm. For two probability measures $P$ and $Q$ over a common measurable space, if $P$ is absolutely continuous with respect to $Q$, the Kullback-Leibler divergence between $P$ and $Q$ is defined as
\begin{equation}
    \text{KL}(P, Q) = 
    \begin{cases}
        \int \log\left( \frac{dP}{dQ} \right) dP, & \text{if } Q \ll P; \\
        \infty, & \text{otherwise},
    \end{cases}
\end{equation}
where $\frac{dP}{dQ}$ is the Radon-Nikodym derivative of $P$ with respect to $Q$, and $Q \ll P$ indicates that $Q$ is absolutely continuous with respect to $P$.

\vspace{2mm}
\noindent\textbf{Setting.} We address the problem of identifying all $\varepsilon$-best arms from a finite set of $K$ arms where $K$ is a (possibly large) positive integer. Each arm $i \in [K]$ has an associated reward distribution with an unknown fixed mean $\mu_i$. Let the mean vector of all arms be denoted as $\boldsymbol{\mu} = (\mu_1, \mu_2, \ldots, \mu_K)$, which can only be estimated through bandit feedback from the selected arms. Without loss of generality, we assume $\mu_1 > \mu_2 \geq \ldots \geq \mu_K$. The gap $\Delta_i = \mu_1 - \mu_i$ (for $i \neq 1$) represents the difference in expected rewards between the optimal arm and arm $i$. To this end, we give a formal definition of the notion of $\varepsilon$-best arm.
\begin{definition}\label{def_all_epsilon} ($\varepsilon$-Best Arm). Given $\varepsilon > 0$, an arm $i$ is called \textit{$\varepsilon$-best} if $\mu_i \geq \mu_1 - \varepsilon$. \end{definition}
Here, we adopt an additive framework to define $\varepsilon$-best arms. There also exists a multiplicative counterpart, where an arm $i$ is considered $\varepsilon$-best if $\mu_i \geq (1 - \varepsilon)\mu_1$. While our study focuses on the additive model, the analysis for the multiplicative model follows similar reasoning. We denote the set of all $\varepsilon$-best arms\footnote{The set of all $\varepsilon$-best arms $G_\varepsilon(\boldsymbol{\mu})$ is also referred to as the good set in this paper.} for a mean vector $\boldsymbol{\mu}$ as  
\begin{equation}
    G_\varepsilon(\boldsymbol{\mu}) \coloneqq \{i : \mu_i \geq \mu_1 - \varepsilon \}. 
\end{equation}

Define $\alpha_\varepsilon \coloneqq \min_{i \in G_\varepsilon(\boldsymbol{\mu})} \left( \mu_i - (\mu_1 - \varepsilon) \right)$ as the distance from the smallest $\varepsilon$-best arm to the threshold $\mu_1 - \varepsilon$. Furthermore, if the complement of $G_\varepsilon(\boldsymbol{\mu})$, denoted as $G_\varepsilon^c(\boldsymbol{\mu})$, is non-empty, we define $\beta_\varepsilon \coloneqq \min_{i \in G_\varepsilon^c(\boldsymbol{\mu})} \left( (\mu_1 - \varepsilon) - \mu_i \right)$ as the closest distance from the threshold to the highest mean value of any arm that is not considered $\varepsilon$-best.

We study this problem under a linear structure, where the mean values depend on an unknown parameter vector $\boldsymbol{\theta} \in \mathbb{R}^d$. Each arm $i$ is associated with a feature vector $\boldsymbol{a}_i \in \mathbb{R}^d$. Let $\mathcal{A} \subset \mathbb{R}^d$ be the set of feature vectors, and let $\boldsymbol{\Psi} \coloneqq [\boldsymbol{a}_1, \boldsymbol{a}_2, \ldots, \boldsymbol{a}_K] \in \mathbb{R}^{K \times d}$ be the feature matrix. With parameter $\bm{\theta}$, the mean value can be represented as $\bm{\mu}(\bm{\theta})$. For simplicity, we will consistently refer to the bandit instance as $\bm{\mu}$. When an arm $A_t$ with corresponding feature vector $\boldsymbol{a}_{A_t} \in \mathcal{A}$ is selected at time $t$, we observe the bandit feedback $X_t$, given by
\begin{equation}\label{eq_linear_model}
X_t = \boldsymbol{a}_{A_t}^\top \boldsymbol{\theta} + \eta_t,
\end{equation}
where $\mu_{A_t} = \boldsymbol{a}_{A_t}^\top \boldsymbol{\theta}$ is the true mean reward of the selected arm, and $\eta_t$ is a noise variable. We also make two additional standard assumptions on the norm of the parameters and the noise distributions \citep{abbasi2011improved}.
\begin{assumption}
    Assume $\max_{i \in [K]} \Vert \boldsymbol{a}_i \Vert_2 \leq L_1$, where $\Vert \cdot \Vert_2$ denotes the $\ell_2$-norm and $L_1$ is a constant. 
\end{assumption}

\begin{assumption}
    The noise $\eta_t$ is conditionally 1-sub-Gaussian, i.e., for any $\lambda \in \mathbb{R}$, 
    \begin{equation}
    \mathbb{E} \left[ e^{\lambda \eta_t} \,\big|\, \boldsymbol{a}_{A_1}, \ldots, \boldsymbol{a}_{A_{t-1}}, \eta_1, \ldots, \eta_{t-1} \right] \leq \exp \left( \frac{\lambda^2}{2} \right).
    \end{equation}
\end{assumption}

\subsection{Probably Approximately Correct Algorithm Framework}\label{subsec_PAC}

Our goal is to identify all $\varepsilon$-best arms with high confidence while minimizing the sampling budget. To achieve this, we employ three main components: stopping rule, sampling rule, and decision rule. 

At each time step $t$, the stopping rule $\tau_\delta$ determines whether to continue or stop the process. If the process continues, an arm is selected according to the sampling rule, and the corresponding random reward is observed. When the process stops at $t = \tau_\delta$, a decision rule provides an estimate $\widehat{\mathcal{I}}_{\tau_\delta}$ of the true solution set $\mathcal{I}(\boldsymbol{\mu})$, which in our problem is the set of all $\varepsilon$-best arms, $ G_\varepsilon(\boldsymbol{\mu})$. 

We define the set of all viable mean vectors $\boldsymbol{\mu}$ as
\begin{equation}
M \coloneqq \left\{ \boldsymbol{\mu} \in \mathbb{R}^K \;\middle|\; \exists \boldsymbol{\theta} \in \mathbb{R}^d, \boldsymbol{\mu} = \boldsymbol{\Psi}\boldsymbol{\theta} \; \land \; \Vert \boldsymbol{a}_i \Vert_2 \leq L_1 \text{ for each } i \in [K] \right\}.
\end{equation}

Here, the set $M$ consists of all possible mean vectors $\boldsymbol{\mu}$ that can be expressed as a linear combination of the parameter vector $\boldsymbol{\theta}$ through the matrix $\boldsymbol{\Psi}$. 

We focus on algorithms that are probably approximately correct with high confidence, referred to as $\delta$-PAC algorithms.

\begin{definition}\label{delta-PAC}
\textup{($\delta$-\textit{PAC} Algorithm).} An algorithm is \textit{$\delta$-PAC} for all $\varepsilon$-best arms identification if it identifies the correct solution set with a probability of at least $1-\delta$ for any problem instance with mean $\boldsymbol{\mu} \in M$, \emph{i.e.},
\begin{equation}
\mathbb{P}_{\boldsymbol{\mu}} \left( \tau_\delta < \infty, \; \widehat{\mathcal{I}}_{\tau_\delta} = G_\varepsilon\left( \boldsymbol{\mu}\right) \right) \geq 1-\delta, \qquad \forall \boldsymbol{\mu} \in M.
\label{def: delta-PAC}
\end{equation}
\end{definition}

Upon stopping, $\delta$-PAC algorithms ensure the identification of all $\varepsilon$-best arms with high confidence. Therefore, our goal is to design a $\delta$-PAC algorithm that minimizes the stopping time, formulated as the following optimization problem.
\begin{align}
    \min &\quad \mathbb{E}_{\boldsymbol{\mu}} \left[ \tau_\delta \right] \\
    \text{s.t.} &\quad \mathbb{P}_{\boldsymbol{\mu}} \left( \tau_\delta < \infty, \; \widehat{\mathcal{I}}_{\tau_\delta} = G_\varepsilon\left( \boldsymbol{\mu}\right) \right) \geq 1-\delta, \qquad \forall \boldsymbol{\mu} \in M.
\end{align}

\subsection{Optimal Design of Experiment}\label{subsec_Optimal Design} 
Linear bandit algorithms can be viewed as an online, adaptive counterpart to the classical optimal design problem. This section develops the key theoretical foundations, emphasizing the confidence bounds for parameter estimation that guide both the design of our algorithm and the subsequent analysis.

\vspace{2mm}
\noindent\textbf{Ordinary Least Squares.}\label{subsubsec_Least Squares Estimators}
Consider a sequence of pulled arms, denoted as $A_1, A_2, \ldots, A_t$, and the corresponding observed rewards $X_1, X_2, \ldots, X_t$. If the feature vectors of these arms, $\boldsymbol{a}_{A_1}, \boldsymbol{a}_{A_2}, \ldots, \boldsymbol{a}_{A_t}$, span the space $\mathbb{R}^d$, the ordinary least squares (OLS) estimator for the parameter $\boldsymbol{\theta}$ is given by
\begin{equation}\label{OLS}
    \hat{\boldsymbol{\theta}}_t = \bm{V}_t^{-1} \sum_{n=1}^t \boldsymbol{a}_{A_n} X_n,
\end{equation}
where $\bm{V}_t = \sum_{n=1}^t \boldsymbol{a}_{A_n} \boldsymbol{a}_{A_n}^\top \in \mathbb{R}^{d \times d}$ represents the information matrix. Using the properties of sub-Gaussian random variables, we can derive a confidence bound for the OLS estimator. This bound, denoted as $B_{t,\delta}$, is detailed in Proposition \ref{proposition_confidence bound for B 1}. The confidence region for the parameter $\boldsymbol{\theta}$ at time step $t$ is given by
\begin{equation}\label{confidence for ellipse}
    \mathcal{C}_{t, \delta} = \left\{ \boldsymbol{\theta} : \Vert \hat{\boldsymbol{\theta}}_t - \boldsymbol{\theta} \Vert_{\bm{V}_t} \leq B_{t,\delta} \right\}.
\end{equation}

\begin{proposition}[\cite{lattimore2020bandit}]\label{proposition_confidence bound for B 1}
For any fixed sampling policy and any given vector $\boldsymbol{x} \in \mathbb{R}^d$, with probability at least $1-\delta$, the following holds. 
\begin{equation}\label{confidence bound BB}
    \left| \boldsymbol{x}^\top \left( \hat{\boldsymbol{\theta}}_t - \boldsymbol{\theta} \right) \right| \leq \Vert \boldsymbol{x} \Vert_{\bm{V}_t^{-1}} B_{t, \delta},
\end{equation}
where the anytime confidence bound $B_{t, \delta}$ is given by
$
    B_{t, \delta} = 2 \sqrt{2 \left( d \log(6) + \log\left( \frac{1}{\delta} \right) \right)}.
$
\end{proposition}

In many practical scenarios, the observed data are not predetermined. To handle this, a martingale-based method can be employed, as described by \citet{abbasi2011improved}, to define an adaptive confidence bound for the OLS estimator. This accounts for the variability introduced by random rewards and adaptive sampling policies. The confidence interval in Proposition \ref{proposition_confidence bound for B 1} highlights the connection between arm allocation policies in linear bandits and experimental design theory \citep{pukelsheim2006optimal}. This connection serves as a fundamental component in constructing our algorithm. 

\vspace{2mm}
\noindent\textbf{Feature Vector Projection.}\label{subsubsec_Projection}
At any time step where estimation needs to be made after sampling, if the feature vectors of the sampled arms do not span $\mathbb{R}^d$, we substitute them with dimensionality-reduced feature vectors \citep{NEURIPS2022_4f9342b7}. Specifically, we project all feature vectors onto the subspace spanned by $\mathcal{A}$. Let $\boldsymbol{B} \in \mathbb{R}^{d \times d'}$ be an orthonormal basis for this subspace, where $d' < d$ is the dimension of the subspace. The new feature vector $\boldsymbol{a}'$ is then given by
$
    \boldsymbol{a}' = \boldsymbol{B}^\top \boldsymbol{a}.
$
In this transformation, $\boldsymbol{B}\boldsymbol{B}^\top$ is a projection matrix, ensuring
\begin{equation}\label{eq_projection}
    \left\langle \boldsymbol{\theta}, \boldsymbol{a} \right\rangle = \left\langle \boldsymbol{\theta}, \boldsymbol{B}\boldsymbol{B}^\top \boldsymbol{a} \right\rangle = \left\langle \boldsymbol{B}^\top \boldsymbol{\theta}, \boldsymbol{B}^\top \boldsymbol{a} \right\rangle = \left\langle \boldsymbol{\theta}', \boldsymbol{a}' \right\rangle.
\end{equation}

Equation \eqref{eq_projection} ensures that the mean values of all arms remain unchanged under the projection. The first equality holds because $\boldsymbol{B}\boldsymbol{B}^\top$ is a projection matrix and $\boldsymbol{a}$ lies in the subspace spanned by $\boldsymbol{B}$ (\emph{i.e.}, $\boldsymbol{B}\boldsymbol{B}^\top \boldsymbol{a} = \boldsymbol{a}$). The second equality follows from the same matrix form $\boldsymbol{\theta}^\top \boldsymbol{B} \boldsymbol{B}^\top \boldsymbol{a}$. The third equality holds by definition. 

\vspace{2mm}
\noindent\textbf{Optimal Design Criteria.}\label{subsubsec_G-Optimal Design}
In contrast to stochastic bandits, where the mean values of the arms are estimated through repeated sampling of each arm, the linear bandit setting allows these values to be inferred from accurate estimation of the underlying parameter vector $\boldsymbol{\theta}$. As a result, pulling a single arm provides information about all arms. 

A key sampling strategy in this context is the \textit{G-optimal design}, which minimizes the maximum variance of the predicted responses across all arms by optimizing the fraction of times each arm is selected. Formally, the G-optimal design problem seeks a probability distribution $\pi$ on $\mathcal{A}$, where $\pi: \mathcal{A} \rightarrow [0,1]$ and $\sum_{\boldsymbol{a} \in \mathcal{A}} \pi(\boldsymbol{a}) = 1$, that minimizes
\begin{equation}\label{equation_G optimal}
    g(\pi) = \max_{\boldsymbol{a} \in \mathcal{A}} \Vert \boldsymbol{a} \Vert_{\bm{V}(\pi)^{-1}}^2,
\end{equation}
where $\bm{V}(\pi) = \sum_{\boldsymbol{a} \in \mathcal{A}} \pi(\boldsymbol{a}) \boldsymbol{a} \boldsymbol{a}^\top$ is the weighted information matrix, analogous to $\bm{V}_t$ in equation~\eqref{OLS}. The G-optimal design~\eqref{equation_G optimal} ensures a tight confidence interval for mean value estimation. However, comparing the relative differences of mean values across different arms is more critical in identifying the best arms, rather than making the best estimation. 

Therefore, we consider an alternative design criterion, the \textit{$\mathcal{XY}$-optimal design}, that directly targets the estimation of these gaps. Consider $\mathcal{S} \subseteq \mathcal{A}$ as a subset of the arm space. We define
\begin{equation}
    \mathcal{Y}(\mathcal{S}) \coloneqq \left\{ \boldsymbol{a} - \boldsymbol{a}^\prime : \forall \boldsymbol{a}, \boldsymbol{a}^\prime \in \mathcal{S}, \boldsymbol{a} \neq \boldsymbol{a}^\prime \right\}
\end{equation}
as the set of vectors representing the differences between each pair of arms in $\mathcal{S}$. The $\mathcal{XY}$-optimal design minimizes
\begin{equation}\label{equation_XY optimal}
    g_{\mathcal{XY}}(\pi) = \max_{\boldsymbol{y} \in \mathcal{Y}(\mathcal{A})} \Vert \boldsymbol{y} \Vert_{\bm{V}(\pi)^{-1}}^2.
\end{equation}

As mentioned previously, the $\mathcal{XY}$-optimal design focuses on minimizing the maximum variance when estimating the differences (gaps) between pairs of arms. By doing so, it ensures differentiation between arms, rather than estimating each arm individually. This criterion is particularly useful when the goal is to identify relative performance rather than absolute quality. 

\section{Lower Bound and Problem Complexity}\label{sec_Lower Bound of Expected Sample Complexity} 

\begin{figure}[htbp]
    \centering
    \includegraphics[scale=1]{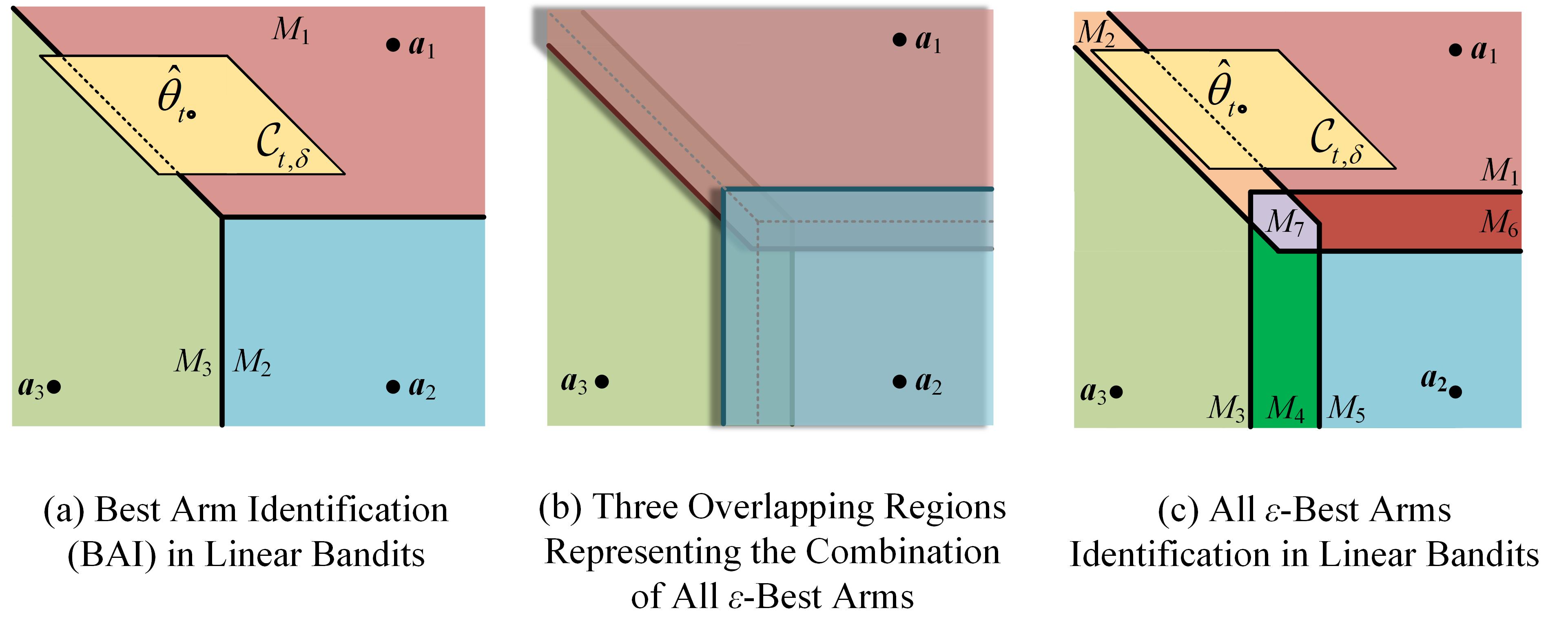}
    \caption{Illustration of the Stopping Condition: Best Arm Identification vs. All $\varepsilon$-Best Arms Identification
    }
\noindent

  \par
  {\footnotesize
    \parbox{\linewidth}{\raggedright
      \textit{Note.} \textit{{(a)} Stopping occurs when the confidence region $\mathcal{C}_{t, \delta}$ for the estimated parameter $\hat{\boldsymbol{\theta}}_t$ contracts entirely within one of the three decision regions $M_i$ in a certain time step $t$. The boundaries between regions are defined by the hyperplanes $\boldsymbol{\vartheta}^\top(\boldsymbol{a}_i - \boldsymbol{a}_j) = 0$. Each dot represents an arm. {(b)} In the case of identifying all $\varepsilon$-best arms, the regions overlap. {(c)} These overlaps partition the space into seven distinct decision regions, increasing the difficulty of identification.}
    }%
  }
  \vspace{-8pt}
    \label{fig: 11}
\end{figure}

In this section, we present a novel information-theoretic lower bound for the problem of identifying all $\varepsilon$-best arms in linear bandits. Building on the approach of \citet{soare2014best}, we extend the lower bound for best arm identification (BAI) to this more general setting. Figure~\ref{fig: 11} visualizes the structure of the stopping condition, with additional graphical insights provided in Section~\ref{subsubsec_Visual Explanation of the Lower Bound}. These visualizations offer geometric intuition for the challenges involved in identifying all $\varepsilon$-best arms in linear bandits. 

The sample complexity of an algorithm is quantified by the number of samples, denoted $\tau_\delta$, required to terminate the process. The goal of the algorithm design is to minimize the expected sample complexity $\mathbb{E}_{\boldsymbol{\mu}} \left[ \tau_\delta \right]$ across the entire set of algorithms $\mathcal{H}$. As introduced in \cite{kaufmann2016complexity}, for $\delta \in (0, 1)$, the non-asymptotic problem complexity of an instance $\boldsymbol{\mu}$ can be defined as
\begin{equation}\label{eq_problem complexity}
    \kappa \left( \boldsymbol{\mu} \right) \coloneqq \inf_{Algo \in \mathcal{H}} \frac{\mathbb{E}_{\boldsymbol{\mu}} \left[ \tau_\delta \right]}{\log\left( \frac{1}{2.4\delta} \right)},
\end{equation}
which is the smallest possible constant such that the expected sample complexity $\mathbb{E}_{\boldsymbol{\mu}} \left[ \tau_\delta \right]$ grows asymptotically in line with $\log\left( \frac{1}{2.4\delta} \right)$. The lower bound of the sample complexity $\mathbb{E}_{\boldsymbol{\mu}} \left[ \tau_\delta \right]$ can be represented in a general form by the following proposition. Building on the analytical framework of Proposition~\ref{general lower bound}, we formulate the $\varepsilon$-best arms identification problem in the linear bandit setting. This enables us to derive both lower and upper bounds on the sample complexity, thereby establishing the near-optimality of our algorithm. 

\begin{proposition}[\cite{qin2025dual}]\label{general lower bound}
For any $\boldsymbol{\mu} \in M$, there exists a set $\mathcal{X} = \mathcal{X}(\boldsymbol{\mu})$ and functions $\{ C_x \}_{x \in \mathcal{X}}$ with $C_x: \mathcal{S}_K \times \mathcal{X} \rightarrow \mathbb{R}_+$ such that
\begin{equation}
    \kappa \left( \boldsymbol{\mu} \right) \geq \left( \Gamma^{\ast}_{\boldsymbol{\mu}} \right)^{-1},
    \label{def: general lower bound inequality}
\end{equation}
where
\begin{equation}\label{def: general lower bound}
    \Gamma^{\ast}_{\boldsymbol{\mu}} = \max_{\boldsymbol{p} \in \mathcal{S}_K} \min_{x \in \mathcal{X}} C_x \left( \boldsymbol{p}; \boldsymbol{\mu} \right ).
\end{equation}
\end{proposition}

In Proposition \ref{general lower bound}, $\mathcal{S}_K$ denotes the $K$-dimensional probability simplex, and $\mathcal{X} = \mathcal{X}(\boldsymbol{\mu})$ is referred to as the culprit set. This set comprises critical subqueries (or comparisons) that must be correctly resolved. An error in any of these comparisons may hinder the identification of the correct set. 

For example, in the case of identifying the best arm, the culprit set is given by $\mathcal{X}= \{ i : i \in [K] \setminus i^\ast \}$, where $i^\ast$ denotes the unique best arm, and each subquery involves distinguishing every arm from the best arm. In the threshold bandit problem, the culprit set consists of all arms, $\mathcal{X} = \{ i : i \in [K] \}$, where each subquery requires accurately determining whether each arm exceeds the threshold. For the task of identifying the best $m$ arms, the culprit set is $\mathcal{X} = \{ (i, j) : i \in \mathcal{I}, j \in \mathcal{I}^c \}$, where $\mathcal{I}$ represents the set of the best $m$ arms, and each subquery entails comparing the mean of each arm in $\mathcal{I}$ with those in the complement set $\mathcal{I}^c$. 

The function $C_x(\boldsymbol{p}; \boldsymbol{\mu})$ represents the population version of the sequential generalized likelihood ratio statistic, which provides an information-theoretic measure of how easily each subquery, corresponding to the culprit $x \in \mathcal{X}$, can be answered. 

In equation~\eqref{def: general lower bound}, the minimum aligns with the intuition that the instance posing the greatest challenge corresponds to the hardest subquery. The outer maximization seeks the optimal allocation of arms $\boldsymbol{p}$ to effectively address this subquery. For a more detailed introduction to this general pure exploration model, please refer to Section \ref{sec_General Pure Exploration Model}. For our setting, we establish the below lower bound. 

\begin{theorem}[Lower Bound]\label{lower bound: All-epsilon in the linear setting}
    Consider a set of arms where arm $i$ follows a normal distribution $\mathcal{N}(\mu_i, 1)$, where $\mu_i = \boldsymbol{a}_{i}^\top \boldsymbol{\theta}$. Any $\delta$-PAC algorithm for identifying all $\varepsilon$-best arms in the linear bandit setting must satisfy 
    \begin{equation}\label{equation: lower bound 3}
        \inf_{Algo \in \mathcal{H}} \frac{\mathbb{E}_{\boldsymbol{\mu}} \left[ \tau_\delta \right]}{\log\left( \frac{1}{2.4\delta} \right)} \geq (\Gamma^\ast_{\boldsymbol{\mu}})^{-1} = \min_{\boldsymbol{p} \in \mathcal{S}_K} \max_{(i, j, m) \in \mathcal{X}} \max \left\{ \frac{2 \Vert \boldsymbol{a}_i - \boldsymbol{a}_j \Vert_{\bm{V}_{\boldsymbol{p}}^{-1}}^2}{\left( \boldsymbol{a}_i^\top \boldsymbol{\theta} - \boldsymbol{a}_j^\top \boldsymbol{\theta} + \varepsilon \right)^2}, \frac{2 \Vert \boldsymbol{a}_1 - \boldsymbol{a}_m \Vert_{\bm{V}_{\boldsymbol{p}}^{-1}}^2}{\left( \boldsymbol{a}_1^\top \boldsymbol{\theta} - \boldsymbol{a}_m^\top \boldsymbol{\theta} - \varepsilon \right)^2} \right\},
    \end{equation}
    where $\mathcal{X} = \{ (i, j, m) : i \in G_\varepsilon(\boldsymbol{\mu}), j \neq i, m \notin G_\varepsilon(\boldsymbol{\mu}) \}$, $\mathcal{S}_K$ is the $K$-dimensional probability simplex.
\end{theorem}

The detailed proof of the above theorem is presented in Section \ref{proof of theorem: lower bound for all-epsilon in the linear bandit}. For the lower bound derivation, we assume normally distributed rewards to obtain a closed-form expression. A similar bound can be derived under sub-Gaussian rewards, though the form is less explicit. 

\begin{remark}[\textbf{Generality of the Lower Bound}]
    We note that the stochastic multi‑armed bandit problem is a special case of the linear bandit problem. By setting $\mathcal{A} = \{ \boldsymbol{e}_1, \boldsymbol{e}_2, \ldots, \boldsymbol{e}_d \}$, where $\boldsymbol{e}_i$ denotes the unit vector, the linear bandit model reduces to a stochastic setting. This relationship allows us to recover the lower bound result for identifying all $\varepsilon$-best arms in stochastic bandits \citep{al2022complexity}. Furthermore, the lower bound in Theorem \ref{lower bound: All-epsilon in the linear setting} extends the lower bound for best arm identification in linear bandits \citep{fiez2019sequential}. This result is recovered by setting $\varepsilon = 0$ and redefining the culprit set as $\mathcal{X}(\boldsymbol{\mu}) = \{ i : i \in [K] \setminus i^\ast \}$, where $i^\ast$ represents the best arm in the context of best arms identification.

\end{remark}

\section{Algorithm and Upper Bound}\label{sec_Algorithm and Upper Bound} 
In this section, we propose the \textit{LinFACT} algorithm (\textbf{Lin}ear \textbf{F}ast \textbf{A}rm \textbf{C}lassification with \textbf{T}hreshold estimation) to identify all $\varepsilon$-best arms in linear bandits efficiently. We then establish upper bounds on the expected sample complexity to demonstrate the optimality of the LinFACT algorithm. Specifically, the upper bound derived from the $\mathcal{XY}$-optimal sampling policy is shown to be instance optimal up to logarithmic factors.\footnote{We refer to the algorithms based on G-optimal sampling and $\mathcal{XY}$-optimal sampling as LinFACT-G and LinFACT-$\mathcal{XY}$, respectively. A detailed comparison between the two approaches is provided in Section~\ref{Essential Difference between G and XY}.}

\subsection{Algorithm}\label{subsec_Linear Fast Arm Classification with Threshold Estimation} 
LinFACT is a phase-based, semi-adaptive algorithm in which the sampling rule remains fixed within each round and is updated only at the end based on the accumulated observations. As the algorithm proceeds through round $r$, LinFACT progressively refines two sets of arms: 
\begin{itemize}
    \item $G_r$: Arms empirically classified as $\varepsilon$-best (good).
    \item $B_r$: Arms empirically classified as not $\varepsilon$-best (bad).
\end{itemize}

This classification process continues until all arms have been assigned to either $G_r$ or $B_r$. Once complete, the decision rule returns $G_r$ as the final set of $\varepsilon$-best arms.

\vspace{2mm}
\noindent\textbf{Sampling Rule.}\label{subsubsec_Sampling Rule} To minimize the sampling budget, we select arms that provide the maximum information about the mean values or the gaps between them. Unlike stochastic multi-armed bandits, where mean values are obtained exclusively by sampling specific arms, linear bandits allow these mean values to be inferred from the estimated parameters. 
In each round, arms are selected based on the G-optimal design (\ref{equation_G optimal}) or the $\mathcal{XY}$-optimal design (\ref{equation_XY optimal}). 

For G-optimal design, LinFACT-G refines an estimate of the true parameter $\boldsymbol{\theta}$ and uses this estimate to maintain an anytime confidence interval, such that for each arm's empirical mean value $\hat{\mu}_i$, we have 
\begin{equation}\label{eq_high_1}
    \mathbb{P} \left( \bigcap_{i \in \mathcal{A}_I(r-1)} \bigcap_{r \in \mathbb{N}} \left| \hat{\mu}_i(r) - \mu_i \right| \leq C_{\delta/K}(r) \right) \geq 1-\delta.
\end{equation}

The active set $\mathcal{A}(r)$ is defined as the set of uneliminated arms, as we continuously eliminate arms as round $r$ progresses. $\mathcal{A}_{I}(r)$ denotes the set of indices corresponding to $\mathcal{A}(r)$. 

This confidence bound indicates that the algorithm maintains a probabilistic guarantee that the true mean value $\mu_i$ is within a certain range of the estimated mean value $\hat{\mu}_i$ for each arm $i$, uniformly over all rounds. The bound shrinks as more data is collected (since the confidence radius $C_{\delta/K}(r)$ decreases with more samples), thereby reducing uncertainty. The anytime confidence width $C_{\delta/K}(r)$ is maintained by the design of the sample budget in each round. We set $C_{\delta/K}(r) = 2^{-r} =: \varepsilon_r$, which is halved with each iteration of the rounds.

In LinFACT-G, the initial budget allocation policy is based on the G-optimal design and is defined as follows
\begin{equation}\label{equation_phase budget G}
\left\{
\begin{array}{l}
T_{r}(\boldsymbol{a}) = \left\lceil \dfrac{2d\pi_{r}(\boldsymbol{a})}{\varepsilon_{r}^2} \log\left( \dfrac{2Kr(r + 1)}{\delta} \right) \right\rceil \\
\\
T_{r} = \sum_{\boldsymbol{a} \in \mathcal{A}(r-1)} T_{r}(\boldsymbol{a})
\end{array},
\right.
\end{equation}
where $T_r$ denotes the total sampling budget allocated in round $r$, and $\pi_r$ is the selection probability distribution over the remaining active arms $\mathcal{A}(r-1)$ from the previous round, obtained via the G-optimal design as defined in equation \eqref{equation_G optimal}. The sampling procedure for each round $r$ is described in Algorithm \ref{alg_LinFACT_sampling}.

\vspace{0.4cm}
\begin{breakablealgorithm}\label{alg_LinFACT_sampling}
\caption{Subroutine: G-Optimal Sampling}
\begin{algorithmic}[1]
\State \textbf{Input:} Projected active set $\mathcal{A}(r-1)$, round $r$, $\delta$.
\State Obtain $\pi_r \in \mathcal{P}(\mathcal{A}(r-1))$ with support size $\text{Supp}(\pi_r) \leq \frac{d(d+1)}{2}$ according to equation \eqref{equation_G optimal}. \label{G_sampling_1}
\ForAll{$\boldsymbol{a} \in \mathcal{A}(r-1)$}
    \Comment{Sampling}
    \State Sample arm $\boldsymbol{a}$ for $T_r(\boldsymbol{a})$ times in round $r$, as specified in equation~\eqref{equation_phase budget G}. \label{G_sampling_2}
\EndFor
\end{algorithmic}
\end{breakablealgorithm}
\vspace{0.6cm}

We also adopt the $\mathcal{XY}$-optimal design because the G-optimal design is less effective for distinguishing between arms. While the G-optimal design minimizes the maximum variance in estimating individual arm means, it does not explicitly focus on the pairwise gaps that are critical for identifying good arms. In contrast, the $\mathcal{XY}$-optimal design is tailored to directly reduce the uncertainty in estimating these gaps.

We now introduce a sampling rule based on the $\mathcal{XY}$-optimal design, as defined in equation~\eqref{equation_XY optimal}. Let $q(\epsilon)$ denote the error introduced by the rounding procedure. We have: 
\begin{equation}\label{equation_phase budget XY}
\left\{
\begin{array}{l}
T_{r} = \max \left\{ \left\lceil \dfrac{2 g_{\mathcal{XY}}\left( \mathcal{Y}(\mathcal{A}(r-1)) \right) (1+\epsilon)}{\varepsilon_r^2} \log \left( \dfrac{2K(K-1)r(r+1)}{\delta} \right) \right\rceil, q\left( \epsilon \right) \right\} \\
\\
T_{r}(\boldsymbol{a}) = \textsc{Round}(\pi_r, T_r)
\end{array}.
\right.
\end{equation}

In contrast to the G-optimal design, the $\mathcal{XY}$-optimal design focuses on bounding the confidence region of the pairwise differences between arms. The following inequality characterizes the corresponding high-probability event.
\begin{equation}
\mathbb{P}\left(  \bigcap_{i \in \mathcal{A}_I(r-1)} \bigcap_{j \in \mathcal{A}_I(r-1),\, j \neq i} \bigcap_{r \in \mathbb{N}} \left| (\hat{\mu}_j(r) - \hat{\mu}_i(r)) - (\mu_j - \mu_i) \right| \leq 2C_{\delta/K}(r) \right) \geq 1-\delta.
\end{equation}

The rounding operation, denoted as \textsc{Round}, uses a $(1+\epsilon)$ approximation algorithm proposed by \citet{allenzhu2017nearoptimal}. The complete sampling procedure is outlined in Algorithm~\ref{alg_LinFACT_sampling_XY}. 

\vspace{0.4cm}
\begin{breakablealgorithm}\label{alg_LinFACT_sampling_XY}
\caption{Subroutine: $\mathcal{XY}$-Optimal Sampling}
\begin{algorithmic}[1]
\State \textbf{Input:} Projected active set $\mathcal{A}(r-1)$, round $r$, $\delta$.
\State Obtain $\pi_r \in \mathcal{P}(\mathcal{A}(r-1))$ according to equation \eqref{equation_XY optimal}.
\ForAll{$\boldsymbol{a} \in \mathcal{A}(r-1)$}
    \Comment{Sampling}
    \State Sample arm $\boldsymbol{a}$ for $T_r(\boldsymbol{a})$ times in round $r$, as specified in equation~ \eqref{equation_phase budget XY}. \label{XY_sampling}
\EndFor
\end{algorithmic}
\end{breakablealgorithm}
\vspace{0.6cm}

\vspace{2mm}
\noindent\textbf{Estimation.}\label{subsubsec_Estimation} At the end of each round, after drawing $T_r(\boldsymbol{a})$ samples from the active set, we compute the empirical estimate of the parameter using standard ordinary least squares (OLS)
\begin{equation}\label{esti_OLS}
    \hat{\boldsymbol{\theta}}_{r} = \bm{V}_{r}^{-1} \sum_{s=1}^{T_{r}} \boldsymbol{a}_{A_s} X_s,
\end{equation}
where
$
    \bm{V}_{r} = \sum_{\boldsymbol{a} \in \mathcal{A}(r-1)} T_{r}(\boldsymbol{a}) \boldsymbol{a} \boldsymbol{a}^\top
$
is the information matrix. The estimator for the mean value of each arm $i \in \mathcal{A}_I(r-1)$ is then
\begin{equation}\label{eq_estimation_1}
    \hat{\mu}_i = \hat{\mu}_i(r) = \boldsymbol{a}_i^\top \hat{\boldsymbol{\theta}}_r.
\end{equation}

\vspace{2mm}
\noindent\textbf{Stopping Rule and Decision Rule.}\label{subsubsec_Stopping Rule and Decision Rule} As round $r$ progresses, LinFACT dynamically updates two sets of arms: $G_r$ and $B_r$, representing arms that are empirically considered $\varepsilon$-best (good) and those that are not (bad), respectively. The algorithm filters arms by maintaining an upper confidence bound $U_r$ and a lower confidence bound $L_r$ around the unknown threshold $\mu_1 - \varepsilon$, along with individual upper and lower confidence bounds for each arm. 

The stopping rule and final decision procedure are described in Algorithm~\ref{alg_LinFACT_stopping}. For each arm $i$ in the active set, LinFACT eliminates the arm if its upper confidence bound falls below the threshold $L_r$ (line~\ref{elimination: good 1}). Conversely, if the lower confidence bound of an arm exceeds $U_r$ (line~\ref{add to Gr}), the arm is added to the set $G_r$. Additionally, any arm already in $G_r$ will be removed from the active set if its upper bound falls below the empirically largest lower bound among all active arms (line~\ref{elimination: good 2}). This ensures that the best arm is always retained in the active set, which is necessary for estimating the threshold $\mu_1 - \varepsilon$. The classification process continues until all arms are categorized, that is, when $G_r \cup B_r = [K]$. At termination, the set $G_r$ is returned as the output of LinFACT, representing the arms identified as $\varepsilon$-best.

\vspace{0.4cm}
\begin{breakablealgorithm}\label{alg_LinFACT_stopping}
\caption{Subroutine: Stopping Rule and Decision Rule}
\begin{algorithmic}[1]
\State \textbf{Input:} Projected active set $\mathcal{A}_I(r-1)$, estimator $\left( \hat{\mu}_i(r) \right)_{i \in \mathcal{A}_I(r-1)}$, round $r$, $\varepsilon$, confidence radius $C_{\delta/K}(r)$.
\State Let $U_r = \max_{i \in \mathcal{A}_I(r-1)} \hat{\mu}_i + C_{\delta/K}(r) - \varepsilon$.
\State Let $L_r = \max_{i \in \mathcal{A}_I(r-1)} \hat{\mu}_i - C_{\delta/K}(r) - \varepsilon$.
\ForAll{$i \in \mathcal{A}_I(r-1)$}
    \Comment{Arm Classification and Elimination}
    \If{$\hat{\mu}_i + C_{\delta/K}(r) < L_r$}
        \State Add $i$ to $B_r$ and eliminate $i$ from $\mathcal{A}_I(r-1)$. \label{elimination: good 1}
    \EndIf
    \If{$\hat{\mu}_i - C_{\delta/K}(r) > U_r$}
        \State Add $i$ to $G_r$. \label{add to Gr}
    \EndIf
    \If{$i \in G_r$ \textbf{and} $\hat{\mu}_i + C_{\delta/K}(r) \leq \max_{j \in \mathcal{A}_I(r-1)} \hat{\mu}_j - C_{\delta/K}(r)$}
        \State Eliminate $i$ from $\mathcal{A}_I(r-1)$. \label{elimination: good 2}
    \EndIf
\EndFor
\If{$G_r \cup B_r = [K]$}
    \Comment{Stopping Condition and Recommendation}
    \State \textbf{Output:} the set $G_r$. \label{stopping condition 1 and 2}
\EndIf
\end{algorithmic}
\end{breakablealgorithm}
\vspace{0.6cm}

\vspace{2mm}
\noindent\textbf{The Complete LinFACT Algorithm.}\label{subsubsec_LinFACT} The complete LinFACT algorithm is presented in Algorithm~\ref{alg_LinFACT}. The procedure proceeds as follows: based on the collected data, the decision-maker updates the parameter estimates and checks whether the stopping condition $\tau_\delta$ is satisfied. If so, the set $G_r$ is returned as the estimated set of all $\varepsilon$-best arms. If the stopping condition is not met, the process continues with further sampling and updates. 

\vspace{0.4cm}
\begin{breakablealgorithm}\label{alg_LinFACT}
\caption{LinFACT Algorithm}
\begin{algorithmic}[1]
\State \textbf{Input:} $\varepsilon$, $\delta$, bandit instance.
\State Initialize $G_0 = \emptyset$, the set of good arms, and $B_0 = \emptyset$, the set of bad arms.
\State Initialize the active set $\mathcal{A}(0) = \mathcal{A}$, and $\mathcal{A}_{I}(0) = [K]$.
\For{$r = 1, 2, \ldots$}
    \State Set $C_{\delta/K}(r) = \varepsilon_r = 2^{-r}$.
    \State Set $G_r = G_{r-1}$ and $B_r = B_{r-1}$.
    \State Project $\mathcal{A}(r-1)$ to a $d_r$-dimensional subspace that $\mathcal{A}(r-1)$ spans.
    \Comment{Projection}
    \If{Using G-optimal Sampling}
        \Comment{Sampling}
        \State Call Algorithm \ref{alg_LinFACT_sampling}.
    \ElsIf{Using $\mathcal{XY}$-optimal Sampling}
        \State Call Algorithm \ref{alg_LinFACT_sampling_XY}.
    \EndIf
    \State Estimate $\left( \hat{\mu}_i(r) \right)_{i \in \mathcal{A}_I(r-1)}$ using equations \eqref{esti_OLS} and \eqref{eq_estimation_1}.
    \Comment{Estimation}
    \State Call Algorithm \ref{alg_LinFACT_stopping}.
    \Comment{Stopping Condition and Decision Rule}
\EndFor
\end{algorithmic}
\end{breakablealgorithm}
\vspace{0.6cm}

\subsection{Upper Bounds of the LinFACT Algorithm}\label{subsec_Upper Bounds on the Sample Complexity of LinFACT Algorithm} 

Theorems~\ref{upper bound: Algorithm 1 G} and~\ref{upper bound: Algorithm 1 XY} establish upper bounds on the sample complexity of the proposed LinFACT algorithm. 

Let $T_G$ and $T_{\mathcal{XY}}$ denote the number of samples required under the G-optimal and $\mathcal{XY}$-optimal designs, respectively. The formal statements of these theorems are given below. 
\begin{theorem}[Upper Bounds, G-Optimal Design]\label{upper bound: Algorithm 1 G}
    For $\xi = \min(\alpha_\varepsilon, \beta_\varepsilon)/16$, there exists an event $\mathcal{E}$ such that $\mathbb{P}(\mathcal{E}) \geq 1-\delta$. On this event, the \textup{LinFACT} algorithm with the G-optimal sampling policy achieves an expected sample complexity upper bound given by
\begin{equation}\label{upper bound for algorithm1: 2}
    \mathbb{E}[T_G \mid \mathcal{E}] 
    = \mathcal{O} 
        \left(d \xi^{-2} \log\!\left( \frac{K}{\delta} \log_{2}(\xi^{-2}) \right) 
        + d^{2} \log(\xi^{-1})\right).
\end{equation}
\end{theorem}

The detailed proof of Theorem~\ref{upper bound: Algorithm 1 G} is presented in Section~\ref{proof of theorem: upper bound of LinFACT G}. However, the LinFACT algorithm based on the G-optimal design does not yield an upper bound that aligns with the lower bound. This limitation is discussed further in Section~\ref{additional insights}. In contrast, we will show that the algorithm using the $\mathcal{XY}$-optimal design achieves an upper bound that matches the lower bound up to a logarithmic factor. 
\begin{theorem}[Upper Bound, $\mathcal{XY}$-Optimal Design]\label{upper bound: Algorithm 1 XY}
    Assume that an instance of arms satisfies $\min_{i \in G_\varepsilon \setminus \{ 1 \}} \Vert \boldsymbol{a}_1 - \boldsymbol{a}_i \Vert^2 \geq L_2$ and $\max_{i \in [K]} \vert \mu_1 - \varepsilon - \mu_i \vert \leq 2$. There exists an event $\mathcal{E}$ such that $\mathbb{P}(\mathcal{E}) \geq 1-\delta$. On this event, the \textup{LinFACT} algorithm with the $\mathcal{XY}$-optimal sampling policy achieves an expected sample complexity upper bound given by
    \begin{equation}\label{upper bound for algorithm1: 3}
    \mathbb{E}[T_{\mathcal{XY}} \mid \mathcal{E}] 
    = \mathcal{O} 
        \left((\Gamma^\ast)^{-1} d \xi^{-1} \log (\xi^{-1}) \log\left( \frac{K}{\delta} \log(\xi^{-2}) \right) 
        + \frac{d}{\epsilon^2} \log(\xi^{-1})\right),
\end{equation}
where $\xi = \min(\alpha_\varepsilon, \beta_\varepsilon)/16$ is the minimum gap of the problem instance, $R_{\textup{upper}} = \max \left\{ \left\lceil \log_2 \frac{4}{\alpha_\varepsilon} \right\rceil, \left\lceil \log_2 \frac{4}{\beta_\varepsilon} \right\rceil \right\}$, and $(\Gamma^\ast)^{-1}$ is the lower bound term defined in Theorem \ref{lower bound: All-epsilon in the linear setting}.
\end{theorem}

The proof of the near-optimal upper bound in Theorem~\ref{upper bound: Algorithm 1 XY} is presented in Section~\ref{proof of theorem: upper bound of LinFACT XY} of the online appendix, where we make it clear how the lower bound helps to establish our upper bound.

\section{Model Misspecification}\label{subsec_Misspecified Linear Bandits}

In this section, we address the challenge of model misspecification, recognizing that real-world problems may deviate from perfect linearity. To account for such deviations, we propose an orthogonal parameterization-based algorithm, \emph{i.e.}, LinFACT-MIS\footnote{LinFACT-MIS builds upon LinFACT-G. We choose to extend LinFACT-G, rather than LinFACT-$\mathcal{XY}$, for reasons of analytical tractability and computational efficiency. Both variants exhibit comparable performance in terms of upper bounds in this setting, while G-optimal design offers a significantly lower computational burden.}, a refined version of LinFACT to address model misspecification. We establish new upper bounds in the misspecified setting and provide insights into how such deviations impact algorithm performance. 

Under model misspecification, we refine the linear model in equation \eqref{eq_linear_model} as
\begin{equation}\label{eq_linear_model_mis}
    X_t = \boldsymbol{a}_{A_t}^\top \boldsymbol{\theta} + \eta_t + \Delta_m(\boldsymbol{a}_{A_t}),
\end{equation}
where $\Delta_m: \mathbb{R}^d \rightarrow \mathbb{R}$ is a misspecification function quantifying the deviation from the true model. 

\begin{assumption}
    Assume $\Vert \boldsymbol{\mu} \Vert_\infty \leq L_\infty$ and $\Vert \boldsymbol{\Delta}_m \Vert_\infty \leq L_m$, where $\Vert \cdot \Vert_\infty$ denotes the infinity norm and the bold $K$-dimensional vector $\boldsymbol{\Delta}_m$ represents the bias term of the misspecified model.
\end{assumption}

Therefore, with this assumption, the set of realizable models is defined as
\begin{equation}\label{eq_mis_realizable_model}
    M \coloneqq \left\{ \boldsymbol{\mu} \in \mathbb{R}^K \,\middle|\, \exists \boldsymbol{\theta} \in \mathbb{R}^d, \exists \boldsymbol{\Delta}_m \in \mathbb{R}^K, \boldsymbol{\mu} = \boldsymbol{\Psi}\boldsymbol{\theta} + \boldsymbol{\Delta}_m \; \land \; \Vert \boldsymbol{\mu} \Vert_\infty \leq L_\infty \; \land \; \Vert \boldsymbol{\Delta}_m \Vert_\infty \leq L_m \right\}.
\end{equation}

The key distinction in the analysis under model misspecification lies in how the estimator $\hat{\boldsymbol{\mu}}_t$ is maintained. Specifically, we construct this estimator by projecting the empirical mean vector $\tilde{\boldsymbol{\mu}}_t$ at time $t$ onto the set of realizable models $M$ via the following optimization
\begin{equation}\label{eq_optimization_esti}
    \hat{\boldsymbol{\mu}}_t \coloneqq \arg\min_{\boldsymbol{\vartheta} \in M} \Vert \boldsymbol{\vartheta} - \tilde{\boldsymbol{\mu}}_t \Vert_{\boldsymbol{D}_{\boldsymbol{N}_t}}^2,
\end{equation}
where $\boldsymbol{N}_t = [N_{t1}, N_{t2}, \ldots, N_{tK}]^\top \in \mathbb{R}^K$ is the vector of sample counts for each arm at time $t$, and $\boldsymbol{D}_{\boldsymbol{N}_t} \in \mathbb{R}^{K \times K}$ is the diagonal matrix with $N_{t1}, N_{t2}, \ldots, N_{tK}$ as its diagonal entries. 
\begin{figure}[htbp]
    \centering
    \includegraphics[scale=1]{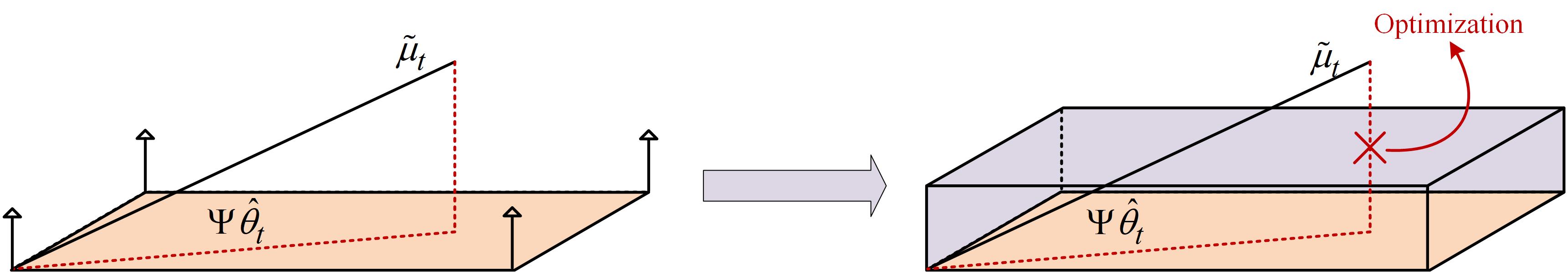}
    \caption{Difference Between Standard OLS and Misspecification-Adjusted Projection Estimates}
\vspace{5pt}
\noindent

  \par
  {\footnotesize
    \parbox{\linewidth}{\raggedright
      \textit{Note.} \textit{The left diagram shows the projection onto the span of pulled arms under a perfect linear model. The right diagram depicts the adjustment required under misspecification, where the projection must account for the deviation.}
    }%
  }
\vspace{-20pt}
    \label{fig: mis_2to3}
\end{figure}

In the absence of misspecification, this projection simplifies to the ordinary least squares (OLS) estimator. However, as shown in Figure~\ref{fig: mis_2to3}, under model misspecification, the estimator can no longer be computed as a simple projection onto a hyperplane and falls outside the scope of standard OLS. Instead, it must be formulated as the optimization problem in equation~\eqref{eq_optimization_esti}, which minimizes a weighted quadratic objective over $K + d$ variables, subject to the constraints $\Vert \boldsymbol{\mu} \Vert_\infty \leq L_\infty$ and $\Vert \boldsymbol{\Delta}_m \Vert_\infty \leq L_m$.

\subsection{Upper Bound with Misspecification}\label{subsec_Upper Bound When the Model Misspecification is Considered}
In this subsection, we present the upper bound on the expected sample complexity for LinFACT-MIS. This analysis highlights the influence of misspecification on the theoretical performance of the algorithm. Let $T_{G_{\textup{mis}}}$ denote the number of samples taken under model misspecification. 
\begin{theorem}[Upper Bound, Misspecification]\label{upper bound: Algorithm 1 G_mis_1}
    Fix $\varepsilon > 0$ and suppose that the magnitude of misspecification satisfies $L_m < \min \left\{ \frac{\alpha_\varepsilon}{2 \sqrt{d} }, \frac{\beta_\varepsilon}{2 \sqrt{d} } \right\}$. For $\xi = \min \left(\alpha_\varepsilon - 2L_m\sqrt{d}, \beta_\varepsilon - 2L_m\sqrt{d}\right)/16$, there exists an event $\mathcal{E}$ such that $\mathbb{P}(\mathcal{E}) \geq 1-\delta$. On this event, LinFACT-MIS terminates and returns the correct solution with an expected sample complexity upper bound given by
    \begin{align}\label{eq_mis_sample_complexity_mis}
        \mathbb{E}_{\boldsymbol{\mu}}\left[T_{G_{\textup{mis}}} \mid \mathcal{E}\right] = \mathcal{O} 
        \left(d \xi^{-2} \log\!\left( \frac{K}{\delta} \log(\xi^{-2}) \right) 
        + d^{2} \log(\xi^{-1})\right).
    \end{align}
\end{theorem}

The proof of this theorem is provided in Section~\ref{sec_Proof of upper bound: Algorithm 1 G_mis_1}. The upper bound in Theorem~\ref{upper bound: Algorithm 1 G}, which assumes no model misspecification, can be viewed as a special case of Theorem~\ref{upper bound: Algorithm 1 G_mis_1} by setting $L_m = 0$. However, the bound in Theorem~\ref{upper bound: Algorithm 1 G_mis_1} becomes invalid when the misspecification magnitude $L_m$ is too large, as the logarithmic terms may involve negative arguments, violating the assumptions required for the bound to hold. If the variation in confidence radius across arms due to misspecification is not accounted for, Theorem~\ref{upper bound: Algorithm 1 G_mis_1} suggests that the sample complexity will increase. In particular, compared to the bound in Theorem \ref{upper bound: Algorithm 1 G}, this result is looser because its denominator terms about $\varepsilon$ decrease from $\alpha_\varepsilon$ and $\beta_\varepsilon$ to $\alpha_\varepsilon-2L_m \sqrt{d}$ and $\beta_\varepsilon-2L_m \sqrt{d}$, respectively.

\subsection{Orthogonal Parameterization}\label{subsec_Orthogonal Parameterization-Based LinFACT and Upper Bound}

In this section, we present an alternative version of LinFACT-MIS based on orthogonal parameterization, designed to improve computational efficiency. 

\begin{figure}[htbp]
    \centering
    \includegraphics[scale=1]{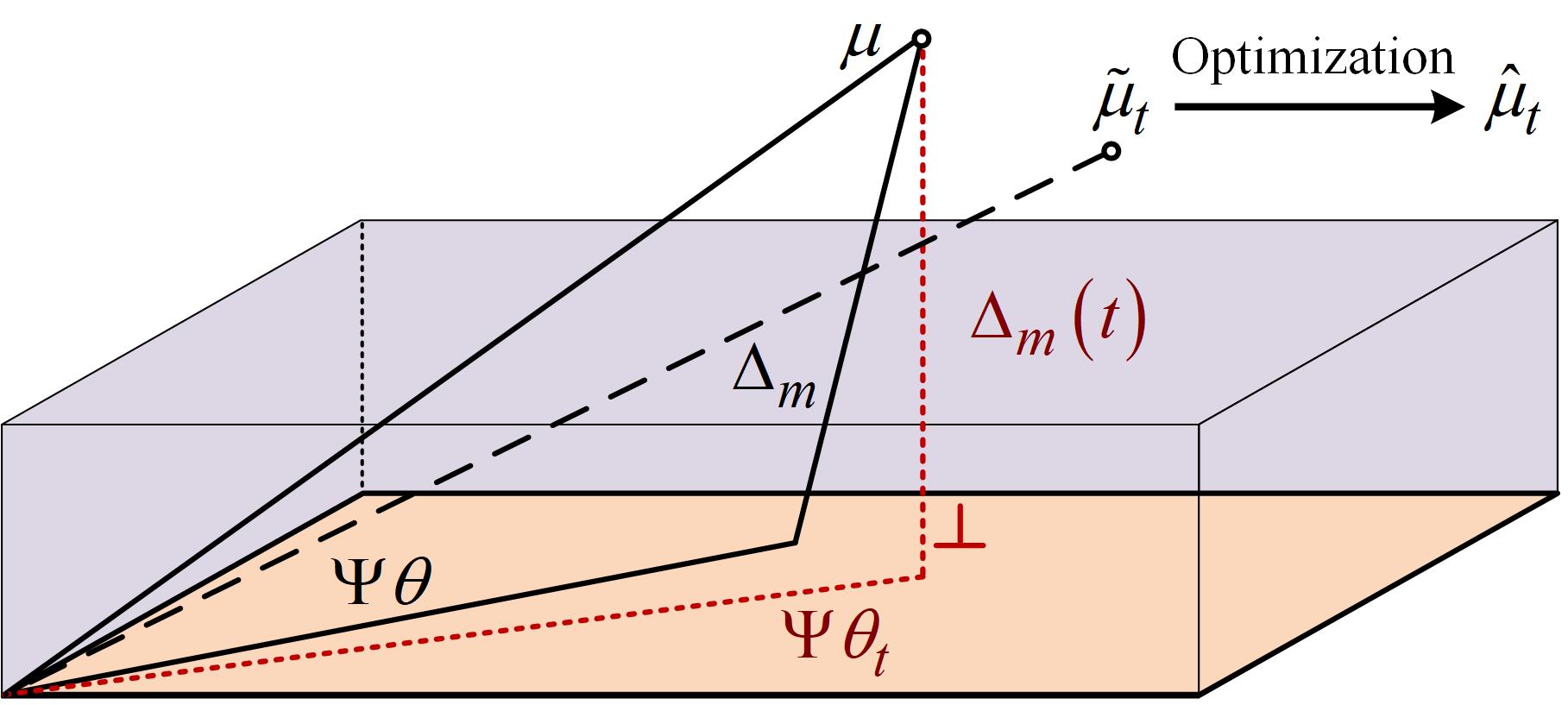}
    \caption{Orthogonal Parameterization and Projection.}
\vspace{5pt}
\noindent

  \par
  {\footnotesize
    \parbox{\linewidth}{\raggedright
      \textit{Note.} \textit{The estimator $\hat{\boldsymbol{\mu}}_t$ is obtained from the empirical mean $\tilde{\boldsymbol{\mu}}_t$ by solving an optimization problem. While the true mean vector $\boldsymbol{\mu}$ can be expressed as the sum of a linear component $\boldsymbol{\Psi\theta}$ and a non-linear model deviation $\boldsymbol{\Delta}_m$, it can also be decomposed at each time step $t$ via orthogonal projection into a linear part $\boldsymbol{\Psi\theta}_t$ on the hyperplane and a residual term $\boldsymbol{\Delta}_m(t)$ orthogonal to it.}
    }%
  }
\vspace{-15pt}
    \label{fig: ortho}
\end{figure}

\vspace{2mm}
\noindent\textbf{Orthogonal Parameterization.}\label{subsubsec_Orthogonal Parameterization} Under model misspecification, traditional confidence bounds for the mean estimator based on $\Vert \hat{\boldsymbol{\theta}}_t - \boldsymbol{\theta} \Vert_{\bm{V}_t}^2$, derived using either martingale-based methods \citep{abbasi2011improved} or covering arguments \citep{lattimore2020bandit}, are no longer directly applicable due to the presence of an additional misspecification term. To improve the concentration of the estimator in this setting, a key strategy is to adopt an orthogonal parameterization of the mean vectors within the realizable model $M$ \citep{reda2021dealing}. 

Rather than centering the confidence region around the true parameter $\boldsymbol{\theta}$, we focus on the quantity $\Vert \hat{\boldsymbol{\theta}}_t - \boldsymbol{\theta}_t \Vert_{\bm{V}_t}^2$, where $\boldsymbol{\theta}_t$ is the orthogonal projection of the true mean vector onto the feature space spanned by the pulled arms at time $t$. This $\boldsymbol{\theta}_t$-centered form corresponds to a self-normalized martingale and thus satisfies the same concentration bounds as in the classical linear bandit setting without misspecification. This approach offers an advantage over prior methods \citep{lattimore2020learning, zanette2020learning}, which require inflating the confidence radius between $\hat{\boldsymbol{\theta}}_t$ and $\boldsymbol{\theta}$ by a factor of $L_m^2 t$, leading to overly conservative bounds in misspecified settings where $L_m \gg 0$.

Specifically, we show that any mean vector $\boldsymbol{\mu} = \boldsymbol{\Psi}\boldsymbol{\theta} + \boldsymbol{\Delta}_m$ can be equivalently expressed at any time $t$ as $\boldsymbol{\mu} = \boldsymbol{\Psi}\boldsymbol{\theta}_t + \boldsymbol{\Delta_{m}(t)}$, where
\begin{equation}
    \boldsymbol{\theta}_t = \left( \boldsymbol{\Psi}_{N_t}^\top \boldsymbol{\Psi}_{N_t} \right)^{-1} \boldsymbol{\Psi}_{N_t}^\top \boldsymbol{D}_{N_t}^{1/2} \boldsymbol{\mu} = \bm{V}_t^{-1} \sum_{s=1}^t \mu_{A_s} \boldsymbol{a}_{A_s}
\end{equation}
is the orthogonal projection of $\boldsymbol{\mu}$ onto the feature space spanned by the columns of $\boldsymbol{\Psi}_{N_t}$, and $\boldsymbol{\Delta_{m}(t)} = \boldsymbol{\mu} - \boldsymbol{\Psi}\boldsymbol{\theta}_t$ is the residual. Here, $\boldsymbol{\Psi}_{N_t} = \boldsymbol{D}_{N_t}^{1/2} \boldsymbol{\Psi}$ is the matrix of feature vectors weighted by the number of times each arm has been sampled up to time $t$, and $\boldsymbol{D}_{N_t}^{1/2}$ is a diagonal matrix with entries corresponding to the square roots of the number of samples for each arm.

\vspace{2mm}
\noindent\textbf{Upper Bound.}\label{subsubsec_Upper Bound for the Orthogonal Parameterization-Based LinFACT} For LinFACT-MIS, the orthogonal parameterization involves updating the sampling rule in Algorithm \ref{alg_LinFACT_sampling} to the sampling rule in Algorithm \ref{alg_LinFACT_sampling_ortho}. The estimation is no longer based on OLS but is achieved by calculating the estimator $\left( \hat{\mu}_i(r) \right)_{i \in \mathcal{A}_I(r-1)}$ with observed data by solving the optimization problem described in equation~\eqref{eq_optimization_esti} directly. 

When model misspecification is accounted for and orthogonal parameterization is applied, the sampling policy is given by:
\begin{equation}\label{equation_phase budget G_mis_ortho}
    \left\{
    \begin{array}{l}
    T_{r}(\boldsymbol{a}) = \left\lceil \dfrac{8d\pi_{r}(\boldsymbol{a})}{\varepsilon_{r}^2} \left( d\log(6) + \log\left(\dfrac{Kr(r + 1)}{\delta}\right) \right) \right\rceil \\
    \\
    T_{r} = \sum_{\boldsymbol{a} \in \mathcal{A}(r-1)} T_r(\boldsymbol{a})
    \end{array}.
    \right.
\end{equation}

Let $T_{G_{\textup{op}}}$ denote the total number of samples required when orthogonal parameterization is used. The corresponding upper bound is stated in the theorem below. 

\vspace{0.4cm}
\begin{breakablealgorithm}\label{alg_LinFACT_sampling_ortho}
\caption{Subroutine: Sampling With Orthogonal Parameterization}
\begin{algorithmic}[1]
\State \textbf{Input:} Projected active set $\mathcal{A}_I(r-1)$, round $r$, $\delta$.
    \State Find the G-optimal design $\pi_r \in \mathcal{P}(\mathcal{A}(r-1))$ with Supp$(\pi_r) \leq \frac{d(d+1)}{2}$ according to equation \eqref{equation_G optimal}.\label{G_sampling_1_ortho}
    \ForAll{$i \in \mathcal{A}_I(r-1)$}
    \Comment{Sampling}
        \State Sample arm $\boldsymbol{a}$ for $T_r(\boldsymbol{a})$ times in round $r$, as specified in equation \eqref{equation_phase budget G_mis_ortho}. \label{G_sampling_2_ortho}
    \EndFor
\end{algorithmic}
\end{breakablealgorithm}
\vspace{0.6cm}
\begin{theorem}[Upper Bound, Orthogonal Parameterization]\label{upper bound: Algorithm 1 G_mis_2}
    Fix $\varepsilon > 0$ and suppose that the magnitude of misspecification satisfies $L_m < \min \left\{ \frac{\alpha_\varepsilon}{2 \left( \sqrt{d} + 2 \right)}, \frac{\beta_\varepsilon}{2 \left( \sqrt{d} + 2 \right)} \right\}$. For $\xi = \min \left(\alpha_\varepsilon - 2L_m(\sqrt{d}+2), \beta_\varepsilon - 2L_m(\sqrt{d}+2)\right)/16$, there exists an event $\mathcal{E}$ such that $\mathbb{P}(\mathcal{E}) \geq 1-\delta$. On this event, LinFACT-MIS terminates and returns the correct solution with an expected sample complexity upper bound given by
    \begin{align}\label{eq_mis_sample_complexity_orho_3}
        \mathbb{E}_{\boldsymbol{\mu}}\left[T_{G_{\textup{op}}} \mid \mathcal{E}\right] = \mathcal{O} 
        \left(d \xi^{-2} \log \left( \frac{K6^d}{\delta} \log(\xi^{-1}) \right) 
        + d^{2} \log(\xi^{-1})\right).
    \end{align}
\end{theorem}

This theorem establishes that the upper bound remains of the same order as in Theorem~\ref{upper bound: Algorithm 1 G_mis_1}, with the detailed proof provided in Section~\ref{sec_Proof of upper bound: Algorithm 1 G_mis_2}. As in Theorem~\ref{upper bound: Algorithm 1 G_mis_1}, the use of orthogonal parameterization does not eliminate the expansion of the upper bound, which remains unavoidable. We provide further intuition in Section~\ref{subsubsec_Improvement to Unknown Misspecification is theoretically Impossible}, arguing that without prior knowledge of the misspecification, it is not possible to recover full performance through algorithmic refinement alone. 

\subsection{Insights for Model Misspecification}\label{subsec_General Insights Related to the Misspecified Linear Bandits in the Pure Exploration Setting}

\vspace{2mm}
\noindent\textit{Lower Bounds in Linear and Stochastic Settings.}\label{subsubsec_Relationship Between the Lower Bounds in the Linear Setting and the Stochastic Setting} The lower bound becomes equivalent to the unstructured lower bound as soon as the misspecification upper bound $L_m > L_{\boldsymbol{\mu}}$, where $L_{\boldsymbol{\mu}}$ is an instance-dependent finite constant. This observation is formalized in Proposition \ref{propos_Relationship Between the Lower Bounds in the Linear Setting and the Stochastic Setting}, whose proof follows the same logic as Lemma 2 in \cite{reda2021dealing}.

\begin{proposition}\label{propos_Relationship Between the Lower Bounds in the Linear Setting and the Stochastic Setting}
    There exists $L_{\boldsymbol{\mu}} \in \mathbb{R}$ with $L_{\boldsymbol{\mu}} \leq \max_{k} \mu_{k} - \min_{k} \mu_{k}$ such that if $L_m > L_{\boldsymbol{\mu}}$, then for any pure exploration task, the lower bound in the linear setting is equal to the unstructured lower bound.
\end{proposition}

\vspace{2mm}
\noindent\textit{Improvement with Unknown Misspecification is Not Possible.}\label{subsubsec_Improvement to Unknown Misspecification is theoretically Impossible} Knowing that a problem is misspecified without access to an upper bound $L_m$ on $\|\boldsymbol{\Delta}_m\|_\infty$ is effectively equivalent to having no structural knowledge of the problem. As a result, improving algorithmic performance under such unknown model misspecification is infeasible. In particular, as shown in cumulative regret settings, sublinear regret guarantees are no longer achievable \citep{ghosh2017misspecified, lattimore2020learning}; similarly, in pure exploration, the theoretical lower bound cannot be attained.

\vspace{2mm}
\noindent\textit{Prior Knowledge for the Misspecification.}\label{subsubsec_Prior Knowledge for the Misspecification} When the upper bound $L_m$ on model misspecification is known in advance, LinFACT-MIS can be modified to account for this deviation. Specifically, we adjust the confidence radius $C_{\delta/K}(r)$ used in computing the lower and upper bounds (\emph{i.e.}, $L_r$ and $U_r$) in Algorithm~\ref{alg_LinFACT}. With this modification, the number of rounds required to complete classification under misspecification, $R_{\text{upper}}^\prime$ and $R_{\text{upper}}^{\prime\prime}$, coincides with $R_{\textup{upper}}$, the corresponding number of rounds under a perfectly linear model. This is achieved by replacing the confidence radius $\varepsilon_r$ with an inflated version $\varepsilon_r + L_m \sqrt{d}$ to compensate for the worst-case deviation due to misspecification. This adjustment preserves the validity of the original analysis and ensures that the same theoretical guarantees are retained. The following proposition formalizes this observation and is proved in Section~\ref{sec_Proof of propos_prior}.

\begin{proposition}\label{propos_prior}
   Suppose the misspecification magnitude $L_m > 0$ is known in advance. Adjusting the confidence radius $C_{\delta/K}(r)$ in Algorithm \ref{alg_LinFACT} at each round $r$ from its original value $\varepsilon_r$ to
\begin{equation}
    \varepsilon_r^\prime = \varepsilon_r + L_m \sqrt{d},
\end{equation}
ensures that the total number of rounds required under misspecification matches that under perfect linearity.
\end{proposition}

\section{Generalized Linear Model}\label{subsec_Generalized Linear Model}

In this section, we extend the linear bandits to a generalized linear model (GLM). In this setting, the reward function no longer follows the standard linear form in equation \eqref{eq_linear_model}, but instead satisfies
\begin{equation}
    \mathbb{E} \left[ X_t \mid A_t \right] = \mu_{\text{link}}(\bm{a}_{A_t}^\top\bm{\theta}),
\end{equation}
where $\mu_{\text{link}}:\mathbb{R}\rightarrow\mathbb{R}$ is the inverse link function. GLMs encompass a class of models that include, but are not limited to, linear models, allowing for various reward distributions beyond the Gaussian. For example, for binary-valued rewards, a suitable choice of $\mu_{\text{link}}$ is $\mu_{\text{link}}(x) = \exp(x)/(1+\exp(x))$, \emph{i.e.}, sigmoid function, leading to the logistic regression model. For integer-valued rewards,  $\mu_{\text{link}}(x) = \exp(x)$ leads to the Poisson regression model. 

To keep this paper self-contained, we briefly review the main properties of GLMs \citep{mccullagh2019generalized}. A univariate probability distribution belongs to a canonical exponential family if its density with respect to a reference measure is given by
\begin{equation}\label{eq_exp_family}
    p_{\omega}(x) = \exp \left( x\omega - b(\omega) + c(x) \right),
\end{equation}
where $\omega$ is a real parameter, $c(\cdot)$ is a real normalization function, and $b(\cdot)$ is assumed to be twice continuously differentiable. This family includes the Gaussian and Gamma distributions when the reference measure is the Lebesgue measure, and the Poisson and Bernoulli distributions when the reference measure is the counting measure on the integers. For a random variable $X$ with density defined in \eqref{eq_exp_family}, $\mathbb{E}(X) = \dot{b}(\omega)$ and $\text{Var}(X) = \ddot{b}(\omega)$, where $\dot{b}$ and $\ddot{b}$ denote the first and second derivatives of $b$, respectively. Since the variance is always positive and $\mu_{\text{link}} = \dot{b}$ represents the inverse link function, $b$ is strictly convex and $\mu_{\text{link}}$ is increasing. 

The canonical GLM assumes that $p_{\bm{\theta}}(X \mid \bm{a}_i) = p_{\bm{a}_i^\top\bm{\theta}}(X)$ for all arms $i$. The maximum likelihood estimator $\hat{\bm{\theta}}_t$, based on $\sigma$-algebra $\mathcal{F}_t = \sigma\left(A_1, X_1, A_2, X_2, \ldots, A_t, X_t\right)$, is defined as the maximizer of the function
\begin{equation}\label{eq_optimization_esti_glm}
    \sum_{s=1}^{t} \log p_{\bm{\theta}}(X_s|\bm{a}_{A_s}) = \sum_{s=1}^{t} X_s \bm{a}_{A_s}^\top \bm{\theta} - b(\bm{a}_{A_s}^\top \bm{\theta}) + c(X_s).
\end{equation}

This function is strictly concave in $\bm{\theta}$. By differentiating, we obtain that $\hat{\bm{\theta}}_t$ is the unique solution of the following estimating equation at time $t$,
\begin{equation}\label{eq_GLM_mle}
    \sum_{s=1}^{t} \left( X_s - \mu_{\text{link}}(\bm{a}_{A_s}^\top \bm{\theta}) \right) \bm{a}_{A_s}=\bm{0}.
\end{equation}

In practice, while the solution to equation \eqref{eq_GLM_mle} does not have a closed-form solution, it can be efficiently found using methods such as iteratively reweighted least squares (IRLS) \citep{wolke1988iteratively}, which employs Newton’s method. Here, $\check{\bm{\theta}}_t$ is a convex combination of $\bm{\theta}$ and its maximum likelihood estimate $\hat{\bm{\theta}}_t$ at time $t$. The existence of $c_{\min}$ can be ensured by performing forced exploration at the beginning of the algorithm, incurring a sampling cost of $O(d)$ \citep{kveton2023randomized}.

\begin{assumption} \label{assmp:GLM}
The derivative of the inverse link function, $\mu_{\text{link}}$, is bounded, \emph{i.e.}, $c_{\min} \leq \dot{\mu}_{\text{link}}(\bm{a}^\top\check{\bm{\theta}}_t)$, for some $c_{\min} \in \mathbb{R}^+$ and all arms. \end{assumption}

Assumption \ref{assmp:GLM} is standard in the GLM literature \citep{li2017provably, azizi2021fixed}, ensuring that the reward function is sufficiently smooth, with $c_{\min}>0$ typically determined by the choice of link function. 

\subsection{Algorithm with GLM}\label{subsubsec_GLM-Based LinFACT}
In this section, we present a refined algorithm for the generalized linear model, referred to as LinFACT-\textup{GLM}. This refinement involves modifying the sampling rule in Algorithm \ref{alg_LinFACT_sampling} and adjusting the estimation method. The designed sampling policy is described by
\begin{equation}\label{equation_phase budget G_GLM}
\left\{
\begin{array}{l}
    T_{r}(\boldsymbol{a}) = \left\lceil \dfrac{2d\pi_{r}(\boldsymbol{a})}{\varepsilon_{r}^2 c_{\min}^2} \log\left(\dfrac{2Kr(r + 1)}{\delta}\right) \right\rceil \\ \\
    T_{r} = \sum_{\boldsymbol{a} \in \mathcal{A}(r-1)} T_r(\boldsymbol{a})
\end{array},
\right.
\end{equation}
where $c_{\min}$ is the known constant controlling the first-order derivative of the inverse link function.

\vspace{0.4cm}
\begin{breakablealgorithm}\label{alg_LinFACT_sampling_GLM}
\caption{Subroutine: G-Optimal Sampling with GLM}
\begin{algorithmic}[1]
    \State \textbf{Input:} Projected active set $\mathcal{A}_I(r-1)$, round $r$, $\delta$.
    \State Find the G-optimal design $\pi_r \in \mathcal{P}(\mathcal{A}(r-1))$ with support size $\text{Supp}(\pi_r) \leq \frac{d(d+1)}{2}$ according to equation \eqref{equation_G optimal}.\label{G_sampling_1_GLM}
    \ForAll{$\boldsymbol{a} \in \mathcal{A}_I(r-1)$}
        \Comment{Sampling}
        \State Sample arm $\boldsymbol{a}$ for $T_r(\boldsymbol{a})$ times in round $r$, as specified in equation~\eqref{equation_phase budget G_GLM}.\label{G_sampling_2_GLM}
    \EndFor
\end{algorithmic}
\end{breakablealgorithm}
\vspace{0.6cm}

In the GLM setting, Ordinary Least Squares (OLS) is also not applicable. Instead, the estimator for each $i \in \mathcal{A}_I(r-1)$ is obtained by solving the optimization problem described in equation~\eqref{eq_optimization_esti_glm}, using the estimating equation in \eqref{eq_GLM_mle} derived from the observed data. 

\subsection{Upper Bound for the GLM-Based LinFACT}\label{subsubsec_Upper Bound for the GLM-Based LinFACT}

Let $T_{G_{\textup{GLM}}}$ denote the number of samples collected under the GLM setting. The following theorem provides an upper bound on the expected sample complexity of LinFACT-\textup{GLM}. 

\begin{theorem}[Upper Bound, Generalized Linear Model]\label{upper bound: Algorithm 1 GLM}
    For $\xi = \min(\alpha_\varepsilon, \beta_\varepsilon)/16$, there exists an event $\mathcal{E}$ such that $\mathbb{P}(\mathcal{E}) \geq 1-\delta$. On this event, the \textup{LinFACT-GLM} algorithm achieves an expected sample complexity upper bound given by
\begin{equation}\label{eq_GLM_sample_complexity_paper}
    \mathbb{E}[T_\textup{GLM} \mid \mathcal{E}] 
    = \mathcal{O} 
        \left(\frac{d}{c_{\min}^2} \xi^{-2} \log\!\left( \frac{K}{\delta} \log_{2}(\xi^{-2}) \right) 
        + d^{2} \log(\xi^{-1})\right).
\end{equation}
\end{theorem}

The upper bound presented in Theorem \ref{upper bound: Algorithm 1 GLM}, which generalizes the model to the GLM setting, can be viewed as an extension of Theorem \ref{upper bound: Algorithm 1 G}. The detailed proof is provided in Section \ref{proof of theorem: upper bound of LinFACT GLM} of the online appendix. 

\section{Numerical Experiments}\label{sec_Simulation Experiments} 
In the numerical experiments, we compare our algorithm, LinFACT, with several baseline methods. These include the Bayesian optimization algorithm based on the knowledge-gradient acquisition function with correlated beliefs for best arm identification (KGCB) proposed by \citet{negoescu2011knowledge}; the gap-based algorithm for best arm identification (BayesGap) introduced by \citet{hoffman2013exploiting}; the track-and-stop algorithm for threshold bandits (Lazy TTS) developed by \citet{rivera2024optimal}; and two gap-based algorithms for top $m$ arm identification, LinGIFA and $m$-LinGapE, presented by \citet{reda2021top}, which represent the state-of-the-art for returning multiple candidates. 

Identifying all $\varepsilon$-best arms is often more challenging than identifying the top $m$ arms or arms above a given threshold. To address this, we adopt a random setting where both the number of $\varepsilon$-best arms and the $\varepsilon$-threshold are randomly sampled. In this setting, top $m$ algorithms and threshold bandit algorithms only have access to the expected reward values, ensuring a fair comparison. Since BayesGap and KGCB operate under a fixed-budget setting, we use the average sample complexity of LinFACT as the budget for comparison and evaluate their performance accordingly. For BAI algorithms, we select arms whose empirical means are within $\varepsilon$ of the empirical best arm once the budget is exhausted.

\begin{figure}[htbp]
  \centering
  \subfloat[Synthetic I - Adaptive Setting\label{fig:synthetic-adaptive-main}]{%
    \includegraphics[width=0.48\textwidth]{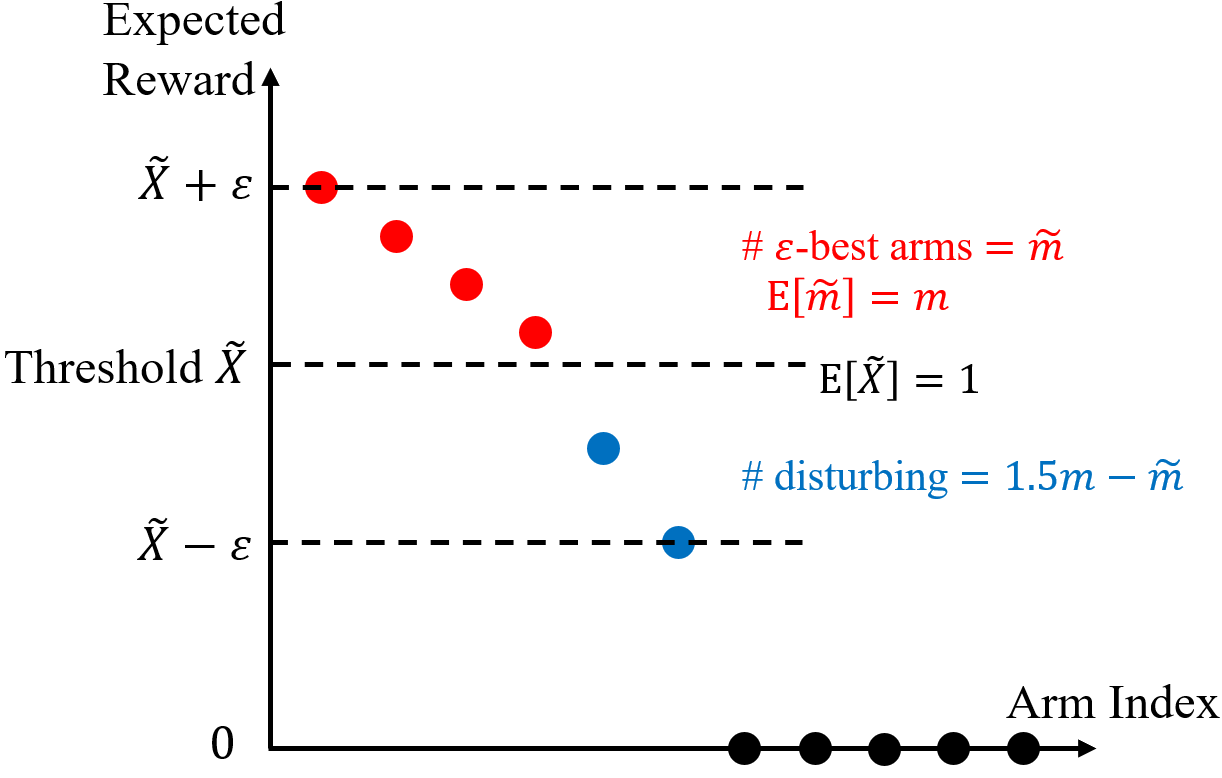}%
  }%
  \hfill
  \subfloat[Synthetic II - Static Setting\label{fig:synthetic-static-main}]{%
    \includegraphics[width=0.48\textwidth]{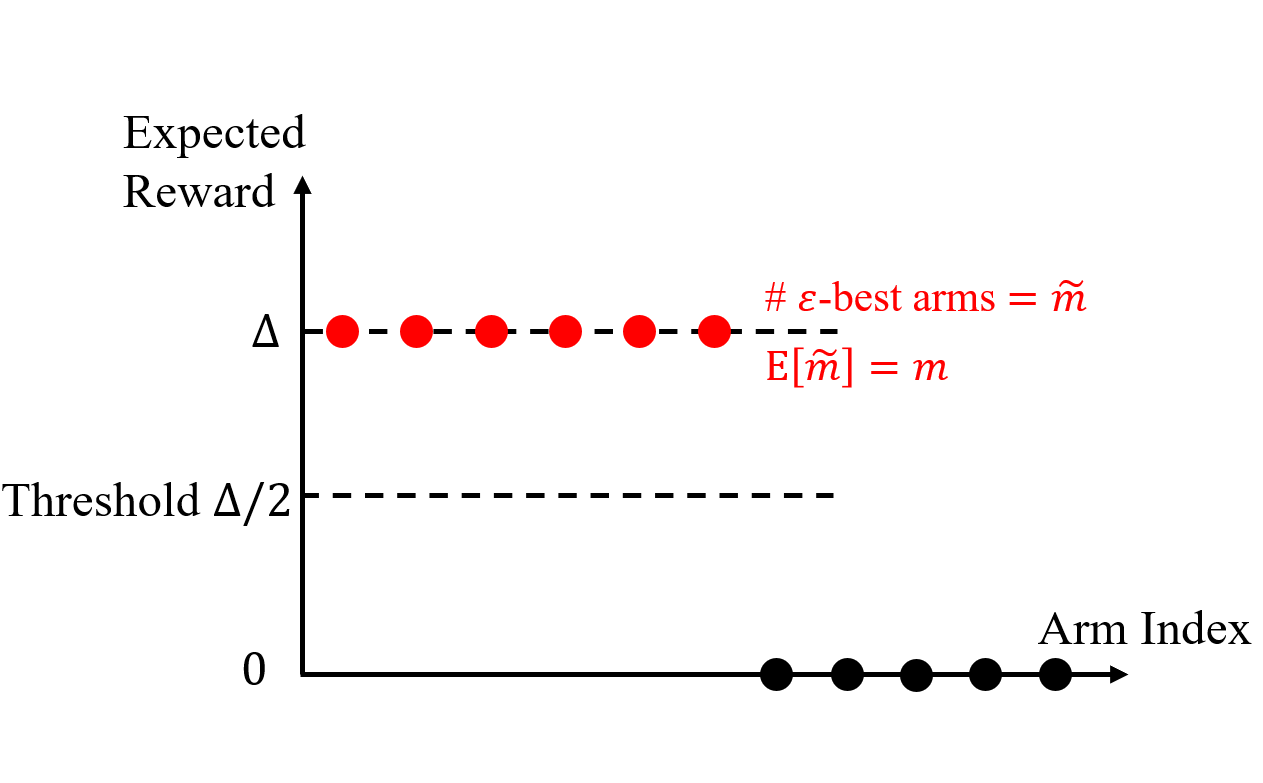}%
  }%

  \caption{Illustration of the Synthetic Experiment Settings}

  \par
  {\footnotesize
    \parbox{\linewidth}{\raggedright
      \textit{Note.} \textit{In the adaptive setting (a), we randomly sample the threshold $\tilde{X}$ from a distribution with mean 1, and independently sample the number of $\varepsilon$-best arms $\tilde{m}$ from a distribution with mean $m$. The arms above and below the threshold are then uniformly drawn from the intervals $[\tilde{X}, \tilde{X}+\varepsilon]$ and $[\tilde{X}-\varepsilon,\tilde{X})$, respectively. In the static setting (b), we fix the threshold at $\Delta/2$ and randomly sample $\tilde{m}$ $\varepsilon$-best arms with reward $\Delta$.}
    }%
  }
\vspace{-8pt}
  \label{fig:synthetic-main}
\end{figure}

\subsection{Synthetic Experiments}\label{sec:exp_setup} 
Following \citet{soare2014best}, \citet{xu2018fully}, and \citet{azizi2021fixed}, we categorize synthetic data into two types: adaptive and static settings. Figure~\ref{fig:synthetic-main} illustrates the construction of these synthetic datasets, with detailed configurations provided in Section~\ref{sec_Detailed Settings for Synthetic Experiments}. A summary of all settings is presented in Table~\ref{tab:synthetic_setting}. 

In the \textit{adaptive setting}, arms are divided into three categories: (1) the arms to be selected (\emph{i.e.}, the all $\varepsilon$-best arms), (2) disturbing arms that are slightly worse, and (3) base arms with zero rewards. The primary challenge for algorithms is to distinguish between the arms in categories (1) and (2), while the base arms in category (3) can be ignored. Adaptive algorithms that effectively leverage shared information to explore similar arms perform well in this setting.

In the \textit{static setting}, arms are divided into two categories: the all $\varepsilon$-best arms and the base arms with zero rewards. In this case, algorithms must distinguish between all arms. Static algorithms that uniformly explore all arms are well-suited for this setting.
\begin{table}[htbp]
    \small
    \centering
    \renewcommand\arraystretch{1.5}
    \caption{Synthetic Experiment Settings}
    \label{tab:synthetic_setting}
    \begin{tabular}{c c c}
        \hline
        \text{Setting Index} & \text{Setting Category} & \text{Setting Details} \\
        \hline
        1 & Adaptive & $(d,\mathbb{E}[m])=(8,4)$, $\varepsilon=0.1$ \\
        2 & Adaptive & $(d,\mathbb{E}[m])=(8,4)$, $\varepsilon=0.2$ \\
        3 & Adaptive & $(d,\mathbb{E}[m])=(8,4)$, $\varepsilon=0.3$ \\
        4 & Adaptive & $(d,\mathbb{E}[m])=(12,4)$, $\varepsilon=0.1$ \\
        5 & Adaptive & $(d,\mathbb{E}[m])=(12,4)$, $\varepsilon=0.2$ \\
        6 & Adaptive & $(d,\mathbb{E}[m])=(12,4)$, $\varepsilon=0.3$ \\
        \hline
        7 & Static & $(d,\Delta) = (8,1)$ \\
        8 & Static & $(d,\Delta) = (12,1)$ \\
        9 & Static & $(d,\Delta) = (16,1)$ \\
        \hline
    \end{tabular}
\end{table}
\vspace{-10.1pt}

\begin{remark}
    A common misconception is that adaptive algorithms universally outperform static ones. While adaptive algorithms are typically efficient at focusing exploration on promising arms in many settings, they can be less effective in static environments. In such cases, adaptive methods may inefficiently allocate samples between both candidate and baseline arms, leading to redundant exploration. In contrast, static algorithms can achieve the objective more efficiently by uniformly allocating samples across all arms, avoiding bias and over-exploration. 
\end{remark}

\vspace{2mm}
\noindent\textbf{Experiment Setup.}
We benchmark our algorithms, LinFACT-G and LinFACT-$\mathcal{XY}$, against BayesGap, KGCB, LinGIFA, m-LinGapE, and Lazy TTS, focusing on both sample complexity and the $F1$ score. 

We conduct each experiment using different data types (adaptive or static), arm dimensions ($d$), and numbers of arms ($K$). For each configuration, we generate 10 values of $m$ from a normal distribution centered at the expected value $\mathbb{E}[m]$ with a variance of 3.0, where $m$ is the input for a top-$m$ algorithm. For each sampled pair $(\tilde{m}, \tilde{X})$, where $\tilde{m}$ denotes the number of $\varepsilon$-best arms and $\tilde{X}$ represents the value of the best arm minus $\varepsilon$, we repeat the experiment 100 times. We then compute the average $F1$ score across the 10 $(\tilde{m}, \tilde{X})$ pairs, resulting in 1,000 total executions per algorithm.

In practice, we observe that KGCB, LinGIFA, m-LinGapE, and Lazy TTS are computationally intensive when the sampling budget is high. The time-consuming nature of KGCB has already been noted in the literature \citep{negoescu2011knowledge}. For Lazy TTS, the algorithm requires repeatedly evaluating an objective function within an optimization problem, where each evaluation has a time complexity of $O(Kd^2+d^3)$. As the optimization process involves a non-negligible number of iterations ($n$), the total time complexity becomes $O(nKd^2)$, making the algorithm inefficient.

For the two top-$m$ algorithms, the computational burden arises from performing matrix inversions for all arms, leading to a total time complexity of $O(Kd^3)$. In contrast, LinFACT achieves a significantly lower total time complexity of $O(Kd^2)$, as the optimization problem within our algorithm can be efficiently solved using a fixed-step gradient descent method. Table~\ref{tab:Synthetic_Time} presents the runtime for synthetic data, demonstrating that our algorithm is at least five times faster than all other methods. Notably, this performance gap is even more pronounced when using real data.

\begin{table}[htbp]
    \small
    \centering
    \renewcommand\arraystretch{1.5}
    \caption{Running Time (seconds) for Different Synthetic Experiments Among Algorithms}
    \begin{tabularx}{\textwidth}{l *{9}{>{\centering\arraybackslash}X}}
        \hline
        \multirow{3}{*}{\makecell{Algorithm \\ Settings}} & \multicolumn{6}{c}{Adaptive Settings} & \multicolumn{3}{c}{Static Settings} \\
        \cline{2-10}
        & \multicolumn{3}{c}{$(d,\mathbb{E}[m])=(8,4)$} & \multicolumn{3}{c}{$(d,\mathbb{E}[m])=(12,4)$} & \multicolumn{3}{c}{$\Delta=1$} \\
        \cline{2-10}
        & $\varepsilon=0.1$ & $\varepsilon=0.2$ & $\varepsilon=0.3$ & $\varepsilon=0.1$ & $\varepsilon=0.2$ & $\varepsilon=0.3$ & $d=8$ & $d=12$ & $d=16$ \\
        \hline
        LinFACT-G & \textbf{0.028} & \textbf{0.030} & \textbf{0.028} & \textbf{0.078} & \textbf{0.078} & \textbf{0.076} & \textbf{0.005} & \textbf{0.007} & \textbf{0.009}  \\ 
        LinFACT-$\mathcal{XY}$ & \underline{0.037} & \underline{0.038} & \underline{0.038} & \underline{0.163} & \underline{0.155} & \underline{0.145} & \underline{0.015} & \underline{0.028} & \underline{0.049}  \\ 
        BayesGap & 0.112 & 0.110 & 0.110 & 0.306 & 0.304 & 0.307 & 0.036 & 0.071 & 0.112  \\ 
        LinGIFA & 0.313 & 0.265 & 0.236 & 1.943 & 1.484 & 1.424 & 0.160 & 0.529 & 1.297  \\ 
        m-LinGapE & 0.195 & 0.166 & 0.157 & 1.362 & 1.053 & 0.984 & 0.116 & 0.352 & 0.932  \\ 
        Lazy TTS & 0.991 & 3.904 & 12.512 & 24.583 & 26.941 & 25.622 & 0.198 & 0.871 & 2.164  \\ 
        KGCB & 1.990 & 1.936 & 1.670 & 6.245 & 6.611 & 6.082 & 0.303 & 0.745 & 1.386  \\ 
        \hline
        
    \end{tabularx}
    \label{tab:Synthetic_Time}
    \vspace{-0.1cm}
    \vspace{5pt}
\noindent

  \par
  {\footnotesize
    \parbox{\linewidth}{\raggedright
      \textit{Note.} \textit{The best result is in \textbf{bold} and the second best is \underline{underlined}.}
    }%
  }

\end{table}

\begin{figure}[htbp]
    \centering
    \includegraphics[width=\textwidth]{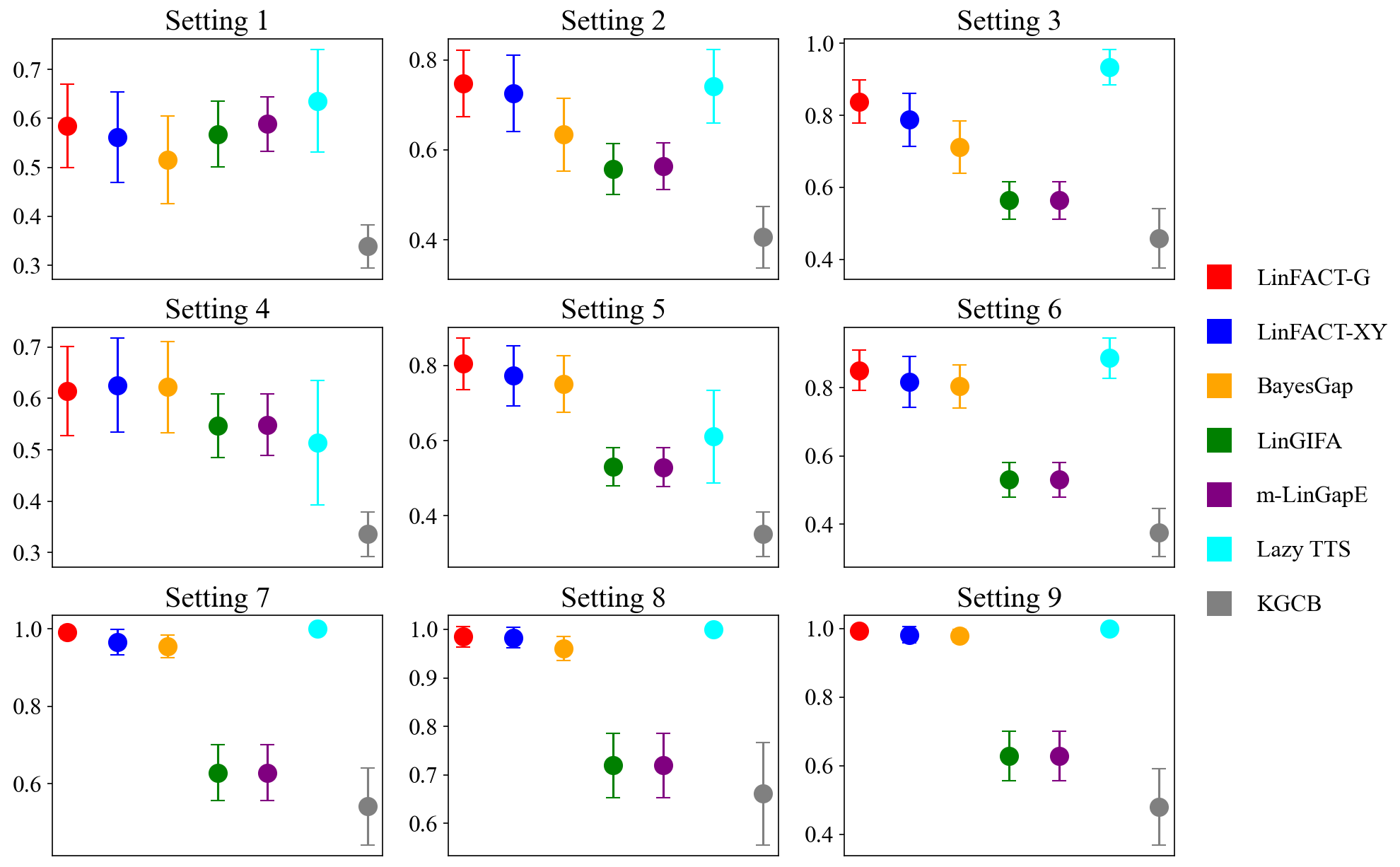}
    \caption{$F1$ Scores for Different Synthetic Experiments Among Algorithms}
\vspace{5pt}
\noindent

  \par
  {\footnotesize
    \parbox{\linewidth}{\raggedright
      \textit{Note.} \textit{The y-axis reports the F1 score, which reflects how accurately each algorithm identifies $\varepsilon$-best arms. LinFACT-G and LinFACT-$\mathcal{XY}$ consistently achieve high F1 scores. While Lazy TTS occasionally attains higher scores, we demonstrate in the next figure that it requires significantly more samples to do so. Detailed configurations of each experimental setting are provided in Table~\ref{tab:synthetic_setting}.}
    }%
  }
\vspace{-1pt}
    \label{fig:Synthetic_F1}
\end{figure}

\begin{figure}[htbp]
    \centering
    \includegraphics[width=.95\textwidth]{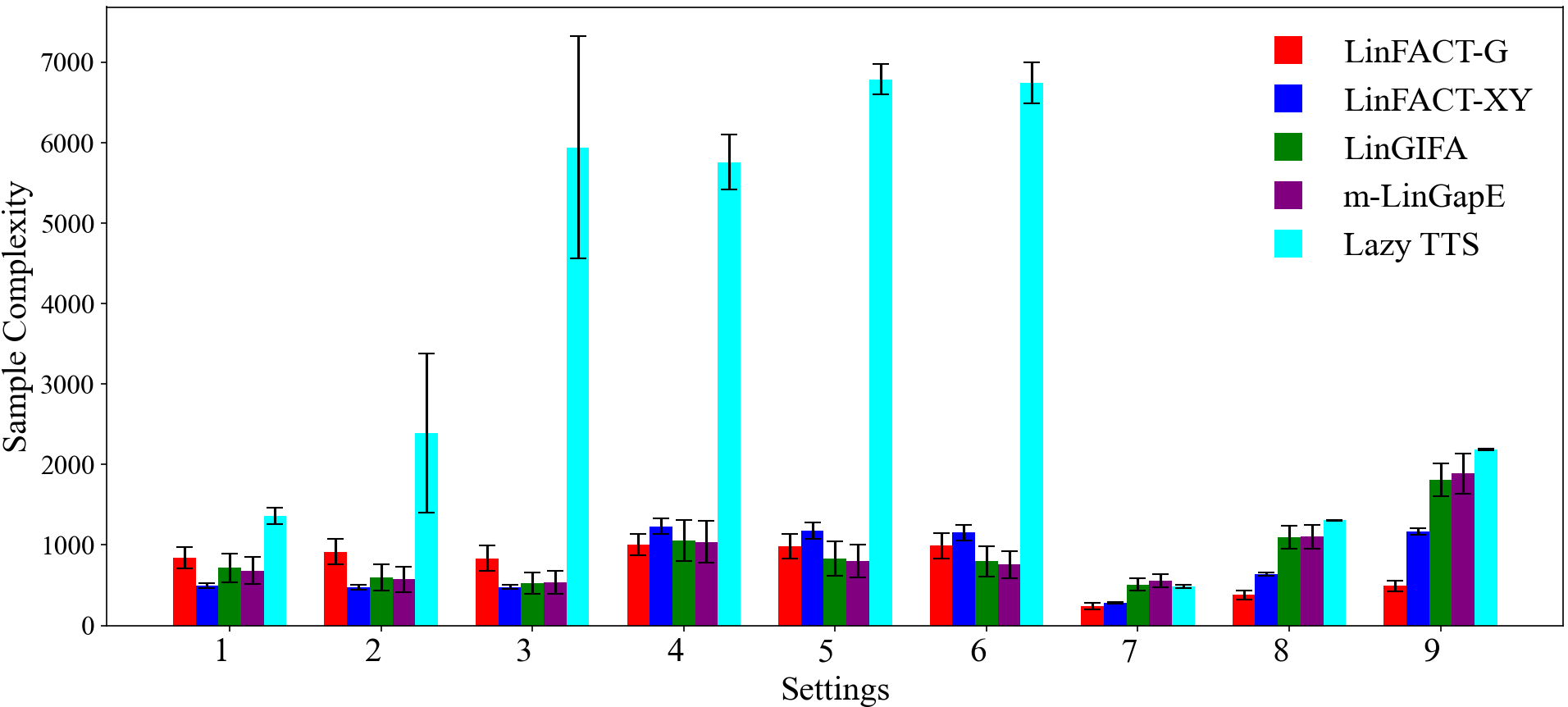}
    \caption{Sample Complexity for Different Synthetic Experiments Among Algorithms}
\vspace{5pt}
\noindent

  \par
  {\footnotesize
    \parbox{\linewidth}{\raggedright
      \textit{Note.} \textit{LinFACT-G and LinFACT-$\mathcal{XY}$ demonstrate sample complexities comparable to LinGIFA and m-LinGapE while achieving higher F1 scores in the adaptive settings (Settings 1 to 6), and consistently outperform all other algorithms in the static settings (Settings 7 to 9). BayesGap and KGCB are excluded from this comparison as they are designed for the fixed-budget setting, and thus their sample complexity is not well-defined.}
    }%
  }
\vspace{-10pt}
    \label{fig:Synthetic_Complexity}
\end{figure}

\vspace{2mm}
\noindent\textbf{Experiment Results.}
Our experimental results are presented in Figures~\ref{fig:Synthetic_F1} and ~\ref{fig:Synthetic_Complexity}. In Figure~\ref{fig:Synthetic_F1}, the vertical axis denotes the $F1$ score, with higher values indicating better algorithm performance. The first row of six plots shows the results under adaptive settings. As $\varepsilon$ increases, the non-optimal (disturbing) arms in the adaptive setting move progressively farther from the optimal arms, making them easier to distinguish. Consequently, the $F1$ score increases from left to right. An exception is BayesGap, which performs best when $\varepsilon=0.2$. This occurs because best-arm identification algorithms, such as BayesGap, struggle to differentiate optimal arms from disturbing ones when they are close ($\varepsilon=0.1$) and fail to fully explore the optimal arms when they are not closely clustered with the best arm ($\varepsilon=0.3$).

Our LinFACT algorithms consistently outperform top-$m$ algorithms, BayesGap, and KGCB. While our algorithms perform slightly worse than Lazy TTS for some cases, they have much lower sample complexity, as shown in Figure~\ref{fig:Synthetic_Complexity}, meaning that Lazy TTS requires substantially more samples to achieve these results. When comparing LinFACT-G and LinFACT-$\mathcal{XY}$, we observe that in adaptive settings, the $F1$ scores are similar, but LinFACT-$\mathcal{XY}$ achieves lower sample complexity. In static settings, however, LinFACT-G attains a higher $F1$ score with a reduced sample complexity. This difference stems from the distinct focus of the two designs: the $\mathcal{XY}$-optimal design prioritizes pulling arms to obtain better estimates along the directions representing differences between arms, while the G-optimal design aims to improve estimates along the directions representing all arms.

\subsection{Experiments with Real Data - Drug Discovery}
\vspace{2mm}
\noindent\textbf{Experiment Setup.}
We adopt the Free-Wilson model \citep{katz1977application} and use real data from a drug discovery task. The Free-Wilson model is a linear framework in which the overall efficacy of a compound is expressed as the sum of the contributions from each substituent on the base molecule, along with the effect of the base molecule itself \citep{negoescu2011knowledge}.

\begin{figure}[htbp]
    \centering
    \includegraphics[width=.9\textwidth]{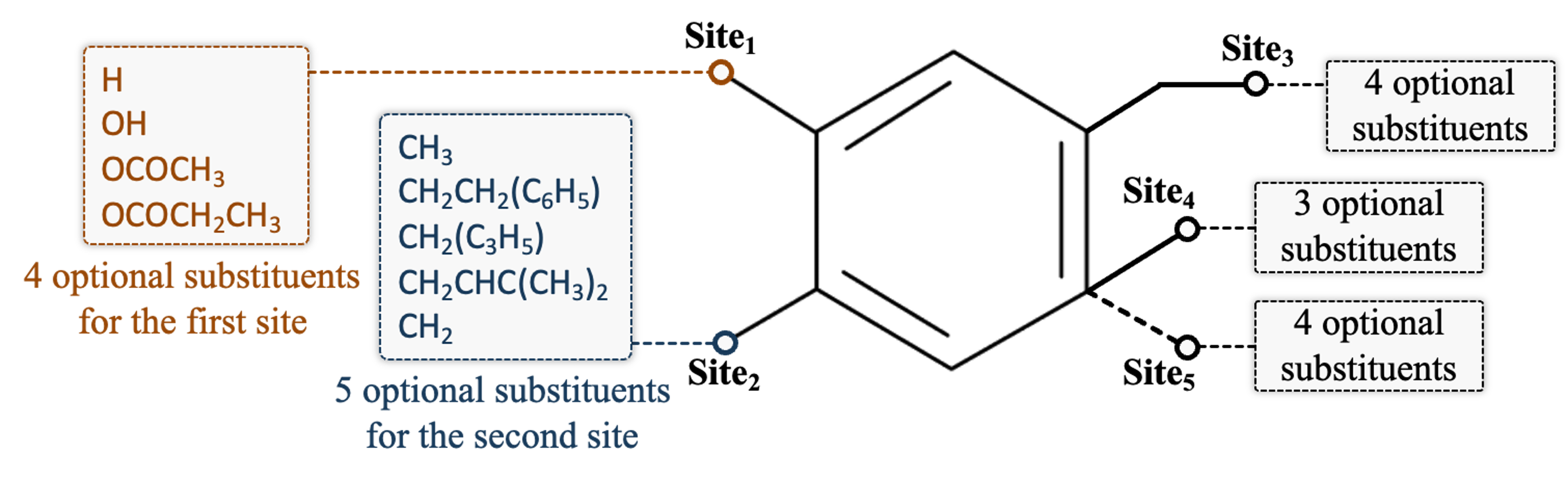}
    \caption{An Example of Molecule and Substituent Locations with $\textup{Site}_1, \ldots,\textup{Site}_5$
    }
    \vspace{5pt}
\noindent

  \par
  {\footnotesize
    \parbox{\linewidth}{\raggedright
      \textit{Note.} \textit{We begin with a base molecule containing multiple attachment sites for chemical substituents. By varying these substituents, we generate a diverse set of compounds and aim to identify those with desirable properties.}
    }%
  }
  \vspace{-20pt}
    \label{fig: drug_example}
\end{figure}

Each compound is modeled as an arm represented by a binary indicator vector. Suppose there are $N$ modification sites, with site $n \in [N]$ offering $l_n$ alternative substituents. Then each arm $\boldsymbol{a}$ lies in $\mathbb{R}^{1 + \sum_{n \in [N]} l_n}$, where the initial entry corresponds to the base molecule (\emph{i.e.}, the intercept term). For each site, the corresponding segment in the vector has exactly one entry set to 1 (indicating the chosen substituent), with the remaining $l_n - 1$ entries set to 0. This results in a total of $\prod_{n \in [N]} l_n$ unique compound configurations.

We conducted our experiments using data as described in \cite{katz1977application}, retaining only the non-zero entries. The base molecule, illustrated in Figure~\ref{fig: drug_example}, contains five sites where substituents can be attached. Each site offers 4, 5, 4, 3, and 4 candidate substituents, respectively, resulting in 960 possible compounds. Each arm $\boldsymbol{a}$ is represented as a vector in $\mathbb{R}^{21}$. 

We benchmarked our algorithms, LinFACT-G and LinFACT-$\mathcal{XY}$, against LinGIFA and m-LinGapE\footnote{KGCB and BayesGap are designed for fixed-budget settings, so their sample complexity is not well-defined. Lazy TTS, due to its prohibitive runtime, was excluded from these experiments.}, evaluating how precision, recall, $F1$ score, and sample complexity vary with the failure probability $\delta$. In this study, the good set was defined to include 20 $\varepsilon$-best arms, which corresponds to $\varepsilon = 4.325$. All methods were tested over 10 trials, with $\delta$ ranging from 0.1 to 0.9 in increments of 0.1. 

\begin{figure}[htbp]
  \centering

  \subfloat[Precision\label{fig:real_precision}]{%
    \includegraphics[width=0.6\textwidth,height=4cm,keepaspectratio]{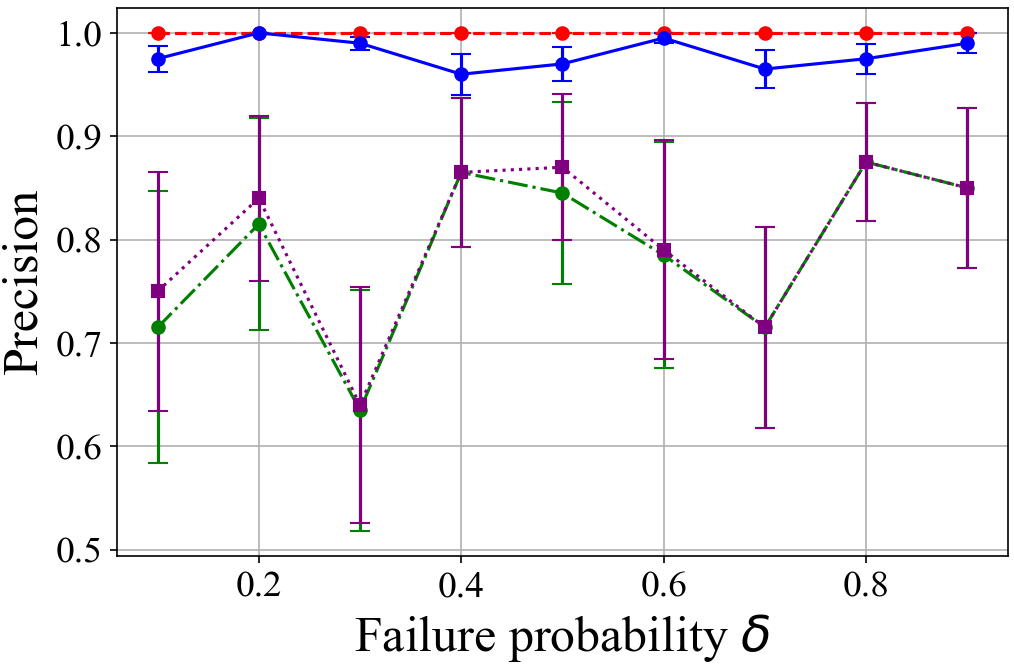}%
  }%
  \qquad
  \subfloat[Recall\label{fig:real_recall}]{%
    \includegraphics[width=0.6\textwidth,height=4cm,keepaspectratio]{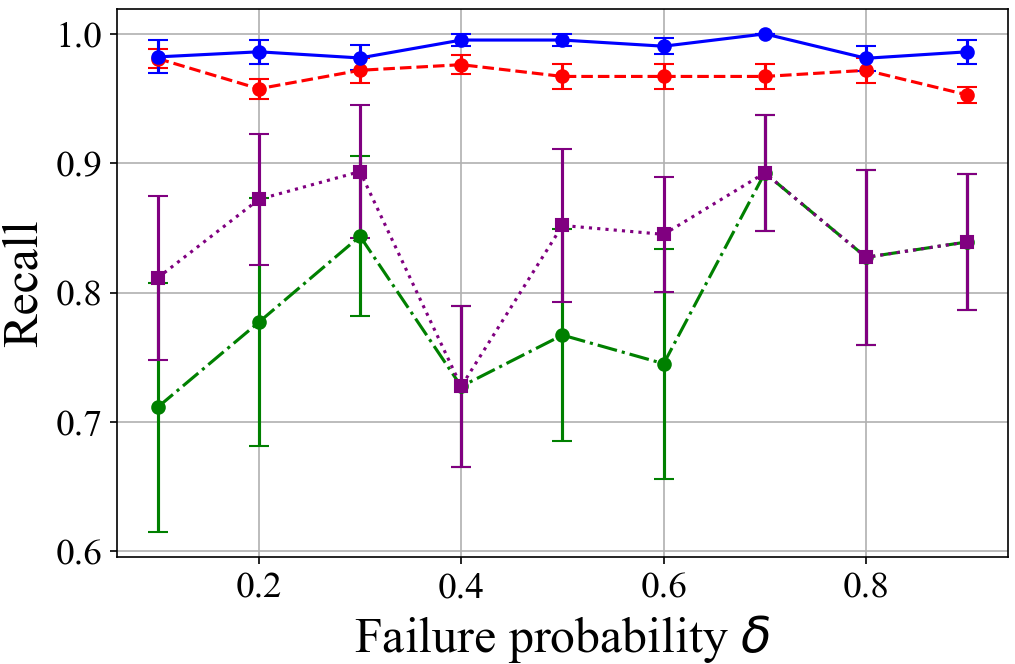}%
  }%

  \par

  \subfloat[$F1$ Score\label{fig:f1}]{%
    \includegraphics[width=0.6\textwidth,height=4cm,keepaspectratio]{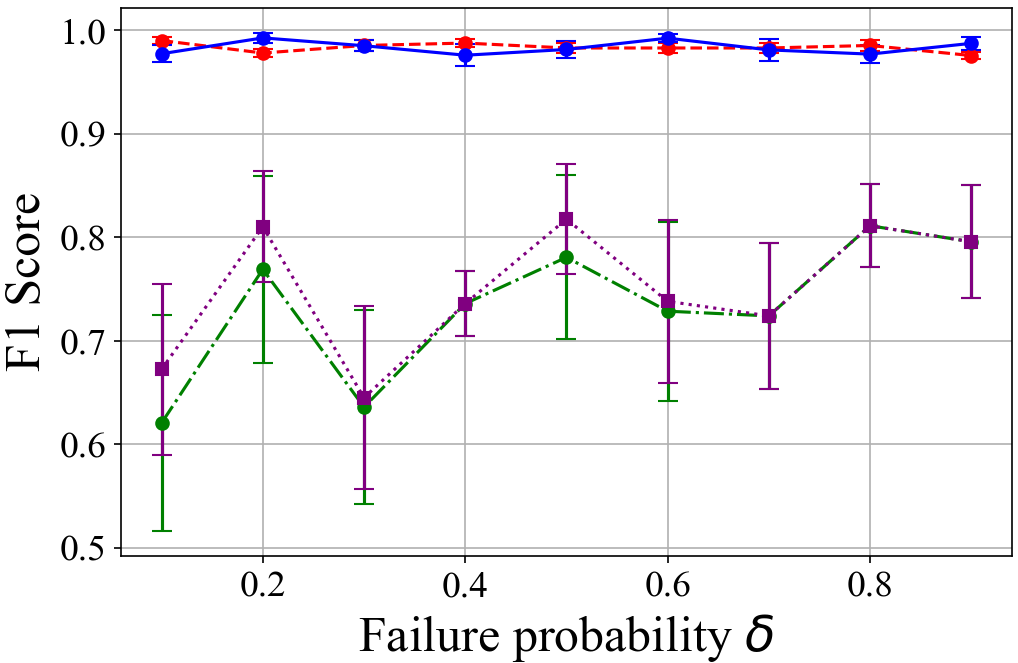}%
  }%
  \qquad
  \subfloat[Sample Complexity\label{fig:sample_complexity_real}]{%
    \includegraphics[width=0.6\textwidth,height=4cm,keepaspectratio]{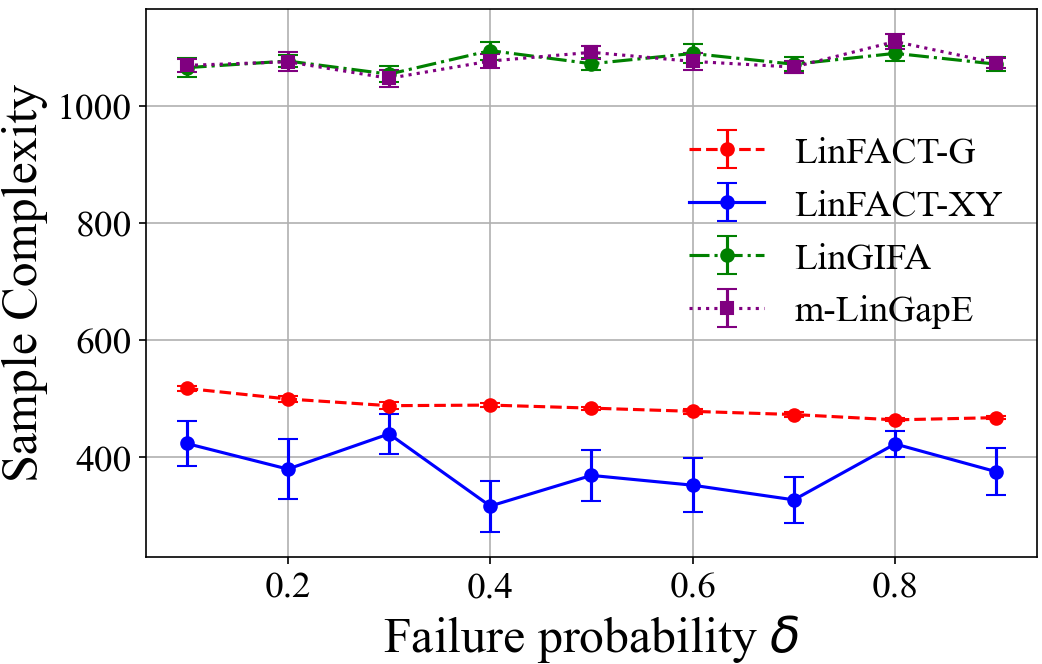}%
  }%

  \caption{Precision, Recall, and $F1$ Score for Various Failure Probabilities $\delta$}

  \par
  {\footnotesize
    \parbox{\linewidth}{\raggedright
      \textit{Note.} \textit{LinFACT-G and LinFACT-$\mathcal{XY}$ consistently demonstrate the best performance across all metrics. LinFACT-$\mathcal{XY}$ achieves the lowest sample complexity while maintaining high accuracy. BayesGap and KGCB are excluded as they operate under a fixed-budget setting. Lazy TTS is also omitted due to excessive runtime.}
    }%
  }
\vspace{-5pt}
  \label{fig:real_data_Drug}
\end{figure}

\noindent\textbf{Experiment results.}
The experimental results are shown in Figure~\ref{fig:real_data_Drug}. LinFACT-G and LinFACT-$\mathcal{XY}$ consistently deliver high precision, recall, and $F1$ scores across various failure probabilities $\delta$, as shown in Figures~\ref{fig:real_precision}–\ref{fig:f1}. In particular, their precision remains close to 1.0, indicating that nearly all selected arms are truly $\varepsilon$-best. Their recall is also robust across $\delta$, suggesting strong coverage of the good set with minimal omission. This balance between high precision and recall leads to $F1$ scores that remain consistently near optimal, even as $\delta$ varies from 0.1 to 0.9. In contrast, LinGIFA and m-LinGapE exhibit larger fluctuations and overall lower values in all three metrics, with especially degraded recall and $F1$ performance at moderate values of $\delta$.

In addition, as illustrated in Figure~\ref{fig:sample_complexity_real}, LinFACT-$\mathcal{XY}$ achieves the lowest sample complexity, followed closely by LinFACT-G, with both outperforming all baseline methods.   These performance disparities highlight the reliability and robustness of LinFACT methods. Finally, LinFACT-G and LinFACT-$\mathcal{XY}$ also demonstrate strong computational efficiency: both complete the task within one minute, whereas LinGIFA and m-LinGapE take about four times longer, and Lazy TTS requires over 30 minutes. 

\section{Conclusion}\label{sec_Conclusion} 

In this paper, we address the challenge of identifying all $\varepsilon$-best arms in linear bandits, motivated by applications such as drug discovery. We establish the first information-theoretic lower bound to characterize the problem's complexity and derive a matching upper bound. Our LinFACT algorithm achieves instance-optimal performance up to a logarithmic factor under the $\mathcal{XY}$-optimal design criterion.

We further extend our analysis to settings with model misspecification and generalized linear models (GLMs), deriving new upper bounds and providing insights into algorithmic behavior under these broader conditions. These results generalize and recover the guarantees from the perfectly linear case as special instances. Our numerical experiments confirm that LinFACT outperforms existing methods in both sample and computational efficiency, while maintaining high accuracy in identifying all $\varepsilon$-best arms. 

\noindent\textbf{Future Research Directions.} First, while our current work focuses on the fixed-confidence setting, many real-world applications operate under a fixed sampling budget, where the objective is to achieve the best outcome within a limited number of trials. Extending the algorithm to this fixed-budget setting and establishing corresponding theoretical guarantees remains an important avenue for exploration. Moreover, deriving fundamental lower bounds in the fixed-budget regime is still an open question. Second, although we have proposed extensions for both misspecified linear bandits and generalized linear models (GLMs), future work could benefit from developing separate algorithms tailored to each setting. Such targeted designs may offer improved performance. 


%
%
%


\bibliographystyle{informs2014} 
\bibliography{main}



\clearpage
\begin{appendices}
\renewcommand{\thepage}{ec\arabic{page}}  
\renewcommand{\thesection}{EC.\arabic{section}}   
\renewcommand{\thetable}{EC.\arabic{table}}   
\renewcommand{\thefigure}{EC.\arabic{figure}}
\renewcommand{\theequation}{EC.\arabic{equation}}
\renewcommand{\thetheorem}{EC.\arabic{theorem}}
\renewcommand{\theproposition}{EC.\arabic{proposition}}
\renewcommand{\thelemma}{EC.\arabic{lemma}}
\renewcommand{\thealgorithm}{EC.\arabic{algorithm}}
\counterwithin{equation}{section}
\counterwithin{table}{section}
\counterwithin{figure}{section}
\counterwithin{lemma}{section}
\setcounter{equation}{0}
\setcounter{section}{0}
\setcounter{page}{1}

\vspace*{0.1cm}
   \begin{center}
      \large\textbf{E-Companion -- Identifying All $\varepsilon$-Best Arms In Linear Bandits With Misspecification}\\
   \end{center}
   \vspace*{0.3cm}

\section{Additional Literature}\label{sec_Literature Review for Misspecified Linear Bandits and Generalized Linear Bandits}
\subsection{Misspecified Linear Bandits.}\label{subsubsec_Pure Exploration for Linear Bandits} 

The linear bandit (LB) problem, introduced by \citeec{abe1999associative}, extends the multi-armed bandits (MABs) framework by incorporating structural relationships among different arms. In the context of best arm identification, \citeec{garivier2016optimal} established a classical lower bound, which was later extended to linear bandits by \citeec{fiez2019sequential} using transportation inequalities.

The foundational study of linear bandits in the pure exploration framework was conducted by \citeec{hoffman2014correlation}, who addressed the best arm identification (BAI) problem in a fixed-budget setting while considering correlations among arm distributions. They proposed BayesGap, a Bayesian variant of the gap-based exploration algorithm \citepec{gabillon2012best}. Although BayesGap outperformed methods that ignore correlations and structural relationships, its limitation of ceasing to pull arms deemed sub-optimal hindered its effectiveness in linear bandit pure exploration.

A key distinction between stochastic MABs and linear bandits is that, in MABs, once an arm’s sub-optimality is confirmed with high probability, it is no longer pulled. In linear bandits, however, even sub-optimal arms can offer valuable information about the parameter vector, improving confidence in estimates and aiding the discrimination of near-optimal arms. This insight has led to the adoption of optimal linear experiment design as a crucial framework for linear bandit pure exploration \citepec{abbasi2011improved, soare2014best, fiez2019sequential, reda2021dealing, yang2021minimax, DBLP:journals/corr/abs-2106-04763}. 

When applying linear models to real data, misspecification inevitably arises in situations where the data deviates from perfect linearity.  The concept of misspecified bandit models was introduced in the context of cumulative regret by \citeec{ghosh2017misspecified}, who demonstrated a significant limitation: any linear bandit algorithm (\emph{e.g.}, OFUL \citepec{abbasi2011improved} or LinUCB \citepec{10.1145/1772690.1772758}), which achieves optimal regret bounds on perfectly linear instances, can suffer linear regret on certain misspecified models. To address this, they proposed a hypothesis-test-based algorithm that avoids linear regret and achieves UCB-type sublinear regret for models with non-sparse deviations from linearity. \citeec{lattimore2020learning} further analyzed misspecification, showing that elimination-based algorithms with G-optimal design perform well under misspecification but incur an additional linear regret term proportional to the misspecification magnitude over the horizon.

In the pure exploration setting, misspecified linear models were first studied in the context of identifying the top $m$ best arms by \citeec{reda2021dealing}, who introduced the MisLid algorithm, leveraging orthogonal parameterization to address misspecification. Subsequent research examined misspecification in ordinal optimization \citepec{doi:10.1287/mnsc.2022.00328}, proposing prospective sampling methods that reduce the impact of misspecification as the sample size increases. Building on the definition of model misspecification and the optimization approach based on orthogonal parameterization from \citeec{reda2021dealing}, we develop new algorithms for identifying all $\varepsilon$-best arms in misspecified linear bandits and establish new upper bounds.

\subsection{Generalized Linear Bandits.}\label{subsubsec_General Linear Bandits} 
The generalized linear bandit (GLB) model \citepec{filippi2010parametric, ahn2020ordinal, kveton2023randomized} extends the multi-armed bandit framework by incorporating generalized linear models (GLMs) \citepec{mccullagh2019generalized} to model expected rewards. Specifically, the expected reward of each arm is given by a known link function applied to the inner product of a feature vector and an unknown parameter vector. Most existing algorithms for generalized linear bandits employ the upper confidence bound (UCB) approach, with randomized GLM algorithms \citepec{chapelle2011empirical, russo2018tutorial, kveton2023randomized} demonstrating superior performance.

In the context of pure exploration, \citeec{azizi2021fixed} introduced the first practical algorithm for best arm identification in generalized linear bandits, supported by theoretical analysis. Their work extends the best arm identification problem from linear models to more complex settings where the relationship between features and rewards follows generalized linear models (GLMs). Building on this foundation, we extend the pure exploration setting from best arm identification (BAI) to identifying all $\varepsilon$-best arms, providing analogous analyses and theoretical results for GLMs. 

\section{General Pure Exploration Model}\label{sec_General Pure Exploration Model} 
In this section, we present a brief discussion about the general pure exploration problem. A more comprehensive explanation can be found in \citeec{qin2025dual}. 

The decision-maker seeks to answer a query concerning the mean parameters $\boldsymbol{\mu}$ by adaptively allocating the sampling budget across the available arms. This query typically involves identifying a subset of arms that satisfy certain criteria, and the goal is to determine the correct answer with high probability. 
Let $\mathcal{I} = \mathcal{I}(\boldsymbol{\mu})$ denote the correct answer, which in our setting corresponds to the set of all $\varepsilon$-best arms. Define $\widetilde{M}$ as the set of parameters that yield a unique answer. Let $\Xi$ represent the collection of all possible answers, and for each $\mathcal{I}^\prime \in \Xi$, define $M_{\mathcal{I}^\prime} \coloneqq \left\{\boldsymbol{\vartheta}\in\widetilde{M}: \mathcal{I}\left(\boldsymbol{\vartheta}\right) = \mathcal{I}^\prime\right\}$ as the set of parameters for which $\mathcal{I}^\prime$ is the correct answer. The overall parameter space of interest is then given by $M \coloneqq \bigcup_{\mathcal{I}^\prime \in \Xi} M_{\mathcal{I}^\prime}$. 

Recall that an algorithm is defined as a triplet $\mathcal{H} = \left ( A_t,\; \tau_\delta,\; \hat{a}_\tau \right )$. The algorithm's sample complexity is quantified by the number of samples, denoted as $\tau_\delta$, at the point of termination. The objective is to formulate algorithms that minimize the expected sample complexity $\mathbb{E}_{\boldsymbol{\mu}} \left [ \tau_\delta \right ]$ across the set $\mathcal{H}$. As stated in \citeec{kaufmann2016complexity}, when $\delta \in \left ( 0, 1 \right )$, the non-asymptotic problem complexity of an instance $\boldsymbol{\mu}$ can be defined as
\begin{equation}\label{eq_problem complexity_ec}
\kappa \left ( \boldsymbol{\mu} \right ) \coloneqq \inf\limits_{Algo \in \mathcal{H}} \frac{\mathbb{E}_{\boldsymbol{\mu}} \left [ \tau_\delta \right ]}{\log\left({1}/{2.4\delta}\right)}.
\end{equation}

This instance-dependent complexity indicates the smallest possible constant such that the expected sample complexity $\mathbb{E}_{\boldsymbol{\mu}} \left [ \tau_\delta \right ]$ scales in alignment with $\log\left({1}/{2.4\delta}\right)$. The problem complexity $\kappa \left ( \boldsymbol{\mu} \right )$ is subject to an information-theoretic lower bound. This lower bound can be expressed as the optimal solution of an allocation problem, which we present in Proposition \ref{general lower bound}. To build this framework, we next introduce three important concepts: culprits, alternative sets, and $C_x$ function.

\subsection{Culprits and Alternative Sets}\label{Culprits} 

Let $\mathcal{X}(\boldsymbol{\mu})$ denote the set of culprits under the true mean vector $\boldsymbol{\mu}$. These culprits are responsible for deviations from the correct answer $\mathcal{I}(\boldsymbol{\mu})$. The structure of $\mathcal{X}(\boldsymbol{\mu})$ varies depending on the specific exploration task, and identifying these culprits is essential for characterizing the problem's complexity and guiding the design of effective algorithms.

To identify the correct answer, an algorithm must distinguish among different instances within the parameter space $M$. Accordingly, for any instance $\boldsymbol{\mu} \in M$, the instance-dependent problem complexity $\kappa(\boldsymbol{\mu})$ is determined by the structure of the corresponding alternative set 
\begin{equation}
\cup_{x \in \mathcal{X}(\boldsymbol{\mu})} \textup{Alt}_x(\boldsymbol{\mu}) = \textup{Alt} \left ( \boldsymbol{\mu} \right ) \coloneqq \left \{ \boldsymbol{\vartheta} \in M: \mathcal{I} \left ( \boldsymbol{\vartheta} \right ) \neq \mathcal{I} \left ( \boldsymbol{\mu} \right ) \right \},
\label{def: alternative set}
\end{equation}
which represents the set of parameters that return a solution that is different from the correct solution $\mathcal{I} \left ( \boldsymbol{\mu} \right )$. 

As an example, consider the task of identifying the single best arm. In this case, the culprit set is $\mathcal{X}(\boldsymbol{\mu}) = [K] \backslash \{ I^\ast(\boldsymbol{\mu})\}$, consisting of all arms except the current best arm. For each culprit $x \in \mathcal{X}(\boldsymbol{\mu})$, if there exists a parameter $\boldsymbol{\vartheta}$ under which arm $x$ has a higher mean than $I^\ast(\boldsymbol{\vartheta})$, then $\boldsymbol{\vartheta}$ leads to an incorrect identification caused by $x$. Each such culprit is associated with an alternative set—namely, the set of parameters that yield a wrong answer due to $x$—given by $\textup{Alt}_x(\boldsymbol{\mu}) = \left \{ \boldsymbol{\vartheta} \in M: \vartheta_x \geq \vartheta_{I^\ast(\boldsymbol{\vartheta})} \right \}$ for $x \in \mathcal{X}(\boldsymbol{\mu})$. 

\subsection{\texorpdfstring{$\boldsymbol{C_x}$ } function}\label{Cx function} 
The task of identifying the correct answer can be formulated as a sequential hypothesis testing problem, which can be addressed using the Sequential Generalized Likelihood Ratio (SGLR) test \citepec{kaufmann2016complexity, kaufmann2021mixture}. The SGLR statistic is defined to test a potentially composite null hypothesis $H_0: (\boldsymbol{\mu} \in \Omega_0)$ against a potentially composite alternative hypothesis $H_1: (\boldsymbol{\mu} \in \Omega_1)$, and is given by: 
\begin{equation}
\textup{SGLR}_t = \frac{\sup_{\boldsymbol{\vartheta} \in \Omega_0 \cup \Omega_1}L(X_1, X_2, \ldots, X_t ; \boldsymbol{\vartheta})}{\sup_{\boldsymbol{\vartheta} \in \Omega_0}L(X_1, X_2, \ldots, X_t ; \boldsymbol{\vartheta})},
\label{def: generalized likelihood ratio statistic}
\end{equation}
where $X_1, X_2, \ldots, X_t$ are the observed values from arm pulls, and $L(\cdot)$ denotes the likelihood function based on these observations and an unknown parameter $\boldsymbol{\vartheta}$. The set $\Omega_0$ corresponds to the restricted parameter space under the null hypothesis, while $\Omega_0 \cup \Omega_1$ defines the full parameter space under consideration, encompassing both the null and alternative hypotheses. These sets correspond to the alternative regions introduced in the previous section. A large value of $\textup{SGLR}_t$ indicates stronger evidence against the null hypothesis and supports rejecting it. 

We consider distributions from a single-parameter exponential family parameterized by their means, following the formulation in \citeec{garivier2016optimal}. This family includes the Bernoulli, Poisson, and Gamma distributions with known shape parameters, as well as the Gaussian distribution with known variance. For each culprit $x \in \mathcal{X}(\hat{\boldsymbol{\mu}})$, where $\hat{\boldsymbol{\mu}}$ is the empirical mean based on observed data, we test the hypotheses: $H_{0, x}: \boldsymbol{\mu} \in \textup{Alt}_x(\hat{\boldsymbol{\mu}})$ versus $H_{1, x}: \boldsymbol{\mu} \notin \textup{Alt}_x(\hat{\boldsymbol{\mu}})$. When $\hat{\boldsymbol{\mu}}(t) \in \Omega_0 \cup \Omega_1$, the generalized likelihood ratio statistic in equation~\eqref{def: generalized likelihood ratio statistic} can be expressed in terms of a self-normalized sum. This leads to a formal expression of the SGLR statistic in Proposition~\ref{SGLR statistic in pure exploration}, derived through maximum likelihood estimation and a reformulation of the KL divergence.

\begin{proposition}[\citeec{kaufmann2021mixture}]
\label{SGLR statistic in pure exploration}
The generalized likelihood ratio statistic for each culprit $x \in \mathcal{X}(\boldsymbol{\mu})$ at time step $t$ is defined as 
\begin{align}
\hat{\Lambda}_{t,x} &= \ln (\textup{SGLR}_t) \nonumber\\
&= \inf\limits_{\boldsymbol{\vartheta} \in \textup{Alt}_x(\hat{\boldsymbol{\mu}}(t))} \sum_{i \in \left [ K \right ]} N_i(t)\textup{KL}(\hat{\mu}_i(t), \vartheta_i),\label{def: SGLR statistic in pure exploration}
\end{align}
where \textup{KL}$(\cdot, \cdot)$ represents the \textup{KL} divergence of the two distributions parameterized by their means, and $N_i(t) = t \cdot p_i$ is the expected number of observations allocated to arm $i \in \left [ K \right ]$ up to time $t$.
\end{proposition}

This proposition links the SGLR test to information-theoretic methods. To quantify the information and confidence required to assert that the true mean does not lie in $\textup{Alt}_x$ for all $x \in \mathcal{X}$, we define the $C_x$ function as the population version of the SGLR statistic, sharing the same form as equation (\ref{def: SGLR statistic in pure exploration}).
\begin{equation}
C_x(\boldsymbol{p}) = C_x(\boldsymbol{p}; \boldsymbol{\mu}) \coloneqq \inf\limits_{\boldsymbol{\vartheta} \in \textup{Alt}_x} \sum_{i \in \left [ K \right ]} p_i\textup{KL}(\mu_i, \vartheta_i).
\label{def: Cx function}
\end{equation}

With the introduction of culprits and the $C_x$ function, we arrive at the optimal allocation problem that defines the lower bound stated in Proposition~\ref{general lower bound}. 
However, computing the lower bound can still be hard since it requires the solution of the minimax problem in equation (\ref{def: general lower bound}). While the KL divergence in equation (\ref{def: Cx function}) is convex for Gaussians, it can be non-convex to minimize the $C_x$ function over the culprit set $\mathcal{X} \left ( \boldsymbol{\mu} \right )$. To solve this problem, based on the following Proposition \ref{smoothness, PDE, and active candidate set}, we can write $\mathcal{X} \left ( \boldsymbol{\mu} \right )$ as a union of several convex sets. The following three equivalent expressions represent different ways of describing the lower bound, making the minimax problem in equation \eqref{def: general lower bound} tractable for every $x \in \mathcal{X}$.
\begin{align}
    \Gamma_{\boldsymbol{\mu}}^* = \max_{\boldsymbol{p} \in \mathcal{S}_K} \inf_{\boldsymbol{\vartheta} \in \operatorname{Alt}(\boldsymbol{\mu})} \sum_{i \in[K]} p_i \textup{KL}\left(\mu_i, \vartheta_i\right) &= \max_{\boldsymbol{p} \in \mathcal{S}_K} \min_{x \in \mathcal{X}} \inf _{\boldsymbol{\vartheta} \in \mathrm{Alt}_x} \sum_{i \in [K]} p_i \textup{KL}\left(\mu_i, \vartheta_i\right) \label{optimal allocation problem_convex_1} \\ 
    &= \max_{\boldsymbol{p} \in \mathcal{S}_K} \min _{x \in \mathcal{X}} \sum_{i \in[K]} p_i \textup{KL}\left(\mu_i, \vartheta_i^x\right) \label{optimal allocation problem_convex_2} \\ 
    &= \max_{\boldsymbol{p} \in \mathcal{S}_K} \min_{x \in \mathcal{X}} C_x(\boldsymbol{p}),\label{optimal allocation problem_convex_3}
\end{align}
where we utilized the existence of a finite union set and a unique minimizer $\boldsymbol{\vartheta}^x$ in Proposition \ref{smoothness, PDE, and active candidate set}.
\begin{proposition}\label{smoothness, PDE, and active candidate set}
Assume that the distribution of each arm belongs to a canonical single-parameter exponential family, parameterized by its mean. Then, for each culprit $x \in \mathcal{X}({\boldsymbol{\mu}})$,
\begin{enumerate}
\item \citepec{wang2021fast} For each problem instance $\boldsymbol{\mu} \in M$, the alternative set $\textup{Alt} \left ( \boldsymbol{\mu} \right )$ is a finite union of convex sets. Namely, there exists a finite collection of convex sets $\left \{ \textup{Alt}_x(\boldsymbol{\mu}): x \in \mathcal{X}(\boldsymbol{\mu}) \right \}$ such that $\textup{Alt}(\boldsymbol{\mu}) = \cup_{x \in \mathcal{X}(\boldsymbol{\mu})} \textup{Alt}_x(\boldsymbol{\mu})$.
\item Given a specific simplex distribution $\boldsymbol{p}$, there exists a unique $\boldsymbol{\vartheta}^x \in \textup{Alt}_x(\boldsymbol{\mu})$ that achieves the infimum in equation~(\ref{def: Cx function}).
\end{enumerate}

\end{proposition}

\proof\\{\textit{Proof.}}
The proof proceeds in two parts. For the first part, the alternative set for any given culprit $x$ in our setting is given by
\begin{equation}
    \textup{Alt}_x(\boldsymbol{\mu}) = \textup{Alt}_{i, j}(\boldsymbol{\mu}) \cup \textup{Alt}_{m}(\boldsymbol{\mu}),
\end{equation}
where $\textup{Alt}_{i, j}(\boldsymbol{\mu})$ and $\textup{Alt}_{m}(\boldsymbol{\mu})$ are defined in \eqref{eq: alternative set condition} and \eqref{eq: Alt_m_simple}, respectively.  Each $\textup{Alt}_x(\boldsymbol{\mu})$ is convex, as convexity is preserved under unions of convex sets. This can be verified by confirming that any convex combination of two points in  $\textup{Alt}_x(\boldsymbol{\mu})$ remains in the set.

For the second part, when the reward distribution belongs to a single-parameter exponential family, the \textup{KL} divergence \textup{KL}$(\chi, \chi^\prime)$ is continuous and strictly convex in $(\chi, \chi^\prime)$. This ensures that the infimum in equation~(\ref{def: Cx function}) is achieved uniquely by a single  $\bm{\vartheta}$.
\hfill\Halmos
\endproof

\subsection{Stopping Rule}\label{Stopping Rule} 
In this section, we introduce the stopping rule, which suggests when to stop the algorithm and returns an answer that gives all $\varepsilon$-best arms with a probability of at least $1-\delta$. This stopping rule is based on the deviation inequalities that are linked to the generalized likelihood ratio test \citepec{kaufmann2021mixture}. 

For each $x_i \in \mathcal{X}(\boldsymbol{\mu})$ with $i \in [\left \lvert \mathcal{X} \right\rvert]$, let $M_i(\boldsymbol{\mu}) = \textup{Alt}_{x_i}(\boldsymbol{\mu})$ denote a partition of the realizable parameter space $M$ introduced in Section \ref{Culprits}, where each partition $M_i(\boldsymbol{\mu})$ is associated with a distinct culprit in $\mathcal{X}(\boldsymbol{\mu})$. This implies that for any $\boldsymbol{\mu} \in M$, the parameter space $M$ can be uniquely partitioned to support a hypothesis testing framework. Let $M_0(\boldsymbol{\mu})$ denote the subset of parameters where $\boldsymbol{\mu}$ resides. If $\boldsymbol{\mu} \in M$, define $i^\ast(\boldsymbol{\mu})$ as the index of the unique element in the partition where the true mean value $\boldsymbol{\mu}$ belongs, and here $i^\ast(\boldsymbol{\mu}) = 0$. In other words, we have $\boldsymbol{\mu} \in M_{0}$ and $\textup{Alt}(\boldsymbol{\mu}) = M \backslash M_{0}$. Since the ordering among suboptimal arms is irrelevant, the sets $M_i(\boldsymbol{\mu})$ for $i \in \left \{ 0,1,2, \ldots, \left \lvert \mathcal{X} \right\rvert \right \}$ form a valid partition of $M$ for each $\boldsymbol{\mu}$. Accordingly, the alternative set can be further defined as
\begin{equation}
\textup{Alt}(\boldsymbol{\mu}) = \cup_{i: \boldsymbol{\mu} \notin M_i(\boldsymbol{\mu})}M_i(\boldsymbol{\mu}) = M \backslash M_{i^\ast(\boldsymbol{\mu})} = M \backslash M_{0}.
\label{def: partition of the alternative set}
\end{equation}

Given a bandit instance $\boldsymbol{\mu}$, we consider a total of $\left \lvert \mathcal{X}(\boldsymbol{\mu}) \right\rvert + 1$ hypotheses, defined as
\begin{equation}
H_0 = (\boldsymbol{\mu} \in M_0(\boldsymbol{\mu})), H_1 = (\boldsymbol{\mu} \in M_1(\boldsymbol{\mu})), \ldots, H_{\left \lvert \mathcal{X} \right\rvert} = (\boldsymbol{\mu} \in M_{\left \lvert \mathcal{X} \right\rvert}(\boldsymbol{\mu})).
\label{def: parallel hypotheses}
\end{equation}

By substituting the true mean vector $\boldsymbol{\mu}$ with its empirical estimate $\hat{\boldsymbol{\mu}}(t)$, the SGLR test becomes data-dependent, relying on the empirical means at each time step $t$. Consequently, the hypotheses tested at time $t$ are also data-dependent. If $\hat{\boldsymbol{\mu}}(t) \in M$, we define $\hat{i}(t) = i^\ast(\hat{\boldsymbol{\mu}}(t))$ as the index of the partition to which $\hat{\boldsymbol{\mu}}(t)$ belongs—that is, $\hat{\boldsymbol{\mu}}(t) \in M_{\hat{i}(t)}$. If instead $\hat{\boldsymbol{\mu}}(t) \notin M$, we set $\hat{\Lambda}_{t,x} = 0$ for all culprits $x$, meaning no hypothesis test is conducted at that time step, and the process continues. In practice, when $\hat{\boldsymbol{\mu}}(t) \notin M$, the algorithm can revert to uniform exploration. Since the true mean vector $\boldsymbol{\mu} \in M$ and $M$ is assumed to be an open set, the law of large numbers guarantees that $\hat{\boldsymbol{\mu}}(t)$ will eventually re-enter the parameter space, i.e., $\hat{\boldsymbol{\mu}}(t) \in M$ after sufficient samples. 

We run $\left| \mathcal{X} \right|$ time-varying SGLR tests in parallel, each testing $H_0$ against $H_i$ for $i \in \left[ \left| \mathcal{X}(\hat{\boldsymbol{\mu}}(t)) \right| \right]$. The procedure stops when any of these tests rejects $H_0$, indicating that the corresponding alternative set is empirically the easiest to reject. At this point, the accepted hypothesis for $\hat{\boldsymbol{\mu}}(t) \in M$ is identified as the most likely to be correct. Given a sequence of exploration rates $(\hat{\beta}_t(\delta))_{t \in \mathbb{N}}$, the SGLR stopping rule in the pure exploration setting is defined as follows:
\begin{equation}
\tau_\delta \coloneqq \inf \left \{ t \in \mathbb{N}: \min_{x \in \mathcal{X}(\hat{\boldsymbol{\mu}}(t))} \hat{\Lambda}_{t,x} > \hat{\beta}_t(\delta) \right \} = \inf \left \{ t \in \mathbb{N}: t \cdot \min_{x \in \mathcal{X}(\hat{\boldsymbol{\mu}}(t))} C_x(\boldsymbol{p}_t;\hat{\boldsymbol{\mu}}(t)) > \hat{\beta}_t(\delta) \right \},
\label{def: stopping rule for pure exploration}
\end{equation}
where the SGLR statistic $\hat{\Lambda}_{t,x}$ is defined in equation \eqref{def: SGLR statistic in pure exploration}. The testing process closely resembles the classical approach, except that the hypotheses are data-dependent and evolve over time. 

We also provide insights from the perspective of the confidence region. Specifically, the event $\left \{ \min_{x \in \mathcal{X}(\hat{\boldsymbol{\mu}}(t))} \hat{\Lambda}_{t,x} > \hat{\beta}_t(\delta) \right \} = \left \{ \mathcal{C}_{t, \delta} \subseteq M_{\hat{i}(t)} \right \}$, where $\mathcal{C}_{t, \delta}$ denotes the confidence region of the mean vector, given by
\begin{equation}
\mathcal{C}_{t, \delta} \coloneqq \left \{ \boldsymbol{\vartheta}: \sum_{i = 1}^{K} N_i(t)\textup{KL}(\hat{\mu}_i(t), \vartheta_i) \leq \hat{\beta}_t(\delta) \right \}.
\label{def: confidence set of stopping rule}
\end{equation}

Notably, although the confidence region $\mathcal{C}_{t, \delta}$ is defined for the mean vector $\boldsymbol{\mu}$, under the assumption of linear structure, it is equivalent to the confidence region for the parameter $\boldsymbol{\theta}$, as discussed in Sections \ref{subsec_Optimal Design} and \ref{subsubsec_Visual Explanation of the Lower Bound}. This equivalence allows the stopping rule to be interpreted as follows: the algorithm halts once the confidence region for the mean vector fully lies within a single partition region, aligning with the graphical interpretation of optimal allocation given in Section \ref{subsubsec_Optimal_Allocation}. 

\section{Difference between G-Optimal Design and $\mathcal{XY}$-Optimal Design}\label{Essential Difference between G and XY}

\begin{figure}[htbp]
    \centering
    \includegraphics[scale = 1]{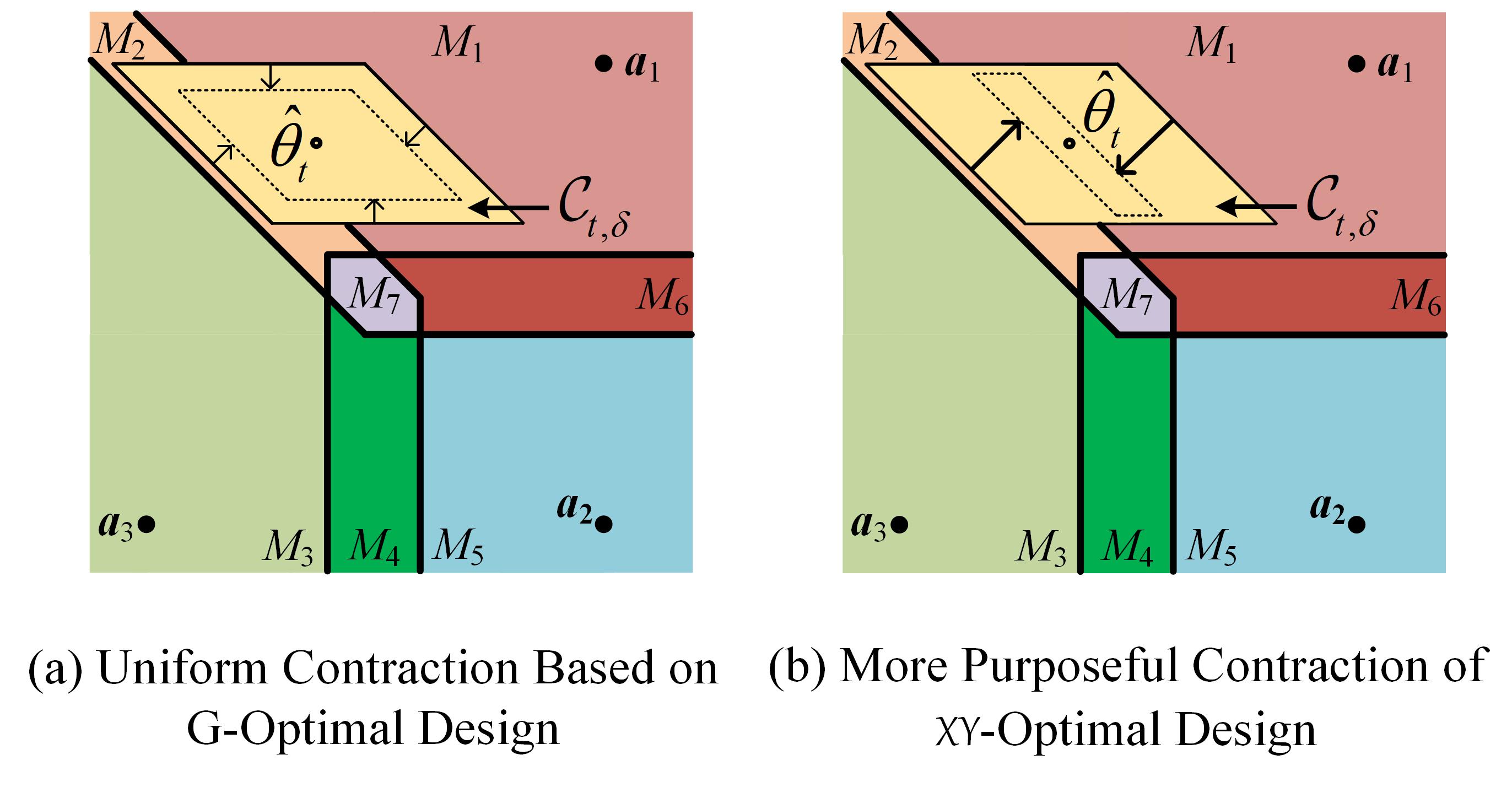}
    \caption{Key Distinction Between G-Optimal and $\mathcal{XY}$-Optimal Designs in Terms of Stopping Criteria}
    \vspace{0.5pt}
    \noindent

  \par
  {\footnotesize
    \parbox{\linewidth}{\raggedright
      \textit{Note.} \textit{The contraction behavior and rate of the confidence region for the parameter $\hat{\boldsymbol{\theta}}_t$ differ: under G-optimal sampling (left), the region shrinks uniformly in all directions, whereas under $\mathcal{XY}$-optimal design (right), it contracts more strategically along directions critical for classification, allowing the confidence region to enter a decision region more rapidly and trigger earlier stopping.}
    }%
  }

    \label{fig: difference_between_G_and_XY.jpg}
\end{figure}

Returning to the intuition and visual explanation in Section~\ref{subsubsec_Visual Explanation of the Lower Bound} and Figure~\ref{fig: 1}, we see that G-optimal sampling inevitably leads to inefficient sampling. Figure~\ref{fig: difference_between_G_and_XY.jpg} illustrates the advantage of adopting the $\mathcal{XY}$-optimal design from the perspective of the stopping rule: the algorithm terminates once the yellow confidence region for $\hat{\boldsymbol{\theta}}$ enters one of the decision regions $M_i$ ($i = 1,\dots,7$). Unlike the isotropic shrinkage of the confidence region under G-optimal design, the $\mathcal{XY}$-optimal design guides the region to contract more aggressively in directions critical for distinguishing arms. Rather than uniformly estimating $\boldsymbol{\theta}$, it prioritizes reducing uncertainty along directions that matter most for classification, leading to more efficient exploration.

\section{Lower Bound for All $\varepsilon$-Best Arms Identification in Linear Bandits}
This section provides both geometric insights regarding the stopping condition and formal proofs establishing the lower bound for identifying all $\varepsilon$-best arms in linear bandit settings.

\subsection{Visual Illustration of the Stopping Condition}\label{subsubsec_Visual Explanation of the Lower Bound}
\begin{figure}[htbp]
    \centering
    \includegraphics[scale=1]{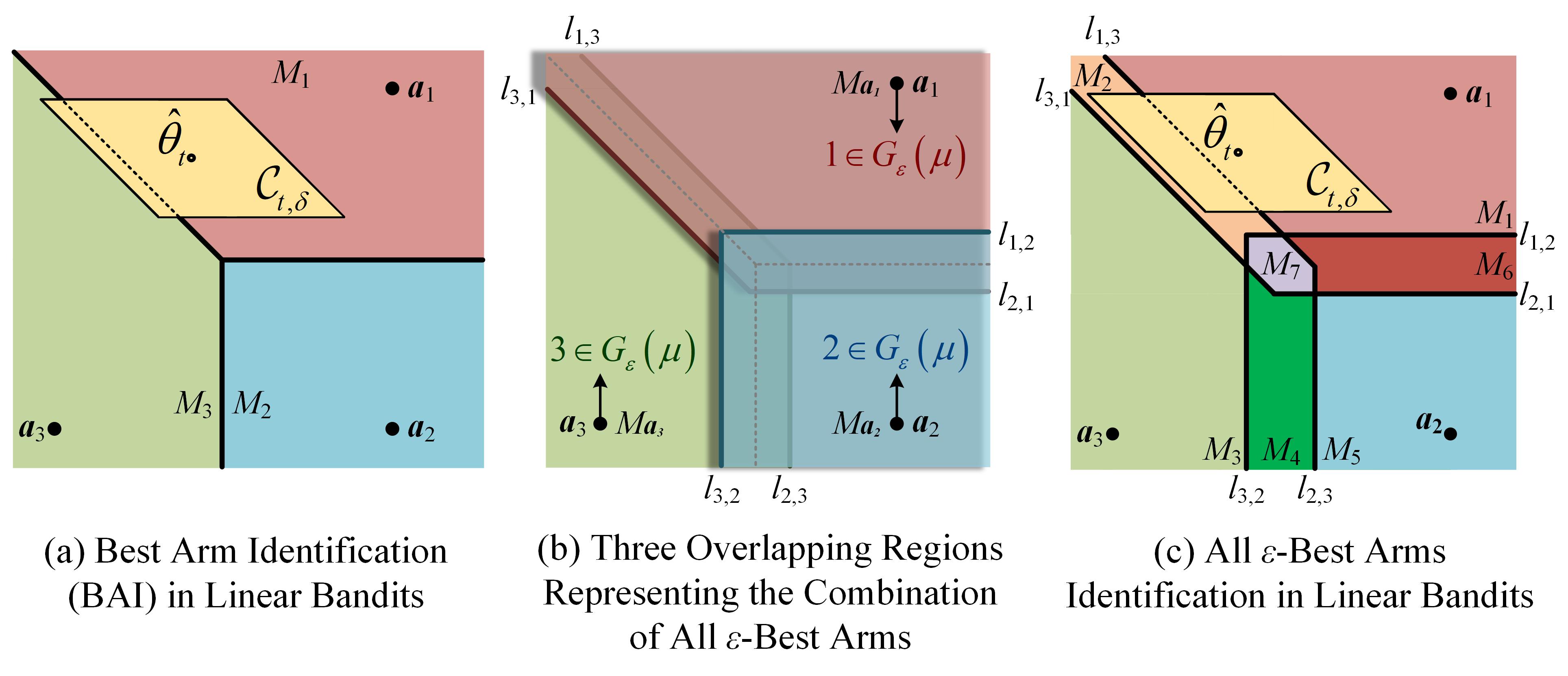}
    \caption{Visual Illustration of Identifying the Best Arm vs. Identifying All $\varepsilon$-Best Arms.
    }
\vspace{5pt}
\noindent

  \par
  {\footnotesize
    \parbox{\linewidth}{\raggedright
      \textit{Note.} \textit{{(a)} Stopping occurs when the confidence region $\mathcal{C}_{t, \delta}$ for the estimated parameter $\hat{\boldsymbol{\theta}}_t$ contracts entirely within one of the three decision regions $M_i$ in a certain time step $t$. The boundaries between regions are defined by the hyperplanes $\boldsymbol{\vartheta}^\top(\boldsymbol{a}_i - \boldsymbol{a}_j) = 0$. Each dot represents an arm. {(b)} In the case of identifying all $\varepsilon$-best arms, the regions overlap. {(c)} Due to these overlaps, the space is partitioned into seven distinct decision regions, increasing the difficulty of identification.}
    }%
  }

    \label{fig: 1}
\end{figure}

The stopping condition is formulated as a hypothesis test conducted as data is collected, which can be interpreted as the process of the parameter confidence region contracting into one of the decision regions (\emph{i.e.}, the set of parameters that yield the same decision). A more detailed version of Figure~\ref{fig: 11} is shown in Figure~\ref{fig: 1}, which illustrates the core idea of the stopping condition for identifying all $\varepsilon$-best arms in linear bandits. The key distinction between identifying all $\varepsilon$-best arms and identifying the single best arm lies in how the decision regions are partitioned. 

Figure~\ref{fig: 1}(a) illustrates the best arm identification process in linear bandits, where $\bm{a}^\ast = \bm{a}^\ast(\boldsymbol{\mu})$ represents the arm with the largest mean value for each bandit instance $\boldsymbol{\mu}$. Let $M_i = \{ \boldsymbol{\vartheta} \in \mathbb{R}^d \mid \bm{a}_i = \bm{a}^\ast \}$ be the set of parameters $\bm{\theta}$ for which $\bm{a}_i$ ($i = 1,2,3$) is the optimal arm. Each $M_i$ forms a cone defined by the intersection of half-spaces.

Figure~\ref{fig: 1}(b) represents an intermediate step, demonstrating the transition from best arm identification to identifying all $\varepsilon$-best arms. Let $M_{\bm{a}_i} = \{ \boldsymbol{\vartheta} \in \mathbb{R}^d \mid i \in G_\varepsilon(\boldsymbol{\mu}) \}$ be the set of parameters $\bm{\theta}$ include arm $i$ in the set $G_\varepsilon(\boldsymbol{\mu})$. $M_{\bm{a}_i}$ is similarly defined by the intersection of half-spaces. The overlap of these three regions forms the decision regions $M_i$ ($i = 1,2,\ldots,7$) in Figure~\ref{fig: 1}(c), which correspond to the seven distinct types of $\varepsilon$-best arms sets. 
Besides, the BAI process in (a) is a special case of the $\varepsilon$-best arms identification in (c), occurring when the gap $\varepsilon$ approaches 0. The following statement provides a detailed explanation of how the decision regions in Figure~\ref{fig: 1}(c) are constructed.

In a $d$-dimensional Euclidean space $\mathbb{R}^d$, hyperplanes $l_{i,j}$ can be defined for any pair of arms, partitioning the space into the following half-spaces:
\begin{align}
    H_{i,j}^+ &= \left\{ \boldsymbol{\vartheta} \in \mathbb{R}^d \mid (\boldsymbol{a}_{i} - \boldsymbol{a}_{j})^\top \boldsymbol{\vartheta} > \varepsilon \right\}, \label{eq:H-} \\ 
    H_{i,j}^- &= \left\{ \boldsymbol{\vartheta} \in \mathbb{R}^d \mid (\boldsymbol{a}_{i} - \boldsymbol{a}_{j})^\top \boldsymbol{\vartheta} \leq \varepsilon \right\}. \label{eq:H+}
\end{align}

The hyperplane $l_{i,j}$, which separates the half-spaces $H_{i,j}^+$ and $H_{i,j}^-$, is perpendicular to the direction vector $\boldsymbol{a}_{i} - \boldsymbol{a}_{j}$. The intersection of these half-spaces, represented by $\bigcap_{i, j \in [K],\, j \neq i} H_{i,j}$, partitions the space into distinct regions. Each region corresponds to a solution set, representing all $\varepsilon$-best arms if the true parameter $\boldsymbol{\theta}$ lies within that region. As the gap $\varepsilon$ approaches 0, the hyperplanes $l_{i,j}$ on both sides of the decision boundaries move closer together, causing some decision regions in Figure~\ref{fig: 1}(c) to shrink until they vanish. The relationship between the true parameter $\boldsymbol{\theta}$ and the half-spaces determines the belonging of arms in the good set $G_\varepsilon(\boldsymbol{\mu})$.

For the case of three arms, the space is divided into three overlapping regions, as shown in Figure~\ref{fig: 1}(b). These regions further generate seven ($2^3 - 1 = 7$) decision regions, denoted $M_1$ through $M_7$, which are summarized in Table~\ref{tab_decisionregion}. 

\begin{table}[htbp]
    \small 
    \centering
    \renewcommand\arraystretch{1.5}
    \caption{Three Overlapping Regions and Seven Decision Regions in the Case of Three Arms}
    \begin{tabular}{@{}c@{\hspace{0.5em}}c@{\hspace{0.5em}}c@{}}
        \hline
        \multicolumn{1}{c}{Decision Region} & 
        \multicolumn{1}{c}{Set Expression} & 
        \multicolumn{1}{c}{$\varepsilon$-Best Arms} \\
        \hline
        $M_1$ & $M_{a_1} \cap M_{a_2}^c \cap M_{a_3}^c$ & $\{1\}$ \\
        $M_2$ & $M_{a_1} \cap M_{a_2}^c \cap M_{a_3}$ & $\{1,3\}$ \\
        $M_3$ & $M_{a_1}^c \cap M_{a_2}^c \cap M_{a_3}$ & $\{3\}$ \\
        $M_4$ & $M_{a_1}^c \cap M_{a_2} \cap M_{a_3}$ & $\{2,3\}$ \\
        $M_5$ & $M_{a_1}^c \cap M_{a_2} \cap M_{a_3}^c$ & $\{2\}$ \\
        $M_6$ & $M_{a_1} \cap M_{a_2} \cap M_{a_3}^c$ & $\{1,2\}$ \\
        $M_7$ & $M_{a_1} \cap M_{a_2} \cap M_{a_3}$ & $\{1,2,3\}$ \\
        \hline
    \end{tabular}
    \label{tab_decisionregion}
\end{table}
\vspace{-5.1pt}

The stopping condition verifies whether the confidence region $\mathcal{C}_{t, \delta}$ is entirely contained within a specific decision region $M_i$. It is important to note that, due to the definition of the confidence region and the property that $\hat{\boldsymbol{\theta}}_t \to \boldsymbol{\theta}$ as $t \to \infty$, any algorithm that continually samples all arms will eventually meet the stopping condition. 

\subsection{Optimal Allocation}\label{subsubsec_Optimal_Allocation}

The goal of the sampling policy is to construct an allocation sequence that drives the confidence set $C_{t, \delta}$ into the optimal region $M^\ast$ as efficiently as possible. Geometrically, this entails selecting arms that cause $C_{t, \delta}$ to contract into the optimal cone $M^\ast$ with minimal sampling effort. The condition $\mathcal{C}_{t, \delta} \subseteq M^\ast$ can be expressed as
\begin{equation}\label{eq_ora_0}
    \text{For all } i \in G_\varepsilon(\boldsymbol{\mu}),\, j \neq i,\, m \notin G_\varepsilon(\boldsymbol{\mu}) \text{ and } \forall\boldsymbol{\vartheta} \in \mathcal{C}_{t, \delta}, \text{ we have } \boldsymbol{\vartheta} \in H_{j,i}^- \text{ and } \boldsymbol{\vartheta} \in H_{1,m}^+.
\end{equation}

In words, every parameter vector $\boldsymbol{\vartheta}$ that remains plausible must preserve all required pairwise orderings: no $\varepsilon$-optimal arm $i$ can be overtaken by any rival $j$, and the best arm 1 must stay ahead of every suboptimal arm $m$. Equivalently, no $\boldsymbol{\vartheta} \in \mathcal{C}_{t, \delta}$ is allowed to flip these comparisons.

The relationships $\boldsymbol{\vartheta} \in H_{j,i}^-$ and $\boldsymbol{\vartheta} \in H_{1,m}^+$ are equivalent to the following inequalities by adding terms on both sides of the inequalities \eqref{eq:H-} and \eqref{eq:H+} and reorganizing:
\begin{equation}\label{eq_ora_1}
    \left\{
    \begin{array}{l}
        (\boldsymbol{a}_i - \boldsymbol{a}_j)^\top (\boldsymbol{\theta} - \boldsymbol{\vartheta}) \leq (\boldsymbol{a}_i - \boldsymbol{a}_j)^\top \boldsymbol{\theta} + \varepsilon \\
        (\boldsymbol{a}_1 - \boldsymbol{a}_m)^\top (\boldsymbol{\theta} - \boldsymbol{\vartheta}) < (\boldsymbol{a}_1 - \boldsymbol{a}_m)^\top \boldsymbol{\theta} - \varepsilon
    \end{array}.
    \right.
\end{equation}

Now, we focus on the confidence region, which can be constructed following Cauchy's inequality and the definition of the confidence ellipse for parameter $\bm{\theta}$ in equation~\eqref{confidence for ellipse}.
\begin{equation}
    \mathcal{C}_{t, \delta} = \left\{ \boldsymbol{\vartheta} \in \mathbb{R}^d \,\middle|\, \forall i \in G_\varepsilon(\boldsymbol{\mu}),\, j \neq i,\, m \notin G_\varepsilon(\boldsymbol{\mu}),\,
    \begin{cases}
        (\boldsymbol{a}_i - \boldsymbol{a}_j)^\top (\boldsymbol{\theta} - \boldsymbol{\vartheta}) \leq \Vert \boldsymbol{a}_i - \boldsymbol{a}_j \Vert_{\bm{V}_t^{-1}} B_{t, \delta} \\
        (\boldsymbol{a}_1 - \boldsymbol{a}_m)^\top (\boldsymbol{\theta} - \boldsymbol{\vartheta}) \leq \Vert \boldsymbol{a}_1 - \boldsymbol{a}_m \Vert_{\bm{V}_t^{-1}} B_{t, \delta}
    \end{cases}
    \right\},
\end{equation}
where $\bm{V}_t$ is the information matrix as defined in equation~\eqref{OLS} and the confidence bound for parameter $\boldsymbol{\theta}$, \emph{i.e.}, $B_{t, \delta}$, can either be a fixed confidence bound as shown in Proposition \ref{proposition_confidence bound for B 1} or a looser adaptive confidence bound introduced in \citeec{abbasi2011improved}. The stopping condition $\mathcal{C}_{t, \delta} \subseteq M^\ast$ can thus be reformulated. For each $i \in G_\varepsilon(\boldsymbol{\mu}),\, j \neq i,\, m \notin G_\varepsilon(\boldsymbol{\mu})$, we have
\begin{equation}\label{eq_oracle_stopping}
    \left\{
    \begin{array}{l}
        \Vert \boldsymbol{a}_i - \boldsymbol{a}_j \Vert_{\bm{V}_t^{-1}} B_{t, \delta} \leq (\boldsymbol{a}_i - \boldsymbol{a}_j)^\top \boldsymbol{\theta} + \varepsilon \\
        \Vert \boldsymbol{a}_1 - \boldsymbol{a}_m \Vert_{\bm{V}_t^{-1}} B_{t, \delta} \leq (\boldsymbol{a}_1 - \boldsymbol{a}_m)^\top \boldsymbol{\theta} - \varepsilon
    \end{array}.
    \right.
\end{equation}

If equation \eqref{eq_oracle_stopping} holds, then for any $\boldsymbol{\theta} \in \mathcal{C}_{t, \delta}$, equation \eqref{eq_ora_1} also holds, implying that $\mathcal{C}_{t, \delta} \subseteq M^\ast$. Given that $(\boldsymbol{a}_i - \boldsymbol{a}_j)^\top \boldsymbol{\theta} + \varepsilon > 0$ and $(\boldsymbol{a}_1 - \boldsymbol{a}_m)^\top \boldsymbol{\theta} - \varepsilon > 0$, and rearranging \eqref{eq_oracle_stopping}, the oracle allocation strategy is determined as follows:
\begin{equation}\label{eq: lower bound_visual_t}
    \{ \bm{a}_{A_n} \}^\ast = \arg \min_{\{ \bm{a}_{A_n} \}} \max_{i \in G_\varepsilon(\boldsymbol{\mu}),\, j \neq i,\, m \notin G_\varepsilon(\boldsymbol{\mu})} \max \left\{ \frac{2 \Vert \boldsymbol{a}_i - \boldsymbol{a}_j \Vert_{\bm{V}_t^{-1}}^2}{\left( \boldsymbol{a}_i^\top \boldsymbol{\theta} - \boldsymbol{a}_j^\top \boldsymbol{\theta} + \varepsilon \right)^2}, \frac{2 \Vert \boldsymbol{a}_1 - \boldsymbol{a}_m \Vert_{\bm{V}_t^{-1}}^2}{\left( \boldsymbol{a}_1^\top \boldsymbol{\theta} - \boldsymbol{a}_m^\top \boldsymbol{\theta} - \varepsilon \right)^2} \right\},
\end{equation}
where $\{ \bm{a}_{A_n} \}= (\bm{a}_{A_1},\bm{a}_{A_1},\ldots,\bm{a}_{A_n}) \in \mathcal{A}^n$ is a sequence of sampled arms and $\{ \bm{a}_{A_n} \}^\ast$ is the oracle allocation strategy\footnote{When the arm elimination is considered, the active set of arms will gradually decrease as arms are removed, becoming a subset of $\mathcal{A}$.}. However, it is more convenient to demonstrate the sample complexity of the problem in terms of the continuous allocation proportion $\bm{p}$ instead of the discrete allocation sequence $\{ \bm{a}_{A_n} \}$. Then, we can have the following optimal allocation proportion. 
\begin{equation}\label{eq: lower bound_visual}
    \boldsymbol{p}^\ast = \arg \min_{\boldsymbol{p} \in \mathcal{S}_K} \max_{i \in G_\varepsilon(\boldsymbol{\mu}),\, j \neq i,\, m \notin G_\varepsilon(\boldsymbol{\mu})} \max \left\{ \frac{2 \Vert \boldsymbol{a}_i - \boldsymbol{a}_j \Vert_{\bm{V}_{\boldsymbol{p}}^{-1}}^2}{\left( \boldsymbol{a}_i^\top \boldsymbol{\theta} - \boldsymbol{a}_j^\top \boldsymbol{\theta} + \varepsilon \right)^2}, \frac{2 \Vert \boldsymbol{a}_1 - \boldsymbol{a}_m \Vert_{\bm{V}_{\boldsymbol{p}}^{-1}}^2}{\left( \boldsymbol{a}_1^\top \boldsymbol{\theta} - \boldsymbol{a}_m^\top \boldsymbol{\theta} - \varepsilon \right)^2} \right\},
\end{equation}
where $\mathcal{S}_K$ denotes the $K$-dimensional probability simplex, and $\bm{V}_{\bm{p}} = \sum_{i=1}^{K}p_i \boldsymbol{a}_i\boldsymbol{a}_i^\top$ is the weighted information matrix, analogous to the Fisher information matrix \citepec{chaloner1995bayesian}. The intuition behind this optimal allocation strategy is as follows: to satisfy the inequality in \eqref{eq_oracle_stopping} as quickly as possible, the ratio of the left-hand side to the right-hand side should be minimized, leading to the outer minimization operation over the allocation probability $\boldsymbol{p}$. The middle maximization operation accounts for the fact that the inequalities in \eqref{eq_oracle_stopping} must hold for each possible triple $(i, j, m)$, which specifies the culprit set $\mathcal{X} = \mathcal{X}(\boldsymbol{\mu})$ (see Proposition \ref{general lower bound} for a formal definition). Furthermore, since both inequalities in \eqref{eq_oracle_stopping} need to be satisfied, we take inner maximization operations to enforce this requirement.

\subsection{Proof of the Lower Bound in Theorem \ref{lower bound: All-epsilon in the linear setting}}\label{proof of theorem: lower bound for all-epsilon in the linear bandit}

\begin{figure}
    \centering
    \includegraphics[scale = 1]{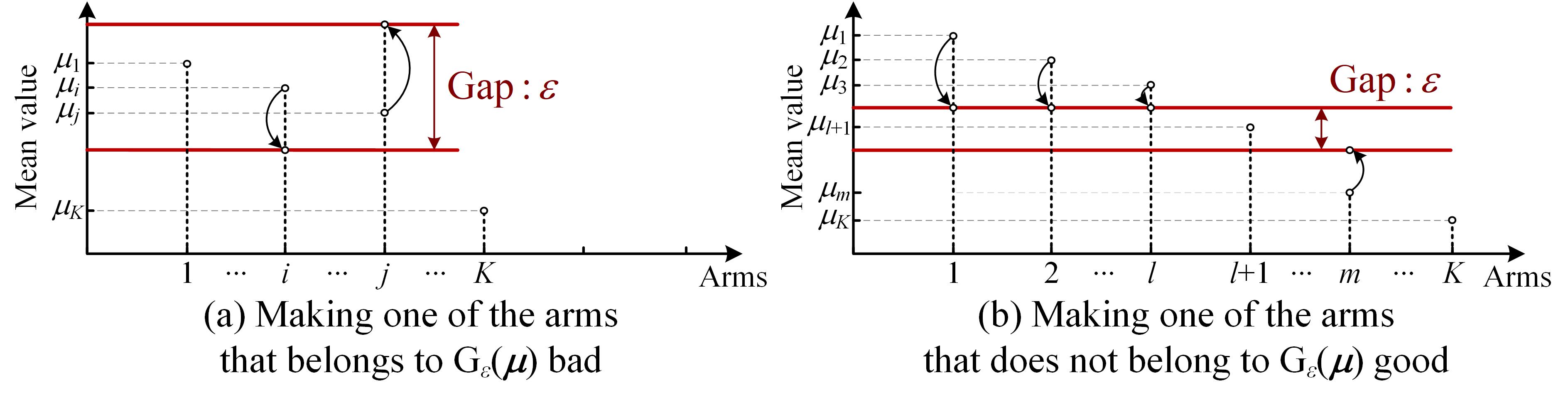}
    \caption{Illustration of Constructing Alternative Set of All $\varepsilon$-Best Arms Identification Lower Bound}
    \vspace{5pt}
    \noindent

  \par
  {\footnotesize
    \parbox{\linewidth}{\raggedright
      \textit{Note.} \textit{(a) Transforming an arm within $G_{\varepsilon}(\boldsymbol{\mu})$ into a non-$\varepsilon$-best arm by decreasing the mean of an $\varepsilon$-best arm $i$ while increasing the mean of another arm $j$. (b) Transforming an arm outside $G_{\varepsilon}(\boldsymbol{\mu})$ into an $\varepsilon$-best arm by increasing the mean of a non-$\varepsilon$-best arm $m$ and simultaneously decreasing the means of higher-performing arms.}
    }%
  }

    \label{fig: 2}
\end{figure}

\proof\\{\textit{Proof.}}
Deriving the lower bound requires constructing the largest possible alternative set and identifying the most challenging instance for distinguishing the arms. Specifically, to construct an alternative problem instance that possesses a distinct set of $\varepsilon$-best arms from the original problem instance $\bm{\mu}$, we systematically perturb the expected rewards of specific arms. This process can be achieved through two ways of construction, as further detailed in Figure~\ref{fig: 2}:

\begin{itemize}
    \item \textit{Lowering an $\varepsilon$-Best Arm:} This approach decreases the expected reward of a currently $\varepsilon$-best arm $i$, while simultaneously increasing the expected reward of another arm $j$. 
    \item \textit{Elevating a Non-$\varepsilon$-Best Arm:} This approach increases the expected reward of a non-$\varepsilon$-best arm $m$, while concurrently reducing the expected rewards of the $\ell$ top-performing arms.
\end{itemize}

For all $\varepsilon$-best arms identification in linear bandits, we establish the correct answer, the culprit set, and the alternative set as follows. The general idea of the proof is to builds these alternatives explicitly and shows which one is the hardest to distinguish. Recall that $\mu_i = \left\langle \boldsymbol{\theta}, \boldsymbol{a}_i \right\rangle$. For notational simplicity, we use $\boldsymbol{\mu}$ to represent the mean vector induced by true parameter $\boldsymbol{\theta}$. 

\begin{itemize}
\item The correct answer $G_\varepsilon(\boldsymbol{\mu}) = \left \{ i:\left\langle \boldsymbol{\theta}, \boldsymbol{a}_i \right\rangle \geq \max_j\left\langle \boldsymbol{\theta}, \boldsymbol{a}_j \right\rangle - \varepsilon \right \}$ for all $i \in [K]$. These are the $\varepsilon$-best arms under the true parameter. 
\item The culprit set $\mathcal{X}(\boldsymbol{\mu}) = \{ (i, j, m, \ell): i \in G_\varepsilon(\boldsymbol{\mu}), j \neq i, m \notin G_\varepsilon(\boldsymbol{\mu}), \ell \in \{1, 2, \ldots, m-1\} \}$. The culprit set identifies potential sources of error. Intuitively, each tuple corresponds to two types of mistakes that can affect the output as mentioned above.
\item The alternative set $\textup{Alt}(\boldsymbol{\mu}) = \cup_{x \in \mathcal{X}(\boldsymbol{\mu})} \textup{Alt}_x(\boldsymbol{\mu})$ and the form of $\textup{Alt}_x(\boldsymbol{\mu})$ is given by
\end{itemize}
\begin{equation}
    \textup{Alt}_x(\boldsymbol{\mu}) = \textup{Alt}_{i, j}(\boldsymbol{\mu}) \cup \textup{Alt}_{m, 
    \; \ell}(\boldsymbol{\mu}) \text{ for all } x \in \mathcal{X}(\boldsymbol{\mu}).
\end{equation}

So for each culprit $x$, we build two kinds of alternative instances, each capable of flipping the answer. The two parts of the alternative sets can be expressed as
\begin{equation}\label{eq: alternative set condition}
    \textup{Alt}_{i, j}(\boldsymbol{\mu}) = \left \{ \boldsymbol{\vartheta}:\left\langle \boldsymbol{\vartheta}, \boldsymbol{a}_i - \boldsymbol{a}_j \right\rangle < -\varepsilon \right \} \text{ for all } \; x \in \mathcal{X}(\boldsymbol{\mu})
\end{equation}
and
\begin{equation}\label{eq: Alt_ml}
    \textup{Alt}_{m, \ell}(\boldsymbol{\mu})  = \left \{ \boldsymbol{\vartheta}: \boldsymbol{\vartheta}^\top \boldsymbol{a}_1 = \dots = \boldsymbol{\vartheta}^\top \boldsymbol{a}_\ell \geq \boldsymbol{\vartheta}^\top \boldsymbol{a}_m + \varepsilon > \boldsymbol{\vartheta}^\top \boldsymbol{a}_{\ell+1} \right \} \text{ for all } \; x \in \mathcal{X}(\boldsymbol{\mu}),
\end{equation}
which are the same as shown in the figure.

The first part $\textup{Alt}_{i, j}(\boldsymbol{\mu})$ forces a good arm $i$ to become at least $\varepsilon$ worse than arm $j$, causing $i$ to fall out of the $\varepsilon$-best set. In contrast, the second part ensures that a suboptimal arm is no more than $\varepsilon$ worse than the top $\ell$ arms with identical mean values, thereby including it in the $\varepsilon$-best set. 

For a given culprit $x=(i,j,m,\ell)$, the first component of the alternative set can be written as
\begin{equation}
    \textup{Alt}_{i, j}(\boldsymbol{\mu}) = \left\{\boldsymbol{\vartheta}_{i,j}(\varepsilon, \bm{p}, \alpha) \mid \boldsymbol{\vartheta}_{i,j}(\varepsilon, \bm{p}, \alpha) = \boldsymbol{\theta} - \frac{\boldsymbol{y}_{i, j}^\top \boldsymbol{\theta} + \varepsilon + \alpha}{\Vert \boldsymbol{y}_{i, j} \Vert_{\bm{V}_{\bm{p}}^{-1}}^2} \bm{V}_{\bm{p}}^{-1} \boldsymbol{y}_{i, j} \right\},
\label{def: proof - alternative set for All-varepsilon in linear bandit}
\end{equation}
where $\bm{V}_{\bm{p}} = \sum_{i=1}^{K}p_i \boldsymbol{a}_i\boldsymbol{a}_i^\top$ is related to the Fisher information matrix \citepec{chaloner1995bayesian},  $\boldsymbol{y}_{i, j} = \boldsymbol{a}_i-\boldsymbol{a}_j$, and $\alpha > 0$ is a perturbation parameter. The form of this alternative set is derived from the solution to the following optimization problem:
\begin{align}
    \mathop{\arg\min}\limits_{\boldsymbol{\vartheta} \in \mathbb{R}^d} \quad & \Vert \boldsymbol{\vartheta} - \boldsymbol{\theta} \Vert_{\bm{V}_{\bm{p}}}^2 \label{nature_vartheta_ob} \\
    \text{s.t.} \quad & \boldsymbol{y}_{i, j}^\top\boldsymbol{\vartheta} = -\varepsilon - \alpha. \label{nature_vartheta_const}
\end{align}

Here, $\alpha$ is introduced to construct the specific alternative set, providing an explicit expression for the alternative parameter $\bm{\vartheta}$. By letting $\alpha \to 0$, we realize the infimum in equation~\eqref{def: Cx function} and equation~\eqref{optimal allocation problem_convex_1}. Then, we have
\begin{equation}
    \boldsymbol{y}_{i, j}^\top \boldsymbol{\vartheta}_{i, j}(\varepsilon, \bm{p}, \alpha) = -\varepsilon - \alpha < - \varepsilon,
\label{def: proof - y theta}
\end{equation}
which satisfies the condition for being a parameter in an alternative set as defined in equation~\eqref{eq: alternative set condition}. Under the Gaussian distribution assumption, \emph{i.e.}, $\mu_i \sim \mathcal{N}(\boldsymbol{a}_i^\top \boldsymbol{\theta}, 1)$, the KL divergence between the mean value associated with the true parameter $\boldsymbol{\theta}$ and that associated with the alternative parameter $\boldsymbol{\vartheta}_{i, j}$ is
\begin{align}
    \textup{KL}(\boldsymbol{a}_i^\top \boldsymbol{\theta}, \boldsymbol{a}_i^\top \boldsymbol{\vartheta}_{i, j}) &= \frac{\left ( \boldsymbol{a}_i^\top \left (\boldsymbol{\theta} - \boldsymbol{\vartheta}_{i, j}(\varepsilon, \bm{p}, \alpha) \right ) \right )^2}{2(1)^2} \nonumber\\
    &= \boldsymbol{y}_{i, j}^\top \bm{V}_{\bm{p}}^{-1} \frac{\left ( \boldsymbol{y}_{i, j}^\top \boldsymbol{\theta} + \varepsilon + \alpha \right )^2 \boldsymbol{a}_i \boldsymbol{a}_i^\top}{ 2 \left ( \Vert \boldsymbol{y}_{i, j} \Vert_{\bm{V}_{\bm{p}}^{-1}}^2 \right )^2} \bm{V}_{\bm{p}}^{-1} \boldsymbol{y}_{i, j}.
\label{def: proof - KL divergence for All-varepsilon and LB}
\end{align}

The last equation follows from substituting the expression for $\boldsymbol{\vartheta}_{i, j}$ given in \eqref{def: proof - alternative set for All-varepsilon in linear bandit}. Then, by Proposition \ref{general lower bound} and the definition of the $C_x$ function in equation~\eqref{def: Cx function}, the lower bound can be expressed as
\begin{align}
    \frac{\mathbb{E}_{\boldsymbol{\mu}} \left [ \tau_\delta \right ]}{\log\left({1}/{2.4 \delta}\right)} &\geq \min_{\bm{p} \in S_K}\max_{\boldsymbol{\vartheta} \in \textup{Alt}(\boldsymbol{\mu})} \frac{1}{\sum_{n = 1}^{K} p_n \textup{KL}(\boldsymbol{a}_n^\top \boldsymbol{\theta},   \boldsymbol{a}_n^\top \boldsymbol{\vartheta})}\nonumber\\
    &\geq \min_{\bm{p} \in S_K}\max_{x \in \mathcal{X}}\sup_{\alpha > 0} \frac{1}{\sum_{n = 1}^{K} p_n \textup{KL}(\boldsymbol{a}_n^\top \boldsymbol{\theta},   \boldsymbol{a}_n^\top \boldsymbol{\vartheta}_{i, j}(\varepsilon, \bm{p}, \alpha))}\nonumber\\
    &= \min_{\bm{p} \in S_K}\max_{x \in \mathcal{X}} \frac{2 \Vert \boldsymbol{y}_{i, j} \Vert_{\bm{V}_{\bm{p}}^{-1}}^2}{\left ( \boldsymbol{y}_{i, j}^\top \boldsymbol{\theta} + \varepsilon \right )^2}.
\end{align}

Note that we let $\alpha \to 0$ establish the result by realizing $\sup_{\alpha > 0}$. From this lower bound, we can also define the $C_x$ function for all $\varepsilon$-best arms identification in linear bandits, which is
\begin{equation}\label{Cx function: 1}
    C_{i, j}(\bm{p}) = \frac{\left ( \boldsymbol{y}_{i, j}^\top \boldsymbol{\theta} + \varepsilon \right )^2}{2 \Vert \boldsymbol{y}_{i, j} \Vert_{\bm{V}_{\bm{p}}^{-1}}^2}.
\end{equation}

For the second part, \emph{i.e.}, $\textup{Alt}_{m, \ell}(\boldsymbol{\mu})$, we assume that the mean value of arm 1 (\emph{i.e.}, the best arm) remains fixed. This assumption partially sacrifices the completeness of the alternative set but allows the alternative set to be expressed in an explicit and symmetric form. This form remains tight. 

Consequently, the culprit set is redefined as $\mathcal{X}(\boldsymbol{\mu}) = \{ (i, j, m): i \in G_\varepsilon(\boldsymbol{\mu}), j \neq i, m \notin G_\varepsilon(\boldsymbol{\mu})  \}$ and the alternative set can be decomposed as $\textup{Alt}_x(\boldsymbol{\mu}) = \textup{Alt}_{i, j}(\boldsymbol{\mu}) \cup \textup{Alt}_{m}(\boldsymbol{\mu})$ for a given culprit $x = (i,j,m)$. Different from equation~\eqref{eq: Alt_ml}, the second part of the alternative set becomes
\begin{equation}\label{eq: Alt_m_simple}
    \textup{Alt}_{m}(\boldsymbol{\mu}) = \left \{ \boldsymbol{\vartheta}:\left\langle \boldsymbol{\vartheta}, \boldsymbol{a}_1 - \boldsymbol{a}_m \right\rangle < -\varepsilon \right \} \text{ for all } \; x \in \mathcal{X}(\boldsymbol{\mu}).
\end{equation}

Similarly, $\textup{Alt}_{m}(\boldsymbol{\mu})$ can be constructed as the set in equation~\eqref{def: proof - alternative set for All-varepsilon in linear bandit}, given by
\begin{equation}
    \textup{Alt}_{m}(\boldsymbol{\mu}) = \left\{\boldsymbol{\vartheta}_{m}(\varepsilon, \bm{p}, \alpha) \mid \boldsymbol{\vartheta}_{m}(\varepsilon, \bm{p}, \alpha) = \boldsymbol{\theta} - \frac{\boldsymbol{y}_{m}^\top \boldsymbol{\theta} + \varepsilon + \alpha}{\Vert \boldsymbol{y}_{m} \Vert_{\bm{V}_{\bm{p}}^{-1}}^2} \bm{V}_{\bm{p}}^{-1} \boldsymbol{y}_{m} \right\},
\label{def: proof - alternative set for All-varepsilon in linear bandit 2}
\end{equation}

\noindent where $\boldsymbol{y}_{m} = \boldsymbol{a}_1 - \boldsymbol{a}_m$ and $\alpha > 0$. Then we have
\begin{equation}
    \boldsymbol{y}_{m}^\top \boldsymbol{\vartheta}_{m}(\varepsilon, \bm{p}, \alpha) = \varepsilon - \alpha < \varepsilon.
\label{def: proof - y theta 2}
\end{equation}

Hence, by following a derivation analogous to that of the first part, we obtain
\begin{equation}\label{Cx function: 2}
    C_{m}(\bm{p}) = \frac{\left ( \boldsymbol{y}_{m}^\top \boldsymbol{\theta} - \varepsilon \right )^2}{2 \Vert \boldsymbol{y}_{m} \Vert_{\bm{V}_{\bm{p}}^{-1}}^2}.
\end{equation}

Deriving the tightest lower bound focuses on constructing alternative bandit instances that thoroughly challenge each $\varepsilon$-best arm configuration. 

The minimum of the information function $C_x$ over all culprits $x \in \mathcal{X}(\boldsymbol{\mu})$ is then given by:
\begin{equation}
    \min \left\{\min_{x \in \mathcal{X}} C_{i, j}(\bm{p}), \min_{x \in \mathcal{X}} C_{m}(\bm{p}) \right\}. 
\end{equation}

Then, by Proposition \ref{general lower bound} and the definition of $C_x$ function, combining equation (\ref{Cx function: 1}) and equation (\ref{Cx function: 2}), the final lower bound can be expressed as
\begin{align}
    \frac{\mathbb{E}_{\boldsymbol{\mu}} \left [ \tau_\delta \right ]}{\log\left({1}/{2.4 \delta}\right)} &\geq \min_{\bm{p} \in S_K}\max_{(i, j, m) \in \mathcal{X}} \max \left\{ \frac{2 \Vert \boldsymbol{y}_{i, j} \Vert_{\bm{V}_{\bm{p}}^{-1}}^2}{\left ( \boldsymbol{y}_{i, j}^\top \boldsymbol{\theta} + \varepsilon \right )^2}, \frac{2 \Vert \boldsymbol{y}_{m} \Vert_{\bm{V}_{\bm{p}}^{-1}}^2}{\left ( \boldsymbol{y}_{m}^\top \boldsymbol{\theta} - \varepsilon \right )^2} \right\} \\
    &= \min_{\bm{p} \in S_K}\max_{(i, j, m) \in \mathcal{X}} \max \left\{ \frac{2 \Vert \boldsymbol{a}_i-\boldsymbol{a}_j \Vert_{\bm{V}_{\bm{p}}^{-1}}^2}{\left ( \boldsymbol{a}_i^\top \boldsymbol{\theta} - \boldsymbol{a}_j^\top \boldsymbol{\theta} + \varepsilon \right )^2}, \frac{2 \Vert \boldsymbol{a}_1 - \boldsymbol{a}_m \Vert_{\bm{V}_{\bm{p}}^{-1}}^2}{\left ( \boldsymbol{a}_1^\top \boldsymbol{\theta} - \boldsymbol{a}_m^\top \boldsymbol{\theta} - \varepsilon \right )^2} \right\}.
\end{align}
\hfill\Halmos
\endproof

\section{Proof of Theorem \ref{upper bound: Algorithm 1 G}}\label{proof of theorem: upper bound of LinFACT G}

The proof of Theorem~\ref{upper bound: Algorithm 1 G} consists of two parts. First, we establish the upper bound given in equation~\eqref{upper bound sum}. Then, we derive the final refined bound in equation~\eqref{upper bound for algorithm1: 2}, removing the summation in our first result.

To establish the result involving the summation in equation~\eqref{upper bound sum}, we define $R_{\max} = \min \left\{r: G_r=G_\varepsilon\right\}$ as the round in which the last $\varepsilon$-best arm is added to $G_r$. We divide the total number of samples into two phases: samples collected up to round $R_{\max}$ and those collected from round $R_{\max} + 1$ until termination (if the algorithm does not stop at $R_{\max}$). The proof proceeds in eight steps, as outlined below.

\subsection{Preliminary: Clean Events $\mathcal{E}_1$ and $\mathcal{E}_2$}\label{algorithm2: preliminary}

We begin by defining two high-probability events, denoted as $\mathcal{E}_1$ and $\mathcal{E}_2$, which we refer to as clean events. 
\begin{equation}
    \mathcal{E}_1 = \left\{ \bigcap_{i \in \mathcal{A}_I(r-1)} \bigcap_{r \in \mathbb{N}} \left| \hat{\mu}_i(r) - \mu_i \right| \leq C_{\delta/K}(r) \right\}. \label{clean event G}
\end{equation}
This event captures the condition that, for each round $r$, the estimated means $\hat{\mu}_i(r)$ of all arms $i$ in the active set $\mathcal{A}_I(r-1)$ lie within a confidence bound $C_{\delta / K}(r)$ of their true means $\mu_i$. It ensures uniform control over estimation errors across all arms and rounds with a prescribed confidence level. 

Then, we introduce the following lemma to provide the confidence region for the estimated parameter $\hat{\boldsymbol{\theta}}$.
\begin{lemma}[\citeec{lattimore2020bandit}]\label{lemma: confidence region for any single fixed arm}
Let confidence level $\delta \in (0,1)$, for each arm $\bm{a} \in \mathcal{A}$, we have
\begin{equation}
    \mathbb{P} \left \{ \left \lvert \left\langle \hat{\boldsymbol{\theta}} - \boldsymbol{\theta}, \bm{a} \right\rangle \right\rvert \geq \sqrt{2 \left \lVert \bm{a} \right\rVert^2_{\bm{V}_t^{-1}} \log \left( \frac{2}{\delta} \right)} \right \} \leq \delta,
\end{equation}
where $\bm{V}_t$ represents the information matrix defined in Section \ref{subsec_Optimal Design}.
\end{lemma}

Let $\pi_r$ denote the optimal allocation proportion computed for round $r$. The corresponding sampling budget is then determined according to equation~\eqref{equation_phase budget G}. Thus, we obtain the information matrix in each round, denoted as $\bm{V}_r$\footnote{$\bm{V}_r$ and $\bm{V}_t$ can be used interchangeably when the context is unambiguous.}.
\begin{equation}\label{matrix_ineq}
    \bm{V}_r = \sum_{\bm{a} \in \text{Supp}(\pi_r)} T_r(\bm{a}) \bm{a} \bm{a}^\top \succeq\footnote{The symbol $\succeq$ represents the relative magnitude relationship between matrices. Specifically, for any matrix $\bm{A}$, $\bm{A} \succeq 0$ means that the matrix $\bm{A}$ is positive semi-definite. $\bm{A} \succeq \bm{B}$ equals $\bm{A}-\bm{B} \succeq 0$.} \frac{2d}{\varepsilon_r^2} \log \left( \frac{2Kr(r+1)}{\delta} \right) \bm{V}(\pi).
\end{equation}
Then, by applying Lemma \ref{lemma: confidence region for any single fixed arm} with $\delta$ replaced by $\delta / (K r (r+1))$, we obtain that for any arm $\bm{a} \in \mathcal{A}(r-1)$, with probability at least $1 - \delta / (K r (r+1))$, we have
\begin{align}\label{equation: confidence region}
    \left| \left\langle \hat{\boldsymbol{\theta}}_r - \boldsymbol{\theta}, \bm{a} \right\rangle \right| &\leq \sqrt{2\left \lVert \bm{a} \right\rVert^2_{\bm{V}_r^{-1}} \log \left( \frac{2Kr(r+1)}{\delta} \right)} \nonumber\\
    &= \sqrt{2\bm{a}^\top\bm{V}_r^{-1}\bm{a} \log \left( \frac{2Kr(r+1)}{\delta} \right)} \nonumber\\
    &\leq \sqrt{2\bm{a}^\top \left( \frac{\varepsilon_r^2}{2d}\frac{1}{\log \left( \frac{2Kr(r+1)}{\delta} \right)}\bm{V}(\pi)^{-1} \right) \bm{a} \log \left( \frac{2Kr(r+1)}{\delta} \right)} \nonumber\\
    &\leq \varepsilon_r,
\end{align}
where the third line follows from the matrix inequality in \eqref{matrix_ineq}, using the auxiliary result in Lemma~\ref{lemma: inversion reverses loewner orders}, and the final line is derived from Lemma~\ref{theorem_KW}.

Thus, by applying the standard result of the G-optimal design in equation~\eqref{equation: confidence region}, we define the confidence radius associated with the clean event $\mathcal{E}_1$ as 
\begin{equation}\label{eq_C_G}
    C_{\delta/K}(r) \coloneqq \varepsilon_r.
\end{equation}

Then, we have
\begin{align}
    \mathbb{P}(\mathcal{E}_1^c) &= \mathbb{P} \left\{ \bigcup_{i \in \mathcal{A}_I(r-1)} \bigcup_{r \in \mathbb{N}} \left| \hat{\mu}_i(r) - \mu_i \right| > C_{\delta/K}(r) \right\}
    \nonumber\\ &\leq \sum_{r=1}^{\infty} \mathbb{P} \left\{ \bigcup_{i \in \mathcal{A}_I(r-1)} \left| \hat{\mu}_i(r) - \mu_i \right| > \varepsilon_r \right\} \nonumber\\ &\leq \sum_{r=1}^{\infty} \sum_{i=1}^{K} \frac{\delta}{Kr(r+1)}\nonumber\\
    &= \delta,
\end{align}
where the second line follows from the union bound, and the third line combines the union bound with equation~\eqref{equation: confidence region}. Therefore, we obtain
\begin{equation}
    P(\mathcal{E}_1) \geq 1 - \delta.
\end{equation}

Now consider another event, $\mathcal{E}_2$, which characterizes the gaps between different arms, defined as
\begin{equation}
    \mathcal{E}_2 = \left \{ \bigcap_{i \in G_{\varepsilon}} \bigcap_{j \in \mathcal{A}_I(r-1)} \bigcap_{r \in \mathbb{N}} \left| (\hat{\mu}_j(r) - \hat{\mu}_i(r)) - (\mu_j - \mu_i) \right| \leq 2\varepsilon_r \right \}.
\end{equation}
This event ensures the gap between the estimated mean rewards of arms $j$ and $i$ is uniformly close to their true gap for all rounds $r$, arms $i \in G_{\varepsilon}$, and arms $j$ in the active set $\mathcal{A}_I(r-1)$.

By (\ref{equation: confidence region}), for $i, j \in \mathcal{A}_I(r-1)$, we have
\begin{align}\label{confidence: arm filter}
    \mathbb{P}\left\{\left|(\hat{\mu}_j - \hat{\mu}_i) - (\mu_j - \mu_i)\right| > 2\varepsilon_r \mid \mathcal{E}_1 \right\} &\leq \mathbb{P}\left\{\left|\hat{\mu}_j - \mu_j \right| + \left|\hat{\mu}_i - \mu_i\right| > 2\varepsilon_r \mid \mathcal{E}_1 \right\} \nonumber\\
    &\leq \mathbb{P}\left\{\left|\hat{\mu}_j - \mu_j \right| > \varepsilon_r \mid \mathcal{E}_1 \right\} + \mathbb{P}\left\{\left|\hat{\mu}_i - \mu_i\right| > \varepsilon_r \mid \mathcal{E}_1 \right\} \nonumber\\
    &= 0,
\end{align}

\noindent which means
\begin{equation}
    \mathbb{P}(\mathcal{E}_2 \mid \mathcal{E}_1) = 1.
\end{equation}

\subsection{Step 1: Correctness}\label{algorithm2: Step 1}
Recall that $G_\varepsilon$ denotes the true set of $\varepsilon$-best arms, and $G_r$ is the empirical good set identified by LinFACT-G in round $r$. Under event $\mathcal{E}_1$, we first show that if there exists a round $r$ such that $G_r \cup B_r = [K]$, then it must be that $G_r = G_\varepsilon$. This implies that under the clean event $\mathcal{E}_1$, the stopping condition of LinFACT-G ensures correct identification of $G_\varepsilon$.

\begin{lemma}\label{claim1_1}
    Under event $\mathcal{E}_1$, we have $G_r \subseteq G_\varepsilon$ for all rounds $r \in \mathbb{N}$.
\end{lemma}

\proof\\{\textit{Proof.}}
We first show that $1 \in \mathcal{A}_I(r)$ for all $r \in \mathbb{N}$; that is, the best arm is never eliminated from the active set $\mathcal{A}(r-1)$ in any round $r$ on the event $\mathcal{E}_1$. For any arm $i$, we have
\begin{equation}
    \hat{\mu}_1(r) + \varepsilon_r \geq \mu_1 \geq \mu_i \geq \hat{\mu}_i(r) - \varepsilon_r > \hat{\mu}_i(r) - \varepsilon_r - \varepsilon,
\end{equation}
which implies that $\hat{\mu}_1(r) + \varepsilon_r > \max_{i \in \mathcal{A}_I(r-1)} \hat{\mu}_i - \varepsilon_r - \varepsilon = L_r$ and $\hat{\mu}_1(r) + \varepsilon_r \geq \max_{i \in \mathcal{A}_I(r-1)} \hat{\mu}_i (r) - \varepsilon_r$. These inequalities confirm that arm 1 will not be removed from the active set $\mathcal{A}_I(r-1)$.

Secondly, we show that at all rounds $r$, $\mu_1 - \varepsilon \in [L_r, U_r]$. Since arm 1 never exists $\mathcal{A}_I(r-1)$,
\begin{equation}
    U_r = \max_{i \in \mathcal{A}_I(r-1)} \hat{\mu}_i + \varepsilon_r -\varepsilon \geq \hat{\mu}_1(r) + \varepsilon_r - \varepsilon \geq \mu_1 - \varepsilon,
\end{equation}

\noindent and for any arm $i$,
\begin{equation}
    \mu_1 - \varepsilon \geq \mu_i - \varepsilon \geq \hat{\mu}_i - \varepsilon_r - \varepsilon.
\end{equation}
Hence, taking the maximum over $i \in \mathcal{A}_I(r-1)$, we obtain
\begin{equation}
    \mu_1 - \varepsilon \geq \max_{i \in \mathcal{A}_I(r-1)}\hat{\mu}_i - \varepsilon_r - \varepsilon = L_r.
\end{equation}

Next, we show that $G_r \subseteq G_{\varepsilon}$ for all $r \geq 1$. By contradiction, if $G_r \not\subseteq G_{\varepsilon}$, then it means that $\exists r \in \mathbb{N}$ and $\exists i \in G_\varepsilon^c \cap G_r$ such that,
\begin{equation}
    \mu_i \geq \hat{\mu}_i - \varepsilon_r \geq U_r \geq \mu_1 - \varepsilon > \mu_i,
\end{equation}
which forms a contradiction.
\hfill\Halmos
\endproof

\begin{lemma}\label{claim1_2}
    Under event $\mathcal{E}_1$, we have $B_r \subseteq G_\varepsilon^c$ for all rounds $r \in \mathbb{N}$.
\end{lemma}

\proof\\{\textit{Proof.}}
Similarly, we proceed by contradiction. Consider the case that a good arm from $G_{\varepsilon}$ is added to $B_r$ for some round $r$. By definition, $B_0 = \emptyset$ and $B_{r-1} \subseteq B_r$ for all $r$. Then there must exist some $r \in \mathbb{N}$ and an $i \in G_{\varepsilon}$ such that $i \in B_r$ and $i \notin B_{r-1}$. Following line \ref{elimination: good 1} of Algorithm \ref{alg_LinFACT_stopping}, this occurs if and only if
\begin{equation}
    \max_{j \in \mathcal{A}_I(r-1)} \hat{\mu}_j - \hat{\mu}_i > 2\varepsilon_r + \varepsilon.
\end{equation}
On the clean event $\mathcal{E}_1$, the above implies $\exists j \in \mathcal{A}_I(r-1)$ such that
\begin{equation}
    \mu_j - \mu_i + 2\varepsilon_r \geq \hat{\mu}_j - \hat{\mu}_i \geq 2\varepsilon_r + \varepsilon,
\end{equation}

\noindent which yields $\mu_j - \mu_i \geq \varepsilon$, contradicting that $i \in G_{\varepsilon}$.
\hfill\Halmos
\endproof

The above Lemma \ref{claim1_1} and Lemma \ref{claim1_2} show that under $\mathcal{E}_1$, $G_r \cup B_r = [K]$ can lead to the result $G_r = G_\varepsilon$ and $B_r = G_\varepsilon^c$. Since $\mathbb{P}\left\{ \mathcal{E}_1 \right\} \geq 1 - \delta$, if LinFACT terminates, it can correctly provide the correct decision rule with a probability of least $1 - \delta$. Up to now, we have demonstrated the correctness of the algorithm's stopping rule. Then we will focus on bounding the sample complexity in the following parts.

\subsection{Step 2: Total Sample Count}\label{algorithm2: Step 2}
To bound the expected sampling budget, we decompose the total number of samples into two parts: the budget used before the round when the last arm is added to the good set $G_r$, and the budget used after this round until termination. The total number of samples drawn by the algorithm can thus be represented as
\begin{align}
T &\leq \sum_{r=1}^{\infty} \mathbbm{1} \left[G_{r-1} \cup B_{r-1} \neq [K] \right] \sum_{\bm{a} \in \mathcal{A}(r-1)}T_r(\bm{a}) \nonumber\\
&= \sum_{r=1}^{\infty} \mathbbm{1} \left[ G_{r-1} \neq G_\varepsilon \right] \mathbbm{1} \left[G_{r-1} \cup B_{r-1} \neq [K] \right] \sum_{\bm{a} \in \mathcal{A}(r-1)}T_r(\bm{a})\label{decomposition of T: part 1} \\
&+ \sum_{r=1}^{\infty} \mathbbm{1} \left[ G_{r-1} = G_\varepsilon \right] \mathbbm{1} \left[G_{r-1} \cup B_{r-1} \neq [K] \right] \sum_{\bm{a} \in \mathcal{A}(r-1)}T_r(\bm{a}).\label{decomposition of T: part 2}
\end{align}

In the following parts, we will bound these two terms separately. We begin by analyzing a single arm $i \in G_\varepsilon$, tracking which sets it belongs to as the rounds progress. 

For \( i \in G_{\varepsilon} \), let \(  R_i \) denote the number of rounds in which arm $i$ is sampled before being added to \( G_r \) in line \ref{add to Gr} of Algorithm \ref{alg_LinFACT_stopping}. For \( i \in G_{\varepsilon}^c \), let \(  R_i \) denote the number of rounds in which arm \( i \) is sampled before being removed from \( \mathcal{A}_I(r-1) \) and added to $B_r$ in line \ref{elimination: good 1} of Algorithm \ref{alg_LinFACT_stopping}. Then, by definition, $R_i$ can be expressed as
\begin{equation}
 R_i = \min \left\{ r:
\begin{cases}
i \in G_r & \text{if } i \in G_{\varepsilon} \\
i \notin \mathcal{A}_I(r) & \text{if } i \in G_{\varepsilon}^c
\end{cases}
\right\} = \min \left\{ r :
\begin{cases}
\hat{\mu}_i - \varepsilon_r \geq U_{r} & \text{if } i \in G_{\varepsilon} \\
\hat{\mu}_i + \varepsilon_r \leq L_{r} & \text{if } i \in G_{\varepsilon}^c
\end{cases}
\right\}.
\end{equation}

\subsubsection*{Bound $ R_i$.}\label{algorithm2: Step 3}

We define a helper function $h(x) = \log_2 \left({1}/{\left \lvert x \right\rvert}\right)$ to facilitate the proof. It can be observed that in round $r$, if $r \geq h(x)$, then $\varepsilon_r = 2^{-r} \leq \left \lvert x \right \rvert$.

\begin{lemma}\label{claim3_1}
    For any $i \in G_\varepsilon$, we have $ R_i \leq \lceil h\left(0.25\left(\varepsilon-\Delta_i\right) \right)\rceil$, where $\Delta_i=\mu_1-\mu_i$. 
\end{lemma}

\proof\\{\textit{Proof.}}
Note that for $i \in G_\varepsilon$, the inequality $4  \varepsilon_r < \mu_i-\left(\mu_1-\varepsilon\right)$ holds when $r > h\left(0.25\left(\varepsilon-\Delta_i\right) \right)$. This implies that for all $j \in \mathcal{A}_I(r-1)$,
\begin{align}
\hat{\mu}_i- \varepsilon_r &\geq \mu_i-2  \varepsilon_r \nonumber\\
& > \mu_1+2  \varepsilon_r-\varepsilon \nonumber\\
& \geq \mu_j+2  \varepsilon_r-\varepsilon \nonumber\\
& \geq \hat{\mu}_j+ \varepsilon_r-\varepsilon.
\end{align}

Thus, in particular, $\hat{\mu}_i- \varepsilon_r > \max _{j \in \mathcal{A}_I(r-1)} \hat{\mu}_j(t)+ \varepsilon_r-\varepsilon=U_r$. Therefore, we conclude that $ R_i \leq \lceil h\left(0.25\left(\varepsilon-\Delta_i\right) \right)\rceil$.
\hfill\Halmos
\endproof

After determining the latest round in which any arm $i \in G_\varepsilon$ is added to $G_r$, we define $R_{\max }$ as the round by which all good arms have been added to $G_r$, i.e., $R_{\max } \coloneqq \min \left\{r: G_r=G_\varepsilon\right\}=\max _{i \in G_\varepsilon}  R_i$. 

\begin{lemma}\label{claim3_2}
    $R_{\max } \leq$ $\lceil h\left(0.25 \alpha_\varepsilon\right)\rceil$.
\end{lemma}

\proof\\{\textit{Proof.}}
Recall that $\alpha_\varepsilon=\min _{i \in G_\varepsilon} \mu_i-\mu_1+\varepsilon=\min _{i \in G_\varepsilon} \varepsilon-\Delta_i$. By Lemma \ref{claim3_1}, $ R_i \leq \lceil h\left(0.25\left(\varepsilon-\Delta_i\right) \right) \rceil$ for $i \in G_\varepsilon$. Furthermore, $h(\cdot)$ is monotonically decreasing if $i \in G_\varepsilon$. Then for any $\delta>0$, $R_{\max }=\max _{i \in G_\varepsilon}  R_i \leq \max _{i \in G_\varepsilon} \lceil h\left(0.25\left(\varepsilon-\Delta_i\right) \right) \rceil = \lceil h(\min _{i \in G_\varepsilon} \varepsilon-\Delta_i) \rceil = \lceil h\left(0.25 \alpha_\varepsilon \right) \rceil$.
\hfill\Halmos
\endproof

\subsubsection*{Bound the Total Samples up to $R_{max}$.}\label{algorithm2: Step 4}
The total number of samples up to round $R_{\max }$ is $\sum_{r=1}^{R_{\max}} \sum_{\bm{a} \in \mathcal{A}(r-1)}T_r(\bm{a})$.
By line \ref{G_sampling_2} of the Algorithm \ref{alg_LinFACT_sampling}, we have
\begin{equation}\label{T_r budget}
    T_{r} \leq \frac{2 d}{\varepsilon_{r}^2} \log \left(\frac{2K r(r+1)}{\delta}\right)+\frac{d(d+1)}{2}.
\end{equation}

Hence,
\begin{align}
    \sum_{r=1}^{R_{\max }} T_r &= \sum_{r=1}^{R_{\max }} \sum_{\bm{a} \in \mathcal{A}(r-1)}T_r(\bm{a}) \nonumber\\
    & \leq \sum_{r=1}^{R_{\max }} \left( \frac{2 d}{\varepsilon_{r}^2} \log \left(\frac{2K r(r+1)}{\delta}\right)+\frac{d(d+1)}{2} \right) \nonumber\\
    & \leq \sum_{r=1}^{R_{\max }} 2^{2r+1} d \log \left(\frac{2K R_{\max }(R_{\max }+1)}{\delta}\right)+\frac{d(d+1)}{2}R_{\max } \nonumber\\
    & \leq c 2^{2R_{\max }+1} d \log \left(\frac{2K R_{\max }(R_{\max }+1)}{\delta}\right)+\frac{d(d+1)}{2}R_{\max },\label{bound: Step 4}
\end{align}

\noindent where $c$ is a universal constant, and recall that $R_{\max } \leq$ $\lceil h\left(0.25 \alpha_\varepsilon\right) \rceil$. The second line follows from equation~\eqref{T_r budget}. The third line applies a scaling argument based on the range of the summation. The last line replaces the term $2r+1$ with the largest round $R_\textup{max}$ and introduces a finite constant $c$.

Next, we bound two terms in equation~\eqref{decomposition of T: part 1} and equation~\eqref{decomposition of T: part 2} separately. The first term, which represents the samples taken before round $R_{\max}$, can be bounded in the following step.

\subsubsection*{Bound (\ref{decomposition of T: part 1}).}\label{algorithm2: Step 5}
Recall that $R_{\max } \leq$ $\lceil h\left(0.25 \alpha_\varepsilon\right) \rceil$ is the round where $G_{R_{\max }} = G_\varepsilon$. We have 
\begin{align}
    &\sum_{r=1}^{\infty} \mathbbm{1} \left[ G_{r-1} \neq G_\varepsilon \right] \mathbbm{1} \left[G_{r-1} \cup B_{r-1} \neq [K] \right] \sum_{\bm{a} \in \mathcal{A}(r-1)}T_r(\bm{a}) \nonumber\\
    &= \sum_{r=1}^{R_{\max }} \mathbbm{1} \left[G_{r-1} \cup B_{r-1} \neq [K] \right] \sum_{\bm{a} \in \mathcal{A}(r-1)}T_r(\bm{a}) \nonumber\\
    &\leq c 2^{2R_{\max }+1} d \log \left(\frac{2K R_{\max }(R_{\max }+1)}{\delta}\right)+\frac{d(d+1)}{2}R_{\max }, \label{upper bound: Step 5}
\end{align}
where the second line follows from the definition of $R_\textup{max}$, as no additional samples are collected after round $R_\textup{max}$. The third line follows directly from equation (\ref{bound: Step 4}).
\subsubsection*{Bound (\ref{decomposition of T: part 2}).}\label{algorithm2: Step 6}
Next, we have 
\begin{align}
    &\sum_{r=1}^{\infty} \mathbbm{1} \left[ G_{r-1} = G_\varepsilon \right] \mathbbm{1} \left[G_{r-1} \cup B_{r-1} \neq [K] \right] \sum_{\bm{a} \in \mathcal{A}(r-1)}T_r(\bm{a}) \nonumber\\
    &= \sum_{r=R_{\max } + 1}^{\infty} \mathbbm{1} \left[ B_{r-1} \neq G_\varepsilon^c \right] \sum_{\bm{a} \in \mathcal{A}(r-1)}T_r(\bm{a}) \nonumber\\
    &\leq \sum_{r=R_{\max } + 1}^{\infty} \left|G_\varepsilon^c \backslash B_{r-1}\right| \sum_{\bm{a} \in \mathcal{A}(r-1)}T_r(\bm{a}) \nonumber\\
    &= \sum_{r=R_{\max } + 1} ^{\infty} \sum_{i \in G_\varepsilon^c} \mathbbm{1}\left[i \notin B_{r-1}\right] \sum_{\bm{a} \in \mathcal{A}(r-1)}T_r(\bm{a}) \nonumber\\
    &= \sum_{i \in G_\varepsilon^c} \sum_{r=R_{\max } + 1} ^{\infty} \mathbbm{1}\left[i \notin B_{r-1}\right] \sum_{\bm{a} \in \mathcal{A}(r-1)}T_r(\bm{a}) \nonumber\\
    &\leq \sum_{i \in G_\varepsilon^c} \sum_{r=1} ^{\infty} \mathbbm{1}\left[i \notin B_{r-1}\right] \sum_{\bm{a} \in \mathcal{A}(r-1)}T_r(\bm{a}) \nonumber\\
    &\leq \sum_{i \in G_\varepsilon^c} \sum_{r=1} ^{\infty} \mathbbm{1}\left[i \notin B_{r-1}\right] \left(\frac{2 d}{\varepsilon_{r}^2} \log \left(\frac{2K r(r+1)}{\delta}\right)+\frac{d(d+1)}{2}\right),\label{upper bound: Step 6}
\end{align}
where the second line follows from the definition of $R_\textup{max}$, as $G_r = G_\varepsilon$ for all rounds beyond $R_\textup{max}$. The third line uses the fact that as long as $G_\varepsilon^c \backslash B_{r-1}$ is not the empty set, the corresponding indicator function is 1. The fourth line considers the indicator function for each arm in $G_\varepsilon$. The fifth and sixth lines exchange the order of the double summation and enlarge the summation range over the rounds $r$. 

\subsection{Step 3: Bound the Expected Total Samples of LinFACT}\label{algorithm2: Step 7}

We now take expectations over the total number of samples drawn for the given bandit instance $\boldsymbol{\mu}$. These expectations are conditioned on the high-probability event $\mathcal{E}_1$.
\begin{align}
    \mathbb{E}_{\boldsymbol{\mu}}\left[T_G \mid \mathcal{E}_1\right]
    &\leq \sum_{r=1}^{\infty} \mathbb{E}_{\boldsymbol{\mu}}\left[\mathbbm{1}\left[G_r \cup B_r \neq[K]\right] \mid \mathcal{E}_1\right] \sum_{\bm{a} \in \mathcal{A}(r-1)}T_r(\bm{a}) \nonumber\\
    &= \sum_{r=1}^{\infty} \mathbb{E}_{\boldsymbol{\mu}}\left[\mathbbm{1}\left[G_{r-1} \neq G_\varepsilon\right] \mathbbm{1}\left[G_{r-1} \cup B_{r-1} \neq[K]\right] \mid \mathcal{E}_1\right] \sum_{\bm{a} \in \mathcal{A}(r-1)}T_r(\bm{a}) \nonumber\\
    &+ \sum_{r=1}^{\infty} \mathbb{E}_{\boldsymbol{\mu}}\left[\mathbbm{1}\left[G_{r-1}=G_\varepsilon\right] \mathbbm{1}\left[G_{r-1} \cup B_{r-1} \neq[K]\right] \mid \mathcal{E}_1\right] \sum_{\bm{a} \in \mathcal{A}(r-1)}T_r(\bm{a}) \nonumber\\
    &\stackrel{\eqref{upper bound: Step 5}}{\leq} c 2^{2R_{\max }+1} d \log \left(\frac{2K R_{\max }(R_{\max }+1)}{\delta}\right)+\frac{d(d+1)}{2}R_{\max } \nonumber\\
    &+ \sum_{r=1}^{\infty} \mathbb{E}_{\boldsymbol{\mu}}\left[\mathbbm{1}\left[G_{r-1}=G_\varepsilon\right] \mathbbm{1}\left[G_{r-1} \cup B_{r-1} \neq[K]\right] \mid \mathcal{E}_1\right] \sum_{\bm{a} \in \mathcal{A}(r-1)}T_r(\bm{a}) \nonumber\\
    &\stackrel{\eqref{upper bound: Step 6}}{\leq} c 2^{2R_{\max }+1} d \log \left(\frac{2K R_{\max }(R_{\max }+1)}{\delta}\right)+\frac{d(d+1)}{2} R_{\max } \nonumber\\
    &+ \sum_{i \in G_\varepsilon^c} \sum_{r=1} ^{\infty} 
    \mathbb{E}_{\boldsymbol{\mu}}\left[ \mathbbm{1}\left[i \notin B_{r-1} \mid \mathcal{E}_1\right] \right] \left(\frac{2 d}{\varepsilon_{r}^2} \log \left(\frac{2K r(r+1)}{\delta}\right)+\frac{d(d+1)}{2}\right).\label{equation: decomposition of T}
\end{align}
The first line follows from the stopping condition of LinFACT-G and the additivity of the expectation. The second line applies the decomposition established in Step 2. The subsequent lines use the results from equation~\eqref{upper bound: Step 5} and equation~\eqref{upper bound: Step 6}.

Next, we bound the last term. For a given $i \in G_\varepsilon^c$ and round $r$, we first bound the probability that $i \notin B_r$. By the Borel-Cantelli lemma, this implies that the probability of $i$ never being added to any $B_r$ is zero.

\begin{lemma}\label{claim8_1}
    For $i \in G_\varepsilon^c$ and $r \geq\left\lceil\log _2\left(\frac{4}{\Delta_i-\varepsilon}\right)\right\rceil$, we have $\mathbb{E}_{\boldsymbol{\mu}}\left[\mathbbm{1}\left[i \notin B_r\right] \mid \mathcal{E}_1\right] = 0$.
\end{lemma}

\proof\\{\textit{Proof.}}

First, we have for any $i \in G_\varepsilon^c$, 
\begin{align}\label{eq_Ev_claim0}
    \mathbb{E}_{\boldsymbol{\mu}}\left[\mathbbm{1}\left[i \notin B_r\right] \mid \mathcal{E}_1\right] &= \mathbb{E}_{\boldsymbol{\mu}}\left[ \mathbbm{1}\left[\max_{j \in \mathcal{A}_I(r-1)} \hat{\mu}_j - \hat{\mu}_i \leq 2\varepsilon_r + \varepsilon\right] \mid \mathcal{E}_1, i \notin B_m (m = \{1, 2, \ldots, r-1\}) \right] \nonumber \\
    &\leq \mathbb{E}_{\boldsymbol{\mu}}\left[ \mathbbm{1}\left[\max_{j \in \mathcal{A}_I(r-1)} \hat{\mu}_j - \hat{\mu}_i \leq 2\varepsilon_r + \varepsilon\right] \mid \mathcal{E}_1\right],
\end{align}
where the first line follows from the fact that arm $i \in G_\varepsilon^c$ will never be added into $B_r$ from round 1 to round $r-1$ if $i \notin B_r$, and the second line accounts for the conditional expectation.

If $i \in B_{r-1}$, then $i \in B_r$ by definition. Otherwise, if $i \notin B_{r-1}$, then under event $\mathcal{E}_1$, for $i \in G_\varepsilon^c$ and $r \geq\left\lceil\log _2\left(\frac{4}{\Delta_i-\varepsilon}\right)\right\rceil$, we have
\begin{equation}
    \max_{j \in \mathcal{A}_I(r-1)} \hat{\mu}_j - \hat{\mu}_i \geq \Delta_i-2^{-r+1} \geq \varepsilon+2\varepsilon_r,
\end{equation}
which implies that $i \in B_r$ by line \ref{elimination: good 1} of the Algorithm \ref{alg_LinFACT_stopping}. In other words, we have established the correctness of the algorithm when line \ref{elimination: good 1} is triggered in Step 1. We now specify the exact condition under which line \ref{elimination: good 1} will occur. In particular, under event $\mathcal{E}_1$, if $i \notin B_{r-1}$, then for all $i \in G_\varepsilon^c$ and $r \geq\left\lceil\log _2\left(\frac{4}{\Delta_i-\varepsilon}\right)\right\rceil$, we have
\begin{equation}\label{expectation about contradictory}
    \mathbb{E}_{\boldsymbol{\mu}}\left[ \mathbbm{1}\left[\max_{j \in \mathcal{A}_I(r-1)} \hat{\mu}_j - \hat{\mu}_i > 2\varepsilon_r + \varepsilon\right] \mid i \notin B_{r-1}, \mathcal{E}_1 \right] = 1.
\end{equation}

Therefore, for all $i \in G_\varepsilon^c$ and $r \geq\left\lceil\log _2\left(\frac{4}{\Delta_i-\varepsilon}\right)\right\rceil$, we have
\begin{align}
    &\mathbb{E}_{\boldsymbol{\mu}}\left[ \mathbbm{1}\left[\max_{j \in \mathcal{A}_I(r-1)} \hat{\mu}_j - \hat{\mu}_i \leq 2\varepsilon_r + \varepsilon\right] \mid \mathcal{E}_1 \right] \nonumber\\
    &= \mathbb{E}_{\boldsymbol{\mu}}\left[ \mathbbm{1}\left[\max_{j \in \mathcal{A}_I(r-1)} \hat{\mu}_j - \hat{\mu}_i \leq 2\varepsilon_r + \varepsilon\right] \mathbbm{1}\left[i \notin B_{r-1}\right] \mid \mathcal{E}_1 \right] \nonumber\\
    &+ \mathbb{E}_{\boldsymbol{\mu}}\left[ \mathbbm{1}\left[\max_{j \in \mathcal{A}_I(r-1)} \hat{\mu}_j - \hat{\mu}_i \leq 2\varepsilon_r + \varepsilon\right] \mathbbm{1}\left[i \in B_{r-1}\right] \mid \mathcal{E}_1 \right]\nonumber\\
    &= \mathbb{E}_{\boldsymbol{\mu}}\left[ \mathbbm{1}\left[\max_{j \in \mathcal{A}_I(r-1)} \hat{\mu}_j - \hat{\mu}_i \leq 2\varepsilon_r + \varepsilon\right] \mathbbm{1}\left[i \notin B_{r-1}\right] \mid \mathcal{E}_1 \right] \nonumber\\
    &=\mathbb{E}_{\boldsymbol{\mu}}\left[\mathbbm{1}\left[\max_{j \in \mathcal{A}_I(r-1)} \hat{\mu}_j - \hat{\mu}_i \leq 2\varepsilon_r + \varepsilon\right] \mathbbm{1}\left[i \notin B_{r-1}\right] \mid i \notin B_{r-1}, \mathcal{E}_1\right] \mathbb{P}\left(i \notin B_{r-1} \mid \mathcal{E}_1\right) \nonumber\\
    &+ \mathbb{E}_{\boldsymbol{\mu}}\left[\mathbbm{1}\left[\max_{j \in \mathcal{A}_I(r-1)} \hat{\mu}_j - \hat{\mu}_i \leq 2\varepsilon_r + \varepsilon\right] \mathbbm{1}\left[i \notin B_{r-1}\right] \mid i \in B_{r-1}, \mathcal{E}_1\right] \mathbb{P}\left(i \in B_{r-1} \mid \mathcal{E}_1\right) \nonumber\\
    &= \mathbb{E}_{\boldsymbol{\mu}}\left[\mathbbm{1}\left[\max_{j \in \mathcal{A}_I(r-1)} \hat{\mu}_j - \hat{\mu}_i \leq 2\varepsilon_r + \varepsilon\right] \mathbbm{1}\left[i \notin B_{r-1}\right] \mid i \notin B_{r-1}, \mathcal{E}_1\right] \mathbb{P}\left(i \notin B_{r-1} \mid \mathcal{E}_1\right) \nonumber\\
    &= \mathbb{E}_{\boldsymbol{\mu}}\left[\mathbbm{1}\left[\max_{j \in \mathcal{A}_I(r-1)} \hat{\mu}_j - \hat{\mu}_i \leq 2\varepsilon_r + \varepsilon\right] \mid i \notin B_{r-1}, \mathcal{E}_1\right] \mathbb{E}_{\boldsymbol{\mu}}\left[\mathbbm{1}\left[i \notin B_{r-1}\right] \mid \mathcal{E}_1\right] \nonumber\\
    &= 0,
\end{align}
where the second line follows from the additivity of expectation. The fourth line follows the deterministic result that $\mathbbm{1}\left[i \notin B_r\right] \mathbbm{1}\left[i \in B_{r-1}\right]=0$. The fifth line is the decomposition based on the conditional expectation. The eighth line uses the fact that the expectation of the indicator function is simply the probability. The last line follows the result in equation~\eqref{expectation about contradictory}. The lemma then follows by combining this result with equation~\eqref{eq_Ev_claim0}. 
\hfill\Halmos
\endproof

\begin{lemma}\label{claim8_2}
    For $i \in G_\varepsilon^c$, we have
\begin{align}
    &\sum_{r=1} ^{\infty} \mathbb{E}_{\boldsymbol{\mu}}\left[\mathbbm{1}\left[i \notin B_{r-1}\right] \mid \mathcal{E}_1\right] \left(\frac{2 d}{\varepsilon_{r}^2} \log \left(\frac{2K r(r+1)}{\delta}\right)+\frac{d(d+1)}{2}\right) \nonumber\\ 
    &\leq \frac{d(d+1)}{2} \log _2\left(\frac{8}{\Delta_i-\varepsilon}\right) + \xi \frac{256d}{(\Delta_i - \varepsilon)^2} \log \left(\frac{2K}{\delta} \log_2 \frac{16}{\Delta_i - \varepsilon} \right).
\end{align}
\end{lemma}
    
\proof\\{\textit{Proof.}}
\begin{align}
& \sum_{r=1}^{\infty} \mathbb{E}_{\boldsymbol{\mu}}\left[\mathbbm{1}\left[i \notin B_{r-1}\right] \mid \mathcal{E}_1\right]\left(\frac{2 d}{\varepsilon_{r}^2} \log \left(\frac{2K r(r+1)}{\delta}\right)+\frac{d(d+1)}{2}\right) \nonumber\\
&= \sum_{r=1}^{\left\lceil\log _2\left(\frac{4}{\Delta_i-\varepsilon}\right)\right\rceil} \mathbb{E}_{\boldsymbol{\mu}}\left[\mathbbm{1}\left[i \notin B_{r-1}\right] \mid \mathcal{E}_1\right]\left(\frac{2 d}{\varepsilon_{r}^2} \log \left(\frac{2K r(r+1)}{\delta}\right)+\frac{d(d+1)}{2}\right) \nonumber\\
&+ \sum_{r=\left\lceil\log _2\left(\frac{4}{\Delta_i-\varepsilon}\right)\right\rceil+1}^{\infty} \mathbb{E}_{\boldsymbol{\mu}}\left[\mathbbm{1}\left[i \notin B_{r-1}\right] \mid \mathcal{E}_1\right]\left(\frac{2 d}{\varepsilon_{r}^2} \log \left(\frac{2K r(r+1)}{\delta}\right)+\frac{d(d+1)}{2}\right) \nonumber\\
&= \sum_{r=1}^{\left\lceil\log _2\left(\frac{4}{\Delta_i-\varepsilon}\right)\right\rceil} \mathbb{E}_{\boldsymbol{\mu}}\left[\mathbbm{1}\left[i \notin B_{r-1}\right] \mid \mathcal{E}_1\right]\left(\frac{2 d}{\varepsilon_{r}^2} \log \left(\frac{2K r(r+1)}{\delta}\right)+\frac{d(d+1)}{2}\right) + 0 \nonumber\\
&\leq \sum_{r=1}^{\left\lceil\log _2\left(\frac{4}{\Delta_i-\varepsilon}\right)\right\rceil} \left(d2^{2r+1} \log \left(\frac{2K r(r+1)}{\delta}\right)+\frac{d(d+1)}{2}\right) \nonumber\\
&\leq \frac{d(d+1)}{2} \log _2\left(\frac{8}{\Delta_i-\varepsilon}\right) + 2d\log \left(\frac{2K}{\delta}\right) \sum_{r=1}^{{\left\lceil\log _2\left(\frac{4}{\Delta_i-\varepsilon}\right)\right\rceil}} 2^{2r} + 4d \sum_{r=1}^{{\left\lceil\log _2\left(\frac{4}{\Delta_i-\varepsilon}\right)\right\rceil}} 2^{2r} \log (r+1) \nonumber\\
&\leq \frac{d(d+1)}{2} \log _2\left(\frac{8}{\Delta_i-\varepsilon}\right) + \xi \frac{256d}{(\Delta_i - \varepsilon)^2} \log \left(\frac{2K}{\delta} \log_2 \frac{16}{\Delta_i - \varepsilon} \right),
\end{align}
Here, $\xi$ is a sufficiently large universal constant. The second line follows from decomposing the summation across all rounds. The fourth line uses Lemma \ref{claim8_1}. The fifth line bounds the expectation of the indicator function to its maximum value of 1. The sixth and seventh lines replace the round $r$ with its maximum value $\left\lceil\log _2\left(\frac{4}{\Delta_i-\varepsilon}\right)\right\rceil$. 
\hfill\Halmos
\endproof

\ 
Summarizing the aforementioned results, we have 
\begin{align}
    \mathbb{E}_{\boldsymbol{\mu}}\left[T_G \mid \mathcal{E}_1\right]
    &\leq c 2^{2R_{\max }+1} d \log \left(\frac{2K R_{\max }(R_{\max }+1)}{\delta}\right)+\frac{d(d+1)}{2}R_{\max } \nonumber\\
    &+ \sum_{i \in G_\varepsilon^c} \sum_{r=1} ^{\infty} 
    \mathbb{E}_{\boldsymbol{\mu}}\left[ \mathbbm{1}\left[i \notin B_{r-1} 
    \mid \mathcal{E}_1 \right] \right] \left(\frac{2 d}{\varepsilon_{r}^2} \log \left(\frac{2K r(r+1)}{\delta}\right)+\frac{d(d+1)}{2}\right),
\end{align}
where
\begin{equation}
    R_{\max } \leq \lceil h\left(0.25 \alpha_\varepsilon\right) \rceil = \log_2 \frac{8}{\alpha_\varepsilon}.
\end{equation}

Also, we have
\begin{align}
& \sum_{r=1}^{\infty} \mathbb{E}_{\boldsymbol{\mu}}\left[\mathbbm{1}\left[i \notin B_{r-1}\right] \mid \mathcal{E}_1\right]\left(\frac{2 d}{\varepsilon_{r}^2} \log \left(\frac{2K r(r+1)}{\delta}\right)+\frac{d(d+1)}{2}\right) \nonumber\\
&\leq \frac{d(d+1)}{2} \log _2\left(\frac{8}{\Delta_i-\varepsilon}\right) + \xi \frac{256d}{(\Delta_i - \varepsilon)^2} \log \left(\frac{2K}{\delta} \log_2 \frac{16}{\Delta_i - \varepsilon} \right).
\end{align}

Then, we arrive at the final result as follows,
\begin{align}
    \mathbb{E}_{\boldsymbol{\mu}}\left[T_G \mid \mathcal{E}_1\right]
    &\leq c 2^{2R_{\max }+1} d \log \left(\frac{2K R_{\max }(R_{\max }+1)}{\delta}\right)+\frac{d(d+1)}{2}R_{\max } \nonumber\\
    &+ \sum_{i \in G_\varepsilon^c} \sum_{r=1} ^{\infty} 
    \mathbb{E}_{\boldsymbol{\mu}}\left[ \mathbbm{1}\left[i \notin B_{r-1} \mid \mathcal{E}_1 \right] \right] \left(\frac{2 d}{\varepsilon_{r}^2} \log \left(\frac{2K r(r+1)}{\delta}\right)+\frac{d(d+1)}{2}\right) \nonumber\\
    &\leq c \frac{256d}{\alpha_\varepsilon^2} \log \left( \frac{2K}{\delta} \log_2 \left( \frac{16}{\alpha_\varepsilon} \right) \right) + \frac{d(d+1)}{2} \log_2 \frac{8}{\alpha_\varepsilon} \nonumber\\
    &+ \sum_{i \in G_\varepsilon^c}\left( \frac{d(d+1)}{2} \log _2\left(\frac{8}{\Delta_i-\varepsilon}\right) + \xi \frac{256d}{(\Delta_i - \varepsilon)^2} \log \left(\frac{2K}{\delta} \log_2 \frac{16}{\Delta_i - \varepsilon} \right) \right),
\end{align}
which can be further expressed as
\begin{align}\label{upper bound sum}
        \mathbb{E}[T_G \mid \mathcal{E}_1] 
        &= \mathcal{O} 
        \left(d \alpha_\varepsilon^{-2} \log\!\left( \frac{K}{\delta} \log (\alpha_\varepsilon^{-2}) \right) 
        + d^{2} \log(\alpha_\varepsilon^{-1})\right) \nonumber\\
        &+ \sum_{i \in G_\varepsilon^c} \left( \mathcal{O}\left( d^2\log (\Delta_i-\varepsilon)^{-1} + d (\Delta_i-\varepsilon)^{-2} \log \left( \frac{K}{\delta} \log (\Delta_i-\varepsilon)^{-2} \right) \right) \right).
    \end{align}

\subsection{A Refined Bound}\label{refined version}
The result obtained in the previous steps involves a summation over the set $G_\varepsilon$, which can be further improved by eliminating this summation. Rather than focusing solely on the round $R_\textup{max}$, defined in Lemma \ref{claim8_1} as the round in which all arms in $G_\varepsilon$ are classified into $G_r$, we now define the round at which all classifications are complete, \emph{i.e.}, $G_r \cup B_r = [K]$.

\begin{lemma}\label{claim_refine_1}
    For $i \in G_\varepsilon$ and $r \geq\left\lceil\log _2\left(\frac{4}{\varepsilon-\Delta_i}\right)\right\rceil$, we have $ \mathbb{E}_{\boldsymbol{\mu}}\left[\mathbbm{1}\left[i \notin G_r\right] \mid \mathcal{E}_1\right] = 0$.
\end{lemma}

\proof\\{\textit{Proof.}}
First, for any $i \in G_\varepsilon$
\begin{align}\label{eq_EvG_claim0}
    \mathbb{E}_{\boldsymbol{\mu}}\left[\mathbbm{1}\left[i \notin G_r\right] \mid \mathcal{E}_1\right] &= \mathbb{E}_{\boldsymbol{\mu}}\left[ \mathbbm{1}\left[\max_{j \in \mathcal{A}_I(r-1)} \hat{\mu}_j - \hat{\mu}_i \geq -2\varepsilon_r + \varepsilon\right] \mid \mathcal{E}_1, i \notin G_m (m = \{1, 2, \ldots, r-1\}) \right] \nonumber \\
    &\leq \mathbb{E}_{\boldsymbol{\mu}}\left[ \mathbbm{1}\left[\max_{j \in \mathcal{A}_I(r-1)} \hat{\mu}_j - \hat{\mu}_i \geq -2\varepsilon_r + \varepsilon\right] \mid \mathcal{E}_1\right].
\end{align}

If $i \in G_{r-1}$, then $i \in G_r$ by definition. Otherwise, if $i \notin G_{r-1}$, then under event $\mathcal{E}_1$, for $i \in G_\varepsilon$ and $r \geq\left\lceil\log _2\left(\frac{4}{\varepsilon-\Delta_i}\right)\right\rceil$, we have
\begin{equation}
    \max_{j \in \mathcal{A}_I(r-1)} \hat{\mu}_j - \hat{\mu}_i \leq \mu_{\arg\max_{j \in \mathcal{A}_I(r-1)}\hat{\mu}_j} - \mu_i +2^{-r+1}\leq \Delta_i+2^{-r+1} \leq \varepsilon-2\varepsilon_r,
\end{equation}
which implies that $i \in G_r$ by line \ref{add to Gr} of the Algorithm \ref{alg_LinFACT_stopping}. In other words, we now specify the exact condition under which line \ref{add to Gr} will occur. In particular, under event $\mathcal{E}_1$, if $i \notin G_{r-1}$, for all $i \in G_\varepsilon$ and $r \geq\left\lceil\log _2\left(\frac{4}{\varepsilon-\Delta_i}\right)\right\rceil$, we have
\begin{equation}\label{expectation about contradictory_xy}
    \mathbb{E}_{\boldsymbol{\mu}}\left[ \mathbbm{1}\left[\max_{j \in \mathcal{A}_I(r-1)} \hat{\mu}_j - \hat{\mu}_i \leq -2\varepsilon_r + \varepsilon\right] \mid i \notin G_{r-1}, \mathcal{E}_1 \right] = 1.
\end{equation}

Deterministically, $\mathbbm{1}\left[i \notin G_r\right] \mathbbm{1}\left[i \in G_{r-1}\right]=0$. Therefore,
\begin{align}
    &\mathbb{E}_{\boldsymbol{\mu}}\left[ \mathbbm{1}\left[\max_{j \in \mathcal{A}_I(r-1)} \hat{\mu}_j - \hat{\mu}_i \geq -2\varepsilon_r + \varepsilon\right] \mid \mathcal{E}_1 \right] \nonumber\\
    &= \mathbb{E}_{\boldsymbol{\mu}}\left[ \mathbbm{1}\left[\max_{j \in \mathcal{A}_I(r-1)} \hat{\mu}_j - \hat{\mu}_i \geq -2\varepsilon_r + \varepsilon\right] \mathbbm{1}\left[i \notin G_{r-1}\right] \mid \mathcal{E}_1 \right] \nonumber\\
    &+ \mathbb{E}_{\boldsymbol{\mu}}\left[ \mathbbm{1}\left[\max_{j \in \mathcal{A}_I(r-1)} \hat{\mu}_j - \hat{\mu}_i \geq -2\varepsilon_r + \varepsilon\right] \mathbbm{1}\left[i \in G_{r-1}\right] \mid \mathcal{E}_1 \right]\nonumber\\
    &= \mathbb{E}_{\boldsymbol{\mu}}\left[ \mathbbm{1}\left[\max_{j \in \mathcal{A}_I(r-1)} \hat{\mu}_j - \hat{\mu}_i \geq -2\varepsilon_r + \varepsilon\right] \mathbbm{1}\left[i \notin G_{r-1}\right] \mid \mathcal{E}_1 \right] \nonumber\\
    &=\mathbb{E}_{\boldsymbol{\mu}}\left[\mathbbm{1}\left[\max_{j \in \mathcal{A}_I(r-1)} \hat{\mu}_j - \hat{\mu}_i \geq -2\varepsilon_r + \varepsilon\right] \mathbbm{1}\left[i \notin G_{r-1}\right] \mid i \notin G_{r-1}, \mathcal{E}_1\right] \mathbb{P}\left(i \notin G_{r-1} \mid \mathcal{E}_1\right) \nonumber\\
    &+ \mathbb{E}_{\boldsymbol{\mu}}\left[\mathbbm{1}\left[\max_{j \in \mathcal{A}_I(r-1)} \hat{\mu}_j - \hat{\mu}_i \geq -2\varepsilon_r + \varepsilon\right] \mathbbm{1}\left[i \notin G_{r-1}\right] \mid i \in G_{r-1}, \mathcal{E}_1\right] \mathbb{P}\left(i \in G_{r-1} \mid \mathcal{E}_1\right) \nonumber\\
    &= \mathbb{E}_{\boldsymbol{\mu}}\left[\mathbbm{1}\left[\max_{j \in \mathcal{A}_I(r-1)} \hat{\mu}_j - \hat{\mu}_i \geq -2\varepsilon_r + \varepsilon\right] \mathbbm{1}\left[i \notin G_{r-1}\right] \mid i \notin G_{r-1}, \mathcal{E}_1\right] \mathbb{P}\left(i \notin G_{r-1} \mid \mathcal{E}_1\right) \nonumber\\
    &= \mathbb{E}_{\boldsymbol{\mu}}\left[\mathbbm{1}\left[\max_{j \in \mathcal{A}_I(r-1)} \hat{\mu}_j - \hat{\mu}_i \geq -2\varepsilon_r + \varepsilon\right] \mid i \notin G_{r-1}, \mathcal{E}_1\right] \mathbb{E}_{\boldsymbol{\mu}}\left[\mathbbm{1}\left[i \notin G_{r-1}\right] \mid \mathcal{E}_1\right] \nonumber\\
    &= 0,
\end{align}
where the second line comes from the additivity of expectation. The fourth line follows the deterministic result that $\mathbbm{1}\left[i \notin G_r\right] \mathbbm{1}\left[i \in G_{r-1}\right]=0$. The fifth line applies a decomposition based on the conditional expectation. The eighth line uses the fact that the expectation of an indicator function equals the corresponding probability. The final line follows from the result in equation~\eqref{expectation about contradictory_xy}. The lemma then follows by combining this result with equation \eqref{eq_EvG_claim0}.
\hfill\Halmos
\endproof

\begin{lemma}\label{claim_refine_2}
    The round at which all classifications are complete and the final answer is returned is $R_{\textup{upper}} = \max \left\{ \left\lceil \log_2 \frac{4}{\alpha_\varepsilon} \right\rceil, \left\lceil \log_2 \frac{4}{\beta_\varepsilon} \right\rceil \right\}$.
\end{lemma}
\proof\\{\textit{Proof.}}
Combining the result of Lemma \ref{claim_refine_1} with that of Lemma \ref{claim8_1}, we have the following: for $i \in G_\varepsilon^c$ and $r \geq\left\lceil\log _2\left(\frac{4}{\Delta_i-\varepsilon}\right)\right\rceil$, we have $\mathbb{E}_{\boldsymbol{\mu}}\left[\mathbbm{1}\left[i \notin B_r\right] \mid \mathcal{E}_1\right] = 0$; for $i \in G_\varepsilon$ and $r \geq\left\lceil\log _2\left(\frac{4}{\varepsilon-\Delta_i}\right)\right\rceil$, we have $\mathbb{E}_{\boldsymbol{\mu}}\left[\mathbbm{1}\left[i \notin G_r\right] \mid \mathcal{E}_1\right] = 0$. Since $\alpha_\varepsilon = \min_{i \in G_\varepsilon}(\varepsilon - \Delta_i)$ and $\beta_\varepsilon = \min_{i \in G_\varepsilon^c}(\Delta_i - \varepsilon)$, for any round $r \geq R_{\textup{upper}}$, all arms have been classified into either $G_r$ or $B_r$, marking the termination of the algorithm.
\hfill\Halmos
\endproof

\begin{lemma}\label{claim_refine_3}
    For the expected sample complexity conditioned on the high-probability event $\mathcal{E}_1$, we have
\begin{align}
    \mathbb{E}_{\boldsymbol{\mu}}\left[T_{\mathcal{XY}} \mid \mathcal{E}_1\right]
    &\leq \zeta \max\left\{ \frac{256d}{\alpha_\varepsilon^2} \log \left( \frac{2K}{\delta} \log_2 \frac{16}{\alpha_\varepsilon} \right), \frac{256d}{\beta_\varepsilon^2} \log \left( \frac{2K}{\delta} \log_2 \frac{16}{\beta_\varepsilon} \right) \right\} + \frac{d(d+1)}{2} R_{\textup{upper}},
\end{align}
where $\zeta$ is a universal constant and $R_{\textup{upper}} = \max \left\{ \left\lceil \log_2 \frac{4}{\alpha_\varepsilon} \right\rceil, \left\lceil \log_2 \frac{4}{\beta_\varepsilon} \right\rceil \right\}$.
\end{lemma}

\proof\\{\textit{Proof.}}
We have
\begin{align}
    \mathbb{E}_{\boldsymbol{\mu}}\left[T_{\mathcal{XY}} \mid \mathcal{E}_1\right]
    &\leq \sum_{r=1}^{\infty} \mathbb{E}_{\boldsymbol{\mu}}\left[\mathbbm{1}\left[G_r \cup B_r \neq[K]\right] \mid \mathcal{E}_1\right] \sum_{\bm{a} \in \mathcal{A}(r-1)}T_r(\bm{a}) \nonumber\\
    &\leq \sum_{r=1}^{R_{\textup{upper}}} \left( d 2^{2r+1} \log \left(\frac{2K r(r+1)}{\delta}\right)+\frac{d(d+1)}{2}\right) \nonumber\\
    &\leq \frac{d(d+1)}{2} R_{\textup{upper}} + 2d\log \left(\frac{2K}{\delta}\right) \sum_{r=1}^{{R_{\textup{upper}}}} 2^{2r} + 4d \sum_{r=1}^{{R_{\textup{upper}}}} 2^{2r} \log (r+1) \nonumber\\
    &\leq 4\log \left[ \frac{2K}{\delta}\left( R_{\textup{upper}} + 1 \right) \right] \sum_{r=1}^{{R_{\textup{upper}}}} d 2^{2r} + \frac{d(d+1)}{2} R_{\textup{upper}} \label{middle result_refined} \\
    &\leq \zeta \max\left\{ \frac{256d}{\alpha_\varepsilon^2} \log \left( \frac{2K}{\delta} \log_2 \frac{16}{\alpha_\varepsilon} \right), \frac{256d}{\beta_\varepsilon^2} \log \left( \frac{2K}{\delta} \log_2 \frac{16}{\beta_\varepsilon} \right) \right\} + \frac{d(d+1)}{2} R_{\textup{upper}}.\label{another angle: decomposition of T - 1}
\end{align}
The second line follows from Lemma \ref{claim_refine_2}. Then, we have
\begin{equation}
    \mathbb{E}[T_G \mid \mathcal{E}] 
    = \mathcal{O} 
        \left(d \xi^{-2} \log\!\left( \frac{K}{\delta} \log_{2}(\xi^{-2}) \right) 
        + d^{2} \log(\xi^{-1})\right),
\end{equation}
where $\xi = \min(\alpha_\varepsilon, \beta_\varepsilon)/16$ is the minimum gap between $\alpha_\varepsilon$ and $\beta_\varepsilon$, indicating the difficulty of the problem instance.

\hfill\Halmos
\endproof

\section{Additional Insights into the Algorithm Optimality}\label{additional insights}

From another perspective, we consider the relationship between the lower bound and the upper bound in the following section and give some additional insights into the algorithm optimality. This relationship serves as the basis for the derivation of a near-optimal upper bound in Theorem \ref{upper bound: Algorithm 1 XY}.

For $\forall i \in \left( \mathcal{A}_I(r-1) \cap G_\varepsilon\right)$ and $\forall j \in \mathcal{A}_I(r-1)$, $j \neq i$, in round $r$, we have
\begin{equation}
    \boldsymbol{y}_{j, i}^\top(\hat{\boldsymbol{\theta}}_r-\boldsymbol{\theta}^{}) \leq 2\varepsilon_r
\end{equation}

\noindent and
\begin{equation}\label{Gr prime}
    \boldsymbol{y}_{j, i}^\top\hat{\boldsymbol{\theta}}_r - \varepsilon \leq \boldsymbol{y}_{j, i}^\top \boldsymbol{\theta}^{} + 2\varepsilon_r - \varepsilon.
\end{equation}

\begin{lemma}\label{claim_add_1}
    Define $G_r^\prime \coloneqq \left\{ \exists j \in \mathcal{A}_I(r-1), j \neq i, i: \boldsymbol{y}_{j,i}^\top\boldsymbol{\theta}^{} - \varepsilon > -4\varepsilon_r \right\}$. We always have $\left( \mathcal{A}_I(r-1) \cap G_\varepsilon \cap G_r^c \right) \subseteq G_r^\prime$.
\end{lemma}

\proof\\{\textit{Proof.}}
For $r = 1$, the lemma follows directly from the assumption in Theorem~\ref{upper bound: Algorithm 1 XY} that $\max_{i \in [K]}\vert \mu_1 - \varepsilon - \mu_i \vert \leq 2$. For $r \geq 2$, we proceed by contradiction. Suppose $i \in \left( \mathcal{A}_I(r-1) \cap G_\varepsilon \cap G_r^c\right) \cap (G_r^\prime)^c$, then for every $j \in \mathcal{A}_I(r-1)$ with $j \neq i$, we have 
\begin{equation}
    \boldsymbol{y}_{j, i}^\top\boldsymbol{\theta}^{} \leq -4\varepsilon_r +\varepsilon.
\end{equation}

\noindent Hence, using equation (\ref{Gr prime}), for every $j \in \mathcal{A}_I(r-1)$ and $j \neq i$, we have
\begin{equation}
    \boldsymbol{y}_{j, i}^\top\hat{\boldsymbol{\theta}}_r - \varepsilon \leq-2\varepsilon_r,
\end{equation}
which is exactly the condition for the algorithm to add arm $i$ into $G_r$ at line \ref{add to Gr} of the algorithm, which yields a contradiction and completes the argument. Moreover, note that when $r \geq \left\lceil \log_2 \frac{4}{\alpha_\varepsilon} \right\rceil$, we have $G_r^\prime \cap G_\varepsilon = \emptyset$. Furthermore, considering that $i \in G_\varepsilon$, we have $\bm{y}_{i, j}^\top\boldsymbol{\theta}^{} + \varepsilon > 0$.
\hfill\Halmos
\endproof

There is, however, one exceptional case that invalidates the above derivation. Specifically, when $i$ is the index of the arm with the largest mean value, \emph{i.e.}, $i = \arg \max_{i \in \mathcal{A}_I(r-1)}$, the proof no longer holds. For this situation to occur, it must be the case that $\arg \max_{i \in \mathcal{A}_I(r-1)} \in G_r^c$, which is equivalent to $\varepsilon \leq 2\varepsilon_r$. Since this condition can only hold for a limited number of rounds, its impact is negligible and can therefore be ignored.

On the other hand, for $i \in \left( \mathcal{A}_I(r-1) \cap G_\varepsilon^c \right)$, in any round $r$, we have
\begin{equation}
    \boldsymbol{y}_{1,i}^\top(\boldsymbol{\theta}^{}-\hat{\boldsymbol{\theta}}_r) \leq 2\varepsilon_r
\end{equation}
and
\begin{equation}\label{Br prime}
    \boldsymbol{y}_{1, i}^\top\hat{\boldsymbol{\theta}}_r - \varepsilon \geq \boldsymbol{y}_{1, i}^\top \boldsymbol{\theta}^{} - 2\varepsilon_r - \varepsilon.
\end{equation}

\begin{lemma}\label{claim_add_2}
    Define $B_r^\prime \coloneqq \left\{ i: \boldsymbol{y}_{1,i}^\top\boldsymbol{\theta}^{} - \varepsilon < 4\varepsilon_r \right\}$. We always have $\left( \mathcal{A}_I(r-1) \cap G_\varepsilon^c\right) \subset B_r^\prime$.
\end{lemma}
    
\proof\\{\textit{Proof.}}
To establish this, first note that when $r = 1$, the lemma directly follows from the assumption in Theorem \ref{upper bound: Algorithm 1 XY} that $\max_{i \in [K]}\vert \mu_1 - \varepsilon - \mu_i \vert \leq 2$. For $r \geq 2$, using the same contradiction argument, assume that $i \in \left( \mathcal{A}_I(r-1) \cap G_\varepsilon^c\right) \cap (B_r^\prime)^c$. Then we must have
\begin{equation}
    \boldsymbol{y}_{1, i}^\top\boldsymbol{\theta}^{} \geq 4\varepsilon_r +\varepsilon.
\end{equation}

Consequently, by equation~(\ref{Br prime}), it follows that
\begin{equation}
    \boldsymbol{y}_{1, i}^\top\hat{\boldsymbol{\theta}}_r - \varepsilon \geq 2\varepsilon_r.
\end{equation}

This is precisely the condition for the algorithm to add arm $i$ into $B_r$ and eliminate it from $\mathcal{A}_I(r-1)$ as specified in line \ref{elimination: good 1} of the algorithm. This contradiction leads to the desired result. Moreover, when $r \geq \left\lceil \log_2 \frac{4}{\beta_\varepsilon} \right\rceil$, we have $B_r^\prime \cap G_\varepsilon^c = \emptyset$. Finally, for $i \in G_\varepsilon^c$, it follows that $\boldsymbol{y}_{1,i}^\top\boldsymbol{\theta}^{} - \varepsilon > 0$.
\hfill\Halmos
\endproof

We now present a critical lemma that establishes the connection between the lower bound in Theorem~\ref{lower bound: All-epsilon in the linear setting} and the upper bound. Define ${\mathcal{G}_{\mathcal{Y}}}$ as the gauge of $\mathcal{Y}(\mathcal{A})$, where $\mathcal{A}$ denotes the initial set of all arm vectors. The details are provided in Lemma~\ref{lemma: inequality iv}. 

\begin{lemma}\label{claim_add_3}
    Considering the lower bound $(\Gamma^\ast)^{-1}$ in Theorem \ref{lower bound: All-epsilon in the linear setting}, we have
\begin{equation}
     (\Gamma^\ast)^{-1} \geq \frac{{\mathcal{G}_{\mathcal{Y}}}^2 L_2}{4R_{\textup{upper}} d L_1}  \sum_{r=1}^{R_{\textup{upper}}} 2^{2r-3} \min_{\bm{p} \in S_K} \max_{i, j \in \mathcal{A}_I(r-1)} \Vert \boldsymbol{y}_{i, j} \Vert_{\bm{V}_{\bm{p}}^{-1}}^2,
\end{equation}
where $R_{\textup{upper}} = \max \left\{ \left\lceil \log_2 \frac{4}{\alpha_\varepsilon} \right\rceil, \left\lceil \log_2 \frac{4}{\beta_\varepsilon} \right\rceil \right\}$, and $L_1$ and $L_2$ are constants. 
\end{lemma}

\proof\\{\textit{Proof.}}
When round $r$ exceeds $R_{\textup{upper}}$, we have $G_r^\prime \cap G_\varepsilon = \emptyset$ and $B_r^\prime \cap G_\varepsilon^c = \emptyset$, implying that $G_r \cup B_r = [K]$ and the algorithm terminates. From Theorem \ref{lower bound: All-epsilon in the linear setting}, we obtain
\begin{align}
    (\Gamma^\ast)^{-1} &= \min_{\bm{p} \in S_K}\max_{(i, j, m) \in \mathcal{X}} \max \left\{ \frac{2 \Vert \boldsymbol{y}_{i, j} \Vert_{\bm{V}_{\bm{p}}^{-1}}^2}{\left ( \boldsymbol{y}_{i, j}^\top \boldsymbol{\theta}^{} + \varepsilon \right )^2}, \frac{2 \Vert \boldsymbol{y}_{1, m} \Vert_{\bm{V}_{\bm{p}}^{-1}}^2}{\left ( \boldsymbol{y}_{1, m}^\top \boldsymbol{\theta}^{} - \varepsilon \right )^2} \right\} \nonumber\\
    &= \min_{\bm{p} \in S_K} \max_{r \leq R_{\textup{upper}}} \max_{i \in G_r^\prime \cap G_\varepsilon} \max_{\substack{j \in \mathcal{A}_I(r-1) \\ j \neq i}} \max_{m \in B_r^\prime \cap G_\varepsilon^c} \max \left\{ \frac{2 \Vert \boldsymbol{y}_{i, j} \Vert_{\bm{V}_{\bm{p}}^{-1}}^2}{\left ( \boldsymbol{y}_{i, j}^\top \boldsymbol{\theta}^{} + \varepsilon \right )^2}, \frac{2 \Vert \boldsymbol{y}_{1, m} \Vert_{\bm{V}_{\bm{p}}^{-1}}^2}{\left ( \boldsymbol{y}_{1, m}^\top \boldsymbol{\theta}^{} - \varepsilon \right )^2} \right\} \nonumber\\
    &\geq \min_{\bm{p} \in S_K} \max_{r \leq R_{\textup{upper}}} \max_{i \in G_r^\prime \cap G_\varepsilon} \max_{\substack{j \in \mathcal{A}_I(r-1) \\ j \neq i}} \max_{m \in B_r^\prime \cap G_\varepsilon^c} \max \left\{ \frac{2 \Vert \boldsymbol{y}_{i, j} \Vert_{\bm{V}_{\bm{p}}^{-1}}^2}{\left (4\varepsilon_r \right )^2}, \frac{2 \Vert \boldsymbol{y}_{1, m} \Vert_{\bm{V}_{\bm{p}}^{-1}}^2}{\left ( 4\varepsilon_r \right )^2} \right\} \nonumber\\
    &\stackrel{\text{(i)}}{\geq} \frac{1}{R_{\textup{upper}}} \min_{\bm{p} \in S_K} \sum_{r=1}^{R_{\textup{upper}}} \max_{i \in G_r^\prime \cap G_\varepsilon} \max_{\substack{j \in \mathcal{A}_I(r-1) \\ j \neq i}} \max_{m \in B_r^\prime \cap G_\varepsilon^c} \max \left\{ \frac{2 \Vert \boldsymbol{y}_{i, j} \Vert_{\bm{V}_{\bm{p}}^{-1}}^2}{\left (4\varepsilon_r \right )^2}, \frac{2 \Vert \boldsymbol{y}_{1, m} \Vert_{\bm{V}_{\bm{p}}^{-1}}^2}{\left ( 4\varepsilon_r \right )^2} \right\} \nonumber\\
    &\stackrel{\text{(ii)}}{\geq} \frac{1}{R_{\textup{upper}}} \sum_{r=1}^{R_{\textup{upper}}} 2^{2r-3} \min_{\bm{p} \in S_K} \max_{i \in G_r^\prime \cap G_\varepsilon} \max_{\substack{j \in \mathcal{A}_I(r-1) \\ j \neq i}} \max_{m \in B_r^\prime \cap G_\varepsilon^c} \max \left\{ \Vert \boldsymbol{y}_{i, j} \Vert_{\bm{V}_{\bm{p}}^{-1}}^2, \Vert \boldsymbol{y}_{1, m} \Vert_{\bm{V}_{\bm{p}}^{-1}}^2 \right\} \nonumber\\
    &\stackrel{\text{(iii)}}{\geq} \frac{1}{R_{\textup{upper}}} \sum_{r=1}^{R_{\textup{upper}}} 2^{2r-3} \min_{\bm{p} \in S_K} \max_{i \in \mathcal{A}_I(r-1) \cap G_\varepsilon \cap G_r^c} \max_{\substack{j \in \mathcal{A}_I(r-1) \\ j \neq i}} \max_{m \in \mathcal{A}_I(r-1) \cap G_\varepsilon^c} \max \left\{ \Vert \boldsymbol{y}_{i, j} \Vert_{\bm{V}_{\bm{p}}^{-1}}^2, \Vert \boldsymbol{y}_{1, m} \Vert_{\bm{V}_{\bm{p}}^{-1}}^2 \right\} \nonumber\\
    &\stackrel{\text{(iv)}}{\geq} \frac{{\mathcal{G}_{\mathcal{Y}}}^2 L_2}{d L_1} \frac{1}{R_{\textup{upper}}} \sum_{r=1}^{R_{\textup{upper}}} 2^{2r-3} \min_{\bm{p} \in S_K} \max_{i \in \mathcal{A}_I(r-1) \cap G_\varepsilon \backslash \{ 1 \}} \max_{m \in \mathcal{A}_I(r-1) \cap G_\varepsilon^c} \max \left\{ \Vert \boldsymbol{y}_{1, i} \Vert_{\bm{V}_{\bm{p}}^{-1}}^2, \Vert \boldsymbol{y}_{1, m} \Vert_{\bm{V}_{\bm{p}}^{-1}}^2 \right\} \nonumber\\
    &\stackrel{\text{(v)}}{\geq} \frac{{\mathcal{G}_{\mathcal{Y}}}^2 L_2}{4R_{\textup{upper}} d L_1}  \sum_{r=1}^{R_{\textup{upper}}} 2^{2r-3} \min_{\bm{p} \in S_K} \max_{\substack{i, j \in \mathcal{A}_I(r-1) \\ i \neq j}} \Vert \boldsymbol{y}_{i, j} \Vert_{\bm{V}_{\bm{p}}^{-1}}^2 \nonumber\\
    &= \frac{{\mathcal{G}_{\mathcal{Y}}}^2 L_2}{32R_{\textup{upper}} d L_1}  \sum_{r=1}^{R_{\textup{upper}}} 2^{2r} g_{\mathcal{XY}}(\mathcal{Y}(\mathcal{A}(r-1))), \label{equation: relationship between the lower bound and the upper bound}
\end{align}
where (i) follows from the fact that the maximum of positive numbers is always greater than or equal to their average, and (ii) uses the fact that the minimum of a sum is greater than or equal to the sum of the minimums. (iii) arises from the set inclusion relationships established in Lemma~\ref{claim_add_1} and Lemma~\ref{claim_add_2}. (iv) is a direct consequence of Lemma~\ref{lemma: inequality iv} with $j=1$. Finally, for (v), note that for any $i,j \in \mathcal{A}_I(r-1)$, we have $\max_{i, j \in \mathcal{A}_I(r-1), i \neq j} \Vert \boldsymbol{y}_{i, j} \Vert_{\bm{V}_{\bm{p}}^{-1}}^2 \leq 4 \max_{i \in \mathcal{A}_I(r-1)\backslash \{ 1 \}} \Vert \boldsymbol{y}_{1, i} \Vert_{\bm{V}_{\bm{p}}^{-1}}^2$.
\hfill\Halmos
\endproof

Moreover, to provide additional insights into the G-optimal design, from equation~\eqref{middle result_refined}, we obtain
\begin{align}
    \mathbb{E}_{\boldsymbol{\mu}}\left[T \mid \mathcal{E}_1\right] &\leq 4 \log \left[ \frac{2K}{\delta}\left( R_{\textup{upper}} + 1 \right) \right] \sum_{r=1}^{{R_{\textup{upper}}}} d 2^{2r} + \frac{d(d+1)}{2} R_{\textup{upper}}.\label{upper bound: additional insights}
\end{align}

From inequalities~(\ref{equation: relationship between the lower bound and the upper bound}) and~(\ref{upper bound: additional insights}), it follows that to align the upper and lower bounds, one must establish $d \leq c \min_{\bm{p} \in S_K} \max_{i, j \in \mathcal{A}_I(r-1), i \neq j} \Vert \boldsymbol{y}_{i, j} \Vert_{\bm{V}_{\bm{p}}^{-1}}^2$ for some universal constant. However, applying the Kiefer–Wolfowitz Theorem from Lemma \ref{theorem_KW} yields only the reverse inequality
\begin{equation}
    \min_{\bm{p} \in S_K} \max_{i, j \in \mathcal{A}_I(r-1), i \neq j} \Vert \boldsymbol{y}_{i, j} \Vert_{\bm{V}_{\bm{p}}^{-1}}^2 \leq 4\min_{\bm{p} \in S_K} \max_{i \in \mathcal{A}_I(r-1)} \Vert \boldsymbol{a}_{i} \Vert_{\bm{V}_{\bm{p}}^{-1}}^2 = 4d.
\end{equation}

This result indicates that LinFACT-G cannot achieve the lower bound established in Theorem~\ref{lower bound: All-epsilon in the linear setting}. 

\section{Proof of Theorem \ref{upper bound: Algorithm 1 XY}}\label{proof of theorem: upper bound of LinFACT XY}
The central idea of this proof is to establish a direct relationship between the lower bound and the key minimax summation terms in the upper bound, thereby enabling the sample complexity to be bounded explicitly in terms of the lower bound term $(\Gamma^\ast)^{-1}$. We first show that the good event $\mathcal{E}_3$ defined below occurs with probability at least $1 - \delta_r$ in each round $r$, where $\delta_r$ denotes the probability that the good event $\mathcal{E}_3$ does not hold in a certain round $r$. By taking the union bound across different rounds, we can complete the proof of Lemma \ref{claim_xy_1} regarding the overall probability of event $\mathcal{E}_3$ occurring, denoted by $\delta$. We then demonstrate that the probability of this good event holding in every round is at least $1 - \delta$. Consequently, we can sum the sample complexity bounds from each round (conditioned on the good event) to obtain the overall bound on the sample complexity.

First, we define the good event $\mathcal{E}_3$: 
\begin{equation}
    \mathcal{E}_3 = \bigcap_{i \in \mathcal{A}_I(r-1)} \bigcap_{\substack{
    j \in \mathcal{A}_I(r-1) \\ j \neq i}} \bigcap_{r \in \mathbb{N}} \left| (\hat{\mu}_j(r) - \hat{\mu}_i(r)) - (\mu_j - \mu_i) \right| \leq 2\varepsilon_r. \label{clean event XY}
\end{equation}

Since arms are sampled according to a preset allocation (\emph{i.e.}, a fixed design), we introduce the following lemma to provide a confidence region for the estimated parameter $\boldsymbol{\theta}$.

\begin{lemma}\label{claim_xy_1}
    Let $\delta \in (0,1)$. Then, it holds that 
    $
      P(\mathcal{E}_3) \geq 1 - \delta.
    $
\end{lemma}
 
\proof\\{\textit{Proof.}}
Since $\hat{\boldsymbol{\theta}}_r$ is a ordinary least squares estimator of $\boldsymbol{\theta}^{}$ and the noise is i.i.d., it follows that $\bm{y}^{\top}\left(\boldsymbol{\theta}^{}-\hat{\boldsymbol{\theta}}_r\right) \text { is }\| \bm{y} \|_{\bm{V}_r^{-1}}^2$-sub-Gaussian for all $\bm{y} \in \mathcal{Y}(\mathcal{A}(r-1))$. Moreover, the guarantees of the rounding procedure ensure that 
\begin{equation}
    \| \bm{y} \|_{\bm{V}_r^{-1}}^2 \leq \left( 1+\epsilon \right) g_{\mathcal{XY}}(\mathcal{Y}(\mathcal{A}(r-1)))/T_r \leq \frac{2^{-2r-1}}{\log \frac{2K(K-1)r(r+1)}{\delta}}
\end{equation}
for all $\bm{y} \in \mathcal{Y}(\mathcal{A}(r-1))$, as ensured by our choice of $T_r$ in equation \eqref{equation_phase budget XY}. Since the right-hand side is deterministic and does not depend on the randomness of the arm rewards, we have that for any $\rho > 0$ and $\bm{y} \in \mathcal{Y}(\mathcal{A}(r-1))$, 
\begin{equation}
    \mathbb{P}\left\{ \vert \bm{y}^{\top}\left(\boldsymbol{\theta}^{}-\hat{\boldsymbol{\theta}}_r\right) \vert > \sqrt{2^{-2r} \frac{\log(2/\rho)}{\log \frac{2K(K-1)r(r+1)}{\delta}}} \right\} \leq \rho.
\end{equation}
Letting $\rho = \frac{\delta}{K(K-1)r(r+1)}$ and applying a union bound over all possible $\bm{y} \in \mathcal{Y}(\mathcal{A}(r-1))$, where $\vert \mathcal{Y}(\mathcal{A}(r-1)) \vert \leq \vert \mathcal{Y}(\mathcal{A}(0)) \vert \leq K(K-1)$, we obtain the desired probability guarantee:
\begin{align}
    \mathbb{P}(\mathcal{E}_3^c) &= \mathbb{P} \left\{ \bigcup_{i \in \mathcal{A}_I(r-1)} \bigcup_{\substack{
    j \in \mathcal{A}_I(r-1) \\ j \neq i}} \bigcup_{r \in \mathbb{N}} \left| (\hat{\mu}_j(r) - \hat{\mu}_i(r)) - (\mu_j - \mu_i) \right| > \varepsilon_r \right\}
    \nonumber\\ &\leq \sum_{r=1}^{\infty} \mathbb{P} \left\{ \bigcup_{i \in \mathcal{A}_I(r-1)} \bigcup_{\substack{
    j \in \mathcal{A}_I(r-1) \\ j \neq i}} \left| (\hat{\mu}_j(r) - \hat{\mu}_i(r)) - (\mu_j - \mu_i) \right| > \varepsilon_r \right\}
    \nonumber\\ &\leq \sum_{r=1}^{\infty} \sum_{i=1}^{K} \sum_{\substack{
    j = 1 \\ j \neq i}}^{K} \frac{\delta}{K(K-1)r(r+1)}\nonumber\\
    &= \delta.
\end{align}
Taking the union bound over all rounds $r \in \mathbb{N}$ completes the proof. 
\hfill\Halmos
\endproof

Thus, by the standard result of the $\mathcal{XY}$-optimal design, we have
\begin{equation}
    C_{\delta/K}(r) \coloneqq \varepsilon_r,
\end{equation}
which matches the expression in equation~\eqref{eq_C_G} for the G-optimal design. 

\begin{lemma}\label{claim_xy_2}
    On the event $\mathcal{E}_3$, the best arm $1 \in \mathcal{A}_I(r)$ for all $r \in \mathbb{N}$.
\end{lemma}

\proof\\{\textit{Proof.}}
If the event $\mathcal{E}_3$ holds, then for any arm $i \in \mathcal{A}_I(r-1)$, we have
\begin{equation}
    \hat{\mu}_i(r) - \hat{\mu}_1(r) \leq \mu_i - \mu_1 + 2\varepsilon_r \leq 2\varepsilon_r < 2\varepsilon_r + \varepsilon,
\end{equation}
which implies that $\hat{\mu}_1(r) + \varepsilon_r > \max_{i \in \mathcal{A}_I(r-1)} \hat{\mu}_i - \varepsilon_r - \varepsilon = L_r$ and $\hat{\mu}_1(r) + \varepsilon_r \geq \max_{i \in \mathcal{A}_I(r-1)} \hat{\mu}_i (r) - C_{\delta/K}(r)$.

These inequalities ensure that arm 1 will not be eliminated from $\mathcal{A}_I(r)$ in LinFACT.
\hfill\Halmos
\endproof

\begin{lemma}\label{claim_xy_3}
    With probability at least $1 - \delta$, and employing an $\varepsilon$-efficient rounding procedure, LinFACT-$\mathcal{X} \mathcal{Y}$ successfully identifies all $\varepsilon$-best arms and achieves instance-optimal sample complexity up to logarithmic factors, as given by
\begin{equation}
    T \leq c \left[d R_{\textup{upper}} \log \left( \frac{2K(R_{\textup{upper}}+1)}{\delta} \right) \right] (\Gamma^\ast)^{-1} + q\left( \epsilon \right) R_{\textup{upper}},
\end{equation}
where $c$ is a universal constant, $R_{\textup{upper}} = \max \left\{ \left\lceil \log_2 \frac{4}{\alpha_\varepsilon} \right\rceil, \left\lceil \log_2 \frac{4}{\beta_\varepsilon} \right\rceil \right\}$. 
\end{lemma}

\proof\\{\textit{Proof.}}
Combining the result of Lemma~\ref{claim_xy_1}, we conclude that with probability at least $1 - \delta$
\begin{align}
    T &\leq \sum_{r=1}^{R_{\textup{upper}}} \max \left\{ \left\lceil \frac{2 g_{\mathcal{XY}}\left( \mathcal{Y}(\mathcal{A}(r-1)) \right) (1+\epsilon)}{\varepsilon_r^2} \log \left( \frac{2K(K-1)r(r+1)}{\delta} \right) \right\rceil, q\left( \epsilon \right) \right\} \nonumber\\
    &\leq \sum_{r=1}^{R_{\textup{upper}}} 2 \cdot 2^{2r} g_{\mathcal{XY}}\left( \mathcal{Y}(\mathcal{A}(r-1)) \right) (1+\epsilon) \log \left( \frac{2K(K-1)r(r+1)}{\delta} \right) + (1 + q\left( \epsilon \right))R_{\textup{upper}} \nonumber\\
    &\leq \left[ 64\left( 1+\epsilon \right) \log \left( \frac{2K(K-1)R_{\textup{upper}}(R_{\textup{upper}}+1)}{\delta} \right) \frac{R_{\textup{upper}}d L_1}{{\mathcal{G}_{\mathcal{Y}}}^2 L_2} \right] (\Gamma^\ast)^{-1} + (1 + q\left( \epsilon \right))R_{\textup{upper}} \nonumber\\
    &\leq \left[ 128\left( 1+\epsilon \right) \log \left( \frac{2K(R_{\textup{upper}}+1)}{\delta} \right) \frac{R_{\textup{upper}}d L_1}{{\mathcal{G}_{\mathcal{Y}}}^2 L_2} \right] (\Gamma^\ast)^{-1} + (1 + q\left( \epsilon \right))R_{\textup{upper}} \nonumber\\
    &\leq c \left[ d R_{\textup{upper}} \log \left( \frac{2K(R_{\textup{upper}}+1)}{\delta} \right) \right] (\Gamma^\ast)^{-1} + q\left( \epsilon \right) R_{\textup{upper}},
\end{align}
where $c$ is a universal constant, $R_{\textup{upper}} = \max \left\{ \left\lceil \log_2 \frac{4}{\alpha_\varepsilon} \right\rceil, \left\lceil \log_2 \frac{4}{\beta_\varepsilon} \right\rceil \right\}$, and the third inequality follows from equation (\ref{equation: relationship between the lower bound and the upper bound}).

Then, let $\xi = \min(\alpha_\varepsilon, \beta_\varepsilon)/16$ be the minimum gap of the problem instance. Considering that the approximation error term $q(\epsilon)$ is in the form of $\mathcal{O}(\frac{d}{\epsilon^2})$ \citep{allen2021near, fiez2019sequential}, we have
\begin{equation}
    \mathbb{E}[T_{\mathcal{XY}} \mid \mathcal{E}] 
    = \mathcal{O} 
        \left(\frac{d}{\Gamma^\ast} \xi^{-1} \log (\xi^{-1}) \log\left( \frac{K}{\delta} \log(\xi^{-2}) \right) 
        + \frac{d}{\epsilon^2} \log(\xi^{-1})\right).
\end{equation}
\hfill\Halmos
\endproof

\section{Proof of Theorem \ref{upper bound: Algorithm 1 G_mis_1}}\label{sec_Proof of upper bound: Algorithm 1 G_mis_1}
\subsection{Step 1: Define the Clean Events}\label{subsec_Rearrange the Clean Event With Misspecification}
The core of the proof lies in similarly defining the round at which all classifications are completed under the misspecified model. To this end, we reconstruct the anytime confidence radius for each arm in round $r$ and redefine the high-probability event over the entire execution of the algorithm. We denote this clean event as $\mathcal{E}_{1m}$.
\begin{equation}\label{clean event mis_1}
    \mathcal{E}_{1m} = \left\{ \bigcap_{i \in \mathcal{A}_I(r-1)} \bigcap_{r \in \mathbb{N}} \left| \hat{\mu}_i(r) - \mu_i \right| \leq \varepsilon_r + L_m \sqrt{d} \right\}. 
\end{equation}

Similarly, since arms are sampled according to a preset allocation (\emph{i.e.}, the optimal design criterion), we invoke Lemma~\ref{lemma: confidence region for any single fixed arm} to derive an adjusted confidence region for the estimated parameter $\hat{\boldsymbol{\theta}}_t$. When following the G-optimal sampling rule, as specified in lines~\ref{G_sampling_1} and~\ref{G_sampling_2} of the pseudocode, we obtain the following result for each round $r$:
\begin{equation}
    \bm{V}_r = \sum_{\bm{a} \in \text{Supp}(\pi_r)} T_r(\bm{a}) \bm{a} \bm{a}^\top \succeq \frac{2d}{\varepsilon_r^2} \log \left( \frac{2Kr(r+1)}{\delta} \right) \bm{V}(\pi).
\end{equation}

To give the confidence radius, we have the following decomposition.
\begin{align}\label{equation: confidence region_mis}
    \left|\left\langle \hat{\boldsymbol{\theta}}_r - \boldsymbol{\theta}, \bm{a} \right\rangle \right| &= \left| \boldsymbol{a}^\top \bm{V}_r^{-1} \sum_{s=1}^{T_r} m(\boldsymbol{a}_{A_s})\boldsymbol{a}_{A_s} + \boldsymbol{a}^\top \bm{V}_r^{-1} \sum_{s=1}^{T_r} \eta_s \boldsymbol{a}_{A_s} \right| \nonumber \\
    &\leq \left| \boldsymbol{a}^\top \bm{V}_r^{-1} \sum_{s=1}^{T_r} \Delta_m(\boldsymbol{a}_{A_s})\boldsymbol{a}_{A_s} \right| + \left| \boldsymbol{a}^\top \bm{V}_r^{-1} \sum_{s=1}^{T_r} \eta_s \boldsymbol{a}_{A_s} \right|,
\end{align}
where the first term is bounded by
\begin{align}\label{eq_mis_1}
    \left| \boldsymbol{a}^\top \bm{V}_r^{-1} \sum_{s=1}^{T_r} \Delta_m(\boldsymbol{a}_{A_s})\boldsymbol{a}_{A_s} \right| &= \left| \boldsymbol{a}^\top \bm{V}_r^{-1} \sum_{\bm{b} \in \mathcal{A}(r-1)} T_r(\bm{b}) \Delta_m(\bm{b}) \bm{b} \right| \nonumber\\
    &\leq L_m \sum_{\bm{b} \in \mathcal{A}(r-1)} T_r(\bm{b}) \left| \boldsymbol{a}^\top \bm{V}_r^{-1} \bm{b} \right| \nonumber\\
    &\leq L_m \sqrt{\left( \sum_{\bm{b} \in \mathcal{A}(r-1)} T_r(\bm{b}) \right) \boldsymbol{a}^\top \left( \sum_{\bm{b} \in \mathcal{A}(r-1)} T_r(\bm{b}) \bm{V}_r^{-1} \bm{b} \bm{b}^\top \bm{V}_r^{-1} \boldsymbol{a} \right)} \nonumber\\
    &= L_m \sqrt{\sum_{\bm{b} \in \mathcal{A}(r-1)} T_r(\bm{b}) \| \boldsymbol{a} \|_{\bm{V}_r^{-1}}^2} \nonumber\\
    &\leq L_m \sqrt{d},
\end{align}
where the first inequality follows from Hölder's inequality, the second from Jensen’s inequality, and the last from the guarantee of the G-optimal exploration policy, which ensures that $\| \boldsymbol{a} \|_{\bm{V}_r^{-1}}^2 \leq {d}/{T_r}$.

The second term is also bounded using Lemma \ref{lemma: confidence region for any single fixed arm} and the result in Lemma \ref{lemma: inversion reverses loewner orders}. For any arm $\bm{a} \in \mathcal{A}(r-1)$, with a probability of at least $1-\delta/{Kr(r+1)}$, we have
\begin{align}\label{eq_mis_2}
    \left| \boldsymbol{a}^\top \bm{V}_r^{-1} \sum_{s=1}^{T_r} \eta_s \boldsymbol{a}_{A_s} \right| &\leq \sqrt{2\left \lVert \bm{a} \right\rVert^2_{\bm{V}_r^{-1}} \log \left( \frac{2Kr(r+1)}{\delta} \right)} \nonumber\\
    &= \sqrt{2\bm{a}^\top\bm{V}_r^{-1}\bm{a} \log \left( \frac{2Kr(r+1)}{\delta} \right)} \nonumber\\
    &\leq \sqrt{2\bm{a}^\top \left( \frac{\varepsilon_r^2}{2d}\frac{1}{\log \left( \frac{2Kr(r+1)}{\delta} \right)}\bm{V}(\pi)^{-1} \right) \bm{a} \log \left( \frac{2Kr(r+1)}{\delta} \right)} \nonumber\\
    &\leq \varepsilon_r.
\end{align}

Thus, with the standard result of the G-optimal design, we also have
\begin{equation}\label{eq_C_G_3}
    C_{\delta/K}(r) \coloneqq \varepsilon_r.
\end{equation}

To establish the probability guarantee for event $\mathcal{E}_{1m}$, we combine the results from equations \eqref{eq_mis_1} and \eqref{eq_mis_2}, yielding
\begin{align}
    \mathbb{P}(\mathcal{E}_{1m}^c) &= \mathbb{P} \left\{ \bigcup_{i \in \mathcal{A}_I(r-1)} \bigcup_{r \in \mathbb{N}} \left| \hat{\mu}_i(r) - \mu_i \right| > \varepsilon_r + L_m \sqrt{d} \right\}
    \nonumber\\
    &\leq \sum_{r=1}^{\infty} \mathbb{P} \left\{ \bigcup_{i \in \mathcal{A}_I(r-1)} \left| \hat{\mu}_i(r) - \mu_i \right| > \varepsilon_r + L_m \sqrt{d} \right\} \nonumber\\
    &\leq \sum_{r=1}^{\infty} \sum_{i=1}^{K} \frac{\delta}{Kr(r+1)}\nonumber\\    &= \delta.
\end{align}

Therefore, taking the union bounds over rounds $r \in \mathbb{N}$, we have
\begin{equation}
    P(\mathcal{E}_{1m}) \geq 1 - \delta.
\end{equation}

Considering an additional event that characterizes the gaps between different arms, defined as follows
\begin{equation}
    \mathcal{E}_{2m} = \bigcap_{i \in G_{\varepsilon}} \bigcap_{j \in \mathcal{A}_I(r-1)} \bigcap_{r \in \mathbb{N}} \left| (\hat{\mu}_j(r) - \hat{\mu}_i(r)) - (\mu_j - \mu_i) \right| \leq 2\varepsilon_r + 2L_m \sqrt{d}.
\end{equation}

By equation (\ref{equation: confidence region_mis}), for $i, j \in \mathcal{A}_I(r-1)$, we have
\begin{align}\label{confidence: arm filter}
    \mathbb{P}\left\{\left|(\hat{\mu}_j - \hat{\mu}_i) - (\mu_j - \mu_i)\right| > 2\varepsilon_r + 2L_m \sqrt{d} \mid \mathcal{E}_{1m} \right\} &\leq \mathbb{P}\left\{\left|\hat{\mu}_j - \mu_j \right| + \left|\hat{\mu}_i - \mu_i\right| > 2\varepsilon_r + 2L_m \sqrt{d} \mid \mathcal{E}_{1m} \right\} \nonumber\\
    &\leq \mathbb{P}\left\{\left|\hat{\mu}_j - \mu_j \right| > \varepsilon_r + L_m \sqrt{d} \mid \mathcal{E}_{1m} \right\} \nonumber\\
    &+ \mathbb{P}\left\{\left|\hat{\mu}_i - \mu_i\right| > \varepsilon_r + L_m \sqrt{d} \mid \mathcal{E}_{1m} \right\} \nonumber\\
    &= 0,
\end{align}
\noindent which implies
\begin{equation}
    \mathbb{P}(\mathcal{E}_{2m} \mid \mathcal{E}_{1m}) = 1.
\end{equation}

\subsection{Step 2: Bound the Expected Sample Complexity}\label{subsec_Bound the Expected Sample Complexity When the Model Misspecification is Considered}
To bound the expected sample complexity under model misspecification, we aim to identify the round in which all arms $i \in G_\varepsilon$ have been included in $G_r$, and the round in which all arms $i \in G_\varepsilon^c$ have been added to $B_r$. 

\begin{lemma}\label{claim_mis_2_1}
    For $i \in G_\varepsilon$ and $L_m < \frac{\alpha_\varepsilon}{2 \sqrt{d}}$, if $r \geq\left\lceil\log _2\left(\frac{4}{\varepsilon-\Delta_i-2L_m\sqrt{d}}\right)\right\rceil$, then we have $ \mathbb{E}_{\boldsymbol{\mu}}\left[\mathbbm{1}\left[i \notin G_r\right] \mid \mathcal{E}_{1m}\right] = 0$.
\end{lemma}

\proof\\{\textit{Proof.}}
First, for any $i \in G_\varepsilon$
\begin{align}\label{eq_EvG_claim0_mis}
    \mathbb{E}_{\boldsymbol{\mu}}\left[\mathbbm{1}\left[i \notin G_r\right] \mid \mathcal{E}_{1m}\right] &= \mathbb{E}_{\boldsymbol{\mu}}\left[ \mathbbm{1}\left[\max_{j \in \mathcal{A}_I(r-1)} \hat{\mu}_j - \hat{\mu}_i \geq -2\varepsilon_r + \varepsilon\right] \mid \mathcal{E}_{1m}, i \notin G_m (m = \{1, 2, \ldots, r-1\}) \right] \nonumber \\
    &\leq \mathbb{E}_{\boldsymbol{\mu}}\left[ \mathbbm{1}\left[\max_{j \in \mathcal{A}_I(r-1)} \hat{\mu}_j - \hat{\mu}_i \geq -2\varepsilon_r + \varepsilon\right] \mid \mathcal{E}_{1m}\right].
\end{align}

If $i \in G_{r-1}$, then $i \in G_r$ by definition. Otherwise, if $i \notin G_{r-1}$, then under event $\mathcal{E}_{1m}$, for $i \in G_\varepsilon$ and $r \geq\left\lceil\log _2\left(\frac{4}{\varepsilon-\Delta_i-2L_m\sqrt{d}}\right)\right\rceil$, we have
\begin{equation}
    \max_{j \in \mathcal{A}_I(r-1)} \hat{\mu}_j - \hat{\mu}_i \leq \mu_{\arg\max_{j \in \mathcal{A}_I(r-1)}\hat{\mu}_j} - \mu_i +2^{-r+1} + 2L_m \sqrt{d} \leq \Delta_i+2^{-r+1} + 2L_m \sqrt{d} \leq \varepsilon-2\varepsilon_r,
\end{equation}
which implies that $i \in G_r$ by line \ref{add to Gr} of the algorithm. In particular, under event $\mathcal{E}_{1m}$, if $i \notin G_{r-1}$, for all $i \in G_\varepsilon$ and $r \geq\left\lceil\log _2\left(\frac{4}{\varepsilon-\Delta_i-2L_m\sqrt{d}}\right)\right\rceil$, we have
\begin{equation}\label{expectation_mis}
    \mathbb{E}_{\boldsymbol{\mu}}\left[ \mathbbm{1}\left[\max_{j \in \mathcal{A}_I(r-1)} \hat{\mu}_j - \hat{\mu}_i \leq -2\varepsilon_r + \varepsilon\right] \mid i \notin G_{r-1}, \mathcal{E}_{1m} \right] = 1.
\end{equation}

Consequently, $\mathbbm{1}\left[i \notin G_r\right] \mathbbm{1}\left[i \in G_{r-1}\right]=0$. Therefore,
\begin{align}
    &\mathbb{E}_{\boldsymbol{\mu}}\left[ \mathbbm{1}\left[\max_{j \in \mathcal{A}_I(r-1)} \hat{\mu}_j - \hat{\mu}_i \geq -2\varepsilon_r + \varepsilon\right] \mid \mathcal{E}_{1m} \right] \nonumber\\
    &= \mathbb{E}_{\boldsymbol{\mu}}\left[ \mathbbm{1}\left[\max_{j \in \mathcal{A}_I(r-1)} \hat{\mu}_j - \hat{\mu}_i \geq -2\varepsilon_r + \varepsilon\right] \mathbbm{1}\left[i \notin G_{r-1}\right] \mid \mathcal{E}_{1m} \right] \nonumber\\
    &+ \mathbb{E}_{\boldsymbol{\mu}}\left[ \mathbbm{1}\left[\max_{j \in \mathcal{A}_I(r-1)} \hat{\mu}_j - \hat{\mu}_i \geq -2\varepsilon_r + \varepsilon\right] \mathbbm{1}\left[i \in G_{r-1}\right] \mid \mathcal{E}_{1m} \right]\nonumber\\
    &= \mathbb{E}_{\boldsymbol{\mu}}\left[ \mathbbm{1}\left[\max_{j \in \mathcal{A}_I(r-1)} \hat{\mu}_j - \hat{\mu}_i \geq -2\varepsilon_r + \varepsilon\right] \mathbbm{1}\left[i \notin G_{r-1}\right] \mid \mathcal{E}_{1m} \right] \nonumber\\
    &=\mathbb{E}_{\boldsymbol{\mu}}\left[\mathbbm{1}\left[\max_{j \in \mathcal{A}_I(r-1)} \hat{\mu}_j - \hat{\mu}_i \geq -2\varepsilon_r + \varepsilon\right] \mathbbm{1}\left[i \notin G_{r-1}\right] \mid i \notin G_{r-1}, \mathcal{E}_{1m}\right] \mathbb{P}\left(i \notin G_{r-1} \mid \mathcal{E}_{1m}\right) \nonumber\\
    &+ \mathbb{E}_{\boldsymbol{\mu}}\left[\mathbbm{1}\left[\max_{j \in \mathcal{A}_I(r-1)} \hat{\mu}_j - \hat{\mu}_i \geq -2\varepsilon_r + \varepsilon\right] \mathbbm{1}\left[i \notin G_{r-1}\right] \mid i \in G_{r-1}, \mathcal{E}_{1m}\right] \mathbb{P}\left(i \in G_{r-1} \mid \mathcal{E}_{1m}\right) \nonumber\\
    &= \mathbb{E}_{\boldsymbol{\mu}}\left[\mathbbm{1}\left[\max_{j \in \mathcal{A}_I(r-1)} \hat{\mu}_j - \hat{\mu}_i \geq -2\varepsilon_r + \varepsilon\right] \mathbbm{1}\left[i \notin G_{r-1}\right] \mid i \notin G_{r-1}, \mathcal{E}_{1m}\right] \mathbb{P}\left(i \notin G_{r-1} \mid \mathcal{E}_{1m}\right) \nonumber\\
    &= \mathbb{E}_{\boldsymbol{\mu}}\left[\mathbbm{1}\left[\max_{j \in \mathcal{A}_I(r-1)} \hat{\mu}_j - \hat{\mu}_i \geq -2\varepsilon_r + \varepsilon\right] \mid i \notin G_{r-1}, \mathcal{E}_{1m}\right] \mathbb{E}_{\boldsymbol{\mu}}\left[\mathbbm{1}\left[i \notin G_{r-1}\right] \mid \mathcal{E}_{1m}\right] \nonumber\\
    &= 0,
\end{align}
where the second line follows from the additivity of expectation. The fourth line follows the deterministic result that $\mathbbm{1}\left[i \notin G_r\right] \mathbbm{1}\left[i \in G_{r-1}\right]=0$. The fifth line is the decomposition based on the conditional expectation. The eighth line comes from the fact that the expectation of the indicator function is simply the probability. The last line follows the result in equation~\eqref{expectation_mis}. The lemma can thus be concluded together with equation \eqref{eq_EvG_claim0_mis}.

In the perfectly linear model, $\varepsilon - \Delta_i > 0$ always holds for any $i \in G_\varepsilon$. However, under model misspecification, the sign of this term within the logarithm must be verified. To ensure positivity for all $i \in G_\varepsilon$, it is necessary that $\alpha_\varepsilon > 2L_m \sqrt{d}$. As the misspecification magnitude $L_m$ approaches $\alpha_\varepsilon / (2 \sqrt{d})$, the upper bound on the expected sample complexity increases sharply, since the misspecification significantly impairs the identification of $\varepsilon$-best arms. Moreover, when $L_m \geq \alpha_\varepsilon / (2 \sqrt{d})$, the sample complexity can no longer be bounded in this form, which is intuitive and consistent with the general insights discussed earlier in Section~\ref{subsec_General Insights Related to the Misspecified Linear Bandits in the Pure Exploration Setting}.
\hfill\Halmos
\endproof

\begin{lemma}\label{claim_mis_2_2}
    For $i \in G_\varepsilon^c$ and $L_m < \frac{\beta_\varepsilon}{2 \sqrt{d}}$, if $r \geq\left\lceil\log _2\left(\frac{4}{\Delta_i-\varepsilon - 2L_m\sqrt{d}}\right)\right\rceil$, then we have $ \mathbb{E}_{\boldsymbol{\mu}}\left[\mathbbm{1}\left[i \notin B_r\right] \mid \mathcal{E}_{1m}\right] = 0$.
\end{lemma}

\proof\\{\textit{Proof.}}
First, we for any $i \in G_\varepsilon^c$, 
\begin{align}\label{eq_Ev_claim0_2}
    \mathbb{E}_{\boldsymbol{\mu}}\left[\mathbbm{1}\left[i \notin B_r\right] \mid \mathcal{E}_{1m}\right] &= \mathbb{E}_{\boldsymbol{\mu}}\left[ \mathbbm{1}\left[\max_{j \in \mathcal{A}_I(r-1)} \hat{\mu}_j - \hat{\mu}_i \leq 2\varepsilon_r + \varepsilon\right] \mid \mathcal{E}_{1m}, i \notin B_m (m = \{1, 2, \ldots, r-1\}) \right] \nonumber \\
    &\leq \mathbb{E}_{\boldsymbol{\mu}}\left[ \mathbbm{1}\left[\max_{j \in \mathcal{A}_I(r-1)} \hat{\mu}_j - \hat{\mu}_i \leq 2\varepsilon_r + \varepsilon\right] \mid \mathcal{E}_{1m}\right].
\end{align}

If $i \in B_{r-1}$, then $i \in B_r$ by definition. Otherwise, if $i \notin B_{r-1}$, then under event $\mathcal{E}_{1m}$, for $i \in G_\varepsilon^c$ and $r \geq\left\lceil\log _2\left(\frac{4}{\Delta_i-\varepsilon - 2L_m\sqrt{d}}\right)\right\rceil$, we have
\begin{equation}
    \max_{j \in \mathcal{A}_I(r-1)} \hat{\mu}_j - \hat{\mu}_i \geq \Delta_i-2^{-r+1} - 2L_m\sqrt{d} \geq \varepsilon+2\varepsilon_r,
\end{equation}
which implies that $i \in B_r$ by line \ref{elimination: good 1} of the algorithm. In particular, under event $\mathcal{E}_{1m}$, if $i \notin B_{r-1}$, for all $i \in G_\varepsilon^c$ and $r \geq\left\lceil\log _2\left(\frac{4}{\Delta_i-\varepsilon - 2L_m\sqrt{d}}\right)\right\rceil$, we have
\begin{equation}\label{expectation_mis_op}
    \mathbb{E}_{\boldsymbol{\mu}}\left[ \mathbbm{1}\left[\max_{j \in \mathcal{A}_I(r-1)} \hat{\mu}_j - \hat{\mu}_i > 2\varepsilon_r + \varepsilon\right] \mid i \notin B_{r-1}, \mathcal{E}_{1m} \right] = 1.
\end{equation}

Deterministically, $\mathbbm{1}\left[i \notin B_r\right] \mathbbm{1}\left[i \in B_{r-1}\right]=0$. Therefore,
\begin{align}
    &\mathbb{E}_{\boldsymbol{\mu}}\left[ \mathbbm{1}\left[\max_{j \in \mathcal{A}_I(r-1)} \hat{\mu}_j - \hat{\mu}_i \leq 2\varepsilon_r + \varepsilon\right] \mid \mathcal{E}_{1m} \right] \nonumber\\
    &= \mathbb{E}_{\boldsymbol{\mu}}\left[ \mathbbm{1}\left[\max_{j \in \mathcal{A}_I(r-1)} \hat{\mu}_j - \hat{\mu}_i \leq 2\varepsilon_r + \varepsilon\right] \mathbbm{1}\left[i \notin B_{r-1}\right] \mid \mathcal{E}_{1m} \right] \nonumber\\
    &+ \mathbb{E}_{\boldsymbol{\mu}}\left[ \mathbbm{1}\left[\max_{j \in \mathcal{A}_I(r-1)} \hat{\mu}_j - \hat{\mu}_i \leq 2\varepsilon_r + \varepsilon\right] \mathbbm{1}\left[i \in B_{r-1}\right] \mid \mathcal{E}_{1m} \right]\nonumber\\
    &= \mathbb{E}_{\boldsymbol{\mu}}\left[ \mathbbm{1}\left[\max_{j \in \mathcal{A}_I(r-1)} \hat{\mu}_j - \hat{\mu}_i \leq 2\varepsilon_r + \varepsilon\right] \mathbbm{1}\left[i \notin B_{r-1}\right] \mid \mathcal{E}_{1m} \right] \nonumber\\
    &=\mathbb{E}_{\boldsymbol{\mu}}\left[\mathbbm{1}\left[\max_{j \in \mathcal{A}_I(r-1)} \hat{\mu}_j - \hat{\mu}_i \leq 2\varepsilon_r + \varepsilon\right] \mathbbm{1}\left[i \notin B_{r-1}\right] \mid i \notin B_{r-1}, \mathcal{E}_{1m}\right] \mathbb{P}\left(i \notin B_{r-1} \mid \mathcal{E}_{1m}\right) \nonumber\\
    &+ \mathbb{E}_{\boldsymbol{\mu}}\left[\mathbbm{1}\left[\max_{j \in \mathcal{A}_I(r-1)} \hat{\mu}_j - \hat{\mu}_i \leq 2\varepsilon_r + \varepsilon\right] \mathbbm{1}\left[i \notin B_{r-1}\right] \mid i \in B_{r-1}, \mathcal{E}_{1m}\right] \mathbb{P}\left(i \in B_{r-1} \mid \mathcal{E}_{1m}\right) \nonumber\\
    &= \mathbb{E}_{\boldsymbol{\mu}}\left[\mathbbm{1}\left[\max_{j \in \mathcal{A}_I(r-1)} \hat{\mu}_j - \hat{\mu}_i \leq 2\varepsilon_r + \varepsilon\right] \mathbbm{1}\left[i \notin B_{r-1}\right] \mid i \notin B_{r-1}, \mathcal{E}_{1m}\right] \mathbb{P}\left(i \notin B_{r-1} \mid \mathcal{E}_{1m}\right) \nonumber\\
    &= \mathbb{E}_{\boldsymbol{\mu}}\left[\mathbbm{1}\left[\max_{j \in \mathcal{A}_I(r-1)} \hat{\mu}_j - \hat{\mu}_i \leq 2\varepsilon_r + \varepsilon\right] \mid i \notin B_{r-1}, \mathcal{E}_{1m}\right] \mathbb{E}_{\boldsymbol{\mu}}\left[\mathbbm{1}\left[i \notin B_{r-1}\right] \mid \mathcal{E}_{1m}\right] \nonumber\\
    &= 0.
\end{align}
The second line follows from the linearity of expectation. The fourth line uses the deterministic fact that $\mathbbm{1}\left[i \notin B_r\right] \mathbbm{1}\left[i \in B_{r-1}\right] = 0$. The fifth line applies the law of total expectation. The eighth line uses the identity that the expectation of an indicator function equals the corresponding probability. The final line follows from equation~\eqref{expectation_mis_op}. The lemma then follows by combining this result with equation~\eqref{eq_Ev_claim0_2}.

Similarly, we must ensure the positivity of the term inside the logarithm. To guarantee that $\Delta_i - \varepsilon - 2L_m \sqrt{d} > 0$ for every $i \in G_\varepsilon^c$, it is necessary that $\beta_\varepsilon > 2L_m \sqrt{d}$.
\hfill\Halmos
\endproof

\begin{lemma}\label{claim_mis_2_3}
    Suppose $L_m < \min \left\{\frac{\alpha_\varepsilon}{2 \sqrt{d}}, \frac{\beta_\varepsilon}{2 \sqrt{d}} \right\}$, then the round by which all arms are classified is $R_{\text{upper}}^\prime = \max \left\{ \left\lceil \log_2 \frac{4}{\alpha_\varepsilon - 2L_m \sqrt{d}} \right\rceil, \left\lceil \log_2 \frac{4}{\beta_\varepsilon - 2L_m \sqrt{d}} \right\rceil \right\}$ under model misspecification.
\end{lemma}

\proof\\{\textit{Proof.}}
Combining the results of Lemmas~\ref{claim_mis_2_1} and \ref{claim_mis_2_2}, we define an auxiliary round $R_m$ to facilitate the derivation of the upper bound.  Specifically, for $i \in G_\varepsilon^c$ and $r \geq\left\lceil\log _2\left(\frac{4}{\Delta_i-\varepsilon - 2L_m \sqrt{d}}\right)\right\rceil$, we have $\mathbb{E}_{\boldsymbol{\mu}}\left[\mathbbm{1}\left[i \notin B_r\right] \mid \mathcal{E}_{1m}\right] = 0$. Similarly, for $i \in G_\varepsilon$ and $r \geq\left\lceil\log _2\left(\frac{4}{\varepsilon-\Delta_i - 2L_m \sqrt{d}}\right)\right\rceil$, we have $\mathbb{E}_{\boldsymbol{\mu}}\left[\mathbbm{1}\left[i \notin G_r\right] \mid \mathcal{E}_{1m}\right] = 0$.

Noting that $\alpha_\varepsilon = \min_{i \in G_\varepsilon}(\varepsilon - \Delta_i)$ and $\beta_\varepsilon = \min_{i \in G_\varepsilon^c}(\Delta_i - \varepsilon)$, it follows that for any round $r \geq R_{\text{upper}}^\prime$, all arms have been included in either $G_r$ or $B_r$, marking the termination of the algorithm. 

\hfill\Halmos
\endproof

\begin{lemma}\label{claim_mis_2_4}
    Under model specification, for the expected sample complexity conditioned on the high-probability event $\mathcal{E}_{1m}$, we have
\begin{align}
    \mathbb{E}_{\boldsymbol{\mu}}\left[T_{G_{\textup{mis}}} \mid \mathcal{E}_{1m}\right] &\leq c \max\left\{ \frac{256d}{(\alpha_\varepsilon - 2L_m \sqrt{d})^2} \log \left( \frac{2K}{\delta} \log_2 \frac{16}{\alpha_\varepsilon - 2L_m \sqrt{d}} \right), \right. \nonumber\\
    &\left. \quad \frac{256d}{(\beta_\varepsilon - 2L_m \sqrt{d})^2} \log \left( \frac{2K}{\delta} \log_2 \frac{16}{\beta_\varepsilon - 2L_m \sqrt{d}} \right) \right\} + \frac{d(d+1)}{2} R_{\text{upper}}^\prime, \label{eq_mis_sample_complexity}
\end{align}
where $c$ is a universal constant and $R_{\text{upper}}^\prime = \max \left\{ \left\lceil \log_2 \frac{4}{\alpha_\varepsilon - 2L_m \sqrt{d}} \right\rceil, \left\lceil \log_2 \frac{4}{\beta_\varepsilon - 2L_m \sqrt{d}} \right\rceil \right\}$.
\end{lemma}

\proof\\{\textit{Proof.}}
We can also decompose $T$ as in equation~\eqref{equation: decomposition of T}, where all expectations are conditioned on the high-probability event $\mathcal{E}_{1m}$, as given by
\begin{align}
    \mathbb{E}_{\boldsymbol{\mu}}\left[T_{G_{\textup{mis}}} \mid \mathcal{E}_{1m}\right]
    &\leq \sum_{r=1}^{\infty} \mathbb{E}_{\boldsymbol{\mu}}\left[\mathbbm{1}\left[G_r \cup B_r \neq[K]\right] \mid \mathcal{E}_{1m}\right] \sum_{\bm{a} \in \mathcal{A}(r-1)}T_r(\bm{a}) \nonumber\\
    &\leq \sum_{r=1}^{R_{\text{upper}}^\prime} \left( d 2^{2r+1} \log \left(\frac{2K r(r+1)}{\delta}\right)+\frac{d(d+1)}{2}\right) \nonumber\\
    &\leq \frac{d(d+1)}{2} R_{\text{upper}}^\prime + 2d\log \left(\frac{2K}{\delta}\right) \sum_{r=1}^{{R_{\text{upper}}^\prime}} 2^{2r} + 4d \sum_{r=1}^{{R_{\text{upper}}^\prime}} 2^{2r} \log (r+1) \nonumber\\
    &\leq 4\log \left[ \frac{2K}{\delta}\left( R_{\text{upper}}^\prime + 1 \right) \right] \sum_{r=1}^{{R_{\text{upper}}^\prime}} d 2^{2r} + \frac{d(d+1)}{2} R_{\text{upper}}^\prime \nonumber\\
    &\leq c \max\left\{ \frac{256d}{(\alpha_\varepsilon - 2L_m \sqrt{d})^2} \log \left( \frac{2K}{\delta} \log_2 \frac{16}{\alpha_\varepsilon - 2L_m \sqrt{d}} \right), \right. \nonumber\\
    &\left. \quad \frac{256d}{(\beta_\varepsilon - 2L_m \sqrt{d})^2} \log \left( \frac{2K}{\delta} \log_2 \frac{16}{\beta_\varepsilon - 2L_m \sqrt{d}} \right) \right\} + \frac{d(d+1)}{2} R_{\text{upper}}^\prime. \label{another angle: decomposition of T - 1_mis_1}
\end{align}

Then, let $\xi = \min \left(\alpha_\varepsilon - 2L_m\sqrt{d}, \beta_\varepsilon - 2L_m\sqrt{d}\right)/16$, we have
\begin{align}
\mathbb{E}_{\boldsymbol{\mu}}\left[T_{G_{\textup{mis}}} \mid \mathcal{E}_{1m}\right] = \mathcal{O} 
        \left(d \xi^{-2} \log\!\left( \frac{K}{\delta} \log(\xi^{-2}) \right) 
        + d^{2} \log(\xi^{-1})\right).
    \end{align}
\hfill\Halmos
\endproof

\section{Proof of Theorem \ref{upper bound: Algorithm 1 G_mis_2}}\label{sec_Proof of upper bound: Algorithm 1 G_mis_2}
\subsection{Step 1: Confidence Radius}\label{subsec_Confidence Radius With Misspecification}
In equation~\eqref{equation: confidence region_mis}, the first term—arising from the unknown model misspecification—is unavoidable without prior knowledge. However, rather than focusing on the distance between the true parameter $\bm{\theta}$ and its estimator $\hat{\bm{\theta}}_t$, we instead compare the orthogonal projection $\bm{\theta}_t$ with $\hat{\bm{\theta}}_t$ in the direction of $\bm{x} \in \mathbb{R}^d$.

Building on the definition of the empirically optimal vector $\hat{\boldsymbol{\mu}}_{o}(r)$, with $(\hat{\bm{\theta}}_o(r), \hat{\bm{\Delta}}_{mo}(r))$ as its associated solution, and the orthogonal parameterization $(\bm{\theta}_r, \boldsymbol{\Delta}_m(r))$, we derive the confidence radius for each arm via the following decomposition. For any $i \in \mathcal{A}_I(r-1)$, let $\hat{\mu}_i(r) = \hat{\mu}_{o,i}(r)$ denote the value of the optimal estimator on arm $i$ in round $r$, then we have
\begin{align}
    \left \lvert \hat{\mu}_i(r) - \mu_i \right\rvert &= \left \lvert\left\langle \hat{\bm{\theta}}_o(r) - \bm{\theta}, \bm{a}_i \right\rangle + \hat{\bm{\Delta}}_{mo}(r) - \Delta_{mi} \right\rvert \nonumber\\
    &\leq \left \lvert\left\langle \hat{\bm{\theta}}_o(r) - \bm{\theta}_r, \bm{a}_i \right\rangle \right\rvert + \left \lvert\left\langle \bm{\theta}_r - \bm{\theta}, \bm{a}_i \right\rangle \right\rvert + \left \lvert \hat{\bm{\Delta}}_{mo}(r) - \Delta_{mi} \right\rvert,
\end{align}
where the third term is bounded by definition as $\left \lvert \hat{\bm{\Delta}}_{mo}(r) - \Delta_{mi} \right\rvert \leq 2L_m$, while the first two terms can be bounded using the auxiliary lemmas provided below. 

\begin{lemma}\label{lemma_para_distance}
Let $r$ be any round such that $\bm{V}_r$ is invertible. Consider the orthogonal parameterization $(\bm{\theta}_r, \boldsymbol{\Delta}_m(r)$ for $\boldsymbol{\mu} = \bm{\Psi}\bm{\theta} + \boldsymbol{\Delta}_m$ with $\|\bm{\Delta_m}\|_\infty \leq L_m$. Then
\begin{equation}
    \|\bm{\theta}_r - \bm{\theta}\|_{\bm{V}_r} \leq L_m \sqrt{T_r},
\end{equation}
where $T_r$ is defined in equation \eqref{equation_phase budget G_mis_ortho}.
\end{lemma}
\proof\\{\textit{Proof.}}
We use the expression $\bm{\theta}_r = \bm{\theta} + \bm{V}_r^{-1} \bm{\Psi}^\top \bm{D}_{\bm{N}_r} \boldsymbol{\Delta}_m$ derived above. Let $\bm{P}_{\bm{N}_r} = \bm{\Psi}_{\bm{N}_r} \left( \bm{\Psi}_{\bm{N}_r}^\top \bm{\Psi}_{\bm{N}_r} \right)^\dagger \bm{\Psi}_{\bm{N}_r}^\top$ be a projection, we have
\begin{align}
    \|\bm{\theta}_r - \bm{\theta}\|_{\bm{V}_r} &= \|\bm{V}_r^{-1} \bm{\Psi}^\top \bm{D}_{\bm{N}_r} \bm{\Delta_m}\|_{\bm{V}_r} \nonumber\\ 
    &= \sqrt{\bm{\Delta_m}^\top \bm{D}_{\bm{N}_r} \bm{\Psi} \bm{V}_r^{-1} \bm{\Psi}^\top \bm{D}_{\bm{N}_r} \bm{\Delta_m}} \nonumber\\
    &= \left\|\bm{D}_{\bm{N}_r}^{1/2} \bm{\Delta_m}\right\|_{\bm{P}_{\bm{N}_r}} \nonumber\\
    &\leq \left\|\bm{D}_{\bm{N}_r}^{1/2} \bm{\Delta_m}\right\| \nonumber\\
    &= \left\|\bm{\Delta_m}\right\|_{D_{\bm{N}_r}} \nonumber\\
    &\leq L_m \sqrt{T_r} .
\end{align}
\hfill\Halmos
\endproof

\begin{lemma}\label{lemma_optimi_distance}
(\citeec{reda2021dealing}, Lemma 10). Let $\hat{\boldsymbol{\mu}}_o(r) = \bm{\Psi}\hat{\bm{\theta}}_o(r) + \hat{\bm{\Delta}}_{mo}(r)$ in round $r$, where $(\hat{\bm{\theta}}_o(r), \hat{\bm{\Delta}}_{mo}(r))$ are the solution of \eqref{eq_optimization_esti}.Then the following relationship holds.
\begin{equation}
    \left\| \hat{\bm{\theta}}_o(r) - \bm{\theta}_r \right\|_{\bm{V}_r}^2 \leq \left\| \hat{\bm{\theta}}_r - \bm{\theta}_r \right\|_{\bm{V}_r}^2.
\end{equation}
\end{lemma}

\begin{lemma}\label{lemma_confidence_C_epsilon}
(\citeec{lattimore2020bandit}, Section 20). Let $\delta \in (0, 1)$. Then, with a probability of at least $1-\delta$, it holds that for all $t \in \mathbb{N}$,
\begin{equation}\label{log6_1}
     \left\| \bm{\hat{\theta}}_t - \bm{\theta} \right\|_{V_t} < 2 \sqrt{2 \left( d \log(6) + \log \left( \frac{1}{\delta} \right) \right)}.
\end{equation}
\end{lemma}

Equation~\eqref{log6_1} is not directly applicable in our setting due to the deviation term introduced by model misspecification, as shown in equation~\eqref{equation: confidence region_mis}. However, by leveraging the orthogonal parameterization, we obtain 
\begin{equation}\label{eq_self-normalized}
    \Vert \hat{\bm{\theta}}_r - \bm{\theta}_r \Vert_{\bm{V}_r}^2 = \Vert \bm{V}_r^{-1} \Psi^\top S_r \Vert_{\bm{V}_r}^2 = \Vert \Psi^\top S_r \Vert_{\bm{V}_r^{-1}}^2,
\end{equation}
where $S_r = \sum_{s=1}^{T_r} X_{A_s} - \mu_{A_s}$ is the standard self-normalized quantity in the linear bandit literature, allowing existing techniques—such as various concentration inequalities—to be applied directly in the presence of model misspecification. This observation leads to the conclusion that, under misspecification, for any round $r \in \mathbb{N}$, it holds with probability at least $1 - \delta$ that
\begin{equation}\label{log6_2}
     \left\| \bm{\hat{\theta}}_r - \bm{\theta}_r \right\|_{V_t} < 2 \sqrt{2 \left( d \log(6) + \log \left( \frac{1}{\delta} \right) \right)}.
\end{equation}
This result serves as the basis for designing the sampling budget and conducting theoretical analyses. 

Together with Lemmas~\ref{lemma_para_distance}, \ref{lemma_optimi_distance}, and \ref{lemma_confidence_C_epsilon}, the distance $\left| \tilde{\mu}_i(r) - \mu_i \right|$ can be bounded as follows: with probability at least $1 - \delta / (K r (r+1))$, we have
\begin{align}
    \left \lvert \tilde{\mu}_i(r) - \mu_i \right\rvert &\leq \left \lvert\left\langle \hat{\bm{\theta}}_o(r) - \bm{\theta}_r, \bm{a}_i \right\rangle \right\rvert + \left \lvert\left\langle \bm{\theta}_r - \bm{\theta}, \bm{a}_i \right\rangle \right\rvert + \left \lvert \hat{\bm{\Delta}}_{mo}(r) - \Delta_{mi} \right\rvert \nonumber\\
    &\leq \left \lVert \hat{\bm{\theta}}_o(r) - \bm{\theta}_r \right\rVert_{\bm{V}_r} \left \lVert \bm{a}_i \right\rVert_{\bm{V}_r^{-1}} + \left \lVert \bm{\theta}_r - \bm{\theta} \right\rVert_{\bm{V}_r} \left \lVert \bm{a}_i \right\rVert_{\bm{V}_r^{-1}} + 2L_m \nonumber\\
    &\leq \left \lVert \hat{\bm{\theta}}_r - \bm{\theta}_r \right\rVert_{\bm{V}_r} \left \lVert \bm{a}_i \right\rVert_{\bm{V}_r^{-1}} + \left \lVert \bm{\theta}_r - \bm{\theta} \right\rVert_{\bm{V}_r} \left \lVert \bm{a}_i \right\rVert_{\bm{V}_r^{-1}} + 2L_m \nonumber\\
    &\leq \left( 2 \sqrt{2 \left( d \log(6) + \log \left( \frac{Kr(r+1)}{\delta} \right) \right)} + L_m \sqrt{T_r}  \right) \sqrt{\frac{d}{T_r}} + 2L_m \nonumber\\
    &\leq \varepsilon_r + (\sqrt{d} + 2) L_m.
\end{align}

Then, we define a new clean event
\begin{equation}\label{clean event mis_2}
    \mathcal{E}_{1m}^\prime = \left\{ \bigcap_{i \in \mathcal{A}_I(r-1)} \bigcap_{r \in \mathbb{N}} \left| \hat{\mu}_i(r) - \mu_i \right| \leq \varepsilon_r + (\sqrt{d} + 2) L_m \right\}.
\end{equation}
This mirrors the event defined in equation~\eqref{clean event mis_1}, allowing us to derive all corresponding results in Section~\ref{sec_Proof of upper bound: Algorithm 1 G_mis_1}.

\subsection{Step 2: Bound the Expected Sample Complexity}\label{subsec_Bound the Expected Sample Complexity When the Model Misspecification is Considered and the Orthogonal Parameterization is Applied}

\begin{lemma}\label{claim_or_3_1}
   For $i \in G_\varepsilon$ and $L_m < \frac{\alpha_\varepsilon}{2 \left( \sqrt{d} + 2 \right)}$, if $r \geq\left\lceil\log _2\left(\frac{4}{\varepsilon-\Delta_i-2L_m\left( \sqrt{d} + 2 \right)}\right)\right\rceil$, then we have $ \mathbb{E}_{\boldsymbol{\mu}}\left[\mathbbm{1}\left[i \notin G_r\right] \mid \mathcal{E}_{1m}^\prime\right] = 0$. 
\end{lemma}

\begin{lemma}\label{claim_or_3_2}
   For $i \in G_\varepsilon^c$ and $L_m < \frac{\beta_\varepsilon}{2 \left( \sqrt{d} + 2 \right)}$, if $r \geq\left\lceil\log _2\left(\frac{4}{\Delta_i-\varepsilon - 2L_m\left( \sqrt{d} + 2 \right)}\right)\right\rceil$, then we have $ \mathbb{E}_{\boldsymbol{\mu}}\left[\mathbbm{1}\left[i \notin B_r\right] \mid \mathcal{E}_{1m}^\prime\right] = 0$. 
\end{lemma}

\begin{lemma}\label{claim_or_3_3}
   For the misspecification magnitude $L_m < \min \left\{\frac{\alpha_\varepsilon}{2 \left( \sqrt{d} + 2 \right)}, \frac{\beta_\varepsilon}{2 \left( \sqrt{d} + 2 \right)} \right\}$, the round $R_{\text{upper}}^{\prime\prime} = \max \left\{ \left\lceil \log_2 \frac{4}{\alpha_\varepsilon - 2L_m \left( \sqrt{d} + 2 \right)} \right\rceil, \left\lceil \log_2 \frac{4}{\beta_\varepsilon - 2L_m \left( \sqrt{d} + 2 \right)} \right\rceil \right\}$ marks the point at which all classifications are completed and the algorithm terminates under model misspecification when using an estimation procedure based on orthogonal parameterization. 
\end{lemma}

The proofs of Lemmas \ref{claim_or_3_1}, \ref{claim_or_3_2}, and \ref{claim_or_3_3} follow similar arguments to those presented in Section~\ref{sec_Proof of upper bound: Algorithm 1 G_mis_1} and are therefore omitted for brevity.

\begin{lemma}\label{claim_or_3_4}
   For the expected sample complexity given the high probability event $\mathcal{E}_{1m}^\prime$, we have
\begin{align}\label{eq_mis_sample_complexity_orthogonal}
    \mathbb{E}_{\boldsymbol{\mu}}\left[T_{G_{\textup{op}}} \mid \mathcal{E}_{1m}^\prime\right] &\leq c \max\bigg\{ \frac{256d}{(\alpha_\varepsilon - 2L_m ( \sqrt{d} + 2 ))^2} \log \left( \frac{K 6^d}{\delta} \log_2 \frac{8}{\alpha_\varepsilon - 2L_m ( \sqrt{d} + 2 )} \right),\nonumber\\
    &\frac{256d}{(\beta_\varepsilon - 2L_m ( \sqrt{d} + 2 ))^2} \log \left( \frac{K 6^d}{\delta} \log_2 \frac{8}{\beta_\varepsilon - 2L_m ( \sqrt{d} + 2 )} \right) \bigg\} + \frac{d(d+1)}{2} R_{\text{upper}}^{\prime\prime},
\end{align}
where $c$ is a universal constant and $R_{\text{upper}}^{\prime\prime} = \max \left\{ \left\lceil \log_2 \frac{4}{\alpha_\varepsilon - 2L_m \left( \sqrt{d} + 2 \right)} \right\rceil, \left\lceil \log_2 \frac{4}{\beta_\varepsilon - 2L_m \left( \sqrt{d} + 2 \right)} \right\rceil \right\}$ denotes the round in which all classifications are completed and the algorithm terminates under model misspecification with orthogonal parameterization. 
\end{lemma}

\proof\\{\textit{Proof.}}
The decomposition of $T$ in equation~(\ref{equation: decomposition of T}) can be reformulated, where the expectations are conditioned on the high-probability event $\mathcal{E}_{1m}^\prime$, given by
\begin{align}
    \mathbb{E}_{\boldsymbol{\mu}}\left[T_{G_{\textup{op}}} \mid \mathcal{E}_{1m}^\prime \right]
    &\leq \sum_{r=1}^{\infty} \mathbb{E}_{\boldsymbol{\mu}}\left[\mathbbm{1}\left[G_r \cup B_r \neq[K]\right] \mid \mathcal{E}_{1m}^\prime \right] \sum_{\bm{a} \in \mathcal{A}(r-1)}T_r(\bm{a}) \nonumber\\
    &\leq \sum_{r=1}^{R_{\text{upper}}^{\prime\prime}} \left( d2^{2r+3} \left( d\log(6) + \log\left(\dfrac{Kr(r + 1)}{\delta}\right) \right) + \frac{d(d+1)}{2}\right) \nonumber\\
    &\leq \frac{d(d+1)}{2} R_{\text{upper}}^{\prime\prime} + 8d\log \left(\frac{K 6^d}{\delta}\right) \sum_{r=1}^{{R_{\text{upper}}^{\prime\prime}}} 2^{2r} + 16d \sum_{r=1}^{{R_{\text{upper}}^{\prime\prime}}} 2^{2r} \log (r+1) \nonumber\\
    &\leq 16\log \left[ \frac{K 6^d}{\delta}\left( R_{\text{upper}}^{\prime\prime} + 1 \right) \right] \sum_{r=1}^{{R_{\text{upper}}^{\prime\prime}}} d 2^{2r} + \frac{d(d+1)}{2} R_{\text{upper}}^{\prime\prime} \\
    &\leq c \max\bigg\{ \frac{256d}{(\alpha_\varepsilon - 2L_m ( \sqrt{d} + 2 ))^2} \log \left( \frac{K 6^d}{\delta} \log_2 \frac{16}{\alpha_\varepsilon - 2L_m ( \sqrt{d} + 2 )} \right),\nonumber\\
    &\frac{256d}{(\beta_\varepsilon - 2L_m ( \sqrt{d} + 2 ))^2} \log \left( \frac{K 6^d}{\delta} \log_2 \frac{16}{\beta_\varepsilon - 2L_m ( \sqrt{d} + 2 )} \right) \bigg\} + \frac{d(d+1)}{2} R_{\text{upper}}^{\prime\prime},\label{another angle: decomposition of T - 1_mis_ortho}
\end{align}
where $c$ is a universal constant. Then, let $\xi = \min \left(\alpha_\varepsilon - 2L_m(\sqrt{d}+2), \beta_\varepsilon - 2L_m(\sqrt{d}+2)\right)/16$, we have
\begin{align}
        \mathbb{E}_{\boldsymbol{\mu}}\left[T_{G_{\textup{op}}} \mid \mathcal{E}\right] = \mathcal{O} 
        \left(d \xi^{-2} \log \left( \frac{K6^d}{\delta} \log(\xi^{-1}) \right) 
        + d^{2} \log(\xi^{-1})\right).
    \end{align}
\hfill\Halmos
\endproof

\section{Proof of Proposition \ref{propos_prior}}\label{sec_Proof of propos_prior}
The proof of this proposition closely follows that of Theorem~\ref{upper bound: Algorithm 1 G_mis_1} in Section~\ref{sec_Proof of upper bound: Algorithm 1 G_mis_1}, with the only modification being the definition of $C_{\delta/K}(r)$, which is now set to $\varepsilon_r + L_m \sqrt{d}$ given that $L_m$ is known in advance. The conclusions in Section~\ref{subsec_Rearrange the Clean Event With Misspecification} remain valid. Additionally, the marginal round in Lemma~\ref{claim_mis_2_1} is updated from $\left\lceil\log _2\left(\frac{4}{\varepsilon-\Delta_i-2L_m\sqrt{d}}\right)\right\rceil$ to $\left\lceil\log _2\left(\frac{4}{\varepsilon-\Delta_i}\right)\right\rceil$. The remainder of the proof follows directly by applying the same reasoning steps. 

\section{Proof of Theorem \ref{upper bound: Algorithm 1 GLM}}\label{proof of theorem: upper bound of LinFACT GLM}
\subsection{Step 1: Rearrange the Clean Event}\label{subsubsec_Step 1: Rearrange the Clean Event With GLM}
Following the derivation approach in Section~\ref{refined version}, the core of the proof is to similarly identify the round in which all classifications are completed under the GLM setting. To this end, we reconstruct the anytime confidence radius for the arms in each round $r$ and define a high-probability event that holds throughout the execution of the algorithm. 

Let $\check{\bm{V}}_r = \sum_{s=1}^{T_r} \dot{\mu}_{\text{link}}(\bm{a}^\top\check{\bm{\theta}}_r)\bm{a}_s\bm{a}_s^\top$, where $\check{\bm{\theta}}_r$ is some convex combination of true parameter $\bm{\theta}$ and parameter $\hat{\bm{\theta}}_r$ based on maximum likelihood estimation (MLE). It can be checked that the unweighted matrix $\bm{V}_r = \sum_{s=1}^{T_r} (\bm{a}^\top\check{\bm{\theta}}_r)\bm{a}_s\bm{a}_s^\top$ in the standard linear model is the special case of this newly defined matrix $\check{\bm{V}}_r$ when the inverse link function $\mu_{\text{link}}(x) = x$.

For each arm $i$, we define the following auxiliary vector
\begin{equation}
    W_i = \left( W_{i,1},W_{i,2},\ldots,W_{i,T_r} \right) = \bm{a}_i^\top \check{\bm{V}}_r^{-1} \left( \bm{a}_{A_1}, \bm{a}_{A_2},\ldots,\bm{a}_{A_{T_r}} \right) \in \mathbb{R}^{T_r},
\end{equation}
and thus we have
\begin{equation}
    \Vert W_i \Vert_2^2 = W_i W_i^\top = \bm{a}_i^\top \check{\bm{V}}_r^{-1} \bm{V}_r \check{\bm{V}}_r^{-1}  \bm{a}_i.
\end{equation}

To give the confidence radius under GLM, we have the following statement for any arm $i \in \mathcal{A}_I(r-1)$ in round $r$.
\begin{align}
    \left \lvert \hat{\mu}_i(r) - \mu_i \right\rvert &= \left \lvert \bm{a}_i^\top (\hat{\bm{\theta}}_r - \bm{\theta}) \right\rvert \nonumber\\
    &= \left \lvert \bm{a}_i^\top \check{\bm{V}}_r^{-1} \sum_{s=1}^{T_r} \bm{a}_{A_s} \eta_s \right\rvert \nonumber\\
    &= \left \lvert \sum_{s=1}^{T_r} W_{i,s}\eta_{A_s} \right\rvert,
\end{align}
where the second equality is established with Lemma 1 of \citeec{kveton2023randomized} and $T_r$, which is defined in equation \eqref{equation_phase budget G_GLM}, is the adjusted sampling budget in each round $r$ for the GLM. Since $(\eta_{A_s})_{s \in T_r}$ are independent, mean zero, 1-sub-Gaussian random variables, then $\sum_{s=1}^{T_r} W_{i,s}\eta_{A_s}$ is a $\Vert W_i \Vert_2$-sub-Gaussian variable for each arm $i$, then we have
\begin{align}\label{eq_GLM_prob_1}
    \mathbb{P}\left( \left \lvert \hat{\mu}_i(r) - \mu_i \right\rvert > \varepsilon_r \right) &\leq 2\text{exp} \left( -\frac{\varepsilon_r^2}{2\Vert W_i \Vert_2^2} \right).
\end{align}

Since $\check{\bm{\theta}}_r$ is not known in the process, we need to find another way to represent this term. By assumption, we know $\dot{\mu}_{\text{link}} \geq c_{\min}$ for some $c_{\min} \in \mathbb{R}^+$ and for all $i \in \mathcal{A}_I(r-1)$. Therefore $c_{\min}^{-1} \bm{V}_r^{-1} \succeq \check{\bm{V}}_r^{-1}$ by definition of $\check{\bm{V}}_r$, and then we have
\begin{align}
    \Vert W_i \Vert_2^2 &= \bm{a}_i^\top \check{\bm{V}}_r^{-1} \bm{V}_r \check{\bm{V}}_r^{-1} \bm{a}_i \nonumber\\
    &\leq \bm{a}_i^\top c_{\min}^{-1} \bm{V}_r^{-1} \bm{V}_r c_{\min}^{-1} \bm{V}_r^{-1} \bm{a}_i \nonumber\\
    &=c_{\min}^{-2} \Vert \bm{a}_i \Vert_{\bm{V}_r^{-1}}^2.
\end{align}

Furthermore, if G-optimal design is considered, we have $\Vert \bm{a}_i \Vert_{\bm{V}_r^{-1}}^2 \leq \frac{d}{T_r}$. Together with equation \eqref{eq_GLM_prob_1}, we have
\begin{align}\label{eq_GLM_prob_2}
    \mathbb{P}\left( \left \lvert \hat{\mu}_i(r) - \mu_i \right\rvert > \varepsilon_r \right) &\leq 2\text{exp} \left( -\frac{\varepsilon_r^2}{2c_{\min}^{-2} \Vert \bm{a}_i \Vert_{\bm{V}_r^{-1}}^2} \right) \nonumber\\
    &\leq 2\text{exp} \left( -\frac{\varepsilon_r^2 c_{\min}^{2}}{2 d} T_r \right).
\end{align}

Finally, considering the definition of $T_r$ in equation \eqref{equation_phase budget G_GLM}, with a probability of at least $1-\delta/{Kr(r+1)}$, we have
\begin{equation}
    \left \lvert \hat{\mu}_i(r) - \mu_i \right\rvert \leq \varepsilon_r.
\end{equation}

Thus, with the standard result of the G-optimal design, we still have
\begin{equation}\label{eq_C_G_2}
    C_{\delta/K}(r) \coloneqq \varepsilon_r,
\end{equation}
with which the events $\mathcal{E}_1$ and $\mathcal{E}_2$ in Section \ref{algorithm2: preliminary} hold with a probability of at least $1-\delta$. 

\subsection{Step 2: Bound the Expected Sample Complexity}\label{subsubsec_Bound the Expected Sample Complexity When the GLM is Assumed}

\begin{lemma}\label{claim_glm_2_1}
   For $i \in G_\varepsilon$, if $r \geq\left\lceil\log _2\left(\frac{4}{\varepsilon-\Delta_i}\right)\right\rceil$, then we have $ \mathbb{E}_{\boldsymbol{\mu}}\left[\mathbbm{1}\left[i \notin G_r\right] \mid \mathcal{E}_{1}\right] = 0$. 
\end{lemma}

\begin{lemma}\label{claim_glm_2_2}
   For $i \in G_\varepsilon^c$, if $r \geq\left\lceil\log _2\left(\frac{4}{\Delta_i-\varepsilon}\right)\right\rceil$, then we have $ \mathbb{E}_{\boldsymbol{\mu}}\left[\mathbbm{1}\left[i \notin B_r\right] \mid \mathcal{E}_{1}\right] = 0$. 
\end{lemma}

\begin{lemma}\label{claim_glm_2_3}
   $R_{\textup{GLM}} = R_{\textup{upper}} = \max \left\{ \left\lceil \log_2 \frac{4}{\alpha_\varepsilon} \right\rceil, \left\lceil \log_2 \frac{4}{\beta_\varepsilon} \right\rceil \right\}$ is the round where all the classifications have been finished and the answer is returned under GLM. 
\end{lemma}

The proofs of Lemmas~\ref{claim_glm_2_1}, \ref{claim_glm_2_2}, and \ref{claim_glm_2_3} closely follow those in Section~\ref{refined version}.

\begin{lemma}\label{claim_glm_2_4}
   For the expected sample complexity with high probability event $\mathcal{E}_{1}$, we have
\begin{align}\label{eq_GLM_sample_complexity}
    \mathbb{E}_{\boldsymbol{\mu}}\left[T_{\textup{GLM}} \mid \mathcal{E}_{1}\right] &\leq c \max\left\{ \frac{256d}{\alpha_\varepsilon^2c_{\min}^{2}} \log \left( \frac{2K}{\delta} \log_2 \frac{16}{\alpha_\varepsilon} \right), \frac{256d}{\beta_\varepsilon^2c_{\min}^{2}} \log \left( \frac{2K}{\delta} \log_2 \frac{16}{\beta_\varepsilon} \right) \right\} + \frac{d(d+1)}{2} R_{\textup{GLM}},
\end{align}
where $c$ is a universal constant, $c_{\min}$ is the known constant controlling the first-order derivative of the inverse link function, and $R_{\textup{GLM}} = R_{\textup{upper}} = \max \left\{ \left\lceil \log_2 \frac{4}{\alpha_\varepsilon} \right\rceil, \left\lceil \log_2 \frac{4}{\beta_\varepsilon} \right\rceil \right\}$ is the round where all the classifications have been finished and the answer is returned under GLM. 
\end{lemma}

\proof\\{\textit{Proof.}}
We also consider the decomposition of $T$ in equation~\eqref{equation: decomposition of T}, where all expectations are conditioned on the high-probability event $\mathcal{E}_{1}$, given by
\begin{align}
    \mathbb{E}_{\boldsymbol{\mu}}\left[T_{\textup{GLM}} \mid \mathcal{E}_{1} \right]
    &\leq \sum_{r=1}^{\infty} \mathbb{E}_{\boldsymbol{\mu}}\left[\mathbbm{1}\left[G_r \cup B_r \neq[K]\right] \mid \mathcal{E}_{1} \right] \sum_{\bm{a} \in \mathcal{A}(r-1)}T_r(\bm{a}) \nonumber\\
    &\leq \sum_{r=1}^{R_{\textup{GLM}}} \left( \dfrac{d2^{2r+1}}{c_{\min}^2} \log\left(\dfrac{2Kr(r + 1)}{\delta}\right) + \frac{d(d+1)}{2} \right) \nonumber\\
    &\leq \frac{d(d+1)}{2} R_{\textup{GLM}} + 2c_{\min}^{-2}d\log \left(\frac{2K}{\delta}\right) \sum_{r=1}^{{R_{\textup{GLM}}}} 2^{2r} + 4c_{\min}^{-2}d \sum_{r=1}^{{R_{\textup{GLM}}}} 2^{2r} \log (r+1) \nonumber\\
    &\leq 4c_{\min}^{-2}\log \left[ \frac{2K}{\delta}\left( R_{\textup{GLM}} + 1 \right) \right] \sum_{r=1}^{{R_{\textup{GLM}}}} d 2^{2r} + \frac{d(d+1)}{2} R_{\textup{GLM}} \label{middle result_refined_GLM} \\
    &\leq c \max\left\{ \frac{256d}{\alpha_\varepsilon^2c_{\min}^{2}} \log \left( \frac{2K}{\delta} \log_2 \frac{16}{\alpha_\varepsilon} \right), \frac{256d}{\beta_\varepsilon^2c_{\min}^{2}} \log \left( \frac{2K}{\delta} \log_2 \frac{16}{\beta_\varepsilon} \right) \right\} + \frac{d(d+1)}{2} R_{\textup{GLM}},\label{another angle: decomposition of T - 1_GLM}
\end{align}
where $c$ is a universal constant. Then, let $\xi = \min(\alpha_\varepsilon, \beta_\varepsilon)/16$ denoting the minimum gap of the problem instance, we have
\begin{equation}
    \mathbb{E}[T_\textup{GLM} \mid \mathcal{E}] 
    = \mathcal{O} 
        \left(\frac{d}{c_{\min}^2} \xi^{-2} \log\!\left( \frac{K}{\delta} \log_{2}(\xi^{-2}) \right) 
        + d^{2} \log(\xi^{-1})\right).
\end{equation}
\hfill\Halmos
\endproof

\section{Detailed Settings for Synthetic Experiments}\label{sec_Detailed Settings for Synthetic Experiments}

We recap the figure here for better clarity. 

\begin{figure}[htbp]
  \centering

  \subfloat[Synthetic I - Adaptive Setting\label{fig:synthetic-adaptive}]{%
    \includegraphics[width=0.48\textwidth]{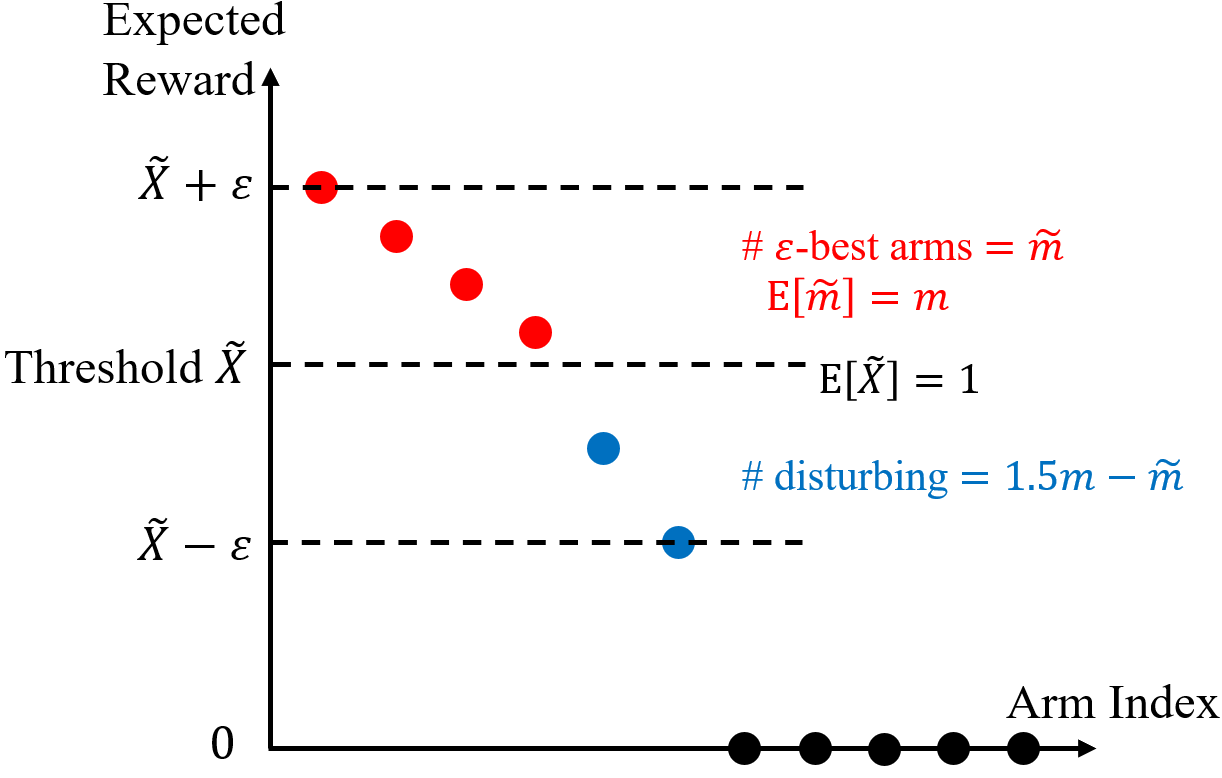}%
  }%
  \hfill
  \subfloat[Synthetic II - Static Setting\label{fig:synthetic-static}]{%
    \includegraphics[width=0.48\textwidth]{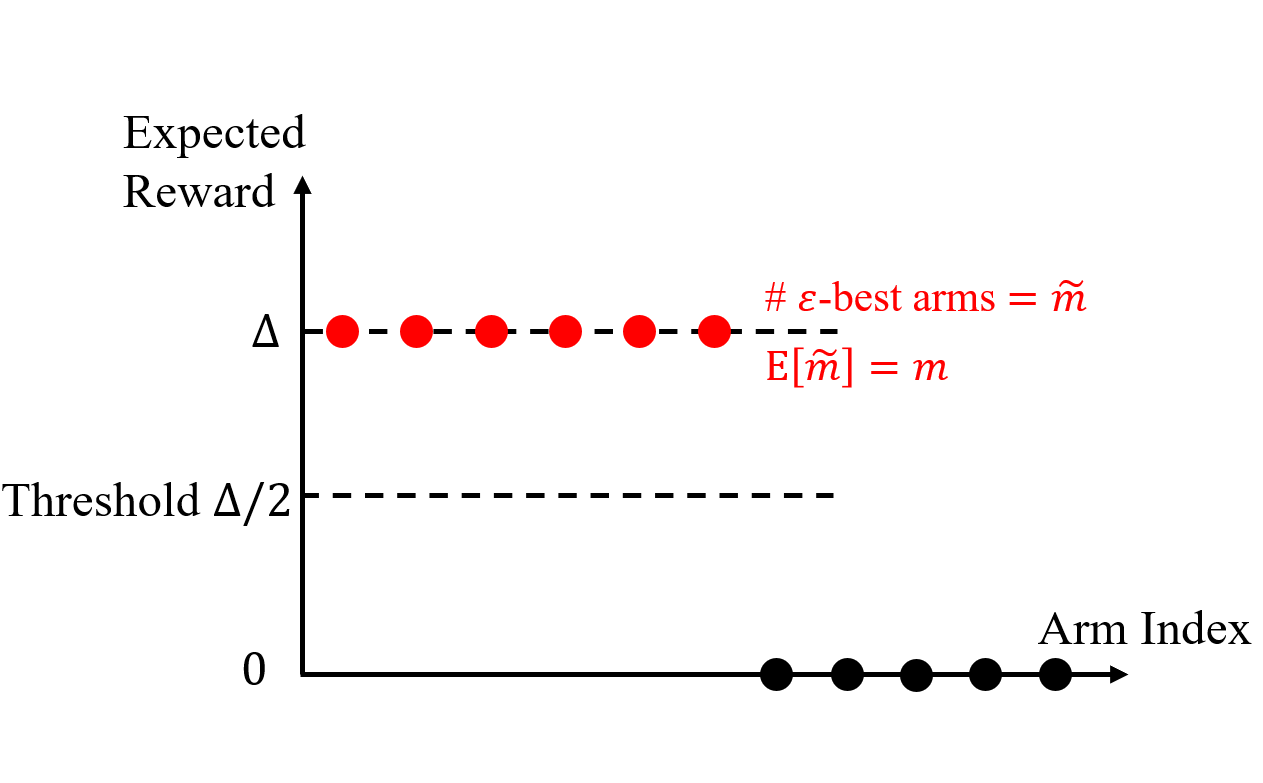}%
  }%

  \caption{Illustration on Synthetic Settings}
  \label{fig:synthetic}
\end{figure}

\subsection{Synthetic I - Adaptive Setting.}

First, we randomly sample $\tilde{m}$, representing the number of $\varepsilon$-best arms, from a distribution with an expected value of $m$ (used as input for a top $m$ algorithm). Next, we randomly sample $\tilde{X}$, representing the best arm reward minus $\varepsilon$, from a distribution with an expected value of $X$ (used as input for a threshold bandit algorithm). We then assign $\tilde{m}$ $\varepsilon$-best arms with expected rewards uniformly distributed between $\tilde{X} + \varepsilon$ and $\tilde{X}$. Additionally, we assign $(1.5m - \tilde{m})$ arms that are not $\varepsilon$-best with expected rewards uniformly distributed between $\tilde{X}$ and $\tilde{X} - \varepsilon$, as illustrated in Figure~\ref{fig:synthetic}. 

Based on these designed arm rewards, we define the linear model parameter as:
$$\bm{\theta}=(\tilde{X}+\varepsilon, \tilde{X}+(\tilde{m}-1)\varepsilon/\tilde{m}, \ldots,\tilde{X}+\varepsilon/\tilde{m}, 0,\ldots,0)^\top.$$

Arms are $d$-dimensional canonical basis $e_1, e_2, \ldots,e_d$ and $(1.5m-\tilde{m})$ additional disturbing arms 
$$\bm{x_i}=\left(\frac{\tilde{X}-(1.5m-\tilde{m}-i)\varepsilon/(1.5m-\tilde{m})}{\tilde{X}+\varepsilon}, 0,\cdots,0, \sqrt{1-\left(\frac{\tilde{X}-(1.5m-\tilde{m}-i)\varepsilon/(1.5m-\tilde{m})}{\tilde{X}+\varepsilon}\right)^2}\right)^\top$$ with $i\in[1.5m-\tilde{m}]$.

In the adaptive setting, pulling one arm can provide information about the distributions of other arms. The optimal policy in this setting should adaptively refine its sampling and stopping strategy based on historical data. This allows the algorithm to focus more on the disturbing arms, making adaptive strategies particularly effective as the algorithm progresses. In our experiments, we set $m=4$, $X=1$, with $d=10$ and $\varepsilon\in\{0.1, 0.2, 0.3\}$. A total of six different problem instances are evaluated to compare the performance of the algorithms. 

\subsection{Synthetic II - Static Setting.}

We consider a static synthetic setting, similar to the one proposed by \citeec{xu2018fully}, where arms are represented as $d$-dimensional canonical basis vectors $e_1, e_2, \ldots, e_d$. We set the parameter vector $\bm{\theta} = (\Delta, \ldots, \Delta, 0, \ldots, 0)^\top$, where $\tilde{m}$ elements are $\Delta$ and $d - \tilde{m}$ elements are 0. In this setting, $\mathbb{E}[\tilde{m}] = m$, and only the value $m$ is provided as input to top $m$ algorithms. Consequently, the true mean values consist of some $\Delta$'s and some $0$'s.

If we set $\varepsilon = \Delta / 2$, as $\Delta$ approaches 0, it becomes difficult to distinguish between the $\varepsilon$-best arms and the arms that are not $\varepsilon$-best. In the static setting, knowledge of the rewards does not alter the sampling strategy, as all arms must be estimated with equal accuracy to effectively differentiate between them. Therefore, a static policy is optimal in this case, and the goal of this setting is to assess the ability of our algorithm to adapt to such static conditions. In our experiment, we set $m = 4$ with $d \in \{8, 12, 16\}$ and $\Delta = 1$. A total of three different problem instances are evaluated to compare the algorithms. 

\section{Auxiliary Results}\label{sec_Auxiliary Results}
The following lemma shows that matrix inversion reverses the order relation.
\begin{lemma}\label{lemma: inversion reverses loewner orders}
    \text{(Inversion Reverses Loewner orders)} Let $\bm{A}, \bm{B} \in \mathbb{R}^{d \times d}$. Suppose that $\bm{A} \succeq \bm{B}$ and $\bm{B}$ is invertible, we have
\begin{equation}
    \bm{A}^{-1} \preceq \bm{B}^{-1}.
\end{equation}
\end{lemma}

\proof\\{\textit{Proof.}}
By definition, to show $\boldsymbol{B}^{-1}-\boldsymbol{A}^{-1}$ is a positive semi-definite matrix, it suffices to show that $\|\bm{x}\|_{\bm{B}^{-1}}^2-\|\bm{x}\|_{\bm{A}^{-1}}^2=\|\bm{x}\|_{\bm{B}^{-1}-\bm{A}^{-1}}^2 \geq 0$ for any $\bm{x} \in \mathbb{R}^d$. Then, by the Cauchy-Schwarz inequality,
\begin{equation}
    \|\bm{x}\|_{\bm{A}^{-1}}^2=\left\langle \bm{x}, \bm{A}^{-1} \bm{x}\right\rangle \leq\|\bm{x}\|_{\bm{B}^{-1}}\left\|\bm{A}^{-1} \bm{x}\right\|_{\bm{B}} \leq\|\bm{x}\|_{\bm{B}^{-1}}\left\|\bm{A}^{-1} \bm{x}\right\|_{\bm{A}}=\|\bm{x}\|_{\bm{B}^{-1}}\|\bm{x}\|_{\bm{A}^{-1}} .
\end{equation}

Hence $\|\bm{x}\|_{\bm{A}^{-1}} \leq\|\bm{x}\|_{\bm{B}^{-1}}$ for all $\bm{x}$, which completes the lemma.
\hfill\Halmos
\endproof

The following lemma establishes an upper bound on the ratio between two optimization problems that incorporate instance-specific information from the bandit setting. 

\begin{lemma}\label{lemma: inequality iv}
    We always have $\mathcal{A}_I = [K]$, \emph{i.e.}, the entire set of arms is under consideration. For any arm $i \in \mathcal{A}_I \cap G_\varepsilon \setminus \{1\}$, we have 
\begin{equation}
    \frac{\min_{\bm{p}\in S_K}\max_{i \in \mathcal{A}_I \cap G_\varepsilon \backslash \{1\}} \Vert \boldsymbol{y}_{1, i} \Vert_{\bm{V}_{\bm{p}}^{-1}}^2 }{\min_{\bm{p}\in S_K}\min_{i \in \mathcal{A}_I \cap G_\varepsilon \backslash \{1\}} \Vert \boldsymbol{y}_{1, i} \Vert_{\bm{V}_{\bm{p}}^{-1}}^2} \leq \frac{d L_1}{{\mathcal{G}_{\mathcal{Y}}}^2 L_2}.
\end{equation}

\end{lemma}

\proof\\{\textit{Proof.}}
For any arm $i \in [K]$, from a perspective of geometry quantity, let conv($\mathcal{A}\cup-\mathcal{A}$) denote the convex hull of symmetric $\mathcal{A}\cup-\mathcal{A}$. Then for any set $\mathcal{Y}\subset \mathbb{R}^d$ define the gauge of $\mathcal{Y}$ as
\begin{equation}
    \mathcal{G}_\mathcal{Y} = \max \left\{ c>0: c\mathcal{Y}\subseteq \text{conv(}\mathcal{A}\cup-\mathcal{A}\text{)} \right\}.
\end{equation}

We then provide a natural upper bound for ${\min_{\bm{p}\in S_K}\max_{i \in \mathcal{A}_I \cap G_\varepsilon \backslash \{1\}} \Vert \boldsymbol{y}_{1, i} \Vert_{\bm{V}_{\bm{p}}^{-1}}^2}$, given by
\begin{align}\label{numerator}
    {\min_{\bm{p}\in S_K}\max_{i \in \mathcal{A}_I \cap G_\varepsilon \backslash \{1\}} \Vert \boldsymbol{y}_{1, i} \Vert_{\bm{V}_{\bm{p}}^{-1}}^2} &\leq {\min_{\bm{p}\in S_K}\max_{y\in \mathcal{Y}(\mathcal{A}_I)} \Vert \boldsymbol{y} \Vert_{\bm{V}_{\bm{p}}^{-1}}^2} \nonumber\\ 
    &= \frac{1}{{\mathcal{G}_{\mathcal{Y}}}^2} {\min_{\bm{p}\in S_K}\max_{y\in \mathcal{Y}(\mathcal{A}_I)} \Vert \boldsymbol{y}{\mathcal{G}_\mathcal{Y}} \Vert_{\bm{V}_{\bm{p}}^{-1}}^2} \nonumber\\
    &\leq \frac{1}{{\mathcal{G}_{\mathcal{Y}}}^2} {\min_{\bm{p}\in S_K}\max_{\bm{a} \in \text{conv(}\mathcal{A}\cup-\mathcal{A}\text{)}} \Vert \bm{a} \Vert_{\bm{V}_{\bm{p}}^{-1}}^2} \nonumber\\
    &= \frac{1}{{\mathcal{G}_{\mathcal{Y}}}^2} {\min_{\bm{p}\in S_K}\max_{i \in \mathcal{A}_I} \Vert \bm{a}_i \Vert_{\bm{V}_{\bm{p}}^{-1}}^2} \nonumber\\
    &\leq \frac{d}{{\mathcal{G}_{\mathcal{Y}}}^2},
\end{align}
where the third line follows from the fact that the maximum value of a convex function on a convex set must occur at a vertex. With the Kiefer-Wolfowitz Theorem for the G-optimal design, the last inequality is achieved. 

Furthermore, for any arm $i \in \mathcal{A}_I \cap G_\varepsilon \backslash \{1\}$, we have 
\begin{align}
    \min_{\bm{p}\in S_K}\min_{i \in \mathcal{A}_I \cap G_\varepsilon \backslash \{1\}} \Vert \boldsymbol{y}_{1, i} \Vert_{\bm{V}_{\bm{p}}^{-1}}^2 &\geq \min_{\bm{p}\in S_K}\min_{i \in \mathcal{A}_I \cap G_\varepsilon \backslash \{1\}} \text{eig}_{\min }(\bm{V}_{\bm{p}}^{-1}) \Vert \boldsymbol{y}_{1, i} \Vert_2^2 \nonumber\\
    &= \min_{\bm{p}\in S_K}\min_{i \in \mathcal{A}_I \cap G_\varepsilon \backslash \{1\}} \frac{1}{\text{eig}_{\max }(\bm{V}_{\bm{p}})} \Vert \boldsymbol{y}_{1, i} \Vert_2^2 \nonumber\\
    &\geq \frac{1}{\max_{i \in \mathcal{A}_I} \Vert \bm{a}_i \Vert_2} \min_{\bm{p}\in S_K}\min_{i \in \mathcal{A}_I \cap G_\varepsilon \backslash \{1\}} \Vert \boldsymbol{y}_{1, i} \Vert_2^2,
\end{align}
where $\text{eig}_{\max }(\cdot)$ and $\text{eig}_{\min }(\cdot)$ are respectively the largest and smallest eigenvalues of a matrix. The first line follows from the Rayleigh Quotient and Rayleigh Theorem. The last line is derived by the relationship $\text{eig}_{\max }(\bm{V}_{\bm{p}}) \leq \max_{i \in \mathcal{A}_I} \Vert \bm{a}_i \Vert_2$. Recall the assumption in Theorem \ref{upper bound: Algorithm 1 XY} that $\min_{i \in G_\varepsilon \backslash \{ 1 \}} \Vert \bm{a}_1 - \bm{a}_i \Vert^2 \geq L_2$ and the assumption in Section \ref{sec_Problem Formulation} that $\Vert \boldsymbol{a}_i \Vert_2 \leq L_1 \text{ for } \forall i \in [K]$, we have
\begin{equation}\label{denominator}
     \min_{\bm{p}\in S_K}\min_{i \in \mathcal{A}_I \cap G_\varepsilon \backslash \{1\}} \Vert \boldsymbol{y}_{1, i} \Vert_{\bm{V}_{\bm{p}}^{-1}}^2 \geq \frac{L_2}{L_1}.
\end{equation}

Finally, combining inequalities (\ref{numerator}) and (\ref{denominator}) completes the lemma.
\hfill\Halmos
\endproof

\begin{lemma}[\citeec{kiefer1960equivalence}]\label{theorem_KW}
If the arm vectors $\boldsymbol{a} \in \mathcal{A}$ span $\mathbb{R}^d$, then for any probability distribution $\pi \in \mathcal{P}(\mathcal{A})$, the following statements are equivalent:
    \begin{enumerate}
        \item $\pi^\ast$ minimizes the function $g(\pi) = \max_{\boldsymbol{a} \in \mathcal{A}} \|\boldsymbol{a}\|_{\bm{V}(\pi)^{-1}}^2$.
\item $\pi^\ast$ maximizes the function $f(\pi) = \log \det \bm{V}(\pi)$.
\item $g(\pi^\ast) = d$.
    \end{enumerate}
     Additionally, there exists a $\pi^\ast$ of $g(\pi)$ such that the size of its support, $\vert \text{Supp}(\pi^\ast) \vert$, is at most $d(d+1)/2$.
\end{lemma}

\end{appendices}

\bibliographystyleec{informs2014}
\bibliographyec{main} %

\end{document}





\RUNTITLE{Identifying All $\varepsilon$-Best Arms In Linear Bandits}

\TITLE{Identifying All $\varepsilon$-Best Arms In Linear Bandits With Misspecification}

\ARTICLEAUTHORS{%
\AUTHOR{Zhekai Li}
\AFF{School of Electronic Information and Electrical Engineering, Shanghai Jiao Tong University, \EMAIL{zhekai\_li.work@sjtu.edu.cn}}
\AUTHOR{Tianyi Ma}
\AFF{University of Michigan – Shanghai Jiao Tong University Joint Institute, \EMAIL{sean\_ma@sjtu.edu.cn }}
\AUTHOR{Cheng Hua}
\AFF{Antai College of Economics and Management, Shanghai Jiao Tong University, \EMAIL{cheng.hua@sjtu.edu.cn}}
\AUTHOR{Ruihao Zhu}
\AFF{SC Johnson College of Business, Cornell University, \EMAIL{ruihao.zhu@cornell.edu}}

} 

\ABSTRACT{%
Motivated by the need to efficiently identify multiple candidates in high trial-and-error cost tasks such as drug discovery, we propose a near-optimal algorithm to identify all $\varepsilon$-best arms (\emph{i.e.}, those at most $\varepsilon$ worse than the optimum). Specifically, we introduce LinFACT, an algorithm designed to optimize the identification of all $\varepsilon$-best arms in linear bandits. We establish a novel information-theoretic lower bound on the sample complexity of this problem and demonstrate that LinFACT achieves instance optimality by matching this lower bound up to a logarithmic factor. A key aspect of our proof integrates the lower bound directly into the scaling process for upper bound derivation, determining the termination round and thus the sample complexity. We also extend our analysis to provide results under model misspecification and within generalized linear model contexts. Numerical experiments, including synthetic and real drug discovery data, demonstrate that LinFACT identifies more promising candidates with reduced sample complexity, offering significant computational efficiency and accelerating early-stage exploratory experiments.

\vspace{0.2cm}

\noindent\textit{Supplemental Material:} The appendix is available \href{https://www.dropbox.com/scl/fo/92hi7tjojsapy0fi2vlos/AOArRiiYy7cqw8_nSTc8R6k?rlkey=tls6yuz4t1avdqmizecsdetoa&st=8o4uzv4p&dl=0}{online}.
}

%


\KEYWORDS{sequential experiment design; pure exploration; optimal design; model misspecification; generalized linear models}

\maketitle

\subsection{Preliminaries}

In this paper, we denote the set of positive integers up to $N$ by $[N] = \{1, \ldots, N\}$. Vectors and matrices are represented using boldface notation. The inner product of two vectors is denoted by $\langle \cdot, \cdot \rangle$. We define the weighted matrix norm $\Vert \boldsymbol{x} \Vert_{\boldsymbol{A}}$ as $\sqrt{\boldsymbol{x}^\top \boldsymbol{A} \boldsymbol{x}}$, where $\boldsymbol{A}$ is a positive semi-definite matrix that weights and scales the norm.

\section{Problem Formulation}\label{sec_Problem Formulation} 

We address the problem of identifying all $\varepsilon$-best arms from a finite set of $K$ arms where $K$ is a (possibly large) positive integer. Each arm $i \in [K]$ has an associated reward distribution with an unknown fixed mean $\mu_i$. Let the mean vector of all arms be denoted as $\boldsymbol{\mu} = (\mu_1, \mu_2, \ldots, \mu_K)$, which can only be estimated through bandit feedback from the selected arms. Without loss of generality, we assume $\mu_1 > \mu_2 \geq \ldots \geq \mu_K$. The gap $\Delta_i = \mu_1 - \mu_i$ (for $i \neq 1$) represents the difference in expected rewards between the optimal arm and arm $i$.

\begin{definition}\label{def_all_epsilon} ($\varepsilon$-Best Arm). Given $\varepsilon > 0$, an arm $i$ is called \textit{$\varepsilon$-best} if $\mu_i \geq \mu_1 - \varepsilon$. \end{definition}

We adopt an additive framework to define $\varepsilon$-best arms. There also exists a multiplicative counterpart, where an arm $i$ is considered $\varepsilon$-best if $\mu_i \geq (1 - \varepsilon)\mu_1$. While our study focuses on the additive model, the analysis for the multiplicative model follows similar reasoning. We denote the set of all $\varepsilon$-best arms for a mean vector $\boldsymbol{\mu}$ as  
\begin{equation}
    G_\varepsilon(\boldsymbol{\mu}) \coloneqq \{i : \mu_i \geq \mu_1 - \varepsilon \}. \footnote{The set of all $\varepsilon$-best arms $G_\varepsilon(\boldsymbol{\mu})$ is also referred to as the good set in this paper.}
\end{equation}

We study this problem under a linear structure, where the mean values depend on an unknown parameter vector $\boldsymbol{\theta}$. Each arm $i$ is associated with a feature vector $\boldsymbol{a}_i \in \mathbb{R}^d$. Let $\mathcal{A} \subset \mathbb{R}^d$ be the set of feature vectors, and let $\boldsymbol{\Psi} \coloneqq [\boldsymbol{a}_1, \boldsymbol{a}_2, \ldots, \boldsymbol{a}_K] \in \mathbb{R}^{K \times d}$ be the feature matrix. When an arm $A_t$ with corresponding feature vector $\boldsymbol{a}_{A_t} \in \mathcal{A}$ is selected at time $t$, we observe the bandit feedback $X_t$, given by
\begin{equation}\label{eq_linear_model}
X_t = \boldsymbol{a}_{A_t}^\top \boldsymbol{\theta} + \eta_t,
\end{equation}
where $\mu_{A_t} = \boldsymbol{a}_{A_t}^\top \boldsymbol{\theta}$ is the true mean reward of the selected arm, and $\eta_t$ is a noise variable. We assume $\max_{i \in [K]} \Vert \boldsymbol{a}_i \Vert_2 \leq L_1$, where $\Vert \cdot \Vert_2$ denotes the $\ell_2$-norm. Additionally, the noise $\eta_t$ is assumed to be conditionally 1-sub-Gaussian, meaning that for any $\lambda \in \mathbb{R}$, 
$
\mathbb{E} \left[ e^{\lambda \eta_t} \,\big|\, \boldsymbol{a}_{A_1}, \ldots, \boldsymbol{a}_{A_{t-1}}, \eta_1, \ldots, \eta_{t-1} \right] \leq \exp \left( \frac{\lambda^2}{2} \right).
$ 
This implies that the noise, conditioned on past data, has zero mean and variance bounded by 1.

\subsection{Probably Approximately Correct}\label{subsec_PAC}

Our goal is to identify all $\varepsilon$-best arms with high confidence while minimizing the sampling budget. To achieve this, we employ three main components: stopping rule, sampling rule, and decision rule. 

At each time step $t$, the stopping rule $\tau_\delta$ determines whether to continue or stop the process. If the process continues, an arm is selected according to the sampling rule, and the corresponding random reward is observed. When the process stops at $t = \tau_\delta$, a decision rule provides an estimate $\widehat{\mathcal{I}}_{\tau_\delta}$ of the true solution set $\mathcal{I}(\boldsymbol{\mu})$, which in our problem is the set of all $\varepsilon$-best arms, $ G_\varepsilon(\boldsymbol{\mu})$. 

We define the set of all viable mean vectors $\boldsymbol{\mu}$ as
\begin{equation}
M \coloneqq \left\{ \boldsymbol{\mu} \in \mathbb{R}^K \;\middle|\; \exists \boldsymbol{\theta} \in \mathbb{R}^d, \boldsymbol{\mu} = \boldsymbol{\Psi}\boldsymbol{\theta} \; \land \; \Vert \boldsymbol{a}_i \Vert_2 \leq L_1 \text{ for each } i \in [K] \right\}.
\end{equation}

Here, the set $M$ consists of all possible mean vectors $\boldsymbol{\mu}$ that can be expressed as a linear combination of the parameter vector $\boldsymbol{\theta}$ through the matrix $\boldsymbol{\Psi}$. Additionally, as mentioned previously, the vectors $\boldsymbol{a}_i$ (representing the rows of $\boldsymbol{\Psi}$) are assumed to have an $\ell_2$-norm not exceeding the bound $L_1$, ensuring that each component vector lies within a specified size constraint. We focus on algorithms that are probably approximately correct with high confidence, referred to as $\delta$-PAC algorithms.

\begin{definition}\label{delta-PAC}
\textup{($\delta$-\textit{PAC} Algorithm).} An algorithm is \textit{$\delta$-PAC} for all $\varepsilon$-best arms identification if it identifies the correct solution set with a probability of at least $1-\delta$ for any problem instance with mean $\boldsymbol{\mu} \in M$, \emph{i.e.},
\begin{equation}
\mathbb{P}_{\boldsymbol{\mu}} \left( \tau_\delta < \infty, \; \widehat{\mathcal{I}}_{\tau_\delta} = G_\varepsilon\left( \boldsymbol{\mu}\right) \right) \geq 1-\delta, \qquad \forall \boldsymbol{\mu} \in M.
\label{def: delta-PAC}
\end{equation}
\end{definition}

Upon stopping, $\delta$-PAC algorithms ensure the identification of all $\varepsilon$-best arms with high confidence. Therefore, our goal is to design a $\delta$-PAC algorithm that minimizes the stopping time, formulated as the following optimization problem.
\begin{align}
    \min &\quad \mathbb{E}_{\boldsymbol{\mu}} \left[ \tau_\delta \right] \\
    \text{s.t.} &\quad \mathbb{P}_{\boldsymbol{\mu}} \left( \tau_\delta < \infty, \; \widehat{\mathcal{I}}_{\tau_\delta} = G_\varepsilon\left( \boldsymbol{\mu}\right) \right) \geq 1-\delta, \qquad \forall \boldsymbol{\mu} \in M.
\end{align}


\subsection{Optimal Design of Experiment}\label{subsec_Optimal Design} 

\paragraph{Ordinary Least Squares.}\label{subsubsec_Least Squares Estimators}
Consider a sequence of pulled arms, denoted as $A_1, A_2, \ldots, A_t$, and the corresponding observed rewards $X_1, X_2, \ldots, X_t$. If the feature vectors of these arms, $\boldsymbol{a}_{A_1}, \boldsymbol{a}_{A_2}, \ldots, \boldsymbol{a}_{A_t}$, span the space $\mathbb{R}^d$, the ordinary least squares (OLS) estimator for the parameter $\boldsymbol{\theta}$ is given by
\begin{equation}\label{OLS}
    \hat{\boldsymbol{\theta}}_t = \boldsymbol{V}_t^{-1} \sum_{n=1}^t \boldsymbol{a}_{A_n} X_n,
\end{equation}
where $\boldsymbol{V}_t = \sum_{n=1}^t \boldsymbol{a}_{A_n} \boldsymbol{a}_{A_n}^\top \in \mathbb{R}^{d \times d}$ represents the information matrix. Using the properties of sub-Gaussian random variables, we can derive a confidence bound for the OLS estimator. This bound, denoted as $B_{t,\delta}$, is detailed in Proposition \ref{proposition_confidence bound for B 1}. The confidence region for the parameter $\boldsymbol{\theta}$ at time step $t$ is given by
\begin{equation}\label{confidence for ellipse}
    \mathcal{C}_{t, \delta} = \left\{ \boldsymbol{\theta} : \Vert \hat{\boldsymbol{\theta}}_t - \boldsymbol{\theta} \Vert_{\boldsymbol{V}_t} \leq B_{t,\delta} \right\}.
\end{equation}

\begin{proposition}\label{proposition_confidence bound for B 1}
For any fixed sampling policy and any given vector $\boldsymbol{x} \in \mathbb{R}^d$, with probability at least $1-\delta$, the following holds. 
\begin{equation}\label{confidence bound BB}
    \left| \boldsymbol{x}^\top \left( \hat{\boldsymbol{\theta}}_t - \boldsymbol{\theta} \right) \right| \leq \Vert \boldsymbol{x} \Vert_{\boldsymbol{V}_t^{-1}} B_{t, \delta},
\end{equation}
where the anytime confidence bound $B_{t, \delta}$ is given by
$
    B_{t, \delta} = 2 \sqrt{2 \left( d \log(6) + \log\left( \frac{1}{\delta} \right) \right)}.
$
\end{proposition}

In many practical scenarios, the observed data are not fixed in advance. To handle this, a martingale-based method can be employed, as described by , to define an adaptive confidence bound for the OLS estimator. This accounts for the variability introduced by random rewards and adaptive sampling policies. The confidence interval in Proposition \ref{proposition_confidence bound for B 1} highlights the connection between arm allocation policies in linear bandits and experimental design theory . This connection serves as a fundamental component in constructing our algorithm. 

\paragraph{Feature Vector Projection.}\label{subsubsec_Projection}
At any time step where estimation needs to be made after sampling, if the feature vectors of the sampled arms do not span $\mathbb{R}^d$, we substitute them with dimensionality-reduced feature vectors . Specifically, we project all feature vectors onto the subspace spanned by $\mathcal{A}$. Let $\boldsymbol{B} \in \mathbb{R}^{d \times d'}$ be an orthonormal basis for this subspace, where $d' < d$ is the dimension of the subspace. The new feature vector $\boldsymbol{a}'$ is then given by
$
    \boldsymbol{a}' = \boldsymbol{B}^\top \boldsymbol{a}.
$
In this transformation, $\boldsymbol{B}\boldsymbol{B}^\top$ is a projection matrix, ensuring
\begin{equation}\label{eq_projection}
    \left\langle \boldsymbol{\theta}, \boldsymbol{a} \right\rangle = \left\langle \boldsymbol{\theta}, \boldsymbol{B}\boldsymbol{B}^\top \boldsymbol{a} \right\rangle = \left\langle \boldsymbol{B}^\top \boldsymbol{\theta}, \boldsymbol{B}^\top \boldsymbol{a} \right\rangle = \left\langle \boldsymbol{\theta}', \boldsymbol{a}' \right\rangle.
\end{equation}

Equation \eqref{eq_projection} ensures that the mean values of all arms remain unchanged under the projection. The first equality holds because $\boldsymbol{B}\boldsymbol{B}^\top$ is a projection matrix and $\boldsymbol{a}$ lies in the subspace spanned by $\boldsymbol{B}$ (\emph{i.e.}, $\boldsymbol{B}\boldsymbol{B}^\top \boldsymbol{a} = \boldsymbol{a}$). The second equality follows from the same matrix form $\boldsymbol{\theta}^\top \boldsymbol{B} \boldsymbol{B}^\top \boldsymbol{a}$. The third equality holds by definition. 

\paragraph{Optimal Design Criteria.}\label{subsubsec_G-Optimal Design}
In contrast to stochastic bandits, where the mean values of the arms are estimated through repeated sampling of each arm, the linear bandit setting allows these values to be inferred from accurate estimation of the underlying parameter vector $\boldsymbol{\theta}$. As a result, pulling a single arm provides information about all arms. 

A key sampling strategy in this context is the \textit{G-optimal design}, which minimizes the maximum variance of the predicted responses across all arms by optimizing the fraction of times each arm is selected. Formally, the G-optimal design problem seeks a probability distribution $\pi$ on $\mathcal{A}$, where $\pi: \mathcal{A} \rightarrow [0,1]$ and $\sum_{\boldsymbol{a} \in \mathcal{A}} \pi(\boldsymbol{a}) = 1$, that minimizes
\begin{equation}\label{equation_G optimal}
    g(\pi) = \max_{\boldsymbol{a} \in \mathcal{A}} \Vert \boldsymbol{a} \Vert_{\boldsymbol{V}(\pi)^{-1}}^2,
\end{equation}
where $\boldsymbol{V}(\pi) = \sum_{\boldsymbol{a} \in \mathcal{A}} \pi(\boldsymbol{a}) \boldsymbol{a} \boldsymbol{a}^\top$ is the weighted information matrix, analogous to $\boldsymbol{V}_t$ in equation~\eqref{OLS}. The G-optimal design~\eqref{equation_G optimal} ensures a tight confidence interval for mean value estimation. However, comparing the relative differences of mean values across different arms is more critical in identifying the best arms, rather than making the best estimation. 

Therefore, we also consider an alternative design criterion, the \textit{$\mathcal{XY}$-optimal design}, that directly targets the estimation of these gaps. Consider $\mathcal{S} \subset \mathcal{A}$ as a subset of the arm space. We define
\begin{equation}
    \mathcal{Y}(\mathcal{S}) \coloneqq \left\{ \boldsymbol{a} - \boldsymbol{a}^\prime : \forall \boldsymbol{a}, \boldsymbol{a}^\prime \in \mathcal{S}, \boldsymbol{a} \neq \boldsymbol{a}^\prime \right\}
\end{equation}
as the set of vectors representing the differences between each pair of arms in $\mathcal{S}$. The $\mathcal{XY}$-optimal design minimizes
\begin{equation}\label{equation_XY optimal}
    g_{\mathcal{XY}}(\pi) = \max_{\boldsymbol{y} \in \mathcal{Y}(\mathcal{A})} \Vert \boldsymbol{y} \Vert_{\boldsymbol{V}(\pi)^{-1}}^2.
\end{equation}

As mentioned previously, the $\mathcal{XY}$-optimal design focuses on minimizing the maximum variance when estimating the differences (gaps) between pairs of arms. By doing so, it ensures differentiation between arms, rather than estimating each arm individually. This criterion is particularly useful when the goal is to identify relative performance rather than absolute quality.

\subsection{Further Notation}\label{subsec_Further Notation} 


Before proceeding, we introduce several additional notations that will be used throughout the analysis. Define
$
    \alpha_\varepsilon \coloneqq \min_{i \in G_\varepsilon} \left( \mu_i - (\mu_1 - \varepsilon) \right)
$ 
as the distance from the smallest $\varepsilon$-best arm to the threshold $\mu_1 - \varepsilon$. Furthermore, if the complement of $G_\varepsilon(\boldsymbol{\mu})$, denoted as $G_\varepsilon^c(\boldsymbol{\mu})$, is non-empty, we define
$
    \beta_\varepsilon \coloneqq \min_{i \in G_\varepsilon^c} \left( (\mu_1 - \varepsilon) - \mu_i \right)
$ 
as the closest distance from the threshold to the highest mean value of any arm that is not considered $\varepsilon$-best. 

For two probability measures $P$ and $Q$ over a common measurable space, if $P$ is absolutely continuous with respect to $Q$, the Kullback-Leibler divergence between $P$ and $Q$ is defined as
\begin{equation}
    \text{KL}(P, Q) = 
    \begin{cases}
        \int \log\left( \frac{dP}{dQ} \right) dP, & \text{if } Q \ll P; \\
        \infty, & \text{otherwise},
    \end{cases}
\end{equation}
where $\frac{dP}{dQ}$ is the Radon-Nikodym derivative of $P$ with respect to $Q$, and $Q \ll P$ indicates that $Q$ is absolutely continuous with respect to $P$.

\section{Lower Bound and Problem Complexity}\label{sec_Lower Bound of Expected Sample Complexity} 

\begin{figure}[htbp]
    \centering
    \includegraphics[scale=1]{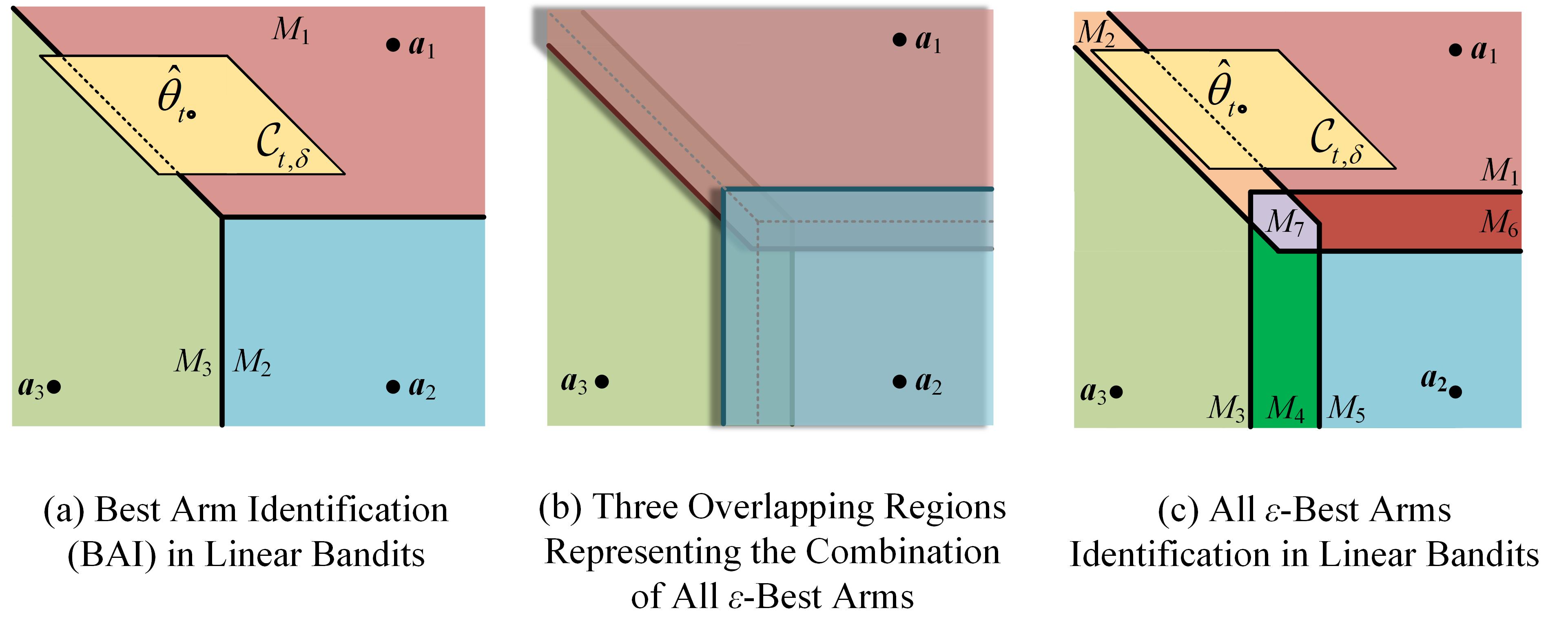}
    \caption{Illustration of the Stopping Condition: Best Arm Identification vs. All $\varepsilon$-Best Arms Identification
    }
\vspace{5pt}
\noindent
\begin{minipage}{\linewidth}
    \footnotesize
    \raggedright
    \textit{Note.} \textit{
        {(a)} Stopping occurs when the confidence region $\mathcal{C}_{t, \delta}$ for the estimated parameter $\hat{\boldsymbol{\theta}}_t$ contracts entirely within one of the three decision regions $M_i$ in a certain time step $t$. The boundaries between regions are defined by the hyperplanes $\boldsymbol{\vartheta}^\top(\boldsymbol{a}_i - \boldsymbol{a}_j) = 0$. Each dot represents an arm. {(b)} In the case of identifying all $\varepsilon$-best arms, the regions overlap. {(c)} Due to these overlaps, the space is partitioned into seven distinct decision regions. Section~\ref{subsubsec_Optimal_Allocation} further outlines an optimal allocation strategy designed to resolve these distinctions. 
    }
\end{minipage}
    \label{fig: 11}
\end{figure}

In this section, we present a novel information-theoretic lower bound for the problem of identifying all $\varepsilon$-best arms in linear bandits. Building on the approach of , we extend the lower bound for best arm identification (BAI) to this more general setting. Figure~\ref{fig: 11} highlights the fundamental structure of the stopping condition, with a complete illustration and additional graphical insights provided in Section~\ref{subsubsec_Visual Explanation of the Lower Bound} of the online appendix. These visualizations offer geometric intuition for the challenges involved in identifying all $\varepsilon$-best arms in linear bandits, showing how overlaps in the parameter space define different sets of $\varepsilon$-best arms. 

\subsection{Lower Bound}\label{subsubsec_All-varepsilon Good Arms Identification in Linear Bandit}

The sample complexity of an algorithm is quantified by the number of samples, denoted as $\tau_\delta$, required to terminate the process. The goal of the algorithm design is to minimize the expected sample complexity $\mathbb{E}_{\boldsymbol{\mu}} \left[ \tau_\delta \right]$ across the entire set of algorithms $\mathcal{H}$. As introduced in , for $\delta \in (0, 1)$, the non-asymptotic problem complexity of an instance $\boldsymbol{\mu}$ can be defined as
\begin{equation}\label{eq_problem complexity}
    \kappa \left( \boldsymbol{\mu} \right) \coloneqq \inf_{Algo \in \mathcal{H}} \frac{\mathbb{E}_{\boldsymbol{\mu}} \left[ \tau_\delta \right]}{\log\left( \frac{1}{2.4\delta} \right)},
\end{equation}
which is the smallest possible constant such that the expected sample complexity $\mathbb{E}_{\boldsymbol{\mu}} \left[ \tau_\delta \right]$ grows asymptotically in line with $\log\left( \frac{1}{2.4\delta} \right)$. The lower bound of the sample complexity $\mathbb{E}_{\boldsymbol{\mu}} \left[ \tau_\delta \right]$ can be represented in a general form by the following proposition.

\begin{proposition}\label{general lower bound}
 For any $\boldsymbol{\mu} \in M$, there exists a set $\mathcal{X} = \mathcal{X}(\boldsymbol{\mu})$ and functions $\{ C_x \}_{x \in \mathcal{X}}$ with $C_x: \mathcal{S}_K \times \mathcal{X} \rightarrow \mathbb{R}_+$ such that
\begin{equation}
    \kappa \left( \boldsymbol{\mu} \right) \geq \left( \Gamma^{\ast}_{\boldsymbol{\mu}} \right)^{-1},
    \label{def: general lower bound inequality}
\end{equation}
where
\begin{equation}\label{def: general lower bound}
    \Gamma^{\ast}_{\boldsymbol{\mu}} = \max_{\boldsymbol{p} \in \mathcal{S}_K} \min_{x \in \mathcal{X}} C_x \left( \boldsymbol{p}; \boldsymbol{\mu} \right ).
\end{equation}
\end{proposition}

In Proposition \ref{general lower bound}, $\mathcal{S}_K$ denotes the $K$-dimensional probability simplex, and $\mathcal{X} = \mathcal{X}(\boldsymbol{\mu})$ is referred to as the culprit set. This set comprises critical subqueries (or comparisons) that must be correctly resolved. An error in any of these comparisons may hinder the identification of the correct set. 

For example, in the case of identifying the best arm, the culprit set is given by $\mathcal{X}= \{ i : i \in [K] \setminus i^\ast \}$, where $i^\ast$ denotes the unique best arm, and each subquery involves distinguishing every arm from the best arm. In the threshold bandit problem, the culprit set consists of all arms, $\mathcal{X} = \{ i : i \in [K] \}$, where each subquery requires accurately determining whether each arm exceeds the threshold. For the task of identifying the best $m$ arms, the culprit set is $\mathcal{X} = \{ (i, j) : i \in \mathcal{I}, j \in \mathcal{I}^c \}$, where $\mathcal{I}$ represents the set of the best $m$ arms, and each subquery entails comparing the mean of each arm in $\mathcal{I}$ with those in the complement set $\mathcal{I}^c$. 

The function $C_x(\boldsymbol{p}; \boldsymbol{\mu})$ represents the population version of the sequential generalized likelihood ratio statistic, which provides an information-theoretic measure of how easily each subquery, corresponding to the culprit $x \in \mathcal{X}$, can be answered. 
In equation~\eqref{def: general lower bound}, the minimum aligns with the intuition that the instance posing the greatest challenge corresponds to the hardest subquery. The outer maximization seeks the optimal allocation of arms $\boldsymbol{p}$ to effectively address this subquery. For a more detailed introduction to this general pure exploration model, please refer to Section \ref{sec_General Pure Exploration Model} in the online appendix. 



\begin{theorem}[Lower Bound]\label{lower bound: All-epsilon in the linear setting}

    Consider a set of arms where arm $i$ follows a normal distribution $\mathcal{N}(\mu_i, 1)$, where $\mu_i = \boldsymbol{a}_{i}^\top \boldsymbol{\theta}$. Any $\delta$-PAC algorithm for identifying all $\varepsilon$-best arms in the linear bandit setting must satisfy 
    \begin{equation}\label{equation: lower bound 3}
        \inf_{Algo \in \mathcal{H}} \frac{\mathbb{E}_{\boldsymbol{\mu}} \left[ \tau_\delta \right]}{\log\left( \frac{1}{2.4\delta} \right)} \geq (\Gamma^\ast_{\boldsymbol{\mu}})^{-1} = \min_{\boldsymbol{p} \in \mathcal{S}_K} \max_{(i, j, m) \in \mathcal{X}} \max \left\{ \frac{2 \Vert \boldsymbol{a}_i - \boldsymbol{a}_j \Vert_{\boldsymbol{V}_{\boldsymbol{p}}^{-1}}^2}{\left( \boldsymbol{a}_i^\top \boldsymbol{\theta} - \boldsymbol{a}_j^\top \boldsymbol{\theta} + \varepsilon \right)^2}, \frac{2 \Vert \boldsymbol{a}_1 - \boldsymbol{a}_m \Vert_{\boldsymbol{V}_{\boldsymbol{p}}^{-1}}^2}{\left( \boldsymbol{a}_1^\top \boldsymbol{\theta} - \boldsymbol{a}_m^\top \boldsymbol{\theta} - \varepsilon \right)^2} \right\},
    \end{equation}
    where $\mathcal{X} = \{ (i, j, m) : i \in G_\varepsilon(\boldsymbol{\mu}), j \neq i, m \notin G_\varepsilon(\boldsymbol{\mu}) \}$, $\mathcal{S}_K$ is the $K$-dimensional probability simplex.
\end{theorem}

The detailed proof of the above theorem is presented in Section \ref{proof of theorem: lower bound for all-epsilon in the linear bandit}. We note that the stochastic bandit problem is a special case of the linear bandit problem. By setting $\mathcal{A} = \{ \boldsymbol{e}_1, \boldsymbol{e}_2, \ldots, \boldsymbol{e}_d \}$, where $\boldsymbol{e}_i$ denotes the unit vector, the linear bandit model reduces to a stochastic setting. This relationship allows us to recover the lower bound result for identifying all $\varepsilon$-best arms in stochastic bandits, which is detailed in Theorem \ref{lower bound: existing 3} in the online appendix. Furthermore, the lower bound in Theorem \ref{lower bound: All-epsilon in the linear setting} extends the lower bound for best arm identification in linear bandits . This result is recovered by setting $\varepsilon = 0$ and redefining the culprit set as $\mathcal{X}(\boldsymbol{\mu}) = \{ i : i \in [K] \setminus i^\ast \}$, where $i^\ast$ represents the best arm in the context of best arms identification.

\section{Algorithm and Upper Bound}\label{sec_Algorithm and Upper Bound} 




In this section, we propose the \textit{LinFACT} algorithm (\textbf{Lin}ear \textbf{F}ast \textbf{A}rm \textbf{C}lassification with \textbf{T}hreshold estimation) to efficiently identify all $\varepsilon$-best arms in linear bandit. We then establish upper bounds on the expected sample complexity to demonstrate the optimality of the LinFACT algorithm. Specifically, the upper bound derived from the $\mathcal{XY}$-optimal sampling policy is shown to be instance optimal up to logarithmic factors.\footnote{We refer to the algorithms based on G-optimal sampling and $\mathcal{XY}$-optimal sampling as LinFACT-G and LinFACT-$\mathcal{XY}$, respectively. A detailed comparison between the two approaches is provided in Section~\ref{Essential Difference between G and XY}.}

\subsection{Algorithm}\label{subsec_Linear Fast Arm Classification with Threshold Estimation} 

LinFACT is a phase-based, semi-adaptive algorithm. The sampling rule remains fixed during each round but is updated based on the accumulated observations at the end of each round. As round $r$ progresses, LinFACT refines two sets of arms:
\begin{itemize}
    \item $G_r$: Arms empirically classified as $\varepsilon$-best (good).
    \item $B_r$: Arms empirically classified as not $\varepsilon$-best (bad).
\end{itemize}

This classification process continues until all arms have been assigned to either $G_r$ or $B_r$. Once complete, the decision rule returns $G_r$ as the final set of $\varepsilon$-best arms.

\paragraph{Sampling Rule.}\label{subsubsec_Sampling Rule} 

To minimize the sampling budget, we select arms that provide the maximum information about the mean values or the gaps between them. Unlike stochastic bandits, where mean values are obtained exclusively by sampling specific arms, linear bandits allow these mean values to be inferred from the estimated parameters. In each round, arms are selected based on the G-optimal design (\ref{equation_G optimal}) or the $\mathcal{XY}$-optimal design (\ref{equation_XY optimal}). 

For G-optimal design, LinFACT-G refines an estimate of the true parameter $\boldsymbol{\theta}$ and uses this estimate to maintain an anytime confidence interval, such that for each arm's empirical mean value $\hat{\mu}_i$, we have 
\begin{equation}\label{eq_high_1}
    \mathbb{P} \left( \bigcap_{i \in \mathcal{A}_I(r-1)} \bigcap_{r \in \mathbb{N}} \left| \hat{\mu}_i(r) - \mu_i \right| \leq C_{\delta/K}(r) \right) \geq 1-\delta.
\end{equation}

The active set $\mathcal{A}(r)$ is defined as the set of uneliminated arms, as we continuously eliminate arms as round $r$ progresses. $\mathcal{A}_{I}(r)$ denotes the set of indices corresponding to the remaining arms in $\mathcal{A}(r)$. 

This confidence bound indicates that the algorithm maintains a probabilistic guarantee that the true mean value $\mu_i$ is within a certain range of the estimated mean value $\hat{\mu}_i$ for each arm $i$, uniformly over all rounds. The bound shrinks as more data is collected (since the confidence radius $C_{\delta/K}(r)$ decreases with more samples), thereby reducing uncertainty. The anytime confidence width $C_{\delta/K}(r)$ is maintained by the design of the sample budget in each round. We set $C_{\delta/K}(r) = 2^{-r} =: \varepsilon_r$, which is halved with each iteration of the rounds.

In LinFACT-G, the initial budget allocation policy is based on the G-optimal design and is defined as follows
\begin{equation}\label{equation_phase budget G}
\left\{
\begin{array}{l}
T_{r}(\boldsymbol{a}) = \left\lceil \dfrac{2d\pi_{r}(\boldsymbol{a})}{\varepsilon_{r}^2} \log\left( \dfrac{2Kr(r + 1)}{\delta} \right) \right\rceil \\
\\
T_{r} = \sum_{\boldsymbol{a} \in \mathcal{A}(r-1)} T_{r}(\boldsymbol{a})
\end{array},
\right.
\end{equation}
where $T_r$ denotes the total sampling budget allocated in round $r$, and $\pi_r$ is the selection probability distribution over the remaining active arms $\mathcal{A}(r-1)$ from the previous round, obtained via the G-optimal design as defined in equation \eqref{equation_G optimal}. The sampling procedure for each round $r$ is described in Algorithm \ref{alg_LinFACT_sampling}.

\vspace{0.2cm}
\begin{breakablealgorithm}\label{alg_LinFACT_sampling}
\caption{Subroutine: G-Optimal Sampling}
\begin{algorithmic}[1]
\State \textbf{Input:} Projected active set $\mathcal{A}_I(r-1)$, round $r$, $\delta$.
\State Obtain $\pi_r \in \mathcal{P}(\mathcal{A}(r-1))$ with support size $\text{Supp}(\pi_r) \leq \frac{d(d+1)}{2}$ according to equation \eqref{equation_G optimal}. \label{G_sampling_1}
\ForAll{$\boldsymbol{a} \in \mathcal{A}_I(r-1)$}
    \Comment{Sampling}
    \State Sample arm $\boldsymbol{a}$ for $T_r(\boldsymbol{a})$ times in round $r$, as specified in equation~\eqref{equation_phase budget G}. \label{G_sampling_2}
\EndFor
\end{algorithmic}
\end{breakablealgorithm}
\vspace{0.2cm}

For $\mathcal{XY}$-optimal design, we adopt this approach because the G-optimal design is not effective for distinguishing between different arms. While G-optimal design minimizes the maximum variance in estimating individual arm means, it does not directly target the pairwise differences that are crucial for identifying all $\varepsilon$-best arms. In contrast, the $\mathcal{XY}$-optimal design focuses on accurately estimating these gaps, making it better suited for this task.

We now introduce a sampling rule based on the $\mathcal{XY}$-optimal design, as defined in equation~\eqref{equation_XY optimal}. Let $q(\epsilon)$ denote the error introduced by the rounding procedure. We have: 
\begin{equation}\label{equation_phase budget XY}
\left\{
\begin{array}{l}
T_{r} = \max \left\{ \left\lceil \dfrac{2 g_{\mathcal{XY}}\left( \mathcal{Y}(\mathcal{A}(r-1)) \right) (1+\epsilon)}{\varepsilon_r^2} \log \left( \dfrac{2K(K-1)r(r+1)}{\delta} \right) \right\rceil, q\left( \epsilon \right) \right\} \\
\\
T_{r}(\boldsymbol{a}) = \text{ROUND}(\pi_r, T_r)
\end{array}.
\right.
\end{equation}

In contrast to the G-optimal design, the $\mathcal{XY}$-optimal design focuses on bounding the confidence region of the pairwise differences between arms. The following inequality characterizes the corresponding high-probability event.
\begin{equation}
\mathbb{P}\left(  \bigcap_{i \in \mathcal{A}_I(r-1)} \bigcap_{j \in \mathcal{A}_I(r-1),\, j \neq i} \bigcap_{r \in \mathbb{N}} \left| (\hat{\mu}_j(r) - \hat{\mu}_i(r)) - (\mu_j - \mu_i) \right| \leq 2C_{\delta/K}(r) \right) \geq 1-\delta.
\end{equation}

The rounding operation, denoted as \textsc{Round}, uses a $(1+\epsilon)$ approximation algorithm proposed by \citet{allenzhu2017nearoptimal}. The complete sampling procedure is outlined in Algorithm~\ref{alg_LinFACT_sampling_XY}. 

\vspace{0.2cm}
\begin{breakablealgorithm}\label{alg_LinFACT_sampling_XY}
\caption{Subroutine: $\mathcal{XY}$-Optimal Sampling}
\begin{algorithmic}[1]
\State \textbf{Input:} Projected active set $\mathcal{A}_I(r-1)$, round $r$, $\delta$.
\State Obtain $\pi_r \in \mathcal{P}(\mathcal{A}(r-1))$ according to equation \eqref{equation_XY optimal}.
\ForAll{$\boldsymbol{a} \in \mathcal{A}_I(r-1)$}
    \Comment{Sampling}
    \State Sample arm $\boldsymbol{a}$ for $T_r(\boldsymbol{a})$ times in round $r$, as specified in equation~ \eqref{equation_phase budget XY}. \label{XY_sampling}
\EndFor
\end{algorithmic}
\end{breakablealgorithm}
\vspace{0.2cm}

\paragraph{Estimation.}\label{subsubsec_Estimation} 

In each round, we sample each arm in the active set $T_r(\boldsymbol{a})$ times and compute the empirical estimate using standard ordinary least squares (OLS) as
\begin{equation}\label{esti_OLS}
    \hat{\boldsymbol{\theta}}_{r} = \boldsymbol{V}_{r}^{-1} \sum_{s=1}^{T_{r}} \boldsymbol{a}_{A_s} X_s,
\end{equation}
where
$
    \boldsymbol{V}_{r} = \sum_{\boldsymbol{a} \in \mathcal{A}(r-1)} T_{r}(\boldsymbol{a}) \boldsymbol{a} \boldsymbol{a}^\top
$
is the information matrix. The estimator for the mean value of each arm $i \in \mathcal{A}_I(r-1)$ is then
\begin{equation}\label{eq_estimation_1}
    \hat{\mu}_i = \hat{\mu}_i(r) = \boldsymbol{a}_i^\top \hat{\boldsymbol{\theta}}_r.
\end{equation}

\paragraph{Stopping Rule and Decision Rule.}\label{subsubsec_Stopping Rule and Decision Rule} 

As round $r$ progresses, LinFACT dynamically updates two sets of arms: $G_r$ and $B_r$, representing arms that are empirically considered $\varepsilon$-best (good) and those that are not (bad), respectively. The algorithm filters arms by maintaining an upper confidence bound $U_r$ and a lower confidence bound $L_r$ around the unknown threshold $\mu_1 - \varepsilon$, along with individual upper and lower confidence bounds for each arm. 

The stopping rule and final decision procedure are described in Algorithm~\ref{alg_LinFACT_stopping}. For each arm $i$ in the active set, LinFACT eliminates the arm if its upper confidence bound falls below the threshold $L_r$ (line~\ref{elimination: good 1}). Conversely, if the lower confidence bound of an arm exceeds $U_r$ (line~\ref{add to Gr}), the arm is added to the set $G_r$. Additionally, any arm already in $G_r$ will be removed from the active set if its upper bound falls below the empirically largest lower bound among all active arms (line~\ref{elimination: good 2}). This ensures that the best arm is always retained in the active set, which is necessary for estimating the threshold $\mu_1 - \varepsilon$. The classification process continues until all arms are categorized, that is, when $G_r \cup B_r = [K]$. At termination, the set $G_r$ is returned as the output of LinFACT, representing the arms identified as $\varepsilon$-best.

\vspace{0.2cm}
\begin{breakablealgorithm}\label{alg_LinFACT_stopping}
\caption{Subroutine: Stopping Rule and Decision Rule}
\begin{algorithmic}[1]
\State \textbf{Input:} Projected active set $\mathcal{A}_I(r-1)$, estimator $\left( \hat{\mu}_i(r) \right)_{i \in \mathcal{A}_I(r-1)}$, round $r$, $\varepsilon$, confidence radius $C_{\delta/K}(r)$.
\State Let $U_r = \max_{i \in \mathcal{A}_I(r-1)} \hat{\mu}_i + C_{\delta/K}(r) - \varepsilon$.
\State Let $L_r = \max_{i \in \mathcal{A}_I(r-1)} \hat{\mu}_i - C_{\delta/K}(r) - \varepsilon$.
\ForAll{$i \in \mathcal{A}_I(r-1)$}
    \Comment{Arm Classification and Elimination}
    \If{$\hat{\mu}_i + C_{\delta/K}(r) < L_r$}
        \State Add $i$ to $B_r$ and eliminate $i$ from $\mathcal{A}_I(r-1)$. \label{elimination: good 1}
    \EndIf
    \If{$\hat{\mu}_i - C_{\delta/K}(r) > U_r$}
        \State Add $i$ to $G_r$. \label{add to Gr}
    \EndIf
    \If{$i \in G_r$ \textbf{and} $\hat{\mu}_i + C_{\delta/K}(r) \leq \max_{j \in \mathcal{A}_I(r-1)} \hat{\mu}_j - C_{\delta/K}(r)$}
        \State Eliminate $i$ from $\mathcal{A}_I(r-1)$. \label{elimination: good 2}
    \EndIf
\EndFor
\If{$G_r \cup B_r = [K]$}
    \Comment{Stopping Condition and Recommendation}
    \State \textbf{Output:} the set $G_r$. \label{stopping condition 1 and 2}
\EndIf
\end{algorithmic}
\end{breakablealgorithm}
\vspace{0.2cm}

\paragraph{The Complete LinFACT Algorithm.}\label{subsubsec_LinFACT} 

The complete LinFACT algorithm is presented in Algorithm~\ref{alg_LinFACT}. The procedure proceeds as follows: based on the collected data, the decision-maker updates the parameter estimates and checks whether the stopping condition $\tau_\delta$ is satisfied. If so, the set $G_r$ is returned as the estimated set of all $\varepsilon$-best arms. If the stopping condition is not met, the process continues with further sampling and updates. 


\vspace{0.2cm}
\begin{breakablealgorithm}\label{alg_LinFACT}
\caption{LinFACT Algorithm}
\begin{algorithmic}[1]
\State \textbf{Input:} $\varepsilon$, $\delta$, bandit instance.
\State Initialize $G_0 = \emptyset$, the set of identified $\varepsilon$-best arms, and $B_0 = \emptyset$, the set of arms that are not $\varepsilon$-best.
\State Initialize the active set $\mathcal{A}(0) = \mathcal{A}$, and $\mathcal{A}_{I}(0) = [K]$.
\For{$r = 1, 2, \ldots$}
    \State Set $C_{\delta/K}(r) = \varepsilon_r = 2^{-r}$.
    \State Set $G_r = G_{r-1}$ and $B_r = B_{r-1}$.
    \State Project $\mathcal{A}(r-1)$ to a $d_r$-dimensional subspace that $\mathcal{A}(r-1)$ spans.
    \Comment{Projection}
    \If{Using G-optimal Sampling}
        \Comment{Sampling}
        \State Call Algorithm \ref{alg_LinFACT_sampling}.
    \ElsIf{Using $\mathcal{XY}$-optimal Sampling}
        \State Call Algorithm \ref{alg_LinFACT_sampling_XY}.
    \EndIf
    \State Estimate $\left( \hat{\mu}_i(r) \right)_{i \in \mathcal{A}_I(r-1)}$ using equations \eqref{esti_OLS} and \eqref{eq_estimation_1}.
    \Comment{Estimation}
    \State Call Algorithm \ref{alg_LinFACT_stopping}.
    \Comment{Stopping Condition and Decision Rule}
\EndFor
\end{algorithmic}
\end{breakablealgorithm}
\vspace{0.2cm}

\subsection{Upper Bounds of the LinFACT Algorithm}\label{subsec_Upper Bounds on the Sample Complexity of LinFACT Algorithm} 

Theorems~\ref{upper bound: Algorithm 1 G} and~\ref{upper bound: Algorithm 1 XY} establish upper bounds on the sample complexity of the proposed LinFACT algorithm. 
Let $T_G$ and $T_{\mathcal{XY}}$ denote the number of samples required under the G-optimal and $\mathcal{XY}$-optimal designs, respectively. The formal statements of these theorems are given below. 

\begin{theorem}[Upper Bound, G-Optimal Design]\label{upper bound: Algorithm 1 G}
    For universal constants $c$, $\xi$, and $\zeta$, there exists an event $\mathcal{E}$ such that $\mathbb{P}(\mathcal{E}) \geq 1-\delta$. On this event, the \textup{LinFACT} algorithm with the G-optimal sampling policy achieves an expected sample complexity upper bound given by
    \begin{align}\label{upper bound for algorithm1: 1}
        \mathbb{E}_{\boldsymbol{\mu}}\left[T_G \mid \mathcal{E}\right]
        &\leq c \cdot \frac{256d}{\alpha_\varepsilon^2} \log \left( \frac{2K}{\delta} \log_2 \left( \frac{16}{\alpha_\varepsilon} \right) \right) + \frac{d(d+1)}{2} \log_2 \left( \frac{8}{\alpha_\varepsilon} \right) \nonumber\\
        &\quad + \sum_{i \in G_\varepsilon^c} \left( \frac{d(d+1)}{2} \log_2 \left( \frac{8}{\Delta_i - \varepsilon} \right) + \xi \cdot \frac{256d}{(\Delta_i - \varepsilon)^2} \log \left( \frac{2K}{\delta} \log_2 \frac{16}{\Delta_i - \varepsilon} \right) \right).
    \end{align}
    
    Another bound, which removes the summation over the complement set $G_\varepsilon^c$, takes the form
    \begin{equation}\label{upper bound for algorithm1: 2}
        \mathbb{E}_{\boldsymbol{\mu}}\left[T_G \mid \mathcal{E}\right] \leq \zeta \cdot \max \left\{ \frac{256d}{\alpha_\varepsilon^2} \log \left( \frac{2K}{\delta} \log_2 \frac{16}{\alpha_\varepsilon} \right), \frac{256d}{\beta_\varepsilon^2} \log \left( \frac{2K}{\delta} \log_2 \frac{16}{\beta_\varepsilon} \right) \right\} + \frac{d(d+1)}{2} R_{\textup{upper}},
    \end{equation}
    where $R_{\textup{upper}} = \max \left\{ \left\lceil \log_2 \frac{4}{\alpha_\varepsilon} \right\rceil, \left\lceil \log_2 \frac{4}{\beta_\varepsilon} \right\rceil \right\}$ is the round in which all classifications have been completed and the answer is returned.
\end{theorem}

However, an algorithm based on the G-optimal design does not achieve a matching upper bound. In contrast, we demonstrate that an algorithm using the $\mathcal{XY}$-optimal design matches the lower bound, optimal up to a logarithmic factor.

\begin{theorem}[Upper Bound, $\mathcal{XY}$-Optimal Design]\label{upper bound: Algorithm 1 XY}
    Assume that an instance of arms satisfies $\min_{i \in G_\varepsilon \setminus \{ 1 \}} \Vert \boldsymbol{a}_1 - \boldsymbol{a}_i \Vert^2 \geq L_2$ and $\max_{i \in [K]} \vert \mu_1 - \varepsilon - \mu_i \vert \leq 2$. There exists an event $\mathcal{E}$ such that $\mathbb{P}(\mathcal{E}) \geq 1-\delta$. On this event, the \textup{LinFACT} algorithm with the $\mathcal{XY}$-optimal sampling policy achieves an expected sample complexity upper bound given by
    \begin{equation}\label{upper bound for algorithm1: 3}
        \mathbb{E}_{\boldsymbol{\mu}}\left[T_{\mathcal{XY}} \mid \mathcal{E}\right] \leq c \cdot \left[ d \cdot R_{\textup{upper}} \cdot \log \left( \frac{2K(R_{\textup{upper}}+1)}{\delta} \right) \right] \cdot (\Gamma^\ast)^{-1} + q\left( \epsilon \right) \cdot R_{\textup{upper}},
    \end{equation}
    where $c$ is a universal constant, $R_{\textup{upper}} = \max \left\{ \left\lceil \log_2 \frac{4}{\alpha_\varepsilon} \right\rceil, \left\lceil \log_2 \frac{4}{\beta_\varepsilon} \right\rceil \right\}$, and $(\Gamma^\ast)^{-1}$ is the lower bound term defined in Theorem \ref{lower bound: All-epsilon in the linear setting}.
\end{theorem}

The detailed proof of Theorem~\ref{upper bound: Algorithm 1 G} is provided in Section~\ref{proof of theorem: upper bound of LinFACT G}. Building on the insights into algorithmic optimality discussed in Section~\ref{additional insights}, the proof of the near-optimal upper bound in Theorem~\ref{upper bound: Algorithm 1 XY} is presented in Section~\ref{proof of theorem: upper bound of LinFACT XY} of the online appendix.

\section{Model Misspecification}\label{subsec_Misspecified Linear Bandits}

In this section, we address the challenge of model misspecification in the linear bandit setting, recognizing that real-world problems may deviate from exact linearity. To account for such deviations, we propose an orthogonal parameterization-based algorithm, \emph{i.e.}, LinFACT-MIS,\footnote{LinFACT-MIS builds upon LinFACT-G. We choose to extend LinFACT-G, rather than LinFACT-$\mathcal{XY}$, for reasons of analytical tractability and computational efficiency. Both variants exhibit comparable performance in terms of upper bounds in this setting, while G-optimal design offers a significantly lower computational burden.} a refined version of LinFACT to address model misspecification. We establish new upper bounds in the misspecified setting and provide insights into how such deviations impact algorithm performance. 

Under model misspecification, we refine the linear model in equation \eqref{eq_linear_model} as
\begin{equation}\label{eq_linear_model_mis}
    X_t = \boldsymbol{a}_{A_t}^\top \boldsymbol{\theta} + \eta_t + \boldsymbol{\Delta_m}(\boldsymbol{a}_{A_t}),
\end{equation}
where $\boldsymbol{\Delta_m}: \mathbb{R}^d \rightarrow \mathbb{R}$ is a misspecification function, and the associated misspecification vector satisfies $\Vert \boldsymbol{\Delta_m} \Vert_\infty \leq L_m$, quantifying the deviation from the true linear model. The set of realizable models is defined as
\begin{equation}\label{eq_mis_realizable_model}
    M \coloneqq \left\{ \boldsymbol{\mu} \in \mathbb{R}^K \,\middle|\, \exists \boldsymbol{\theta} \in \mathbb{R}^d, \exists \boldsymbol{\Delta_m} \in \mathbb{R}^K, \boldsymbol{\mu} = \boldsymbol{\Psi}\boldsymbol{\theta} + \boldsymbol{\Delta_m} \; \land \; \Vert \boldsymbol{\mu} \Vert_\infty \leq L_\infty \; \land \; \Vert \boldsymbol{\Delta_m} \Vert_\infty \leq L_m \right\}.
\end{equation}

The key distinction in the analysis under model misspecification lies in how the estimator $\hat{\boldsymbol{\mu}}_t$ is maintained. Specifically, we construct this estimator by projecting the empirical mean vector $\tilde{\boldsymbol{\mu}}_t$ at time $t$ onto the set of realizable models $M$ via the following optimization
\begin{equation}\label{eq_optimization_esti}
    \hat{\boldsymbol{\mu}}_t \coloneqq \arg\min_{\boldsymbol{\vartheta} \in M} \Vert \boldsymbol{\vartheta} - \tilde{\boldsymbol{\mu}}_t \Vert_{\boldsymbol{D}_{\boldsymbol{N}_t}}^2,
\end{equation}
where $\boldsymbol{N}_t = [N_{t1}, N_{t2}, \ldots, N_{tK}]^\top \in \mathbb{R}^K$ is the vector of sample counts for each arm at time $t$, and $\boldsymbol{D}_{\boldsymbol{N}_t} \in \mathbb{R}^{K \times K}$ is the diagonal matrix with $N_{t1}, N_{t2}, \ldots, N_{tK}$ as its diagonal entries.


\begin{figure}[htbp]
    \centering
    \includegraphics[scale=1]{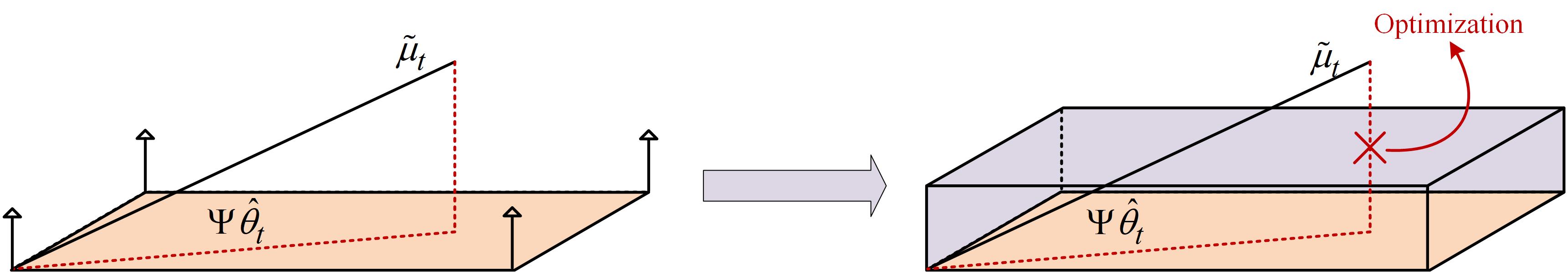}
    \caption{Difference Between Standard OLS and Misspecification-Adjusted Projection Estimates}
\vspace{5pt}
\noindent
\begin{minipage}{\linewidth}
    \footnotesize
    \raggedright
    \textit{Note.} \textit{The left diagram shows the projection onto the span of pulled arms under a perfect linear model. The right diagram depicts the adjustment required under misspecification, where the projection must account for the deviation.
    }
\end{minipage}
    \label{fig: mis_2to3}
\end{figure}

In the absence of misspecification, this projection simplifies to the ordinary least squares (OLS) estimator. However, as shown in Figure~\ref{fig: mis_2to3}, under model misspecification the estimator can no longer be computed as a simple projection onto a hyperplane and falls outside the scope of standard OLS. Instead, it must be formulated as the optimization problem in equation~\eqref{eq_optimization_esti}, which minimizes a weighted quadratic objective over $K + d$ variables, subject to the constraints $\Vert \boldsymbol{\mu} \Vert_\infty \leq L_\infty$ and $\Vert \boldsymbol{\Delta_m} \Vert_\infty \leq L_m$.

\subsection{Upper Bound with Misspecification}\label{subsec_Upper Bound When the Model Misspecification is Considered}

In this subsection, we present the upper bound on the expected sample complexity for LinFACT-G under model misspecification. This analysis highlights the influence of misspecification on the theoretical performance of the algorithm. Let $T_{G_{\textup{mis}}}$ denote the number of samples taken using the G-optimal design under model misspecification. 

\begin{theorem}[Upper Bound, Misspecification]\label{upper bound: Algorithm 1 G_mis_1}
    Fix $\varepsilon > 0$ and assume that the magnitude of misspecification satisfies $L_m \leq \min \left\{ \frac{\alpha_\varepsilon}{2 \sqrt{d}}, \frac{\beta_\varepsilon}{2 \sqrt{d}} \right\}$. For universal constant $c$, there exists an event $\mathcal{E}$ such that $\mathbb{P}(\mathcal{E}) \geq 1-\delta$. On this event, LinFACT with the G-optimal sampling policy terminates and returns the set $G_\varepsilon$ with an expected sample complexity upper bound given by
    \begin{align}\label{eq_mis_sample_complexity_mis}
        \mathbb{E}_{\boldsymbol{\mu}}\left[T_{G_{\textup{mis}}} \mid \mathcal{E}\right] &\leq c \cdot \max\bigg\{ \frac{256d}{(\alpha_\varepsilon - 2L_m \sqrt{d})^2} \log \left( \frac{2K}{\delta} \log_2 \frac{16}{\alpha_\varepsilon - 2L_m \sqrt{d}} \right), \nonumber\\
        &\quad \frac{256d}{(\beta_\varepsilon - 2L_m \sqrt{d})^2} \log \left( \frac{2K}{\delta} \log_2 \frac{16}{\beta_\varepsilon - 2L_m \sqrt{d}} \right) \bigg\} + \frac{d(d+1)}{2} R_{\text{upper}}^\prime,
    \end{align}
    where $R_{\text{upper}}^\prime = \max \left\{ \left\lceil \log_2 \frac{4}{\alpha_\varepsilon - 2L_m \sqrt{d}} \right\rceil, \left\lceil \log_2 \frac{4}{\beta_\varepsilon - 2L_m \sqrt{d}} \right\rceil \right\}$ is the round in which all classifications have been completed and the answer is returned under model misspecification.
\end{theorem}

The proof of this theorem is provided in Section~\ref{sec_Proof of upper bound: Algorithm 1 G_mis_1}. The upper bound in Theorem~\ref{upper bound: Algorithm 1 G}, which assumes no model misspecification, can be viewed as a special case of Theorem~\ref{upper bound: Algorithm 1 G_mis_1} by setting $L_m = 0$. However, the bound in this theorem ceases to hold when the misspecification magnitude $L_m$ is too large, as the expressions inside the logarithmic terms become negative, violating the conditions necessary for the bound to apply. If the variation in confidence radius across arms due to misspecification is ignored and no modifications are made to the algorithm, Theorem~\ref{upper bound: Algorithm 1 G_mis_1} implies that the resulting sample complexity will increase.

\subsection{Refined Upper Bound Using Orthogonal Parameterization}\label{subsec_Orthogonal Parameterization-Based LinFACT and Upper Bound}

In this section, we present an enhanced version of LinFACT based on orthogonal parameterization, designed to improve both theoretical guarantees and computational efficiency. 

\paragraph{Orthogonal Parameterization.}\label{subsubsec_Orthogonal Parameterization}

\begin{figure}[htbp]
    \centering
    \includegraphics[scale=1]{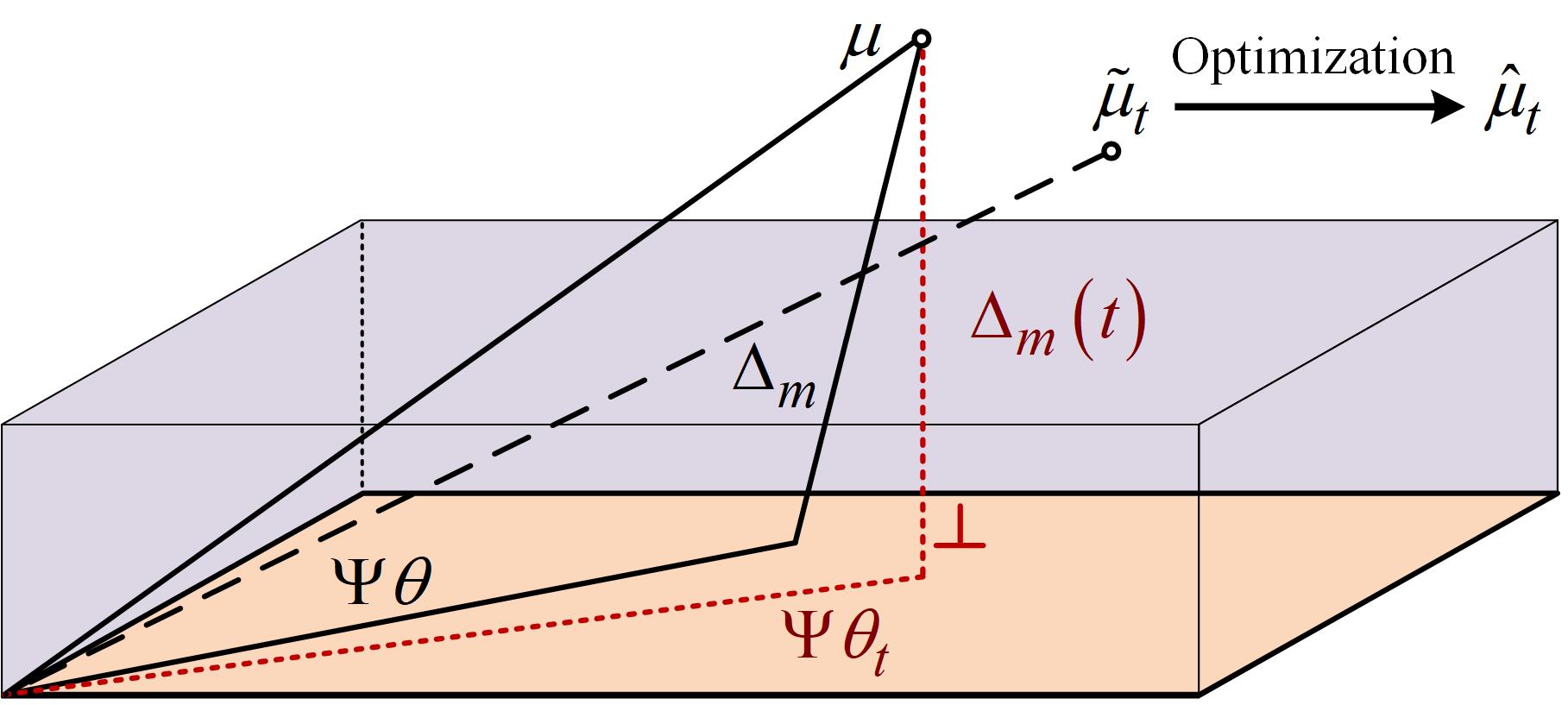}
    \caption{Orthogonal Parameterization and Projection.}
\vspace{5pt}
\noindent
\begin{minipage}{\linewidth}
    \footnotesize
    \raggedright
    \textit{Note.} \textit{The estimator $\hat{\boldsymbol{\mu}}_t$ is obtained from the empirical mean $\tilde{\boldsymbol{\mu}}_t$ by solving an optimization problem. While the true mean vector $\boldsymbol{\mu}$ can be expressed as the sum of a linear component $\boldsymbol{\Psi\theta}$ and a non-linear model deviation $\boldsymbol{\Delta}_m$, it can also be decomposed at each time step $t$ via orthogonal projection into a linear part $\boldsymbol{\Psi\theta}_t$ on the hyperplane and a residual term $\boldsymbol{\Delta}_m(t)$ orthogonal to it. 
    }
\end{minipage}
    \label{fig: ortho}
\end{figure}

Specifically, we show that any mean vector $\boldsymbol{\mu} = \boldsymbol{\Psi}\boldsymbol{\theta} + \boldsymbol{\Delta_m}$ can be equivalently expressed at any time $t$ as $\boldsymbol{\mu} = \boldsymbol{\Psi}\boldsymbol{\theta}_t + \boldsymbol{\Delta_{m}(t)}$, where
\begin{equation}
    \boldsymbol{\theta}_t = \left( \boldsymbol{\Psi}_{N_t}^\top \boldsymbol{\Psi}_{N_t} \right)^{-1} \boldsymbol{\Psi}_{N_t}^\top \boldsymbol{D}_{N_t}^{1/2} \boldsymbol{\mu} = \boldsymbol{V}_t^{-1} \sum_{s=1}^t \mu_{A_s} \boldsymbol{a}_{A_s}
\end{equation}
is the orthogonal projection of $\boldsymbol{\mu}$ onto the feature space spanned by the columns of $\boldsymbol{\Psi}_{N_t}$, and $\boldsymbol{\Delta_{m}(t)} = \boldsymbol{\mu} - \boldsymbol{\Psi}\boldsymbol{\theta}_t$ is the residual. Here, $\boldsymbol{\Psi}_{N_t} = \boldsymbol{D}_{N_t}^{1/2} \boldsymbol{\Psi}$ is the matrix of feature vectors weighted by the number of times each arm has been sampled up to time $t$, and $\boldsymbol{D}_{N_t}^{1/2}$ is a diagonal matrix with entries corresponding to the square roots of the number of samples for each arm.


\paragraph{Refined Upper Bound.}\label{subsubsec_Upper Bound for the Orthogonal Parameterization-Based LinFACT}

For LinFACT, the orthogonal parameterization-based refinement involves updating the sampling rule in Algorithm \ref{alg_LinFACT_sampling} to the sampling rule in Algorithm \ref{alg_LinFACT_sampling_ortho}. The estimation is no longer based on OLS but is achieved by calculating the estimator $\left( \hat{\mu}_i(r) \right)_{i \in \mathcal{A}_I(r-1)}$ with observed data by solving the optimization problem described in equation~\eqref{eq_optimization_esti} directly. 

When model misspecification is accounted for and orthogonal parameterization is applied, the sampling policy is given by:
\begin{equation}\label{equation_phase budget G_mis_ortho}
    \left\{
    \begin{array}{l}
    T_{r}(\boldsymbol{a}) = \left\lceil \dfrac{8d\pi_{r}(\boldsymbol{a})}{\varepsilon_{r}^2} \left( d\log(6) + \log\left(\dfrac{Kr(r + 1)}{\delta}\right) \right) \right\rceil \\
    \\
    T_{r} = \sum_{\boldsymbol{a} \in \mathcal{A}(r-1)} T_r(\boldsymbol{a})
    \end{array}.
    \right.
\end{equation}



Let $T_{G_{\textup{op}}}$ denote the total number of samples required when orthogonal parameterization is used. The corresponding upper bound is stated in the theorem below. 

\vspace{0.2cm}
\begin{breakablealgorithm}\label{alg_LinFACT_sampling_ortho}
\caption{Subroutine: Sampling With Orthogonal Parameterization}
\begin{algorithmic}[1]
\State \textbf{Input:} Projected active set $\mathcal{A}_I(r-1)$, round $r$, $\delta$.
    \State Find the G-optimal design $\pi_r \in \mathcal{P}(\mathcal{A}(r-1))$ with Supp$(\pi_r) \leq \frac{d(d+1)}{2}$ according to equation \eqref{equation_G optimal}.\label{G_sampling_1_ortho}
    \ForAll{$i \in \mathcal{A}_I(r-1)$}
    \Comment{Sampling}
        \State Sample arm $\boldsymbol{a}$ for $T_r(\boldsymbol{a})$ times in round $r$, as specified in equation \eqref{equation_phase budget G_mis_ortho}. \label{G_sampling_2_ortho}
    \EndFor
\end{algorithmic}
\end{breakablealgorithm}
\vspace{0.2cm}

\begin{theorem}[Upper Bound, Orthogonal Parameterization]\label{upper bound: Algorithm 1 G_mis_2}
    Fix $\varepsilon > 0$ and assume that the magnitude of misspecification satisfies $L_m < \min \left\{ \frac{\alpha_\varepsilon}{2 \left( \sqrt{d} + 2 \right)}, \frac{\beta_\varepsilon}{2 \left( \sqrt{d} + 2 \right)} \right\}$. For universal constant $c$, there exists an event $\mathcal{E}$ such that $\mathbb{P}(\mathcal{E}) \geq 1-\delta$. On this event, LinFACT-MIS terminates and returns the set $G_\varepsilon$ with an expected sample complexity upper bound given by
    \begin{align}\label{eq_mis_sample_complexity_orho_2}
        \mathbb{E}_{\boldsymbol{\mu}}\left[T_{G_{\textup{op}}} \mid \mathcal{E}\right] &\leq c \cdot \max\bigg\{ \frac{256d}{(\alpha_\varepsilon - 2L_m (\sqrt{d} + 2 ))^2} \log \left( \frac{K 6^d}{\delta} \log_2 \frac{16}{\alpha_\varepsilon - 2L_m (\sqrt{d} + 2 )} \right), \nonumber\\
        &\quad \frac{256d}{(\beta_\varepsilon - 2L_m (\sqrt{d} + 2 ))^2} \log \left( \frac{K 6^d}{\delta} \log_2 \frac{16}{\beta_\varepsilon - 2L_m (\sqrt{d} + 2 )} \right) \bigg\} + \frac{d(d+1)}{2} R_{\text{upper}}^{\prime\prime},
    \end{align}
    where $R_{\text{upper}}^{\prime\prime} = \max \left\{ \left\lceil \log_2 \frac{4}{\alpha_\varepsilon - 2L_m \left( \sqrt{d} + 2 \right)} \right\rceil, \left\lceil \log_2 \frac{4}{\beta_\varepsilon - 2L_m \left( \sqrt{d} + 2 \right)} \right\rceil \right\}$ is the round in which the answer is returned.
\end{theorem}

This theorem shows that the upper bound remains of the same order as in Theorem~\ref{upper bound: Algorithm 1 G_mis_1}, with the detailed proof provided in Section~\ref{sec_Proof of upper bound: Algorithm 1 G_mis_2}. However, there is a noticeable worsening in the constant terms. We provide further intuition in Section~\ref{subsubsec_Improvement to Unknown Misspecification is theoretically Impossible}, where we argue that without prior knowledge of the misspecification, it is fundamentally impossible to recover full performance through algorithmic refinement alone.

\subsection{Insights for Model Misspecification}\label{subsec_General Insights Related to the Misspecified Linear Bandits in the Pure Exploration Setting}

In this section, we provide insights regarding misspecified linear bandits in the pure exploration setting.

\paragraph{Lower Bounds in Linear and Stochastic Settings.}\label{subsubsec_Relationship Between the Lower Bounds in the Linear Setting and the Stochastic Setting}

\begin{proposition}\label{propos_Relationship Between the Lower Bounds in the Linear Setting and the Stochastic Setting}
    There exists $L_{\bm{\mu}} \in \mathbb{R}$ with $L_{\bm{\mu}} \leq \max_{k} \mu_{k} - \min_{k} \mu_{k}$ such that if $L_m > L_{\bm{\mu}}$, then for any pure exploration task, the lower bound in the linear setting is equal to the unstructured lower bound.
\end{proposition}

\paragraph{Improvement with Unknown Misspecification is Theoretically Impossible.}\label{subsubsec_Improvement to Unknown Misspecification is theoretically Impossible}

\paragraph{Prior Knowledge for the Misspecification.}\label{subsubsec_Prior Knowledge for the Misspecification}

When the upper bound $L_m$ on model misspecification is known in advance, LinFACT can be modified to account for this deviation. Specifically, we adjust the confidence radius $C_{\delta/K}(r)$ used in computing the lower and upper bounds (\emph{i.e.}, $L_r$ and $U_r$) in Algorithm~\ref{alg_LinFACT}. With this modification, the number of rounds required to complete classification under misspecification, $R_{\text{upper}}^\prime$ and $R_{\text{upper}}^{\prime\prime}$, coincides with $R_{\textup{upper}}$, the corresponding number of rounds under a perfectly linear model. This is achieved by replacing the confidence radius $\varepsilon_r$ with an inflated version $\varepsilon_r + L_m \sqrt{d}$ to compensate for the worst-case deviation due to misspecification. This adjustment preserves the validity of the original analysis and ensures that the same theoretical guarantees are retained. The following proposition formalizes this observation and is proved in Section~\ref{sec_Proof of propos_prior}.

\begin{proposition}\label{propos_prior}
   Suppose the misspecification magnitude $L_m > 0$ is known in advance. Adjusting the confidence radius $C_{\delta/K}(r)$ in Algorithm \ref{alg_LinFACT} at each round $r$ from its original value $\varepsilon_r$ to
\begin{equation}
    \varepsilon_r^\prime = \varepsilon_r + L_m \sqrt{d},
\end{equation}
ensures that the total number of rounds required under misspecification matches that under perfect linearity.
\end{proposition}

\section{Generalized Linear Model}\label{subsec_Generalized Linear Model}

In this section, we extend the linear bandit to a generalized linear model (GLM). In this setting, the reward function no longer follows the standard linear form in equation \eqref{eq_linear_model}, but instead satisfies
\begin{equation}
    \mathbb{E} \left[ X_t \mid A_t \right] = \mu_{\text{link}}(\bm{a}_{A_t}^\top\bm{\theta}),
\end{equation}
where $\mu_{\text{link}}:\mathbb{R}\rightarrow\mathbb{R}$ is the inverse link function. GLMs encompass a class of models that include, but are not limited to, linear models, allowing for various reward distributions beyond the Gaussian. For example, for binary-valued rewards, a suitable choice of $\mu_{\text{link}}$ is $\mu_{\text{link}}(x) = \exp(x)/(1+\exp(x))$, \emph{i.e.}, sigmoid function, leading to the logistic regression model. For integer-valued rewards,  $\mu_{\text{link}}(x) = \exp(x)$ leads to the Poisson regression model. 

\begin{equation}\label{eq_exp_family}
    p_{\omega}(x) = \exp \left( x\omega - b(\omega) + c(x) \right),
\end{equation}
where $\omega$ is a real parameter, $c(\cdot)$ is a real normalization function, and $b(\cdot)$ is assumed to be twice continuously differentiable. This family includes the Gaussian and Gamma distributions when the reference measure is the Lebesgue measure, and the Poisson and Bernoulli distributions when the reference measure is the counting measure on the integers. For a random variable $X$ with density defined in \eqref{eq_exp_family}, $\mathbb{E}(X) = \dot{b}(\omega)$ and $\text{Var}(X) = \ddot{b}(\omega)$, where $\dot{b}$ and $\ddot{b}$ denote the first and second derivatives of $b$, respectively. Since the variance is always positive and $\mu_{\text{link}} = \dot{b}$ represents the inverse link function, $b$ is strictly convex and $\mu_{\text{link}}$ is increasing. 

The canonical GLM assumes that $p_{\bm{\theta}}(X \mid \bm{a}_i) = p_{\bm{a}_i^\top\bm{\theta}}(X)$ for all arms $i$. The maximum likelihood estimator $\hat{\bm{\theta}}_t$, based on $\sigma$-algebra $\mathcal{F}_t = \sigma\left(A_1, X_1, A_2, X_2, \ldots, A_t, X_t\right)$, is defined as the maximizer of the function
\begin{equation}
    \sum_{s=1}^{t} \log p_{\bm{\theta}}(X_s|\bm{a}_{A_s}) = \sum_{s=1}^{t} X_s \bm{a}_{A_s}^\top \bm{\theta} - b(\bm{a}_{A_s}^\top \bm{\theta}) + c(X_s).
\end{equation}

This function is strictly concave in $\bm{\theta}$. By differentiating, we obtain that $\hat{\bm{\theta}}_t$ is the unique solution of the following estimating equation at time $t$,
\begin{equation}\label{eq_GLM_mle}
    \sum_{s=1}^{t} \left( X_s - \mu_{\text{link}}(\bm{a}_{A_s}^\top \bm{\theta}) \right) \bm{a}_{A_s}=\bm{0}.
\end{equation}


\subsection{Algorithm with GLM}\label{subsubsec_GLM-Based LinFACT}
In this section, we present a refined version of LinFACT for the generalized linear model, referred to as LinFACT-\textup{GLM}. This refinement involves modifying the sampling rule in Algorithm \ref{alg_LinFACT_sampling} and adjusting the estimation method. The designed sampling policy is described by
\begin{equation}\label{equation_phase budget G_GLM}
\left\{
\begin{array}{l}
    T_{r}(\boldsymbol{a}) = \left\lceil \dfrac{2d\pi_{r}(\boldsymbol{a})}{\varepsilon_{r}^2 c_{\min}^2} \log\left(\dfrac{2Kr(r + 1)}{\delta}\right) \right\rceil \\ \\
    T_{r} = \sum_{\boldsymbol{a} \in \mathcal{A}(r-1)} T_r(\boldsymbol{a})
\end{array},
\right.
\end{equation}
where $c_{\min}$ is the known constant controlling the first-order derivative of the inverse link function.

\vspace{0.2cm}
\begin{breakablealgorithm}\label{alg_LinFACT_sampling_GLM}
\caption{Subroutine: G-Optimal Sampling with GLM}
\begin{algorithmic}[1]
    \State \textbf{Input:} Projected active set $\mathcal{A}_I(r-1)$, round $r$, $\delta$.
    \State Find the G-optimal design $\pi_r \in \mathcal{P}(\mathcal{A}(r-1))$ with support size $\text{Supp}(\pi_r) \leq \frac{d(d+1)}{2}$ according to equation \eqref{equation_G optimal}.\label{G_sampling_1_GLM}
    \ForAll{$\boldsymbol{a} \in \mathcal{A}_I(r-1)$}
        \Comment{Sampling}
        \State Sample arm $\boldsymbol{a}$ for $T_r(\boldsymbol{a})$ times in round $r$, as specified in equation~\eqref{equation_phase budget G_GLM}.\label{G_sampling_2_GLM}
    \EndFor
\end{algorithmic}
\end{breakablealgorithm}
\vspace{0.2cm}

In the GLM setting, Ordinary Least Squares (OLS) is no longer applicable. Instead, the estimator for each $i \in \mathcal{A}_I(r-1)$ is obtained by solving the optimization problem described in equation~\eqref{eq_optimization_esti}, using the estimating equation in \eqref{eq_GLM_mle} derived from the observed data. 

\subsection{Upper Bound for the GLM-Based LinFACT}\label{subsubsec_Upper Bound for the GLM-Based LinFACT}

Let $T_{G_{\textup{GLM}}}$ denote the number of samples collected under the GLM setting. The following theorem provides an upper bound on the expected sample complexity of LinFACT-\textup{GLM}. 

\begin{theorem}[Upper Bound, Generalized Linear Model]\label{upper bound: Algorithm 1 GLM}
    For a universal constant $c$, there exists an event $\mathcal{E}$ such that $\mathbb{P}(\mathcal{E}) \geq 1-\delta$. On this event, the \textup{LinFACT-GLM} algorithm achieves an expected sample complexity upper bound given by
    \begin{align}\label{eq_GLM_sample_complexity_paper}
        \mathbb{E}_{\boldsymbol{\mu}}\left[T_{\textup{GLM}} \mid \mathcal{E}\right] &\leq c \cdot \max \left\{ \frac{256d}{\alpha_\varepsilon^2 c_{\min}^{2}} \log \left( \frac{2K}{\delta} \log_2 \frac{16}{\alpha_\varepsilon} \right), \frac{256d}{\beta_\varepsilon^2 c_{\min}^{2}} \log \left( \frac{2K}{\delta} \log_2 \frac{16}{\beta_\varepsilon} \right) \right\} + \frac{d(d+1)}{2} R_{\textup{GLM}},
    \end{align}
    where 
    $
        R_{\textup{GLM}} = \max \left\{ \left\lceil \log_2 \frac{4}{\alpha_\varepsilon} \right\rceil, \left\lceil \log_2 \frac{4}{\beta_\varepsilon} \right\rceil \right\}
    $
    is the round in which the answer is returned.
\end{theorem}

The upper bound presented in Theorem \ref{upper bound: Algorithm 1 GLM}, which generalizes the model to the GLM setting, can be viewed as an extension of Theorem \ref{upper bound: Algorithm 1 G}. The detailed proof is provided in Section \ref{proof of theorem: upper bound of LinFACT GLM} of the online appendix. 

\section{Numerical Experiments}\label{sec_Simulation Experiments} 


Identifying all $\varepsilon$-best arms is often more challenging than identifying the top $m$ arms or arms above a given threshold. In real-world scenarios, the values of $m$ and the threshold typically require prior knowledge, which may not be readily available, unlike the value of $\varepsilon$. To address this, we adopt a random setting where both the number of $\varepsilon$-best arms and the $\varepsilon$-threshold are randomly sampled. In this setting, top $m$ algorithms and threshold bandit algorithms only have access to the expected reward values, ensuring a fair comparison.

Since BayesGap and KGCB operate in a fixed-budget setting, we set the average sample complexity of LinFACT as the budget for comparison and evaluate their performance accordingly. For BAI algorithms, we select the arms whose empirical means are at most $\varepsilon$ worse than the empirical optimum after the budget is exhausted. 


\subsection{Synthetic Experiments}\label{sec:exp_setup} 

In the \textit{adaptive setting}, arms are divided into three categories: (1) the arms to be selected (\emph{i.e.}, the all $\epsilon$-best arms), (2) disturbing arms that are slightly worse, and (3) base arms with zero rewards. The primary challenge for algorithms is to distinguish between the arms in categories (1) and (2), while the base arms in category (3) can be ignored. Adaptive algorithms that effectively leverage shared information to explore similar arms perform well in this setting.

\begin{figure}[htbp]
    \centering
    \begin{subfigure}{0.48\textwidth}
        \centering
        \includegraphics[width=\textwidth]{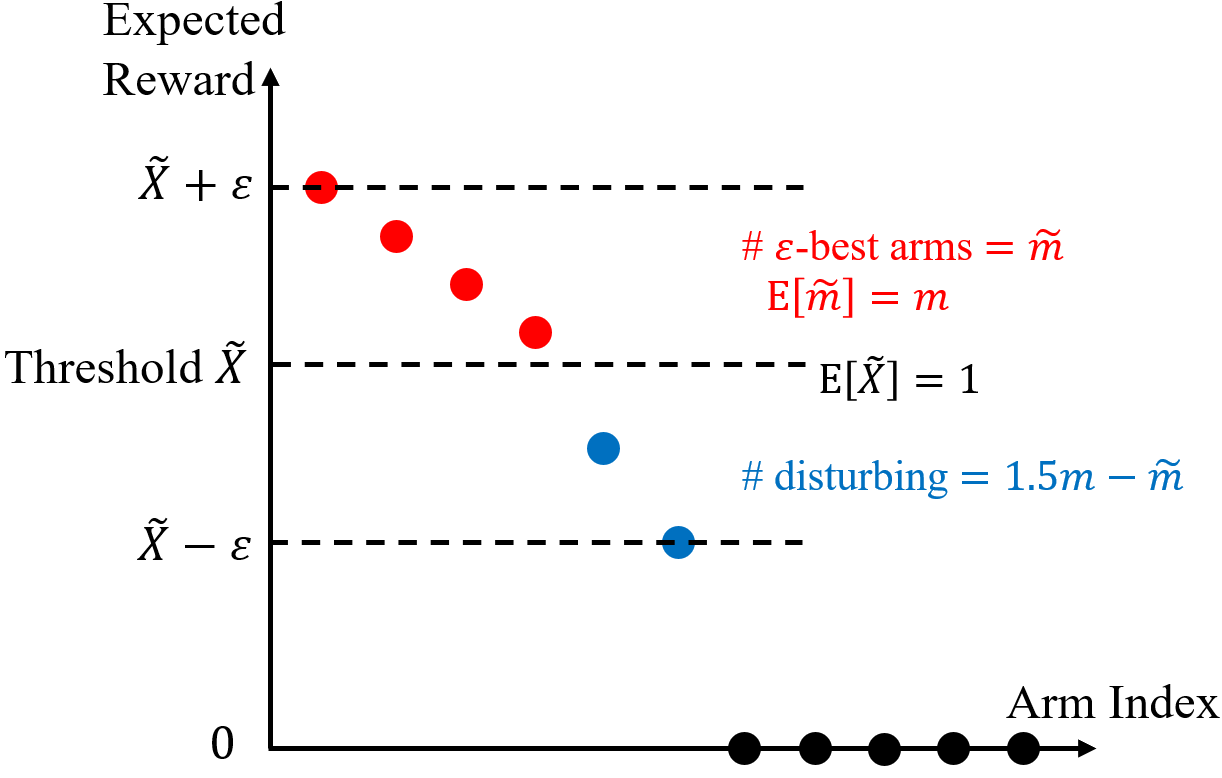}  
        \caption{Synthetic I - Adaptive Setting}
        \label{fig:synthetic-adaptive-main}
    \end{subfigure}
    \hfill
    \begin{subfigure}{0.48\textwidth}
        \centering
        \includegraphics[width=\textwidth]{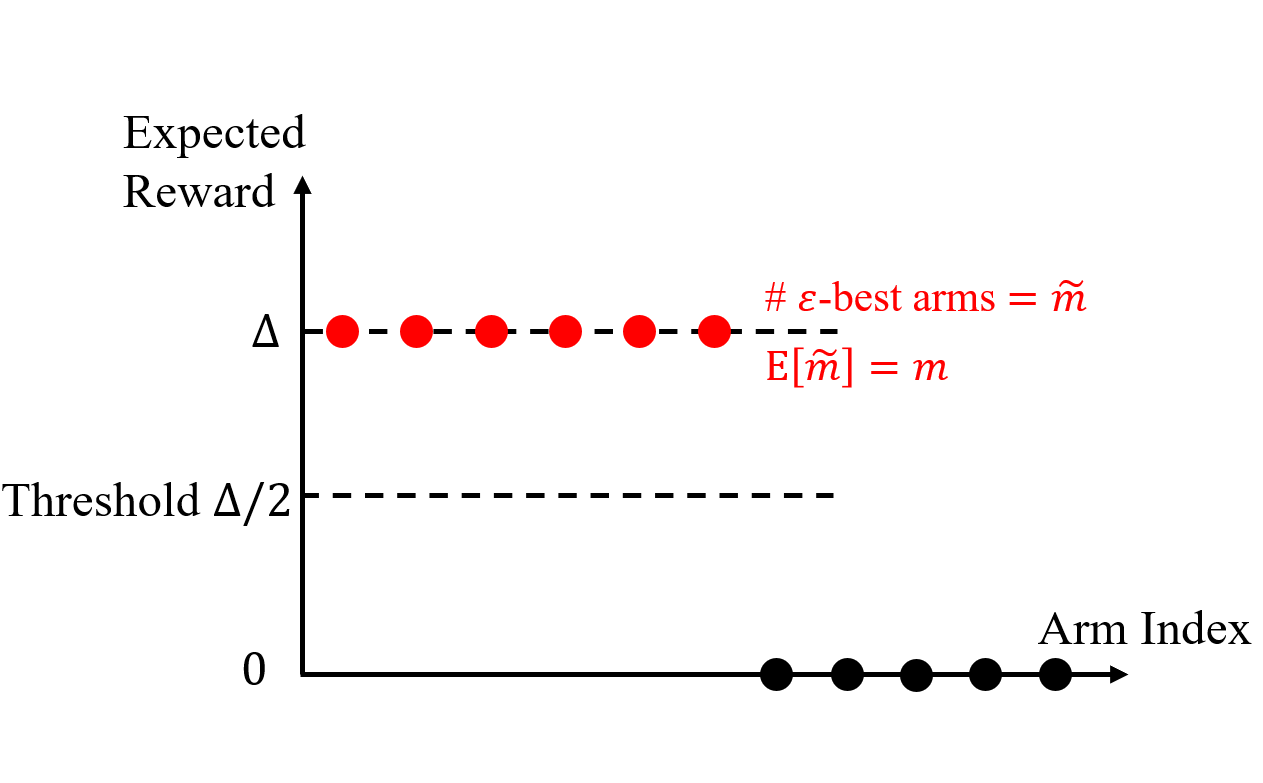}  
        \caption{Synthetic II - Static Setting}
        \label{fig:synthetic-static-main}
    \end{subfigure}
    
    \caption{Illustration on Synthetic Settings}
    \label{fig: synthetic-main}
\end{figure}

In the \textit{static setting}, arms are divided into two categories: the all $\epsilon$-best arms and the base arms with zero rewards. In this case, algorithms must distinguish between all arms. Static algorithms that uniformly explore all arms are well-suited for this setting. The construction of synthetic data is illustrated in Figure~\ref{fig: synthetic-main}, with detailed settings provided in Section \ref{sec_Detailed Settings for Synthetic Experiments} of the online appendix.

\begin{remark}
    A common misconception is that adaptive algorithms always outperform static algorithms. In typical scenarios, including adaptive settings, adaptive algorithms efficiently explore similar candidate arms, reducing sample complexity. However, in static settings, adaptive algorithms may waste excessive samples exploring both candidate and base arms, while static algorithms can achieve the objective more efficiently by uniformly exploring all arms without bias.
\end{remark}

\subsubsection{Experiment Setup.}
We benchmark our algorithms, LinFACT-G and LinFACT-$\mathcal{XY}$, against BayesGap, KGCB, LinGIFA, m-LinGapE, and Lazy TTS, focusing on $F1$ score and sample complexity.
We conduct each experiment using different types of data (adaptive or static), varying arm dimensions ($d$), and arm numbers ($K$). For each configuration, we generate 10 normally distributed values of $m$ around the expected value $\mathbb{E}[m]$, with a variance of $3.0$. For each randomly generated pair $(\tilde{m},\tilde{X})$, we repeat the experiment 100 times. We then calculate the average $F1$ score across 10 different $(\tilde{m},\tilde{X})$ pairs, resulting in a total of 1,000 executions for each algorithm.

For the two top-$m$ algorithms, the computational burden arises from performing matrix inversions for all arms, leading to a total time complexity of $O(Kd^3)$. In contrast, LinFACT achieves a significantly lower total time complexity of $O(Kd^2)$, as the optimization problem within our algorithm can be efficiently solved using a fixed-step gradient descent method. Table~\ref{tab:Synthetic_Time} presents the runtime for synthetic data, demonstrating that our algorithm is at least five times faster than all other methods. Notably, this performance gap is even more pronounced when using real data.

\begin{table}[htbp]
    \small
    \centering
    \renewcommand\arraystretch{1.5}
    \caption{Running Time for Different Synthetic Experiments Among Algorithms}
    \begin{tabularx}{\textwidth}{l *{6}{>{\centering\arraybackslash}X}}
        \hline
        \multirow{3}{*}{\makecell{Algorithm \\ Settings}} & \multicolumn{3}{c}{Adaptive Settings} & \multicolumn{3}{c}{Static Settings} \\
        \cline{2-7}
        & \multicolumn{3}{c}{$(d,E[m])=(10,4)$} & \multicolumn{3}{c}{$\Delta=1$} \\
        \cline{2-7}
        & $\varepsilon=0.1$ & $\varepsilon=0.2$ & $\varepsilon=0.3$ & $d=8$ & $d=12$ & $d=16$ \\
        \hline
        LinFACTE-G & 0.028 & 0.03 & 0.028 & 0.005 & 0.007 & 0.009 \\
        LinFACTE-$\mathcal{XY}$ & 0.037 & 0.038 & 0.038 & 0.015 & 0.028 & 0.049 \\
        BayesGap & 0.112 & 0.110 & 0.110 & 0.036 & 0.071 & 0.112 \\
        LinGIFA & 0.313 & 0.265 & 0.236 & 0.160 & 0.529 & 1.297 \\
        m-LinGapE & 0.195 & 0.166 & 0.157 & 0.116 & 0.352 & 0.932 \\
        Lazy TTS & 0.991 & 3.904 & 12.51 & 0.190 & 0.870 & 2.160 \\
        KGCB & 1.990 & 1.936 & 1.670 & 0.303 & 0.745 & 1.386 \\
        \hline
    \end{tabularx}
    \label{tab:Synthetic_Time}
    \vspace{-0.4cm}
\end{table}

    

\subsubsection{Experiment Results.}

\begin{figure}[htbp]
    \centering
    \includegraphics[width=\textwidth]{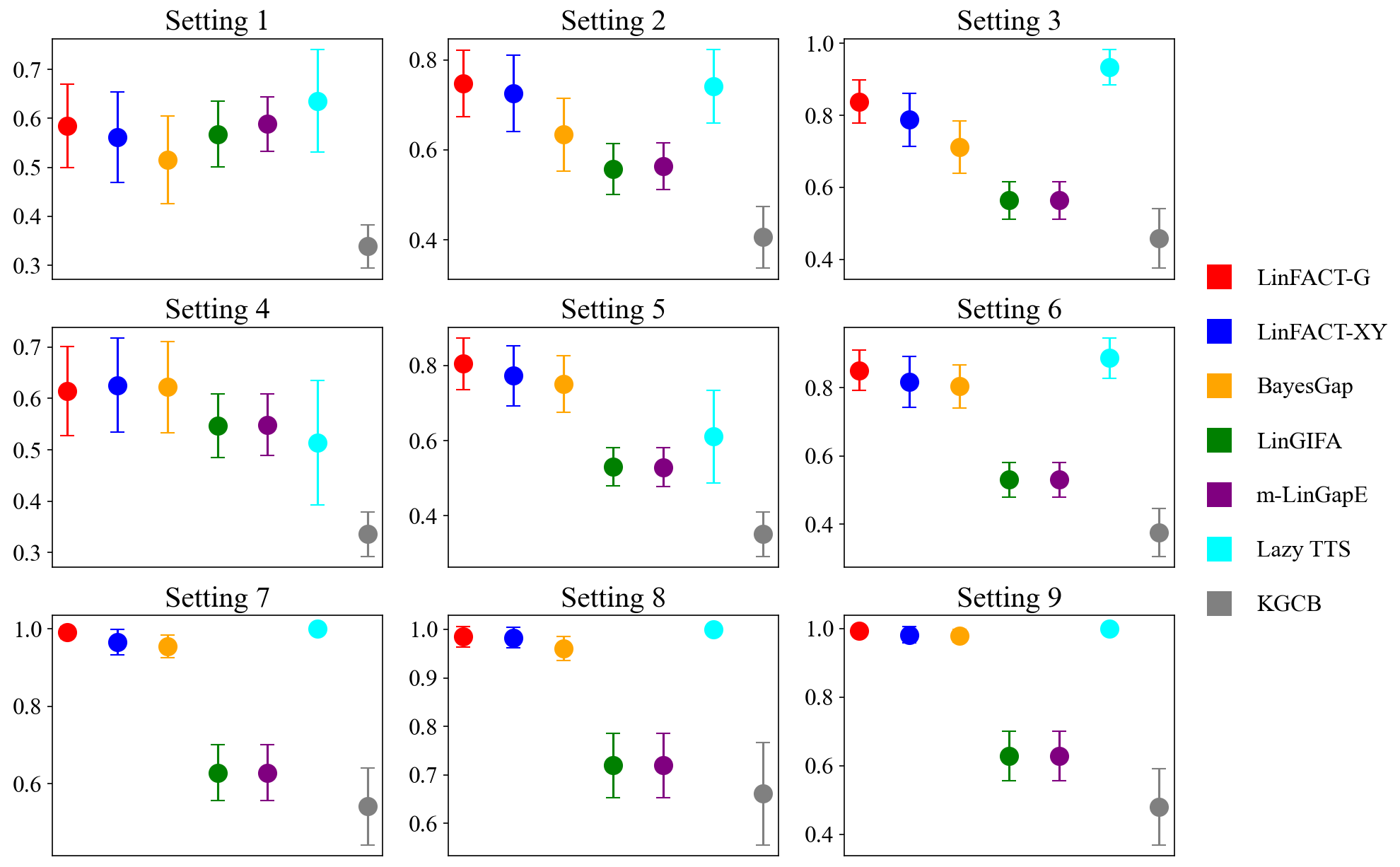}
    \caption{$F1$ Scores for Different Synthetic Experiments Among Algorithms
    }
    \label{fig:Synthetic_F1}
\end{figure}

\begin{figure}[htbp]
    \centering
    \includegraphics[width=.9\textwidth]{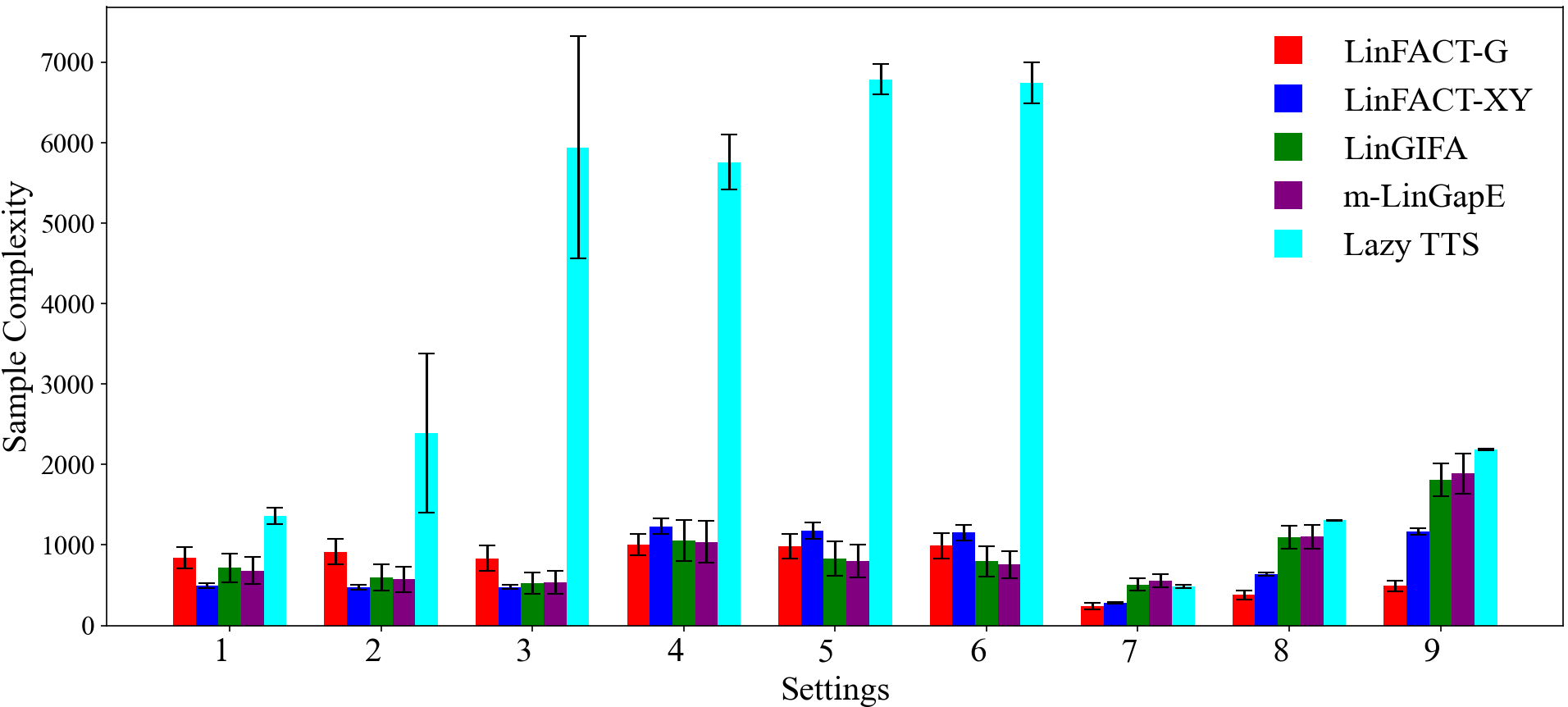}
    \caption{Sample Complexity for Different Synthetic Experiments Among Algorithms
    }
    \label{fig:Synthetic_Complexity}
\end{figure}

Our experimental results are presented in Figure~\ref{fig:Synthetic_F1} and Figure~\ref{fig:Synthetic_Complexity}. In Figure~\ref{fig:Synthetic_F1}, the vertical axis denotes the $F1$ score, with higher values indicating better algorithm performance. The first row of six plots shows the results under adaptive settings. As $\varepsilon$ increases, the non-optimal (disturbing) arms in the adaptive setting move progressively farther from the optimal arms, making them easier to distinguish. Consequently, the $F1$ score increases from left to right. An exception is BayesGap, which performs best when $\varepsilon=0.2$. This occurs because best-arm identification algorithms, such as BayesGap, struggle to differentiate optimal arms from disturbing ones when they are close ($\varepsilon=0.1$) and fail to fully explore the optimal arms when they are not closely clustered with the best arm ($\varepsilon=0.3$).

Our LinFACT algorithms consistently outperform top-$m$ algorithms, BayesGap, and KGCB. While our algorithms perform slightly worse than Lazy TTS for some cases, they have much lower sample complexity, as shown in Figure~\ref{fig:Synthetic_Complexity}, meaning that Lazy TTS requires substantially more samples to achieve these results.

When comparing LinFACT-G and LinFACT-$\mathcal{XY}$, we observe that in adaptive settings, the $F1$ scores are similar, but LinFACT-$\mathcal{XY}$ achieves lower sample complexity. In static settings, however, LinFACT-G attains a higher $F1$ score with a reduced sample complexity. This difference stems from the distinct focus of the two designs: the $\mathcal{XY}$-optimal design prioritizes pulling arms to obtain better estimates along the directions representing differences between arms, while the G-optimal design aims to improve estimates along the directions representing all arms.


\subsection{Experiments with Real Data - Drug Discovery}
\subsubsection{Experiment Setup.}

\begin{figure}[htbp]
    \centering
    \includegraphics[width=.95\textwidth]{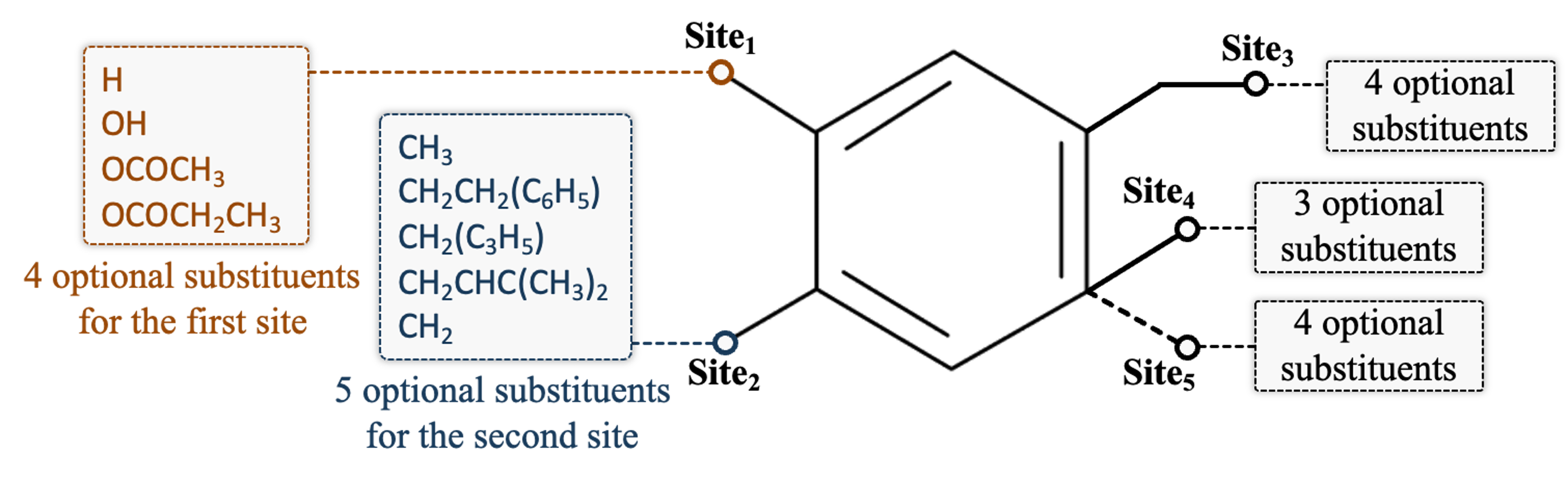}
    \caption{An Example of Molecule and Substituent Locations with $\textup{Site}_1, \ldots,\textup{Site}_5$
    }
    \label{fig: drug_example}
\end{figure}

Each compound is modeled as an arm represented by a binary indicator vector. Suppose there are $N$ modification sites, with site $n \in [N]$ offering $l_n$ alternative substituents. Then each arm $\boldsymbol{a}$ lies in $\mathbb{R}^{1 + \sum_{n \in [N]} l_n}$, where the initial entry corresponds to the base molecule (\emph{i.e.}, the intercept term). For each site, the corresponding segment in the vector has exactly one entry set to 1 (indicating the chosen substituent), with the remaining $l_n - 1$ entries set to 0. This results in a total of $\prod_{n \in [N]} l_n$ unique compound configurations.

We benchmarked our algorithm, LinFACT-G, against KGCB, BayesGap, LinGIFA, m-LinGapE, and Lazy TTS, focusing on the precision and recall changing with failure probability $\delta$. For this study, the ``correct" set of 20 $\varepsilon$-best arms was defined with $\varepsilon=4.325$, treating the output as a classification result. All algorithms were tested across 50 trials, with the failure probability $\delta$ varying from 0.1 to 0.9 in increments of 0.1.

\subsubsection{Experiment results.}

\begin{figure}[htbp]
\centering
\begin{subfigure}{0.52\textwidth}
    \includegraphics[height=4.5cm]{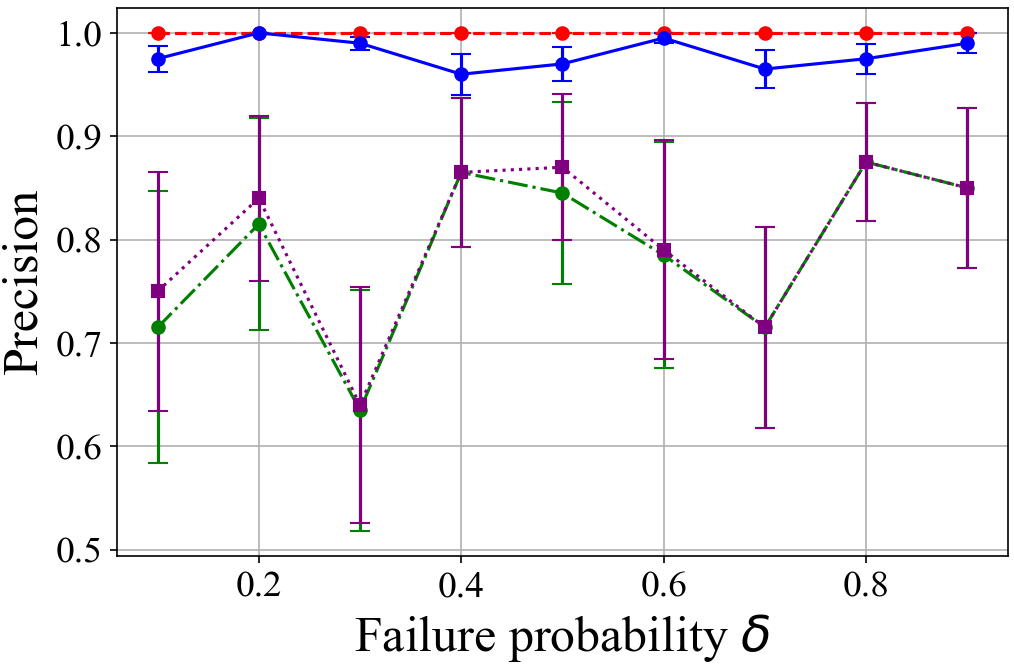}
    \caption{Precision}
    \label{fig:real_precision}
\end{subfigure}
\begin{subfigure}{0.42\textwidth}
    \includegraphics[height=4.5cm]{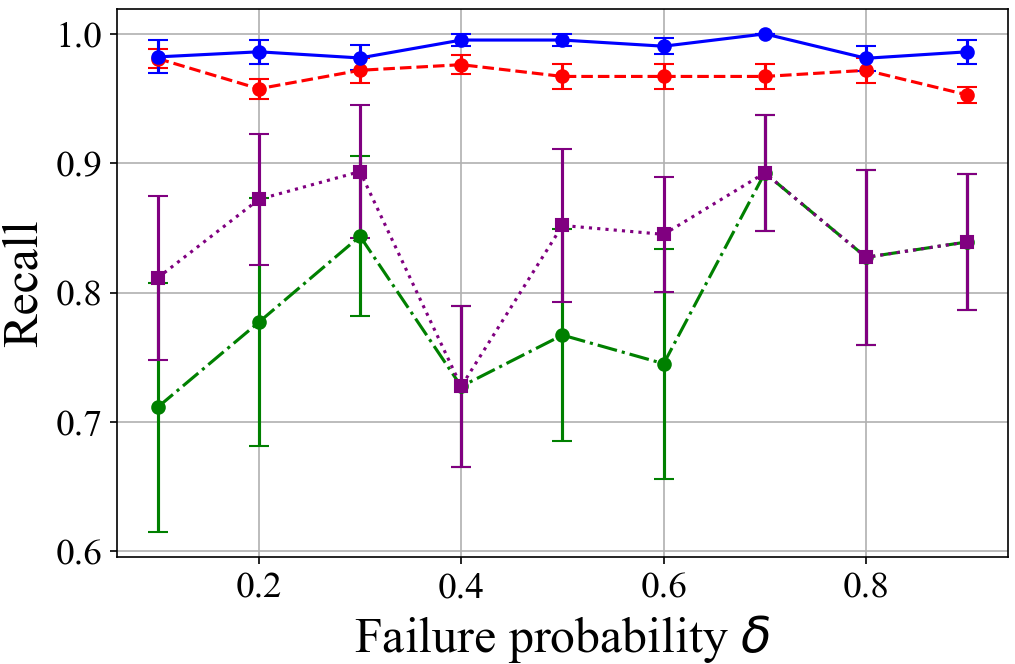}
    \caption{Recall}
    \label{fig:real_recall}
\end{subfigure}
\caption{Precision and Recall for Various Failure Probabilities $\delta$} 
\label{fig:real_data_Drug}
\end{figure}

The experimental results are shown in Figure~\ref{fig:real_data_Drug}. LinFACT-G consistently achieves high precision and recall. In contrast, KGCB performs poorly, with both metrics remaining below 0.4 across all $\delta$ values, likely due to the imposed runtime threshold. Beyond its accuracy, LinFACT-G also demonstrates strong computational efficiency: it completes the task within one minute, whereas BayesGap, LinGIFA, m-LinGapE, and Lazy TTS take approximately four times longer, and Knowledge Gradient requires over two hours. This efficiency makes LinFACT-G particularly well-suited for real-time applications.



\section{Conclusion}\label{sec_Conclusion} 

In this paper, we address the challenge of identifying all $\varepsilon$-best arms in linear bandits, motivated by applications such as drug discovery. We establish the first information-theoretic lower bound to characterize the problem's complexity and derive a matching upper bound. Our LinFACT algorithm achieves instance-optimal performance up to a logarithmic factor under the $\mathcal{XY}$-optimal design criterion.

We further extend our analysis to settings with model misspecification and generalized linear models (GLMs), deriving new upper bounds and providing insights into algorithmic behavior under these broader conditions. These results generalize and recover the guarantees from the perfectly linear case as special instances.

Extensive numerical experiments confirm that LinFACT outperforms existing methods in both sample and computational efficiency, while maintaining high accuracy in identifying all $\varepsilon$-best arms.




%
%
%





\clearpage
\begin{appendices}
\renewcommand{\thepage}{ec\arabic{page}}  
\renewcommand{\thesection}{EC.\arabic{section}}   
\renewcommand{\thetable}{EC.\arabic{table}}   
\renewcommand{\thefigure}{EC.\arabic{figure}}
\renewcommand{\theequation}{EC.\arabic{equation}}
\renewcommand{\thetheorem}{EC.\arabic{theorem}}
\renewcommand{\theproposition}{EC.\arabic{proposition}}
\renewcommand{\thelemma}{EC.\arabic{lemma}}
\newcommand{\thealgorithm}{\arabic{algorithm}}
\renewcommand{\thealgorithm}{EC.\arabic{algorithm}}
\counterwithin{equation}{section}
\counterwithin{table}{section}
\counterwithin{figure}{section}
\counterwithin{lemma}{section}
\setcounter{equation}{0}
\setcounter{section}{0}
\setcounter{page}{1}

\vspace*{0.1cm}
   \begin{center}
      \large\textbf{E-Companion -- Identifying All $\varepsilon$-Best Arms In Linear Bandits With Misspecification}\\
   \end{center}
   \vspace*{0.3cm}

\section{Additional Literature}\label{sec_Literature Review for Misspecified Linear Bandits and Generalized Linear Bandits}
\subsection{Misspecified Linear Bandits.}\label{subsubsec_Pure Exploration for Linear Bandits} 

The linear bandit (LB) problem, introduced by \citet{abe1999associative}, extends the multi-armed bandits (MABs) framework by incorporating structural relationships among different arms. In the context of best arm identification, \citet{garivier2016optimal} established a classical lower bound, which was later extended to linear bandits by \citet{fiez2019sequential} using transportation inequalities.

The foundational study of linear bandits in the pure exploration framework was conducted by \citet{hoffman2014correlation}, who addressed the best arm identification (BAI) problem in a fixed-budget setting while considering correlations among arm distributions. They proposed BayesGap, a Bayesian variant of the gap-based exploration algorithm \citep{gabillon2012best}. Although BayesGap outperformed methods that ignore correlations and structural relationships, its limitation of ceasing to pull arms deemed sub-optimal hindered its effectiveness in linear bandit pure exploration.

A key distinction between stochastic MABs and linear bandits is that, in MABs, once an arm’s sub-optimality is confirmed with high probability, it is no longer pulled. In linear bandits, however, even sub-optimal arms can offer valuable information about the parameter vector, improving confidence in estimates and aiding the discrimination of near-optimal arms. This insight has led to the adoption of optimal linear experiment design as a crucial framework for linear bandit pure exploration \citep{abbasi2011improved, soare2014best, fiez2019sequential, reda2021dealing, yang2021minimax, DBLP:journals/corr/abs-2106-04763}. 

When applying linear models to real data, misspecification inevitably arises in situations where the data deviates from perfect linearity.  The concept of misspecified bandit models was introduced in the context of cumulative regret by \citet{ghosh2017misspecified}, who demonstrated a significant limitation: any linear bandit algorithm (\emph{e.g.}, OFUL \citep{abbasi2011improved} or LinUCB \citep{10.1145/1772690.1772758}), which achieves optimal regret bounds on perfectly linear instances, can suffer linear regret on certain misspecified models. To address this, they proposed a hypothesis-test-based algorithm that avoids linear regret and achieves UCB-type sublinear regret for models with non-sparse deviations from linearity. \citet{lattimore2020learning} further analyzed misspecification, showing that elimination-based algorithms with G-optimal design perform well under misspecification but incur an additional linear regret term proportional to the misspecification magnitude over the horizon.

In the pure exploration setting, misspecified linear models were first studied in the context of identifying the top $m$ best arms by \citet{reda2021dealing}, who introduced the MisLid algorithm, leveraging orthogonal parameterization to address misspecification. Subsequent research examined misspecification in ordinal optimization \citep{doi:10.1287/mnsc.2022.00328}, proposing prospective sampling methods that reduce the impact of misspecification as the sample size increases. Building on the definition of model misspecification and the optimization approach based on orthogonal parameterization from \cite{reda2021dealing}, we develop new algorithms for identifying all $\varepsilon$-best arms in misspecified linear bandits and establish new upper bounds.

\subsection{Generalized Linear Bandits.}\label{subsubsec_General Linear Bandits} 

The generalized linear bandit (GLB) model \citep{filippi2010parametric, ahn2020ordinal, kveton2023randomized} extends the multi-armed bandit framework by incorporating generalized linear models (GLMs) \citep{mccullagh2019generalized} to model expected rewards. Specifically, the expected reward of each arm is given by a known link function applied to the inner product of a feature vector and an unknown parameter vector. Most existing algorithms for generalized linear bandits employ the upper confidence bound (UCB) approach, with randomized GLM algorithms \citep{chapelle2011empirical, russo2018tutorial, kveton2023randomized} demonstrating superior performance.

In the context of pure exploration, \citet{azizi2021fixed} introduced the first practical algorithm for best arm identification in generalized linear bandits, supported by theoretical analysis. Their work extends the best arm identification problem from linear models to more complex settings where the relationship between features and rewards follows generalized linear models (GLMs). Building on this foundation, we extend the pure exploration setting from best arm identification (BAI) to identifying all $\varepsilon$-best arms, providing analogous analyses and theoretical results for GLMs. 

\section{General Pure Exploration Model}\label{sec_General Pure Exploration Model} 
In this section, we present a comprehensive framework that describes the general pure exploration problem, where the decision-maker aims to address a query related to the mean parameters $\boldsymbol{\mu}$, achieved through adaptive allocation of the sampling budget to the set of arms. This query task involves finding a specific group of arms based on various criteria and we are trying to find the right answer with high probability. Recall that $\mathcal{I} = \mathcal{I}\left(\boldsymbol{\mu}\right)$ is the correct answer, the set with all $\varepsilon$-best arms. Let $\widetilde{M}$ denote the set of parameters associated with a unique answer. Let $\Xi$ be the set of all possible answers, and for each $\mathcal{I}^\prime\in\Xi$, define $M_{\mathcal{I}^\prime} \coloneqq \left\{\boldsymbol{\vartheta}\in\widetilde{M}: \mathcal{I}\left(\boldsymbol{\vartheta}\right) = \mathcal{I}^\prime\right\}$ as the set of parameters with $\mathcal{I}^\prime$ as the correct answer. The parameter space $M \coloneqq \cup_{\mathcal{I}^\prime\in\Xi}M_{\mathcal{I}^\prime}$ is the focus of the problem.

Recall that an algorithm is defined as a triple $\mathcal{H} = \left ( A_t,\; \tau_\delta,\; \hat{a}_\tau \right )$. The algorithm's sample complexity is quantified by the number of samples, denoted as $\tau_\delta$, at the point of termination. The objective is to formulate algorithms that minimize the expected sample complexity $\mathbb{E}_{\boldsymbol{\mu}} \left [ \tau_\delta \right ]$ across the set $\mathcal{H}$. As stated in \cite{kaufmann2016complexity}, when $\delta \in \left ( 0, 1 \right )$, the non-asymptotic problem complexity of an instance $\boldsymbol{\mu}$ can be defined as
\begin{equation}\label{eq_problem complexity_ec}
\kappa \left ( \boldsymbol{\mu} \right ) \coloneqq \inf\limits_{Algo \in \mathcal{H}} \frac{\mathbb{E}_{\boldsymbol{\mu}} \left [ \tau_\delta \right ]}{\log\left({1}/{2.4\delta}\right)}.
\end{equation}

This instance-dependent complexity indicates the smallest possible constant such that the expected sample complexity $\mathbb{E}_{\boldsymbol{\mu}} \left [ \tau_\delta \right ]$ scales in alignment with $\log\left({1}/{2.4\delta}\right)$. The problem complexity $\kappa \left ( \boldsymbol{\mu} \right )$ is subject to an information-theoretic lower bound. This lower bound can be expressed as the optimal solution of an allocation problem, which we show in Proposition \ref{general lower bound}. To build this framework, we will first introduce three important concepts: culprits, alternative sets, and $C_x$ function.

\subsection{Culprits and Alternative Sets}\label{Culprits} 
We define $\mathcal{X}(\boldsymbol{\mu})$ as the set of culprits when the true mean vector is $\boldsymbol{\mu}$. These culprits cause the deviation of the correct answer of $\boldsymbol{\mu}$ from $\mathcal{I}(\boldsymbol{\mu})$. The set of culprits differs for each exploration task, and identifying culprits is crucial as it allows for the characterization of the problem complexity and the design of algorithms.

An algorithm that finds the correct answer must distinguish different problem instances in the parameter space $M$. Therefore, for any instance $\boldsymbol{\mu} \in M$, the instance-dependent problem complexity $\kappa \left ( \boldsymbol{\mu} \right )$ is related to the alternative set
\begin{equation}
\cup_{x \in \mathcal{X}(\boldsymbol{\mu})} \textup{Alt}_x(\boldsymbol{\mu}) = \textup{Alt} \left ( \boldsymbol{\mu} \right ) \coloneqq \left \{ \boldsymbol{\vartheta} \in M: \mathcal{I} \left ( \boldsymbol{\vartheta} \right ) \neq \mathcal{I} \left ( \boldsymbol{\mu} \right ) \right \},
\label{def: alternative set}
\end{equation}
which represents the set of parameters that return a solution that is different from the correct solution $\mathcal{I} \left ( \boldsymbol{\mu} \right )$. 

As an example, for the task of identifying the single best arm, $\mathcal{X}(\boldsymbol{\mu}) = [K] \backslash \left \{ I^\ast(\boldsymbol{\mu}) \right \}$, which is the set of all arms without the best arm. For each culprit $x \in \mathcal{X}(\boldsymbol{\mu})$, if arm $x$ is a better arm under another mean vector $\boldsymbol{\vartheta}$ than the best arm $I^\ast(\boldsymbol{\vartheta})$, then $\boldsymbol{\vartheta}$ is the parameter that can bring a wrong answer caused by $x$. Associated with each culprit $x \in \mathcal{X}(\boldsymbol{\mu})$ is an alternative set, which is the set of parameters that causes the identification of the wrong answer. We have $\textup{Alt}_x(\boldsymbol{\mu}) = \left \{ \boldsymbol{\vartheta} \in M: \vartheta_x \geq \vartheta_{I^\ast(\boldsymbol{\vartheta})} \right \}$ for $x \in \mathcal{X}(\boldsymbol{\mu})$.

\subsection{\texorpdfstring{$\boldsymbol{C_x}$ } ffunction}\label{Cx function} 
The process of finding the correct answer is solved by a sequential hypothesis test using the test of sequential generalized likelihood ratio (\textup{SGLR}) \citep{kaufmann2016complexity, kaufmann2021mixture}. A \textup{SGLR} statistic is defined for testing a potentially composite null hypothesis $H_0: (\boldsymbol{\mu} \in \Omega_0)$ versus a potentially composite alternative hypothesis $H_1: (\boldsymbol{\mu} \in \Omega_1)$ by
\begin{equation}
\textup{SGLR}_t = \frac{\sup_{\boldsymbol{\vartheta} \in \Omega_0 \cup \Omega_1}L(X_1, X_2, \ldots, X_t ; \boldsymbol{\vartheta})}{\sup_{\boldsymbol{\vartheta} \in \Omega_0}L(X_1, X_2, \ldots, X_t ; \boldsymbol{\vartheta})},
\label{def: generalized likelihood ratio statistic}
\end{equation}
where $X_1, X_2, \ldots, X_t$ are observed rewards from arm pulling, and $L(\cdot)$ is the likelihood function with these observed data and some unknown parameter $\boldsymbol{\vartheta}$. $\Omega_0$ represents the restricted parameter space under the null hypothesis and $\Omega_0 \cup \Omega_1$ represents the parameter space under the alternative hypothesis, encompassing the full, unrestricted model space that allows for the greatest flexibility in fitting the data to the model. These parameter spaces are the alternative sets defined in the previous section. A large value of $\textup{SGLR}_t$ means we are more confident in rejecting the alternative hypothesis.

We limit our setting to a single-parameter exponential family parameterized by its mean, as in \cite{garivier2016optimal}, which includes Bernoulli distribution, Poisson distribution, Gamma distributions with known shape parameter, or Gaussian distribution with known variance. 

We need to test the following hypotheses: $H_{0, x}: \boldsymbol{\mu} \in \textup{Alt}_x(\hat{\boldsymbol{\mu}})$ against $H_{1, x}: \boldsymbol{\mu} \notin \textup{Alt}_x(\hat{\boldsymbol{\mu}})$ for each culprit $x \in \mathcal{X}(\hat{\boldsymbol{\mu}})$, where $\hat{\boldsymbol{\mu}}$ is the empirical mean based on observed data. If $\hat{\boldsymbol{\mu}}(t) \in \Omega_0 \cup \Omega_1$, equation (\ref{def: generalized likelihood ratio statistic}) can be shown in the form of self-normalized sum, resulting in the formal expression of SGLR statistic in Proposition \ref{SGLR statistic in pure exploration} through maximum likelihood estimation (MLE) and rewriting the KL divergence.
\begin{proposition}
\label{SGLR statistic in pure exploration}
The generalized likelihood ratio statistic with respect to each culprit $x \in \mathcal{X}(\boldsymbol{\mu})$ and each time step $t$ for the pure exploration setting is defined as
\begin{align}
\hat{\Lambda}_{t,x} &= \ln (\textup{SGLR}_t) \nonumber\\
&= \inf\limits_{\boldsymbol{\vartheta} \in \textup{Alt}_x(\hat{\boldsymbol{\mu}}(t))} \sum_{i \in \left [ K \right ]} N_i(t)\textup{KL}(\hat{\mu}_i(t), \vartheta_i),\label{def: SGLR statistic in pure exploration}
\end{align}
where \textup{KL}$(\cdot, \cdot)$ represents the \textup{KL} divergence of the two distributions parameterized by their means, $L(\cdot)$ represents the likelihood function, and $N_i(t) = t \cdot p_i$ is the expected number of observations allocated to each arm $i \in \left [ K \right ]$ up to time step $t$.
\end{proposition}

The proof of the proposition is presented in Section \ref{proof of proposition: SGLR statistic in pure exploration}.  This proposition connects the SGLR test with information-theoretic methods. To quantify the amount of information and confidence we have to assert that the unknown true mean value does not belong to $\textup{Alt}_x$ for all $x \in \mathcal{X}$, we define the $C_x$ function as the population version of the SGLR statistic, which is the same form as equation (\ref{def: SGLR statistic in pure exploration}).
\begin{equation}
C_x(\boldsymbol{p}) = C_x(\boldsymbol{p}; \boldsymbol{\mu}) \coloneqq \inf\limits_{\boldsymbol{\vartheta} \in \textup{Alt}_x} \sum_{i \in \left [ K \right ]} p_i\textup{KL}(\mu_i, \vartheta_i).
\label{def: Cx function}
\end{equation}

Then with the introduction of culprits and $C_x$ function, we define the following optimal allocation problem in Proposition \ref{general lower bound}.

However, computing the lower bound can still be hard since it requires the solution of the minimax problem in equation (\ref{def: general lower bound}). While the KL divergence in equation (\ref{def: Cx function}) is convex for Gaussians, it can be non-convex to minimize the $C_x$ function over the culprit set $\mathcal{X} \left ( \boldsymbol{\mu} \right )$. To solve this problem, depending on the following Proposition \ref{smoothness, PDE, and active candidate set}, we can write $\mathcal{X} \left ( \boldsymbol{\mu} \right )$ as a union of several convex sets. The following three equivalent expressions represent different ways of describing the lower bound, making the minimax problem in equation \eqref{def: general lower bound} tractable with the introduction of culprit set $\mathcal{X}$ and the corresponding alternative set $\textup{Alt}_x$ for every $x \in \mathcal{X}$.
\begin{align}
    \Gamma_{\boldsymbol{\mu}}^* = \max_{\boldsymbol{p} \in \mathcal{S}_K} \inf_{\boldsymbol{\vartheta} \in \operatorname{Alt}(\boldsymbol{\mu})} \sum_{i \in[K]} p_i \textup{KL}\left(\mu_i, \vartheta_i\right) &= \max_{\boldsymbol{p} \in \mathcal{S}_K} \min_{x \in \mathcal{X}} \inf _{\boldsymbol{\vartheta} \in \mathrm{Alt}_x} \sum_{i \in [K]} p_i \textup{KL}\left(\mu_i, \vartheta_i\right)\label{equation_alternative} \\ \label{optimal allocation problem_convex_1}
    &= \max_{\boldsymbol{p} \in \mathcal{S}_K} \min _{x \in \mathcal{X}} \sum_{i \in[K]} p_i \textup{KL}\left(\mu_i, \vartheta_i^x\right) \\ \label{optimal allocation problem_convex_2}
    &= \max_{\boldsymbol{p} \in \mathcal{S}_K} \min_{x \in \mathcal{X}} C_x(\boldsymbol{p}),\label{optimal allocation problem_convex_3}
\end{align}
where we used the existence of finite union set and unique minimizer $\boldsymbol{\vartheta}^x$ in Proposition \ref{smoothness, PDE, and active candidate set}.
\begin{proposition}\label{smoothness, PDE, and active candidate set}
Assuming that the arms distribution follows a canonical single-parameter exponential family parameterized by its mean value. Then, for each culprit $x \in \mathcal{X}({\boldsymbol{\mu}})$,
\begin{enumerate}
\item \citep{wang2021fast} (Assumption 1) For each problem instance $\boldsymbol{\mu} \in M$, the alternative set $\textup{Alt} \left ( \boldsymbol{\mu} \right )$ is a finite union of convex sets. Namely, there exists a finite collection of convex sets $\left \{ \textup{Alt}_x(\boldsymbol{\mu}): x \in \mathcal{X}(\boldsymbol{\mu}) \right \}$ such that $\textup{Alt}(\boldsymbol{\mu}) = \cup_{x \in \mathcal{X}(\boldsymbol{\mu})} \textup{Alt}_x(\boldsymbol{\mu})$.
\item Given a specific simplex distribution $\boldsymbol{p}$, there exists a unique $\boldsymbol{\vartheta}^x \in \textup{Alt}_x(\boldsymbol{\mu})$ that achieves the infimum in equation~(\ref{def: Cx function}).
\end{enumerate}

\end{proposition}
The proof of this proposition is provided in Section \ref{proof of proposition: assumption on decomposition} to show that our setting, \emph{i.e.}, all $\varepsilon$-best arms identification with linear bandits, satisfies these two assumptions. 

\subsection{Stopping Rule}\label{Stopping Rule} 
In this section, we introduce the stopping rule, which suggests when to stop the algorithm and returns an answer that gives all $\varepsilon$-best arms with a probability of at least $1-\delta$. This stopping rule is based on the deviation inequalities that are linked to the generalized likelihood ratio test \citep{kaufmann2021mixture}. 

For each $x_i \in \mathcal{X}(\boldsymbol{\mu})$ and $i \in [\left \lvert \mathcal{X} \right\rvert]$, let $M_i(\boldsymbol{\mu}) = \textup{Alt}_{x_i}(\boldsymbol{\mu})$ be a partition of the whole realizable parameter space $M$ considered in Section \ref{Culprits}, with each partition $M_i(\boldsymbol{\mu})$ being characterized by a culprit in $\mathcal{X}(\boldsymbol{\mu})$. This means for each $\boldsymbol{\mu} \in M$, we can construct a unique partition of the parameter space $M$ to form our hypothesis test. $M_0(\boldsymbol{\mu})$ denotes the set of parameters where $\boldsymbol{\mu}$ resides. If $\boldsymbol{\mu} \in M$, define $i^\ast(\boldsymbol{\mu})$ as the index of the unique element in the partition where the true mean value $\boldsymbol{\mu}$ belongs and here $i^\ast(\boldsymbol{\mu}) = 0$. In other words, we have $\boldsymbol{\mu} \in M_{0}$ and $\textup{Alt}(\boldsymbol{\mu}) = M \backslash M_{0}$. Since we do not care about the order among suboptimal arms, $M_i(\boldsymbol{\mu})$ for $i \in \left \{ 0,1,2, \ldots, \left \lvert \mathcal{X} \right\rvert \right \}$ can form a partition of $M$ for each $\boldsymbol{\mu}$. The alternative set can thus be further defined as
\begin{equation}
\textup{Alt}(\boldsymbol{\mu}) = \cup_{i: \boldsymbol{\mu} \notin M_i(\boldsymbol{\mu})}M_i(\boldsymbol{\mu}) = M \backslash M_{i^\ast(\boldsymbol{\mu})} = M \backslash M_{0}.
\label{def: partition of the alternative set}
\end{equation}

Given a bandit instance $\boldsymbol{\mu}$, we consider $\left \lvert \mathcal{X}(\boldsymbol{\mu}) \right\rvert + 1$ hypotheses, given by
\begin{equation}
H_0 = (\boldsymbol{\mu} \in M_0(\boldsymbol{\mu})), H_1 = (\boldsymbol{\mu} \in M_1(\boldsymbol{\mu})), \ldots, H_{\left \lvert \mathcal{X} \right\rvert} = (\boldsymbol{\mu} \in M_{\left \lvert \mathcal{X} \right\rvert}(\boldsymbol{\mu})).
\label{def: parallel hypotheses}
\end{equation}

Substitute true mean value $\boldsymbol{\mu}$ with the empirical value $\hat{\boldsymbol{\mu}}(t)$, the SGLR test depends on the empirical value $\hat{\boldsymbol{\mu}}(t)$ in each time $t$. The hypotheses tested at time $t$ are data-dependent. If $\hat{\boldsymbol{\mu}}(t) \in M$, define $\hat{i}(t) = i^\ast(\hat{\boldsymbol{\mu}}(t))$ as the index of the partition to which $\hat{\boldsymbol{\mu}}(t)$ belong; in other words, $\hat{\boldsymbol{\mu}}(t) \in M_{\hat{i}(t)}$; otherwise if $\hat{\boldsymbol{\mu}}(t) \notin M$, $\hat{\Lambda}_{t,x} = 0$ for each pair of hypotheses in this time step and the process does not stop. In practice, if $\hat{\boldsymbol{\mu}}(t) \notin M$, any designed algorithm can be replaced by the uniform exploration. With this forced exploration, as true mean value $\boldsymbol{\mu} \in M$ (which is an open set by assumption), the law of large numbers ensures that at some point the empirical mean value will fall back into our parameter space again, \emph{i.e.}, $\hat{\boldsymbol{\mu}}(t) \in M$.

We run $\left \lvert \mathcal{X} \right\rvert$ time-varying SGLR tests in parallel, each of which tests $H_0$ against $H_i$ for $i \in [\left \lvert \mathcal{X}(\hat{\boldsymbol{\mu}}(t)) \right\rvert]$. We stop when one of the tests rejects $H_0$. This means that we have empirically found the alternative set that is most easily rejected and we have identified the accepted hypothesis for $\hat{\boldsymbol{\mu}}(t) \in M$ with the highest probability of being right. Given a sequence of exploration rate $(\hat{\beta}_t(\delta))_{t \in \mathbb{N}}$, the SGLR stopping rule for the pure exploration setting is defined as
\begin{equation}
\tau_\delta \coloneqq \inf \left \{ t \in \mathbb{N}: \min_{x \in \mathcal{X}(\hat{\boldsymbol{\mu}}(t))} \hat{\Lambda}_{t,x} > \hat{\beta}_t(\delta) \right \} = \inf \left \{ t \in \mathbb{N}: t \cdot \min_{x \in \mathcal{X}(\hat{\boldsymbol{\mu}}(t))} C_x(\boldsymbol{p}_t;\hat{\boldsymbol{\mu}}(t)) > \hat{\beta}_t(\delta) \right \},
\label{def: stopping rule for pure exploration}
\end{equation}
where the SGLR statistic $\hat{\Lambda}_{t,x}$ is defined in equation \eqref{def: SGLR statistic in pure exploration}. The testing process is quite similar to the classical one except that the hypotheses in each round are data-dependent, changing with time.

We also give some insight from the perspective of the confidence region. It can be noted that $\left \{ \min_{x \in \mathcal{X}(\hat{\boldsymbol{\mu}}(t))} \hat{\Lambda}_{t,x} > \hat{\beta}_t(\delta) \right \} = \left \{ \mathcal{C}_{t, \delta} \subseteq M_{\hat{i}(t)} \right \}$, where $\mathcal{C}_{t, \delta}$ is the confidence region of mean value, given by
\begin{equation}
\mathcal{C}_{t, \delta} \coloneqq \left \{ \boldsymbol{\vartheta}: \sum_{i = 1}^{K} N_i(t)\textup{KL}(\hat{\mu}_i(t), \vartheta_i) \leq \hat{\beta}_t(\delta) \right \}.
\label{def: confidence set of stopping rule}
\end{equation}

Notably, although the confidence region $\mathcal{C}_{t, \delta}$ here is the set for the mean value $\bm{\mu}$, when the perfect linearity is assumed, it is equivalent to the confidence region of parameter $\bm{\theta}$ as defined and mentioned in Section \ref{subsec_Optimal Design} and Section \ref{subsubsec_Visual Explanation of the Lower Bound}. The stopping rule can thus be interpreted as follows: the algorithm stops when the confidence region of mean value parameters shrinks into one part of the partition, which exactly follows the same graphical interpretation as the optimal allocation in Section \ref{subsubsec_Optimal_Allocation}. Based on this stopping framework, different research has been conducted to explore the design of the exploration rate $\hat{\beta}_t(\delta)$ and concentration inequalities, achieving the {$\delta$-PAC} performance for any sampling rule \citep{kaufmann2021mixture}. The anytime and decomposition characteristics of the stopping rule make it possible for our algorithm to concentrate on the sampling rule and be used in the fixed-budget setting directly.

\section{Proof of Proposition \ref{smoothness, PDE, and active candidate set}}\label{proof of proposition: assumption on decomposition}

\proof\\{\textit{Proof.}}
The proof of assumptions in this proposition is twofold. For the first assumption, the alternative set for any given culprit $x$ in our setting can be described as
\begin{equation}
    \textup{Alt}_x(\boldsymbol{\mu}) = \textup{Alt}_{i, j}(\boldsymbol{\mu}) \cup \textup{Alt}_{m}(\boldsymbol{\mu}).
\end{equation}

The definitions of $\textup{Alt}_{i, j}(\boldsymbol{\mu})$ and $\textup{Alt}_{m}(\boldsymbol{\mu})$ are shown in equation~\eqref{eq: alternative set condition} and equation~\eqref{eq: Alt_m_simple} separately. $\textup{Alt}_x(\boldsymbol{\mu})$ is a convex set for all $x$, which means the alternative set $\textup{Alt}(\boldsymbol{\mu})$ is a finite union of convex sets. Convexity can be readily tested by checking if any convex combination of any two points in $\textup{Alt}_x(\boldsymbol{\mu})$ is still in $\textup{Alt}_x(\boldsymbol{\mu})$.

For the second assumption, we know that when the reward distribution follows a single-parameter exponential family, the \textup{KL} divergence \textup{KL}$(\chi, \chi^\prime)$ is continuous in $(\chi, \chi^\prime)$ and strictly convex in $(\chi, \chi^\prime)$, which leads to a unique $\bm{\vartheta}$ that achieves the infimum in equation~(\ref{def: Cx function}).
\hfill\Halmos
\endproof

\section{Difference between G-Optimal Design and $\mathcal{XY}$-Optimal Design}\label{Essential Difference between G and XY}

\begin{figure}[htbp]
    \centering
    \includegraphics[scale = 1]{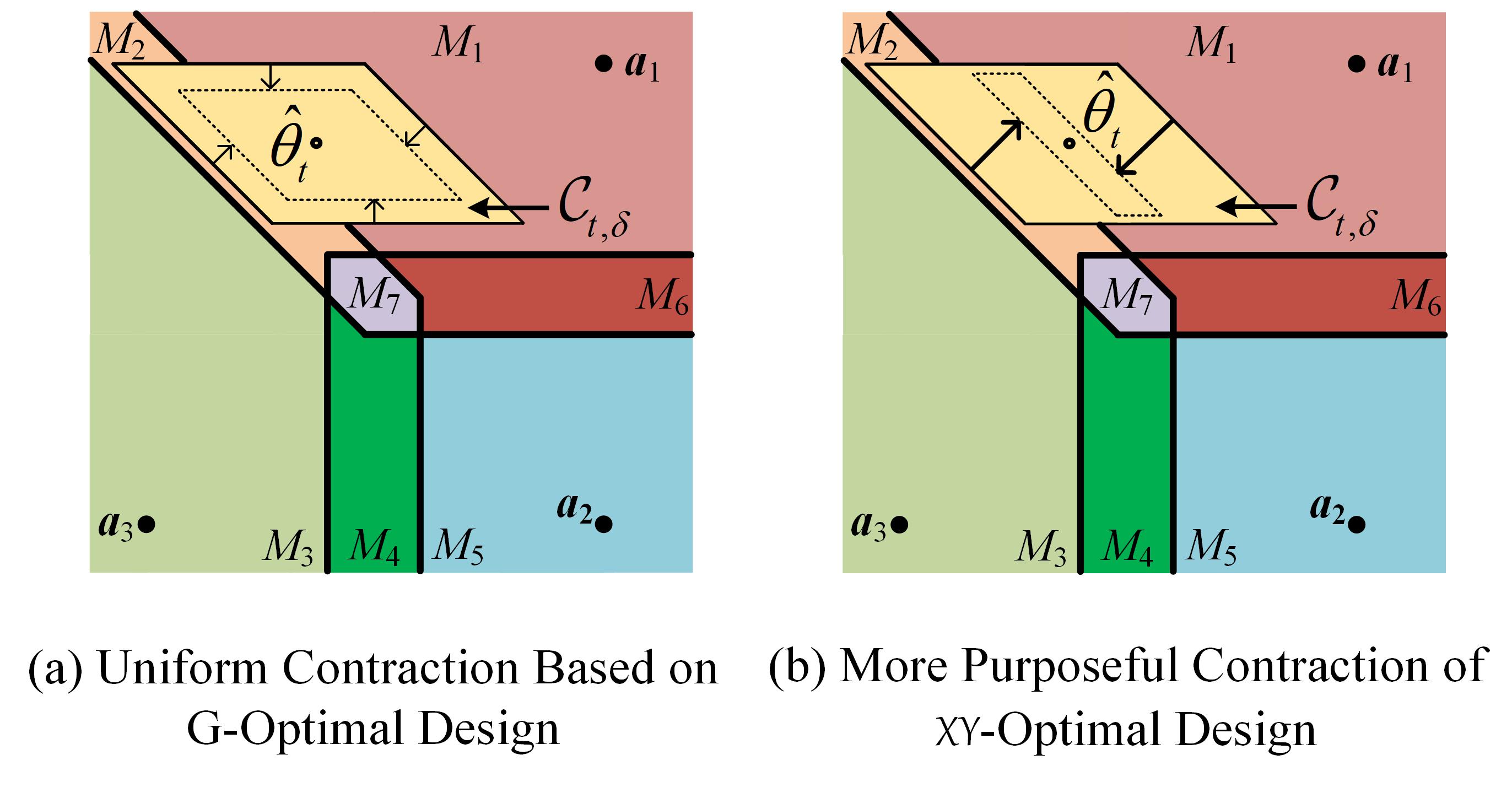}
    \caption{Main Difference Between G-optimal Design and $\mathcal{XY}$-optimal Design From the Perspective of the Stopping Rule.}
    \vspace{0.5pt}
    \noindent
\begin{minipage}{\linewidth}
    \footnotesize
    \raggedright
    \textit{Note.} \textit{The contraction method and speed of the confidence region for parameter $\hat{\bm{\theta}}_t$ are quite different: the left G-optimal sampling shrinks uniformly, while the right $\mathcal{XY}$-optimal design shrinks more purposefully, accelerating along directions that facilitate classification stopping and ensure the confidence region falls within a decision area.
    }
\end{minipage}
    \label{fig: difference_between_G_and_XY.jpg}
\end{figure}

Back to the intuition and visual explanation provided in section \ref{subsubsec_Visual Explanation of the Lower Bound} and Figure \ref{fig: 1}, it can be found that the G-optimal sampling-based LinFACT, \emph{i.e.}, LinFACT-G, will certainly fail in achieving the most efficient sampling. Specifically, here Figure~\ref{fig: difference_between_G_and_XY.jpg} is an example that shows the merits of introducing the $\mathcal{XY}$-optimal design from the perspective of the stopping rule. As long as the yellow confidence region of parameter $\hat{\bm{\theta}}$ shrinks into one of these decision regions $M_i$ ($i=1,2,\dots,7$), the algorithm is terminated. From this figure we can find that compared to the uniform contraction of the confidence region $\mathcal{C}_{t,\delta}$ based on the G-optimal design, the confidence region designed by the $\mathcal{XY}$-optimal design shrinks more purposefully in certain directions that are harder and need more exploration budget, being more efficient and direct in distinguishing different arms. This sampling policy does not provide a uniformly accurate estimate of the true parameter $\bm{\theta}$ itself; rather, it focuses on pulling arms that reduce the uncertainty in the estimation of $\bm{\theta}$ in the directions of interest.

\section{Proof of Proposition \ref{SGLR statistic in pure exploration}}\label{proof of proposition: SGLR statistic in pure exploration}

\proof\\{\textit{Proof.}}
This proposition can be proved in a more general setting, that is, the exponential family. Here we provide our results in the case where the distributions of arms belong to a canonical exponential family, \emph{i.e.},
\begin{equation}\label{def: exponential family}
    \mathcal{P} = \left\{ \left( \nu_\chi \right)_{\chi \in \Theta}: \frac{d\nu_\chi}{d\xi} = \exp\left( \chi x-b\left(\chi\right) \right) \right\},
\end{equation}
where $\Theta \in \mathbb{R}$, $\xi$ is some reference measure on $\mathbb{R}$, and $b: \Theta \to \mathbb{R}$ is a convex, twice differentiable function. A distribution $\nu_\chi \in \mathcal{P}$ can be parameterized by its mean value $\dot{b}(\chi)$. Then a \textup{SGLR} statistic is defined for testing a potentially composite null hypothesis $H_0: (\boldsymbol{\mu} \in \Omega_0)$ versus a potentially composite alternative hypothesis $H_1: (\boldsymbol{\mu} \in \Omega_1)$ by
\begin{equation}
\textup{SGLR}_t = \frac{\sup_{\chi \in \Omega_0 \cup \Omega_1}L(X_1, X_2, \ldots, X_t ; \chi)}{\sup_{\chi \in \Omega_0}L(X_1, X_2, \ldots, X_t ; \chi)},
\end{equation}
where $X_1, X_2, \ldots, X_t$ are observed rewards from arm pulling up to time step $t$, and $L(\cdot)$ is the likelihood function with these observed data and some unknown parameter $\chi \in \Omega_0$. The log-likelihood statistic $\hat{\Lambda}_{t,x}$ for a given culprit $x$ can thus be represented as
\begin{equation}
\hat{\Lambda}_{t,x} = \ln (\textup{SGLR}_t) &= \ln \frac{\sup_{\chi \in \Omega_0 \cup \Omega_1}L(X_1, X_2, \ldots, X_t ; \chi)}{\sup_{\chi \in \Omega_0}L(X_1, X_2, \ldots, X_t ; \chi)}.
\end{equation}

Considering the numerator, for the exponential family, the empirical mean value is the maximum likelihood estimation of the parameter and the unknown parameter $\chi \in \Omega_0$ has a one-to-one correspondence relationship with mean value $\mu_i$ for each arm $i \in [K]$. Then we have
\begin{equation}
    \sup_{\chi \in \Omega_0 \cup \Omega_1}L(X_1, X_2, \ldots, X_t ; \chi) = L(\hat{\chi}).
\end{equation}

Besides, the restricted parameter space under the null hypothesis $\Omega_0$ can be defined as $\textup{Alt}_x(\hat{\boldsymbol{\mu}}(t))$ for any given culprit $x$ at time step $t$. Then we have
\begin{align}
\hat{\Lambda}_{t,x} &= \ln \frac{\sup_{\chi \in \Omega_0 \cup \Omega_1}L(X_1, X_2, \ldots, X_t ; \chi)}{\sup_{\chi \in \Omega_0}L(X_1, X_2, \ldots, X_t ; \chi)} \nonumber\\
&= \inf_{\chi \in \Omega_0} \ln \frac{L(\hat{\chi})}{L(X_1, X_2, \ldots, X_t ; \chi)} \nonumber\\
&= \inf_{\chi \in \Omega_0} \ln \frac{\exp \left[ \sum_{i \in [K]} \left( N_i \hat{\chi}_i \frac{\sum_{j \in [N_i]} X_{ij}}{N_i} - N_i b\left( \hat{\chi}_i \right) \right) \right]}{\exp \left[ \sum_{i \in [K]} \left( N_i {\chi}_i \frac{\sum_{j \in [N_i]} X_{ij}}{N_i} - N_i b\left( {\chi}_i \right) \right) \right]} \nonumber\\
&= \inf_{\chi \in \Omega_0} t \sum_{i \in [K]} p_i \left[ b(\chi_i)-b(\hat{\chi}_i) + \hat{\mu}_i(\hat{\chi}_i - \chi_i) \right] \nonumber\\
&= \inf_{\chi \in \Omega_0} t \sum_{i \in [K]} p_i \left[ b(\chi_i)-b(\hat{\chi}_i) - \dot{b}(\hat{\mu}_i)(\chi_i - \hat{\chi}_i) \right] \nonumber\\
&= \inf\limits_{\chi \in \Omega_0} \sum_{i \in \left [ K \right ]} N_i(t)\textup{KL}(\hat{\chi}_i, {\chi}_i)\nonumber\\
&= \inf\limits_{\boldsymbol{\vartheta} \in \textup{Alt}_x(\hat{\boldsymbol{\mu}}(t))} \sum_{i \in \left [ K \right ]} N_i(t)\textup{KL}(\hat{\mu}_i(t), \vartheta_i),
\end{align}
where $N_i$ represents the number of samples for arm $i$ and $X_{ij}$ represents observed value for arm $i$ up to time $t$.
\hfill\Halmos
\endproof

\section{Lower Bound for All $\varepsilon$-Best Arms Identification in Stochastic Bandits}\label{subsubsec_Lower Bound for the Stochastic Bandit} 

In this section, we present the lower bound results for the all $\varepsilon$-best arms identification problem in the stochastic setting. These results serve as a foundation for extending the analysis to the linear bandit case, which we discuss in Section \ref{subsubsec_All-varepsilon Good Arms Identification in Linear Bandit}.

\begin{figure}
    \centering
    \includegraphics[scale = 1]{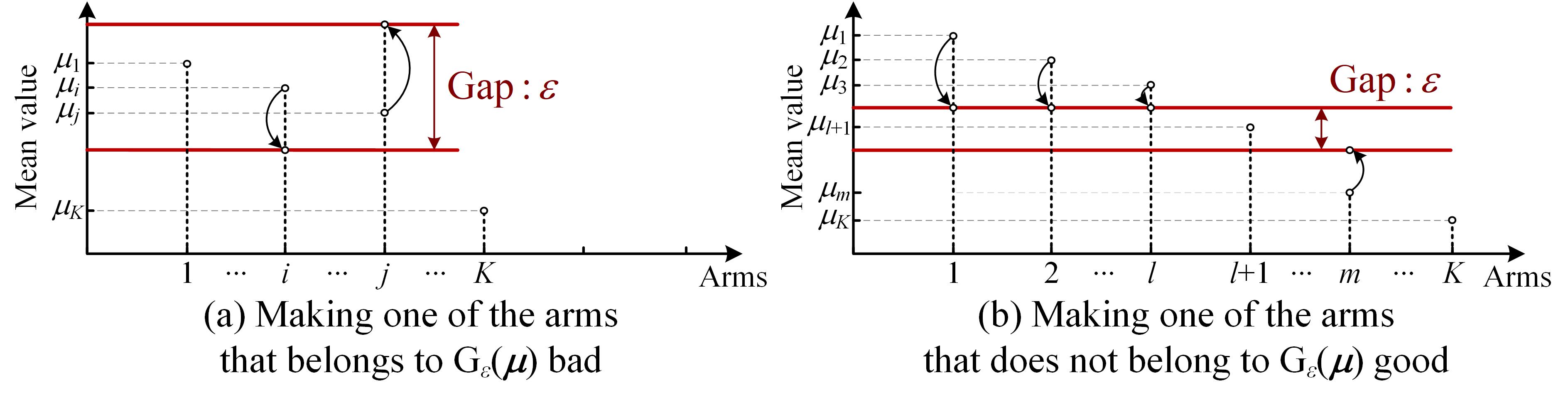}
    \caption{Diagram for Constructing Alternative Set of All $\varepsilon$-Best Arms Identification Lower Bound}
    \vspace{5pt}
    \noindent
\begin{minipage}{\linewidth}
    \footnotesize
    \raggedright
    \textit{Note.} \textit{
       (a) Transforming an arm within $G_{\varepsilon}(\boldsymbol{\mu})$ into a non-$\varepsilon$-best arm. This is done by decreasing the mean value of a currently $\varepsilon$-best arm $i$ while increasing the mean of another arm $j$ until $i$ no longer meets the $\varepsilon$-best criterion. (b) Transforming an arm outside $G_{\varepsilon}(\boldsymbol{\mu})$ into an $\varepsilon$-best arm. This is achieved by increasing the mean value of an arm $m$ that is not in $G_{\varepsilon}(\boldsymbol{\mu})$ while lowering the means of the arms with higher values, bringing $m$ into the $\varepsilon$-best set.
    }
\end{minipage}
    \label{fig: 2}
\end{figure}

Identifying the lower bound involves constructing the largest possible alternative set and pinpointing the most challenging one to identify. For all $\varepsilon$-best arms identification in stochastic bandit, \cite{al2022complexity} presents the tightest lower bound up to now. Their characterization is based on two cases shown in Figure~\ref{fig: 2}. 

We recall the following major components and we need to make the forms of these components clear to give the current lower bound result of all $\varepsilon$-best arms identification in stochastic bandit.
\begin{itemize}
\item The correct answer $G_\varepsilon(\bm{\mu})$ corresponding to the preset exploration task.
\item The culprit set $\mathcal{X}(\boldsymbol{\mu})$.
\item The alternative set $\textup{Alt}(\boldsymbol{\mu})$ based on the culprit set $\mathcal{X}(\boldsymbol{\mu})$, which satisfies the existence of optimal solutions and the union structure in Proposition \ref{smoothness, PDE, and active candidate set}.
\item The $C_x(\boldsymbol{p})$ function that quantifies the amount of information we need to reject the wrong hypothesis.
\end{itemize}

The culprit set and the alternative set can be defined as
\begin{align}
    \mathcal{X}(\boldsymbol{\mu}) &= \{ (i, j, m, \ell): i \in G_\varepsilon(\bm{\mu}), j \neq i, m \notin G_\varepsilon(\bm{\mu}), \ell \in [\vert 1, m-1 \vert] \}, \label{equation_X_epsilon}\\
    \textup{Alt}_x(\boldsymbol{\mu}) &= \textup{Alt}_{i, j}(\boldsymbol{\mu}) \cup \textup{Alt}_{m, \ell}(\boldsymbol{\mu}) \text{ for all } x \in \mathcal{X}(\boldsymbol{\mu}), \\
    \textup{Alt}_{i, j}(\boldsymbol{\mu}) &= \left \{ \boldsymbol{\mu}: \mu_i - \mu_j < -\varepsilon \right \} \text{ for all } \; x \in \mathcal{X}(\boldsymbol{\mu}), \\
    \textup{Alt}_{m, \ell}(\boldsymbol{\mu}) &= \left \{ \boldsymbol{\mu}: \mu_\ell \geq \mu_\varepsilon^{m, \ell}{p} + \varepsilon > \mu_{\ell+1} \} \right \} \text{ for all } \; x \in \mathcal{X}(\boldsymbol{\mu}),\label{eqaution_Alt_ml}
\end{align}
where
\begin{equation}
    \mu_{\varepsilon}^{m, \ell}(\bm{p}) \coloneqq \frac{p_m \mu_m+\sum_{i=1}^{\ell} p_i\left(\mu_i-\varepsilon\right)}{p_m+\sum_{i=1}^{\ell} p_i}
\end{equation}
is the optimal mean value of arm $m$ in the alternative bandit instance for parameters $(m, \ell)$. This threshold is designed to minimize the infimum of the KL divergence-based objective $C_{m, \ell}(\bm{p})$. 

The core of deriving the tightest lower bound involves constructing alternative bandit instances that comprehensively test each $\varepsilon$-best arm instance. An alternative bandit instance $\bm{\vartheta} \in \operatorname{Alt}(\boldsymbol{\mu})$ differs in expected mean reward from the original bandit instance $\boldsymbol{\mu}$. Each culprit $x \in \mathcal{X}(\boldsymbol{\mu})$ is decomposed into $\mathrm{Alt}_{i, j}(\boldsymbol{\mu})$ and $\mathrm{Alt}_{m, \ell}(\boldsymbol{\mu})$, corresponding to two distinct ways for creating alternative bandit instances, one by transforming an arm within $G_{\varepsilon}(\boldsymbol{\mu})$ into a non-$\varepsilon$-best arm, and another one by transforming an arm outside $G_{\varepsilon}(\boldsymbol{\mu})$ into an $\varepsilon$-best arm, as depicted in Figure~\ref{fig: 2}. The minimum of the function $C_x$ across all culprit sets $x \in \mathcal{X}(\boldsymbol{\mu})$ can thus be expressed as: 
\begin{equation}
    \min \left\{\min_{x \in \mathcal{X}} C_{i, j}(\bm{p}), \min_{x \in \mathcal{X}} C_{m, \ell}(\bm{p}) \right\}. 
\end{equation}


Next, in Theorem \ref{lower bound: existing 3}, we introduce the lower bound for all $\varepsilon$-best arms identification in stochastic bandits.

\begin{theorem}\label{lower bound: existing 3}\textup{[\cite{al2022complexity}] }
Fix $\delta, \varepsilon>0$ and consider $K$ arms. Assume the value of the $i^{\text {th }}$ arm be distributed according to $\mathcal{N}\left(\mu_i, 1\right)$. Then any $\delta$-PAC algorithm for the additive setting satisfies
\begin{align}
    \frac{\mathbb{E}_{\boldsymbol{\mu}} \left [ \tau_\delta \right ]}{\log\left({1}/{2.4 \delta}\right)} &\geq \min_{p \in \mathcal{S}_K} \max \left\{ \frac{1}{\min_{x \in \mathcal{X}} C_{i, j}(\bm{p})}, \frac{1}{\min_{x \in \mathcal{X}} C_{m, \ell}(\bm{p})} \right\} \nonumber\\
    &= \min_{p \in \mathcal{S}_K} \max_{x \in \mathcal{X}} \max \left\{ \frac{1}{C_{i, j}(\bm{p})}, \frac{1}{ C_{m, \ell}(\bm{p})} \right\} \nonumber\\
    &= 2 \min_{p \in \mathcal{S}_K} \max \Bigg\{  \frac{1/p_i+1/p_j}{(\mu_i - \mu_j + \varepsilon)^2}, \nonumber\\
    & \frac{1}{\inf_{\boldsymbol{\vartheta} \in \textup{Alt}_{m, \ell}(\boldsymbol{\mu})} \left[ \sum_{i=1}^\ell (\mu_i - \mu_{\varepsilon}^{m, \ell}(\bm{p}) - \varepsilon)^2 p_i + (\mu_m - \mu_{\varepsilon}^{m, \ell}(\bm{p}))^2 p_m \right]} \Bigg\}, \label{All epsilon lower bound}
\end{align}
where the indices $i$, $j$, $m$, and $l$, which specify the culprit set, are introduced in equation (\ref{equation_X_epsilon}). The alternative set $\textup{Alt}_{m, \ell}(\boldsymbol{\mu})$ is defined in equation (\ref{eqaution_Alt_ml})

\end{theorem}

The above theorem recovers the previous lower bound result for identifying all $\varepsilon$-best arms in stochastic bandits, as proposed by \cite{mason2020finding}. However, their result is less tight due to the incompleteness of the alternative set. Specifically, we have:
\begin{enumerate}
    \item Setting $j = 1$, $j \neq i$, $i \in G_\varepsilon(\bm{\mu})$, $\ell=1$ and $\mu_{\varepsilon}^{m, \ell}(\bm{p}) = \mu_1 - \varepsilon$, the first and second terms in Theorem \ref{lower bound: existing 3} match the first term in Theorem 2.1 of \cite{mason2020finding}.
    \item Setting $i = k$ and $j \neq k$, the first term in Theorem \ref{lower bound: existing 3} matches the second term in Theorem 2.1 of \cite{mason2020finding}.
\end{enumerate}

\section{Lower Bound for All $\varepsilon$-Best Arms Identification in Linear Bandits}
This section provides both geometric insights regarding the stopping condition and formal proofs establishing the lower bound for identifying all $\varepsilon$-best arms in linear bandit settings.

\subsection{Visual Illustration of the Stopping Condition}\label{subsubsec_Visual Explanation of the Lower Bound}

\begin{figure}[htbp]
    \centering
    \includegraphics[scale=1]{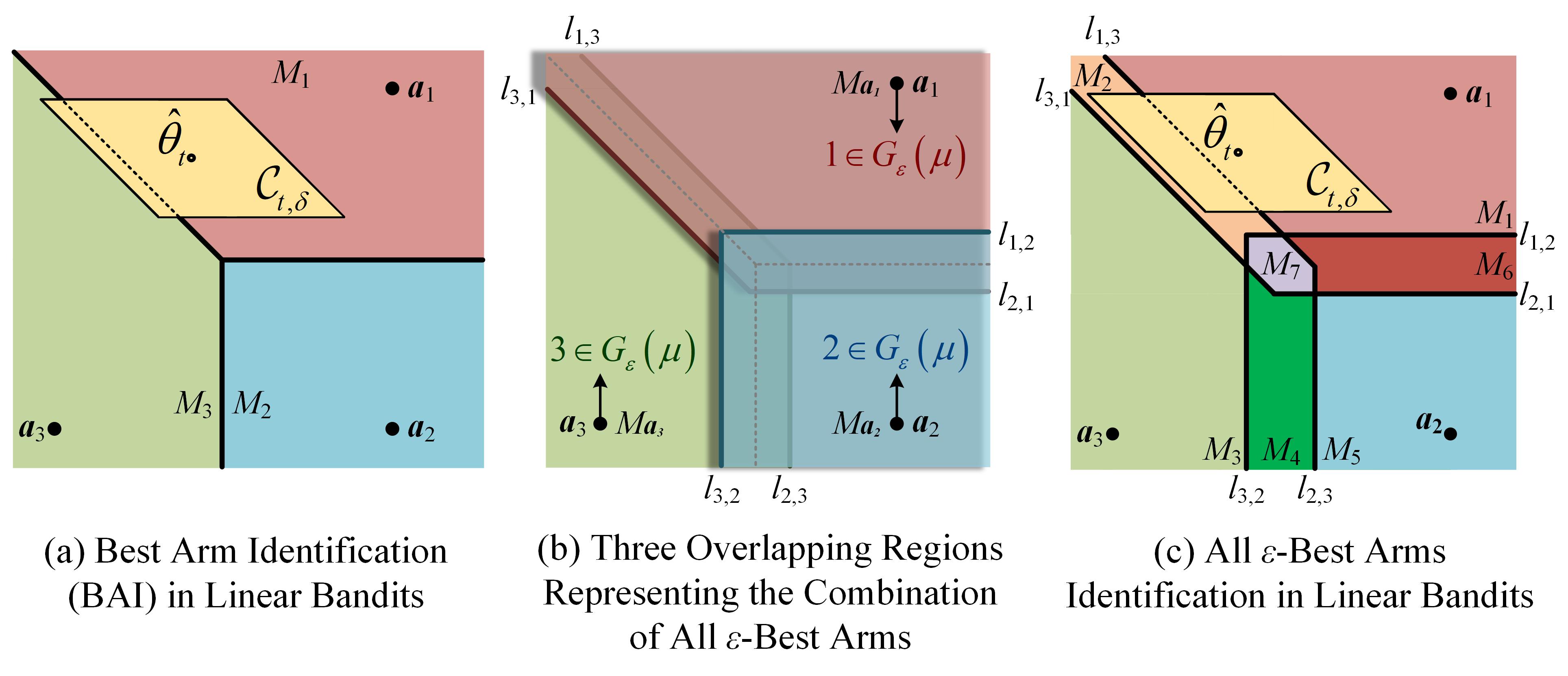}
    \caption{Visual Illustration of Identifying the Best Arm vs. Identifying All $\varepsilon$-Best Arms.
    }
\vspace{5pt}
\noindent
\begin{minipage}{\linewidth}
    \footnotesize
    \raggedright
    \textit{Note.} \textit{
        {(a)} Stopping occurs when the confidence region $\mathcal{C}_{t, \delta}$ for the estimated parameter $\hat{\boldsymbol{\theta}}_t$ contracts entirely within one of the three decision regions $M_i$ in a certain time step $t$. The boundaries between regions are defined by the hyperplanes $\boldsymbol{\vartheta}^\top(\boldsymbol{a}_i - \boldsymbol{a}_j) = 0$. Each dot represents an arm. {(b)} In the case of identifying all $\varepsilon$-best arms, the regions overlap. {(c)} Due to these overlaps, the space is partitioned into seven distinct decision regions. 
    }
\end{minipage}
    \label{fig: 1}
\end{figure}

The stopping condition is formulated as a hypothesis test conducted as data is collected, which can be interpreted as the process of the parameter confidence region contracting into one of the decision regions (\emph{i.e.}, the set of parameters that yield the same decision). Figure~\ref{fig: 1} illustrates the core concept of the stopping condition involved in identifying all $\varepsilon$-best arms in linear bandits. The primary difference between identifying all $\varepsilon$-best arms and identifying the single best arm lies in the way decision regions are partitioned.

Figure~\ref{fig: 1}(a) illustrates the best arm identification process in linear bandits, where $\bm{a}^\ast = \bm{a}^\ast(\bm{\mu})$ represents the arm with the largest mean value for each bandit instance $\bm{\mu}$. Let $M_i = \{ \boldsymbol{\vartheta} \in \mathbb{R}^d \mid \bm{a}_i = \bm{a}^\ast \}$ be the set of parameters $\bm{\theta}$ for which $\bm{a}_i$ ($i = 1,2,3$) is the optimal arm. Each $M_i$ forms a cone defined by the intersection of half-spaces.

Figure~\ref{fig: 1}(b) represents an intermediate step, demonstrating the transition from best arm identification to identifying all $\varepsilon$-best arms. Let $M_{\bm{a}_i} = \{ \boldsymbol{\vartheta} \in \mathbb{R}^d \mid i \in G_\varepsilon(\bm{\mu}) \}$ be the set of parameters $\bm{\theta}$ include arm $i$ in the set $G_\varepsilon(\bm{\mu})$. $M_{\bm{a}_i}$ is similarly defined by the intersection of half-spaces. The overlap of these three regions forms the decision regions $M_i$ ($i = 1,2,\ldots,7$) in Figure~\ref{fig: 1}(c), which correspond to the seven distinct types of $\varepsilon$-best arms sets. 
Besides, the BAI process in (a) is a special case of the $\varepsilon$-best arms identification in (c), occurring when the gap $\varepsilon$ approaches 0. The following statement provides a detailed explanation of how the decision regions in Figure~\ref{fig: 1}(c) are constructed.

In a $d$-dimensional Euclidean space $\mathbb{R}^d$, hyperplanes $l_{i,j}$ can be defined for any pair of arms, partitioning the space into the following half-spaces:
\begin{align}
    H_{i,j}^+ &= \left\{ \boldsymbol{\vartheta} \in \mathbb{R}^d \mid (\boldsymbol{a}_{i} - \boldsymbol{a}_{j})^\top \boldsymbol{\vartheta} > \varepsilon \right\}, \\ 
    H_{i,j}^- &= \left\{ \boldsymbol{\vartheta} \in \mathbb{R}^d \mid (\boldsymbol{a}_{i} - \boldsymbol{a}_{j})^\top \boldsymbol{\vartheta} \leq \varepsilon \right\}.
\end{align}

The hyperplane $l_{i,j}$, which separates the half-spaces $H_{i,j}^+$ and $H_{i,j}^-$, is perpendicular to the direction vector $\boldsymbol{a}_{i} - \boldsymbol{a}_{j}$. The intersection of these half-spaces, represented by $\bigcap_{i, j \in [K],\, j \neq i} H_{i,j}$, partitions the space into distinct regions. Each region corresponds to a solution set, representing all $\varepsilon$-best arms if the true parameter $\boldsymbol{\theta}$ lies within that region. As the gap $\varepsilon$ approaches 0, the hyperplanes $l_{i,j}$ on both sides of the decision boundaries move closer together, causing the decision regions $M_i$ (for $i = 2,4,6,7$ in Figure~\ref{fig: 1}(c)) to shrink until they vanish. The relationship between the true parameter $\boldsymbol{\theta}$ and the half-spaces determines the membership of arms in the good set $G_\varepsilon(\boldsymbol{\mu})$.

For the case of three arms, the space is divided into three overlapping regions, as shown in Figure~\ref{fig: 1}(b). These regions further generate seven ($2^3 - 1 = 7$) decision regions, denoted $M_1$ through $M_7$, which are summarized in Table~\ref{tab_decisionregion}. 

\begin{table}[htbp]
    \small 
    \centering
    \renewcommand\arraystretch{1.5}
    \caption{Three Overlapping Regions and Seven Decision Regions in the Case of Three Arms}
    \begin{tabular}{@{}c@{\hspace{0.5em}}c@{\hspace{0.5em}}c@{}}  
        \hline
        \multicolumn{1}{c}{Decision Region} & 
        \multicolumn{1}{c}{Set Expression} & 
        \multicolumn{1}{c}{$\varepsilon$-Best Arms} \\
        \hline
        $M_1$ & $M_{a_1} \cap M_{a_2}^c \cap M_{a_3}^c$ & $\{1\}$ \\
        $M_2$ & $M_{a_1} \cap M_{a_2}^c \cap M_{a_3}$ & $\{1,3\}$ \\
        $M_3$ & $M_{a_1}^c \cap M_{a_2}^c \cap M_{a_3}$ & $\{3\}$ \\
        $M_4$ & $M_{a_1}^c \cap M_{a_2} \cap M_{a_3}$ & $\{2,3\}$ \\
        $M_5$ & $M_{a_1}^c \cap M_{a_2} \cap M_{a_3}^c$ & $\{2\}$ \\
        $M_6$ & $M_{a_1} \cap M_{a_2} \cap M_{a_3}^c$ & $\{1,2\}$ \\
        $M_7$ & $M_{a_1} \cap M_{a_2} \cap M_{a_3}$ & $\{1,2,3\}$ \\
        \hline
    \end{tabular}
    \label{tab_decisionregion}
\end{table}


The stopping condition verifies whether the confidence region $\mathcal{C}_{t, \delta}$ is entirely contained within a specific decision region $M_i$. It is important to note that, due to the definition of the confidence region and the property that $\hat{\boldsymbol{\theta}}_t \to \boldsymbol{\theta}$ as $t \to \infty$, any algorithm that continually samples all arms will eventually meet the stopping condition. However, if the region $\mathcal{C}_{t, \delta}$ overlaps with multiple decision regions, it becomes ambiguous to identify all the arms in $G_\varepsilon(\boldsymbol{\mu})$. In contrast, when all possible values of $\hat{\bm{\theta}}_t$ are included with high probability within the correct decision region $M^\ast$, we can accurately identify all $\varepsilon$-best arms. 



\subsection{Optimal Allocation}\label{subsubsec_Optimal_Allocation}

The objective of the sampling policy is to define an allocation sequence that leads to $B_{t, \delta} \subset M^\ast$ as quickly as possible. From a geometrical point of view, this corresponds to choosing arms such that the confidence set $B_{t, \delta}$ shrinks into the optimal cone $M^\ast$ with the smallest number of samples. Condition $\mathcal{C}_{t, \delta} \subset M^\ast$ can be represented as 
\begin{equation}\label{eq_ora_0}
    \text{For all } i \in G_\varepsilon(\boldsymbol{\mu}),\, j \neq i,\, m \notin G_\varepsilon(\boldsymbol{\mu}) \text{ and } \forall\boldsymbol{\vartheta} \in \mathcal{C}_{t, \delta}, \text{ we have } \boldsymbol{\vartheta} \in H_{j,i}^- \text{ and } \boldsymbol{\vartheta} \in H_{1,m}^+,
\end{equation}
which means that for every pair of arms $i$ and $j$, where $i$ is an $\varepsilon$-best arm and a different arm $j$, and for any arm $m$ that is not $\varepsilon$-best, every parameter vector $\boldsymbol{\vartheta}$ within the confidence region $\mathcal{C}_{t, \delta}$ lies within the half-space $H_{j,i}^-$ and $H_{1,m}^+$.

The relationships $\boldsymbol{\vartheta} \in H_{j,i}^-$ and $\boldsymbol{\vartheta} \in H_{1,m}^+$ are equivalent to the following inequalities by adding terms on both sides of the inequalities and reorganizing:
\begin{equation}\label{eq_ora_1}
    \left\{
    \begin{array}{l}
        (\boldsymbol{a}_i - \boldsymbol{a}_j)^\top (\boldsymbol{\theta} - \boldsymbol{\vartheta}) \leq (\boldsymbol{a}_i - \boldsymbol{a}_j)^\top \boldsymbol{\theta} + \varepsilon \\
        (\boldsymbol{a}_1 - \boldsymbol{a}_m)^\top (\boldsymbol{\theta} - \boldsymbol{\vartheta}) < (\boldsymbol{a}_1 - \boldsymbol{a}_m)^\top \boldsymbol{\theta} - \varepsilon
    \end{array}.
    \right.
\end{equation}

Now, we focus on the confidence region, which can be constructed following Cauchy's inequality and the definition of the confidence ellipse for parameter $\bm{\theta}$ in equation~\eqref{confidence for ellipse}.
\begin{equation}
    \mathcal{C}_{t, \delta} = \left\{ \boldsymbol{\vartheta} \in \mathbb{R}^d \,\middle|\, \forall i \in G_\varepsilon(\boldsymbol{\mu}),\, j \neq i,\, m \notin G_\varepsilon(\boldsymbol{\mu}),\,
    \begin{cases}
        (\boldsymbol{a}_i - \boldsymbol{a}_j)^\top (\boldsymbol{\theta} - \boldsymbol{\vartheta}) \leq \Vert \boldsymbol{a}_i - \boldsymbol{a}_j \Vert_{\boldsymbol{V}_t^{-1}} B_{t, \delta} \\
        (\boldsymbol{a}_1 - \boldsymbol{a}_m)^\top (\boldsymbol{\theta} - \boldsymbol{\vartheta}) \leq \Vert \boldsymbol{a}_1 - \boldsymbol{a}_m \Vert_{\boldsymbol{V}_t^{-1}} B_{t, \delta}
    \end{cases}
    \right\},
\end{equation}
where $\boldsymbol{V}_t$ is the information matrix as defined in equation~\eqref{OLS} and the confidence bound for parameter $\boldsymbol{\theta}$, \emph{i.e.}, $B_{t, \delta}$, can either be a fixed confidence bound as shown in Proposition \ref{proposition_confidence bound for B 1} or a looser adaptive confidence bound introduced in \cite{abbasi2011improved}. The stopping condition $\mathcal{C}_{t, \delta} \subset M^\ast$ can thus be reformulated. For each $i \in G_\varepsilon(\boldsymbol{\mu}),\, j \neq i,\, m \notin G_\varepsilon(\boldsymbol{\mu})$, we have
\begin{equation}\label{eq_oracle_stopping}
    \left\{
    \begin{array}{l}
        \Vert \boldsymbol{a}_i - \boldsymbol{a}_j \Vert_{\boldsymbol{V}_t^{-1}} B_{t, \delta} \leq (\boldsymbol{a}_i - \boldsymbol{a}_j)^\top \boldsymbol{\theta} + \varepsilon \\
        \Vert \boldsymbol{a}_1 - \boldsymbol{a}_m \Vert_{\boldsymbol{V}_t^{-1}} B_{t, \delta} \leq (\boldsymbol{a}_1 - \boldsymbol{a}_m)^\top \boldsymbol{\theta} - \varepsilon
    \end{array}.
    \right.
\end{equation}

If equation \eqref{eq_oracle_stopping} holds, then for any $\boldsymbol{\theta} \in \mathcal{C}_{t, \delta}$, equation \eqref{eq_ora_1} also holds, implying that $\mathcal{C}_{t, \delta} \subset M^\ast$. Given that $(\boldsymbol{a}_i - \boldsymbol{a}_j)^\top \boldsymbol{\theta} + \varepsilon > 0$ and $(\boldsymbol{a}_1 - \boldsymbol{a}_m)^\top \boldsymbol{\theta} - \varepsilon > 0$, the oracle allocation strategy is determined as follows:
\begin{equation}\label{eq: lower bound_visual_t}
    \{ \bm{a}_{A_n} \}^\ast = \arg \min_{\{ \bm{a}_{A_n} \}} \max_{i \in G_\varepsilon(\boldsymbol{\mu}),\, j \neq i,\, m \notin G_\varepsilon(\boldsymbol{\mu})} \max \left\{ \frac{2 \Vert \boldsymbol{a}_i - \boldsymbol{a}_j \Vert_{\boldsymbol{V}_t^{-1}}^2}{\left( \boldsymbol{a}_i^\top \boldsymbol{\theta} - \boldsymbol{a}_j^\top \boldsymbol{\theta} + \varepsilon \right)^2}, \frac{2 \Vert \boldsymbol{a}_1 - \boldsymbol{a}_m \Vert_{\boldsymbol{V}_t^{-1}}^2}{\left( \boldsymbol{a}_1^\top \boldsymbol{\theta} - \boldsymbol{a}_m^\top \boldsymbol{\theta} - \varepsilon \right)^2} \right\},
\end{equation}
where $\{ \bm{a}_{A_n} \}= (\bm{a}_{A_1},\bm{a}_{A_1},\ldots,\bm{a}_{A_n}) \in \mathcal{A}^n$ is a sequence of sampled arms and $\{ \bm{a}_{A_n} \}^\ast$ is the oracle allocation strategy\footnote{When the arm elimination is considered, the active set of arms will gradually decrease as arms are removed, becoming a subset of $\mathcal{A}$.}. However, considering that it is more convenient to demonstrate the sample complexity of the problem in terms of the continuous allocation proportion $\bm{p}$ instead of the discrete allocation sequence $\{ \bm{a}_{A_n} \}$. Then, we can have the following optimal allocation proportion. 
\begin{equation}\label{eq: lower bound_visual}
    \boldsymbol{p}^\ast = \arg \min_{\boldsymbol{p} \in \mathcal{S}_K} \max_{i \in G_\varepsilon(\boldsymbol{\mu}),\, j \neq i,\, m \notin G_\varepsilon(\boldsymbol{\mu})} \max \left\{ \frac{2 \Vert \boldsymbol{a}_i - \boldsymbol{a}_j \Vert_{\boldsymbol{V}_{\boldsymbol{p}}^{-1}}^2}{\left( \boldsymbol{a}_i^\top \boldsymbol{\theta} - \boldsymbol{a}_j^\top \boldsymbol{\theta} + \varepsilon \right)^2}, \frac{2 \Vert \boldsymbol{a}_1 - \boldsymbol{a}_m \Vert_{\boldsymbol{V}_{\boldsymbol{p}}^{-1}}^2}{\left( \boldsymbol{a}_1^\top \boldsymbol{\theta} - \boldsymbol{a}_m^\top \boldsymbol{\theta} - \varepsilon \right)^2} \right\},
\end{equation}
where $\mathcal{S}_K$ denotes the $K$-dimensional probability simplex, and $\bm{V}_{\bm{p}} = \sum_{i=1}^{K}p_i \boldsymbol{a}_i\boldsymbol{a}_i^\top$ is the weighted information matrix, analogous to the Fisher information matrix \citep{chaloner1995bayesian}. The intuition behind this optimal allocation strategy is as follows: to satisfy the inequality in \eqref{eq_oracle_stopping} as quickly as possible, the ratio of the left-hand side to the right-hand side should be minimized, leading to the outer minimization operation over the allocation probability $\boldsymbol{p}$. The middle maximization operation accounts for the fact that the inequalities in \eqref{eq_oracle_stopping} must hold for each possible triple $(i, j, m)$, which specifies the culprit set $\mathcal{X} = \mathcal{X}(\boldsymbol{\mu})$ (see Proposition \ref{general lower bound} for a formal definition). Furthermore, since both inequalities in \eqref{eq_oracle_stopping} need to be satisfied, we take inner maximization operations to enforce this requirement.

\subsection{Proof of the Lower Bound in Theorem \ref{lower bound: All-epsilon in the linear setting}}\label{proof of theorem: lower bound for all-epsilon in the linear bandit}

\proof\\{\textit{Proof.}}
For all $\varepsilon$-best arms identification in linear bandits, we establish the correct answer $\mathcal{I}$, the culprit set $\mathcal{X}(\boldsymbol{\mu})$, and the alternative set $\textup{Alt}(\boldsymbol{\mu})$ as follows.

\begin{itemize}
\item The correct answer $G_\varepsilon(\bm{\mu}) \coloneqq \left \{ i:\left\langle \boldsymbol{\theta}, \boldsymbol{a}_i \right\rangle \geq \max_i\left\langle \boldsymbol{\theta}, \boldsymbol{a}_i \right\rangle - \varepsilon \right \}$ for all $i \in [K]$. 
\item The culprit set $\mathcal{X}(\boldsymbol{\mu}) = \{ (i, j, m, \ell): i \in G_\varepsilon(\bm{\mu}), j \neq i, m \notin G_\varepsilon(\bm{\mu}), \ell \in [\vert 1, m-1 \vert] \}$.
\item The alternative set $\textup{Alt}(\boldsymbol{\mu}) = \cup_{x \in \mathcal{X}(\boldsymbol{\mu})} \textup{Alt}_x(\boldsymbol{\mu})$ and the form of $\textup{Alt}_x(\boldsymbol{\mu})$ is given by
\end{itemize}
\begin{equation}
    \textup{Alt}_x(\boldsymbol{\mu}) = \textup{Alt}_{i, j}(\boldsymbol{\mu}) \cup \textup{Alt}_{m, 
    \; \ell}(\boldsymbol{\mu}) \text{ for all } x \in \mathcal{X}(\boldsymbol{\mu}).
\end{equation}

The two parts of the alternative sets can be expressed as
\begin{equation}\label{eq: alternative set condition}
    \textup{Alt}_{i, j}(\boldsymbol{\mu}) = \textup{Alt}_{i, j}(\boldsymbol{\theta}) = \left \{ \boldsymbol{\vartheta}:\left\langle \boldsymbol{\vartheta}, \boldsymbol{a}_i - \boldsymbol{a}_j \right\rangle < -\varepsilon \right \} \text{ for all } \; x \in \mathcal{X}(\boldsymbol{\mu}),
\end{equation}
and
\begin{equation}\label{eq: Alt_ml}
    \textup{Alt}_{m, \ell}(\boldsymbol{\mu}) = \textup{Alt}_{m, \ell}(\boldsymbol{\theta}) = \left \{ \boldsymbol{\vartheta}: \boldsymbol{\vartheta}^\top \boldsymbol{a}_\ell \geq \mu_\varepsilon^{m, \ell}{p} + \varepsilon \geq \boldsymbol{\vartheta}^\top \boldsymbol{a}_{\ell+1} \right \} \text{ for all } \; x \in \mathcal{X}(\boldsymbol{\mu}),
\end{equation}
where
\begin{equation}
    \mu_{\varepsilon}^{m, \ell}(\bm{p}) \coloneqq \frac{p_m \mu_m+\sum_{i=1}^{\ell} p_i\left(\mu_i-\varepsilon\right)}{p_m+\sum_{i=1}^{\ell} p_i}.
\end{equation}

For a given culprit $x=(i,j,m,\ell)$, the alternative set $\textup{Alt}_x(\boldsymbol{\mu}) = \textup{Alt}_{i, j}(\boldsymbol{\mu}) \cup \textup{Alt}_{m,\ell}(\boldsymbol{\mu})$ is composed of two distinct parts, \emph{i.e.}, $\textup{Alt}_{i, j}(\boldsymbol{\mu})$ and $\textup{Alt}_{m,\ell}(\boldsymbol{\mu})$.

For the first part, \emph{i.e.}, $\textup{Alt}_{i, j}(\boldsymbol{\mu})$, we have
\begin{equation}
    \textup{Alt}_{i, j}(\boldsymbol{\mu}) = \left\{\boldsymbol{\vartheta}_{i,j}(\varepsilon, \bm{p}, \alpha) \mid \boldsymbol{\vartheta}_{i,j}(\varepsilon, \bm{p}, \alpha) = \boldsymbol{\theta} - \frac{\boldsymbol{y}_{i, j}^\top \boldsymbol{\theta} + \varepsilon + \alpha}{\Vert \boldsymbol{y}_{i, j} \Vert_{\bm{V}_{\bm{p}}^{-1}}^2} \bm{V}_{\bm{p}}^{-1} \boldsymbol{y}_{i, j} \right\},
\label{def: proof - alternative set for All-varepsilon in linear bandit}
\end{equation}
where $\bm{V}_{\bm{p}} = \sum_{i=1}^{K}p_i \boldsymbol{a}_i\boldsymbol{a}_i^\top$ is related to the Fisher information matrix \citep{chaloner1995bayesian} and $\boldsymbol{y}_{i, j} = \boldsymbol{a}_i-\boldsymbol{a}_j$ and $\alpha > 0$.  Here $\alpha$ is introduced to construct the specific alternative set with an explicit expression of alternative parameter $\bm{\vartheta}$, By letting $\alpha \to 0$, we are realizing the innermost infimum in equation~\eqref{def: Cx function} and equation~\eqref{optimal allocation problem_convex_1}. Then we have
\begin{equation}
    \boldsymbol{y}_{i, j}^\top \boldsymbol{\vartheta}_{i, j}(\varepsilon, \bm{p}, \alpha) = -\varepsilon - \alpha < - \varepsilon,
\label{def: proof - y theta}
\end{equation}
which satisfies the condition of being the parameter in an alternative set as shown in equation~\eqref{eq: alternative set condition}. Under Gaussian distribution assumption, \emph{i.e.}, $\mu_i \sim \mathcal{N}(\boldsymbol{a}_i^\top \boldsymbol{\theta}, 1)$. Then the KL divergence between the mean value based on the true parameter $\boldsymbol{\theta}$ against the mean value based on the alternative parameter $\boldsymbol{\vartheta}_{i, j}$ is
\begin{align}
    \textup{KL}(\boldsymbol{a}_i^\top \boldsymbol{\theta}, \boldsymbol{a}_i^\top \boldsymbol{\vartheta}_{i, j}) &= \frac{\left ( \boldsymbol{a}_i^\top \left (\boldsymbol{\theta} - \boldsymbol{\vartheta}_{i, j}(\varepsilon, \bm{p}, \alpha) \right ) \right )^2}{2(1)^2} \nonumber\\
    &= \boldsymbol{y}_{i, j}^\top \bm{V}_{\bm{p}}^{-1} \frac{\left ( \boldsymbol{y}_{i, j}^\top \boldsymbol{\theta} + \varepsilon + \alpha \right )^2 \boldsymbol{a}_i \boldsymbol{a}_i^\top}{ 2 \left ( \Vert \boldsymbol{y}_{i, j} \Vert_{\bm{V}_{\bm{p}}^{-1}}^2 \right )^2} \bm{V}_{\bm{p}}^{-1} \boldsymbol{y}_{i, j}.
\label{def: proof - KL divergence for All-varepsilon and LB}
\end{align}

The last equation is obtained by taking the expression for $\boldsymbol{\vartheta}_{i, j}$ as in \eqref{def: proof - alternative set for All-varepsilon in linear bandit}. Then by Proposition \ref{general lower bound} and the definition of $C_x$ function in equation~\eqref{def: Cx function} the lower bound can be calculated as
\begin{align}
    \frac{\mathbb{E}_{\boldsymbol{\mu}} \left [ \tau_\delta \right ]}{\log\left({1}/{2.4 \delta}\right)} &\geq \min_{\bm{p} \in S_K}\max_{\boldsymbol{\vartheta} \in \textup{Alt}(\boldsymbol{\mu})} \frac{1}{\sum_{n = 1}^{K} p_n \textup{KL}(\boldsymbol{a}_n^\top \boldsymbol{\theta},   \boldsymbol{a}_n^\top \boldsymbol{\vartheta})}\nonumber\\
    &\geq \min_{\bm{p} \in S_K}\max_{x \in \mathcal{X}}\sup_{\alpha > 0} \frac{1}{\sum_{n = 1}^{K} p_n \textup{KL}(\boldsymbol{a}_n^\top \boldsymbol{\theta},   \boldsymbol{a}_n^\top \boldsymbol{\vartheta}_{i, j}(\varepsilon, \bm{p}, \alpha))}\nonumber\\
    &= \min_{\bm{p} \in S_K}\max_{x \in \mathcal{X}} \frac{2 \Vert \boldsymbol{y}_{i, j} \Vert_{\bm{V}_{\bm{p}}^{-1}}^2}{\left ( \boldsymbol{y}_{i, j}^\top \boldsymbol{\theta} + \varepsilon \right )^2}.
\end{align}

From this lower bound we can also define the $C_x$ function for all $\varepsilon$-best arms identification in linear bandits, which is
\begin{equation}\label{Cx function: 1}
    C_{i, j}(\bm{p}) = \frac{\left ( \boldsymbol{y}_{i, j}^\top \boldsymbol{\theta} + \varepsilon \right )^2}{2 \Vert \boldsymbol{y}_{i, j} \Vert_{\bm{V}_{\bm{p}}^{-1}}^2}.
\end{equation}

Notice that we let $\alpha \to 0$ establish the result by realizing $\sup_{\alpha > 0}$. Furthermore, the form of the alternative set in the first part comes from the solution of an optimization problem, given by
\begin{align}
    \mathop{\arg\min}\limits_{\boldsymbol{\vartheta} \in \mathbb{R}^d} \quad & \Vert \boldsymbol{\vartheta} - \boldsymbol{\theta} \Vert_{\bm{V}_{\bm{p}}}^2 \label{nature_vartheta_ob} \\
    \text{s.t.} \quad & \boldsymbol{y}_{i, j}^\top\boldsymbol{\vartheta} = -\varepsilon - \alpha. \label{nature_vartheta_const}
\end{align}



For the second part, \emph{i.e.}, $\textup{Alt}_{m, \ell}(\boldsymbol{\mu})$. If we use the alternative set introduced in equation~\eqref{eq: Alt_ml} directly, the result is complex and does not have an explicit form. Thus, we fine-tune the formula of the alternative set depicted in Figure~\ref{fig: 2}, simplifying its form while ensuring the completeness of the alternative set and the tightness of the lower bound. Compared to equation~\eqref{eq: Alt_ml}, the result in Theorem \ref{lower bound: All-epsilon in the linear setting} assumes that the mean value of arm 1 (\emph{i.e.}, the arm with the largest mean value), remains unchanged, partly sacrificing the completeness of the alternative set in equation (\ref{equation: lower bound 3}) regarding its second term but is presented in an explicit and symmetric form. This form is also tight.

Therefore, the culprit set is adjusted as $\mathcal{X}(\boldsymbol{\mu}) = \{ (i, j, m): i \in G_\varepsilon(\bm{\mu}), j \neq i, m \notin G_\varepsilon(\bm{\mu})  \}$ and the alternative set can be decomposed as $\textup{Alt}_x(\boldsymbol{\mu}) = \textup{Alt}_{i, j}(\boldsymbol{\mu}) \cup \textup{Alt}_{m}(\boldsymbol{\mu})$ for a given culprit $x = (i,j,m)$. Then different from equation~\eqref{eq: Alt_ml}, we have the second part of the alternative set as
\begin{equation}\label{eq: Alt_m_simple}
    \textup{Alt}_{m}(\boldsymbol{\mu}) = \textup{Alt}_{m}(\boldsymbol{\theta}) = \left \{ \boldsymbol{\vartheta}:\left\langle \boldsymbol{\vartheta}, \boldsymbol{a}_1 - \boldsymbol{a}_m \right\rangle < -\varepsilon \right \} \text{ for all } \; x \in \mathcal{X}(\boldsymbol{\mu}).
\end{equation}

$\textup{Alt}_{m}(\boldsymbol{\mu})$ can be similarly constructed as the set in equation~\eqref{def: proof - alternative set for All-varepsilon in linear bandit}, given by
\begin{equation}
    \textup{Alt}_{m}(\boldsymbol{\mu}) = \left\{\boldsymbol{\vartheta}_{m}(\varepsilon, \bm{p}, \alpha) \mid \boldsymbol{\vartheta}_{m}(\varepsilon, \bm{p}, \alpha) = \boldsymbol{\theta} - \frac{\boldsymbol{y}_{m}^\top \boldsymbol{\theta} + \varepsilon + \alpha}{\Vert \boldsymbol{y}_{m} \Vert_{\bm{V}_{\bm{p}}^{-1}}^2} \bm{V}_{\bm{p}}^{-1} \boldsymbol{y}_{m} \right\},
\label{def: proof - alternative set for All-varepsilon in linear bandit 2}
\end{equation}

\noindent where $\boldsymbol{y}_{m} = \boldsymbol{a}_1 - \boldsymbol{a}_m$ and $\alpha > 0$. Then we have
\begin{equation}
    \boldsymbol{y}_{m}^\top \boldsymbol{\vartheta}_{m}(\varepsilon, \bm{p}, \alpha) = \varepsilon - \alpha < \varepsilon.
\label{def: proof - y theta 2}
\end{equation}

Hence following the derivation for the first part, we have
\begin{equation}\label{Cx function: 2}
    C_{m}(\bm{p}) = \frac{\left ( \boldsymbol{y}_{m}^\top \boldsymbol{\theta} - \varepsilon \right )^2}{2 \Vert \boldsymbol{y}_{m} \Vert_{\bm{V}_{\bm{p}}^{-1}}^2}.
\end{equation}

Then similarly by Proposition \ref{general lower bound} and the definition of $C_x$ function, combining equation (\ref{Cx function: 1}) and equation (\ref{Cx function: 2}), the final lower bound can be calculated as
\begin{align}
    \frac{\mathbb{E}_{\boldsymbol{\mu}} \left [ \tau_\delta \right ]}{\log\left({1}/{2.4 \delta}\right)} &\geq \min_{\bm{p} \in S_K}\max_{(i, j, m) \in \mathcal{X}} \max \left\{ \frac{2 \Vert \boldsymbol{y}_{i, j} \Vert_{\bm{V}_{\bm{p}}^{-1}}^2}{\left ( \boldsymbol{y}_{i, j}^\top \boldsymbol{\theta} + \varepsilon \right )^2}, \frac{2 \Vert \boldsymbol{y}_{m} \Vert_{\bm{V}_{\bm{p}}^{-1}}^2}{\left ( \boldsymbol{y}_{m}^\top \boldsymbol{\theta} - \varepsilon \right )^2} \right\} \\
    &= \min_{\bm{p} \in S_K}\max_{(i, j, m) \in \mathcal{X}} \max \left\{ \frac{2 \Vert \boldsymbol{a}_i-\boldsymbol{a}_j \Vert_{\bm{V}_{\bm{p}}^{-1}}^2}{\left ( \boldsymbol{a}_i^\top \boldsymbol{\theta} - \boldsymbol{a}_j^\top \boldsymbol{\theta} + \varepsilon \right )^2}, \frac{2 \Vert \boldsymbol{a}_1 - \boldsymbol{a}_m \Vert_{\bm{V}_{\bm{p}}^{-1}}^2}{\left ( \boldsymbol{a}_1^\top \boldsymbol{\theta} - \boldsymbol{a}_m^\top \boldsymbol{\theta} - \varepsilon \right )^2} \right\}.
\end{align}
\hfill\Halmos
\endproof

\section{Proof of Theorem \ref{upper bound: Algorithm 1 G}}\label{proof of theorem: upper bound of LinFACT G}

The complete proof of Theorem \ref{upper bound: Algorithm 1 G} is divided into two parts. In the first part, we provide the proof for the first upper bound presented in equation (\ref{upper bound for algorithm1: 1}). In the second part, we show the second upper bound presented in equation (\ref{upper bound for algorithm1: 2}), where we remove the summation form and make it closer to the lower bound. 



For the first result with a summation as shown in equation (\ref{upper bound for algorithm1: 1}), we estimate $R_{\max} = \min \left\{r: G_r=G_\varepsilon\right\}$, the round in which we add the final arm from $G_\varepsilon$ to $G_r$. We divide the total number of samples drawn into two parts: those taken up to round $R_{\max}$ and those taken from $R_{\max} + 1$ until termination, if the algorithm does not stop in round $R_{\max}$. The proof is split into eight steps detailed as follows.

\subsection{Preliminary: Clean Event $\mathcal{E}_1$ and $\mathcal{E}_2$}\label{algorithm2: preliminary}


\proof\\{\textit{Proof.}}
We first define clean events $\mathcal{E}_1$ and $\mathcal{E}_1$, referred to as the events that are highly probable.
\begin{equation}
    \mathcal{E}_1 = \left\{ \bigcap_{i \in \mathcal{A}_I(r-1)} \bigcap_{r \in \mathbb{N}} \left| \hat{\mu}_i(r) - \mu_i \right| \leq C_{\delta/K}(r) \right\}. \label{clean event G}
\end{equation}

It represents the occurrence where the estimated means $\hat{\mu}_i(r)$ for each arm $i$ in the active set $\mathcal{A}_I(r-1)$ remain within a confidence bound $C_{\delta / K}(r)$ of their true means $\mu_i$ at each round $r$. It is designed to control the probability of error across all arms and rounds at a specified confidence level. 

Then, we introduce the following lemma to provide the confidence region for the estimated parameter $\hat{\boldsymbol{\theta}}$.
\begin{lemma}\label{lemma: confidence region for any single fixed arm}
Let confidence level $\delta \in (0,1)$, for each arm $\bm{a} \in \mathcal{A}$, we have
\begin{equation}
    \mathbb{P} \left \{ \left \lvert \left\langle \hat{\boldsymbol{\theta}} - \boldsymbol{\theta}, \bm{a} \right\rangle \right\rvert \geq \sqrt{2 \left \lVert \bm{a} \right\rVert^2_{\bm{V}_t^{-1}} \log \left( \frac{2}{\delta} \right)} \right \} \leq \delta,
\end{equation}
where $\bm{V}_t$, which is constructed with collected data, represents the information matrix defined in Section \ref{subsec_Optimal Design}.

\end{lemma}

For the G-optimal sampling rule, let $\pi_r$ represent the calculated optimal allocation proportion for each round $r$, the sampling budget is then designed as equation~\eqref{equation_phase budget G}. Therefore, we have information matrix in each round defined as\footnote{The symbol $\succeq$ represents the relative magnitude relationship between matrices. Specifically, for any matrix $\bm{A}$, $\bm{A} \succeq 0$ means that the matrix $\bm{A}$ is positive semi-definite. $\bm{A} \succeq \bm{B}$ equals $\bm{A}-\bm{B} \succeq 0$.} 
\begin{equation}\label{matrix_ineq}
    \bm{V}_r = \sum_{\bm{a} \in \text{Supp}(\pi_r)} T_r(\bm{a}) \bm{a} \bm{a}^\top \succeq \frac{2d}{\varepsilon_r^2} \log \left( \frac{2Kr(r+1)}{\delta} \right) \bm{V}(\pi),
\end{equation}
where $\bm{V}(\pi)$, which is defiend in Section \ref{subsubsec_G-Optimal Design}, represents the weighted information matrix. Then, applying Lemma \ref{lemma: confidence region for any single fixed arm}, where $\delta$ is substituted with $\delta/{Kr(r+1)}$. For any arm $\bm{a} \in \mathcal{A}(r-1)$, with a probability of at least $1-\delta/{Kr(r+1)}$, we have
\begin{align}\label{equation: confidence region}
    \left| \left\langle \hat{\boldsymbol{\theta}}_r - \boldsymbol{\theta}, \bm{a} \right\rangle \right| &\leq \sqrt{2\left \lVert \bm{a} \right\rVert^2_{\bm{V}_r^{-1}} \log \left( \frac{2Kr(r+1)}{\delta} \right)} \nonumber\\
    &= \sqrt{2\bm{a}^\top\bm{V}_r^{-1}\bm{a} \log \left( \frac{2Kr(r+1)}{\delta} \right)} \nonumber\\
    &\leq \sqrt{2\bm{a}^\top \left( \frac{\varepsilon_r^2}{2d}\frac{1}{\log \left( \frac{2Kr(r+1)}{\delta} \right)}\bm{V}(\pi)^{-1} \right) \bm{a} \log \left( \frac{2Kr(r+1)}{\delta} \right)} \nonumber\\
    &\leq \varepsilon_r,
\end{align}
where the first line comes from Lemma \ref{lemma: confidence region for any single fixed arm}, the third line follows the matrix inequality in \eqref{matrix_ineq} with the help of auxiliary lemma \ref{lemma: inversion reverses loewner orders}, and the last line comes from Lemma \ref{theorem_KW}.

Thus with the standard result of the G-optimal design in equation \eqref{equation: confidence region}, the confidence radius of clean event $\mathcal{E}_1$ can be defined as 
\begin{equation}\label{eq_C_G}
    C_{\delta/K}(r) \coloneqq \varepsilon_r.
\end{equation}

Then, we have
\begin{align}
    \mathbb{P}(\mathcal{E}_1^c) &= \mathbb{P} \left\{ \bigcup_{i \in \mathcal{A}_I(r-1)} \bigcup_{r \in \mathbb{N}} \left| \hat{\mu}_i(r) - \mu_i \right| > C_{\delta/K}(r) \right\}
    \nonumber\\ &\leq \sum_{r=1}^{\infty} \mathbb{P} \left\{ \bigcup_{i \in \mathcal{A}_I(r-1)} \left| \hat{\mu}_i(r) - \mu_i \right| > \varepsilon_r \right\} \nonumber\\ &\leq \sum_{r=1}^{\infty} \sum_{i=1}^{K} \frac{\delta}{Kr(r+1)}\nonumber\\
    &= \delta,
\end{align}
where the first line is the definition, the second line follows the union bounds over $r$ rounds, and the third line comes from the union bounds over all arms in the active set based on the result for any arm provided by equation \eqref{equation: confidence region}. Therefore, we have 
\begin{equation}
    P(\mathcal{E}_1) \geq 1 - \delta.
\end{equation}

Considering another event $\mathcal{E}_2$ describing the gaps between different arms, given by
\begin{equation}
    \mathcal{E}_2 = \left \{ \bigcap_{i \in G_{\varepsilon}} \bigcap_{j \in \mathcal{A}_I(r-1)} \bigcap_{r \in \mathbb{N}} \left| (\hat{\mu}_j(r) - \hat{\mu}_i(r)) - (\mu_j - \mu_i) \right| \leq 2\varepsilon_r \right \}.
\end{equation}

This event ensures the gap between the estimated mean rewards of arms $j$ and $i$ is uniformly close to their true gap for all rounds $r$, arms $i \in G_{\varepsilon}$, and arms $j$ in the active set $\mathcal{A}_I(r-1)$.

By equation (\ref{equation: confidence region}), for $i, j \in \mathcal{A}_I(r-1)$, we have
\begin{align}\label{confidence: arm filter}
    \mathbb{P}\left\{\left|(\hat{\mu}_j - \hat{\mu}_i) - (\mu_j - \mu_i)\right| > 2\varepsilon_r \mid \mathcal{E}_1 \right\} &\leq \mathbb{P}\left\{\left|\hat{\mu}_j - \mu_j \right| + \left|\hat{\mu}_i - \mu_i\right| > 2\varepsilon_r \mid \mathcal{E}_1 \right\} \nonumber\\
    &\leq \mathbb{P}\left\{\left|\hat{\mu}_j - \mu_j \right| > \varepsilon_r \mid \mathcal{E}_1 \right\} + \mathbb{P}\left\{\left|\hat{\mu}_i - \mu_i\right| > \varepsilon_r \mid \mathcal{E}_1 \right\} \nonumber\\
    &= 0,
\end{align}

\noindent which means
\begin{equation}
    \mathbb{P}(\mathcal{E}_2 \mid \mathcal{E}_1) = 1.
\end{equation}
\hfill\Halmos
\endproof

\subsection{Step 1: Correctness}\label{algorithm2: Step 1}
\ 
Recall that $G_\varepsilon$ is the true good set and $G_r$ is the empirical good set in round $r$, where the arms are classified by LinFACT-G as belonging to $G_\varepsilon$. On event $\mathcal{E}_1$, we first prove that if there exists a round $r$, at which $G_r \cup B_r = [K]$, then $G_r = G_\varepsilon$. As we can see, for this stopping condition of LinFACT-G, it will return the set $G_\varepsilon$ correctly on the clean event $\mathcal{E}_1$. 

\begin{lemma}\label{claim1_1}
    On event $\mathcal{E}_1$, for all round $r \in \mathbb{N}$, $G_r \subset G_\varepsilon$.
\end{lemma}

\proof\\{\textit{Proof.}}
Firstly we show $1 \in \mathcal{A}_I(r)$ for all $r \in \mathbb{N}$, that is, the best arm is never eliminated from the active set $\mathcal{A}(r-1)$ in any round $r$ on event $\mathcal{E}_1$. Note for any arm $i$
\begin{equation}
    \hat{\mu}_1(r) + \varepsilon_r \geq \mu_1 \geq \mu_i \geq \hat{\mu}_i(r) - \varepsilon_r > \hat{\mu}_i(r) - \varepsilon_r - \varepsilon,
\end{equation}
which shows that $\hat{\mu}_1(r) + \varepsilon_r > \max_{i \in \mathcal{A}_I(r-1)} \hat{\mu}_i - \varepsilon_r - \varepsilon = L_r$ and $\hat{\mu}_1(r) + \varepsilon_r \geq \max_{i \in \mathcal{A}_I(r-1)} \hat{\mu}_i (r) - \varepsilon_r$ showing that $1$ will never exit the active set $\mathcal{A}_I(r-1)$.

Secondly, we show that at all rounds $r$, $\mu_1 - \varepsilon \in [L_r, U_r]$. Since arm 1 never exists $\mathcal{A}_I(r-1)$,
\begin{equation}
    U_r = \max_{i \in \mathcal{A}_I(r-1)} \hat{\mu}_i + \varepsilon_r -\varepsilon \geq \hat{\mu}_1(r) + \varepsilon_r - \varepsilon \geq \mu_1 - \varepsilon,
\end{equation}

\noindent and for any arm $i$,
\begin{equation}
    \mu_1 - \varepsilon \geq \mu_i - \varepsilon \geq \hat{\mu}_i - \varepsilon_r - \varepsilon.
\end{equation}

Hence,
\begin{equation}
    \mu_1 - \varepsilon \geq \max_{i \in \mathcal{A}_I(r-1)}\hat{\mu}_i - \varepsilon_r - \varepsilon = L_r.
\end{equation}

Next, we show that $G_r \subset G_{\varepsilon}$ for all $r \geq 1$. By contradiction, if $G_r \not\subset G_{\varepsilon}$, then it means that $\exists r \in \mathbb{N}$ and $\exists i \in G_\varepsilon^c \cap G_r$ such that,
\begin{equation}
    \mu_i \geq \hat{\mu}_i - \varepsilon_r \geq U_r \geq \mu_1 - \varepsilon > \mu_i,
\end{equation}
which forms a contradiction.
\hfill\Halmos
\endproof

\begin{lemma}\label{claim1_2}
    On event $\mathcal{E}_1$, for all $r \in \mathbb{N}$, $B_r \subset G_\varepsilon^c$.
\end{lemma}

\proof\\{\textit{Proof.}}
Similarly, we construct a contradiction. Consider the case that a good arm from $G_{\varepsilon}$ is added to $B_r$ for some round $r$. By definition, $B_0 = \emptyset$ and $B_{r-1} \subset B_r$ for all $r$. Then there must exist some $r \in \mathbb{N}$ and an $i \in G_{\varepsilon}$ such that $i \in B_r$ and $i \notin B_{r-1}$. Following line \ref{elimination: good 1} of Algorithm \ref{alg_LinFACT_stopping}, this occurs if and only if
\begin{equation}
    \max_{j \in \mathcal{A}_I(r-1)} \hat{\mu}_j - \hat{\mu}_i > 2\varepsilon_r + \varepsilon.
\end{equation}

On the clean event $\mathcal{E}_1$, the above implies $\exists j \in \mathcal{A}_I(r-1)$ such that
\begin{equation}
    \mu_j - \mu_i + 2\varepsilon_r \geq \hat{\mu}_j - \hat{\mu}_i \geq 2\varepsilon_r + \varepsilon,
\end{equation}

\noindent which shows that $\mu_j - \mu_i \geq \varepsilon$, contradicting the assertion that $i \in G_{\varepsilon}$.
\hfill\Halmos
\endproof

The above Lemma \ref{claim1_1} and Lemma \ref{claim1_2} show that under $\mathcal{E}_1$, $G_r \cup B_r = [K]$ can lead to the result $G_r = G_\varepsilon$ and $B_r = G_\varepsilon^c$. Since $\mathbb{P}\left\{ \mathcal{E}_1 \right\} \geq 1 - \delta$, if LinFACT terminates, it can correctly provide the correct decision rule with a probability of least $1 - \delta$, being a $\delta$-PAC algorithm.

Up to now, we have proved the correctness of the stopping rule of the algorithm. Then we will focus on bounding the sample complexity in the following parts.

\subsection{Step 2: Expression for total number of samples drown}\label{algorithm2: Step 2}


For \( i \in G_{\varepsilon} \), let \(  R_i \) denote the random variable of the number of rounds that arm \( i \) is sampled before it is added to \( G_r \) in line \ref{add to Gr} of Algorithm \ref{alg_LinFACT_stopping}. For \( i \in G_{\varepsilon}^c \), let \(  R_i \) denote the random variable of the number of rounds that arm \( i \) is sampled before it is eliminated from \( \mathcal{A}_I(r-1) \) and added to $B_r$ in line \ref{elimination: good 1} of Algorithm \ref{alg_LinFACT_stopping}. Then by definition, $R_i$ can be defined as
\begin{equation}
 R_i = \min \left\{ r:
\begin{cases}
i \in G_r & \text{if } i \in G_{\varepsilon} \\
i \notin \mathcal{A}_I(r) & \text{if } i \in G_{\varepsilon}^c
\end{cases}
\right\} = \min \left\{ r :
\begin{cases}
\hat{\mu}_i - \varepsilon_r \geq U_{r} & \text{if } i \in G_{\varepsilon} \\
\hat{\mu}_i + \varepsilon_r \leq L_{r} & \text{if } i \in G_{\varepsilon}^c
\end{cases}
\right\}.
\end{equation}



The total number of samples drawn by the algorithm can be represented as
\begin{align}
T &\leq \sum_{r=1}^{\infty} \mathbbm{1} \left[G_{r-1} \cup B_{r-1} \neq [K] \right] \sum_{\bm{a} \in \mathcal{A}(r-1)}T_r(\bm{a}) \nonumber\\
&= \sum_{r=1}^{\infty} \mathbbm{1} \left[ G_{r-1} \neq G_\varepsilon \right] \mathbbm{1} \left[G_{r-1} \cup B_{r-1} \neq [K] \right] \sum_{\bm{a} \in \mathcal{A}(r-1)}T_r(\bm{a})\label{decomposition of T: part 1} \\
&+ \sum_{r=1}^{\infty} \mathbbm{1} \left[ G_{r-1} = G_\varepsilon \right] \mathbbm{1} \left[G_{r-1} \cup B_{r-1} \neq [K] \right] \sum_{\bm{a} \in \mathcal{A}(r-1)}T_r(\bm{a}).\label{decomposition of T: part 2}
\end{align}

\subsection{Step 3: Bound $ R_i$ for $i \in G_\varepsilon$}\label{algorithm2: Step 3}

A helper function $h(x)$ is defined as $h(x) = \log_2 \left({1}/{\left \lvert x \right\rvert}\right)$ to assist the proof. We can observe that in round $r$, if $r \geq h(x)$, then $\varepsilon_r = 2^{-r} \leq \left \lvert x \right \rvert$.

\begin{lemma}\label{claim3_1}
    For $i \in G_\varepsilon$, we have that $ R_i \leq \lceil h\left(0.25\left(\varepsilon-\Delta_i\right) \right)\rceil$.
\end{lemma}

\proof\\{\textit{Proof.}}
Note that for $i \in G_\varepsilon$, $4  \varepsilon_r < \mu_i-\left(\mu_1-\varepsilon\right)$ is true when $r > h\left(0.25\left(\varepsilon-\Delta_i\right) \right)$, implies that for all $j \in \mathcal{A}_I(r-1)$,
\begin{align}
\hat{\mu}_i- \varepsilon_r &\geq \mu_i-2  \varepsilon_r \nonumber\\
& > \mu_1+2  \varepsilon_r-\varepsilon \nonumber\\
& \geq \mu_j+2  \varepsilon_r-\varepsilon \nonumber\\
& \geq \hat{\mu}_j+ \varepsilon_r-\varepsilon.
\end{align}

Thus, in particular, $\hat{\mu}_i- \varepsilon_r > \max _{j \in \mathcal{A}_I(r-1)} \hat{\mu}_j(t)+ \varepsilon_r-\varepsilon=U_r$. Then we have that $ R_i \leq \lceil h\left(0.25\left(\varepsilon-\Delta_i\right) \right)\rceil$.
\hfill\Halmos
\endproof

After finding the largest round for an arm $i \in G_\varepsilon$ to be added into $G_r$, we then define a round $R_{\max }$ when all good arms have been added into $G_r$.

\begin{lemma}\label{claim3_2}
    Define $R_{\max } \coloneqq \min \left\{r: G_r=G_\varepsilon\right\}=\max _{i \in G_\varepsilon}  R_i$, we have that $R_{\max } \leq$ $\lceil h\left(0.25 \alpha_\varepsilon\right)\rceil$.
\end{lemma}

\proof\\{\textit{Proof.}}
Recall that $\alpha_\varepsilon=\min _{i \in G_\varepsilon} \mu_i-\mu_1+\varepsilon=\min _{i \in G_\varepsilon} \varepsilon-\Delta_i$. By Lemma \ref{claim3_1} in this section, $ R_i \leq \lceil h\left(0.25\left(\varepsilon-\Delta_i\right) \right) \rceil$ for $i \in G_\varepsilon$. Furthermore, $h(\cdot)$ is monotonically decreasing in its argument if $i \in G_\varepsilon$. Then for any $\delta>0$, $R_{\max }=\max _{i \in G_\varepsilon}  R_i \leq \max _{i \in G_\varepsilon} \lceil h\left(0.25\left(\varepsilon-\Delta_i\right) \right) \rceil = \lceil h(\min _{i \in G_\varepsilon} \varepsilon-\Delta_i) \rceil = \lceil h\left(0.25 \alpha_\varepsilon \right) \rceil$.
\hfill\Halmos
\endproof

\subsection{Step 4: Bound the total samples at round $r = R_{\max }$}\label{algorithm2: Step 4}

The total number of samples up to round $R_{\max }$ is $\sum_{r=1}^{R_{\max}} \sum_{\bm{a} \in \mathcal{A}(R_{\max}-1)}T_r(\bm{a})$.
By line \ref{G_sampling_2} of the Algorithm \ref{alg_LinFACT_sampling}, we have
\begin{equation}\label{T_r budget}
    T_{r} \leq \frac{2 d}{\varepsilon_{r}^2} \log \left(\frac{2K r(r+1)}{\delta}\right)+\frac{d(d+1)}{2}.
\end{equation}

Hence,
\begin{align}
    \sum_{r=1}^{R_{\max }} T_r &= \sum_{r=1}^{R_{\max }} \sum_{\bm{a} \in \mathcal{A}(r-1)}T_r(\bm{a}) \nonumber\\
    & \leq \sum_{r=1}^{R_{\max }} \left( \frac{2 d}{\varepsilon_{r}^2} \log \left(\frac{2K r(r+1)}{\delta}\right)+\frac{d(d+1)}{2} \right) \nonumber\\
    & \leq \sum_{r=1}^{R_{\max }} 2^{2r+1} d \log \left(\frac{2K R_{\max }(R_{\max }+1)}{\delta}\right)+\frac{d(d+1)}{2}R_{\max } \nonumber\\
    & \leq c 2^{2R_{\max }+1} d \log \left(\frac{2K R_{\max }(R_{\max }+1)}{\delta}\right)+\frac{d(d+1)}{2}R_{\max },\label{bound: Step 4}
\end{align}

\noindent where $c$ is a universal constant and recall that $R_{\max } \leq$ $\lceil h\left(0.25 \alpha_\varepsilon\right) \rceil$. The second line comes from equation~\eqref{T_r budget}. The third line comes from the scaling based on the range of the summation. The last line scales the exponential term $2r+1$ to the largest rounded $R_\textup{max}$ and introduces a finite constant $c$.

Next, we bound two terms in equation~\eqref{decomposition of T: part 1} and equation~\eqref{decomposition of T: part 2} separately. The first term, which represents the samples taken before round $R_{\max}$, can be bounded in the following step.

\subsection{Step 5: Bound equation (\ref{decomposition of T: part 1})}\label{algorithm2: Step 5}

Recall that $R_{\max } \leq$ $\lceil h\left(0.25 \alpha_\varepsilon\right) \rceil$ is the round where $G_{R_{\max }} = G_\varepsilon$. Using the result from previous Step 4, we may bound the total number of samples taken through this round, controlling equation (\ref{decomposition of T: part 1}).

\begin{lemma}\label{claim5_1}
    \begin{align}
    &\sum_{r=1}^{\infty} \mathbbm{1} \left[ G_{r-1} \neq G_\varepsilon \right] \mathbbm{1} \left[G_{r-1} \cup B_{r-1} \neq [K] \right] \sum_{\bm{a} \in \mathcal{A}(r-1)}T_r(\bm{a}) \nonumber\\
    &\leq c 2^{2R_{\max }+1} d \log \left(\frac{2K R_{\max }(R_{\max }+1)}{\delta}\right)+\frac{d(d+1)}{2}R_{\max },\label{upper bound: Step 5}
\end{align}
where $R_{\max } \leq$ $\lceil h\left(0.25 \alpha_\varepsilon\right) \rceil$.

\end{lemma}

\proof\\{\textit{Proof.}}
By definition of $R_{\max }$ and equation (\ref{bound: Step 4}) in Step 4,
\begin{align}
    &\sum_{r=1}^{\infty} \mathbbm{1} \left[ G_{r-1} \neq G_\varepsilon \right] \mathbbm{1} \left[G_{r-1} \cup B_{r-1} \neq [K] \right] \sum_{\bm{a} \in \mathcal{A}(r-1)}T_r(\bm{a}) \nonumber\\
    &= \sum_{r=1}^{R_{\max }} \mathbbm{1} \left[G_{r-1} \cup B_{r-1} \neq [K] \right] \sum_{\bm{a} \in \mathcal{A}(r-1)}T_r(\bm{a}) \nonumber\\
    &\leq c 2^{2R_{\max }+1} d \log \left(\frac{2K R_{\max }(R_{\max }+1)}{\delta}\right)+\frac{d(d+1)}{2}R_{\max },
\end{align}
where the second line comes from the definition of $R_\textup{max}$ since there will be no more samples collected after round $R_\textup{max}$. The third line directly comes from equation (\ref{bound: Step 4}).

\hfill\Halmos
\endproof

The first part of the total sample complexity is bounded. The second part collected after round $R_\textup{max}$, as represented by equation~\eqref{decomposition of T: part 2}, is bounded by the following Step 6.

\subsection{Step 6: Bound equation (\ref{decomposition of T: part 2})}\label{algorithm2: Step 6}

\begin{align}
    &\sum_{r=1}^{\infty} \mathbbm{1} \left[ G_{r-1} = G_\varepsilon \right] \mathbbm{1} \left[G_{r-1} \cup B_{r-1} \neq [K] \right] \sum_{\bm{a} \in \mathcal{A}(r-1)}T_r(\bm{a}) \nonumber\\
    &= \sum_{r=R_{\max } + 1}^{\infty} \mathbbm{1} \left[ B_{r-1} \neq G_\varepsilon^c \right] \sum_{\bm{a} \in \mathcal{A}(r-1)}T_r(\bm{a}) \nonumber\\
    &\leq \sum_{r=R_{\max } + 1}^{\infty} \left|G_\varepsilon^c \backslash B_{r-1}\right| \sum_{\bm{a} \in \mathcal{A}(r-1)}T_r(\bm{a}) \nonumber\\
    &= \sum_{r=R_{\max } + 1} ^{\infty} \sum_{i \in G_\varepsilon^c} \mathbbm{1}\left[i \notin B_{r-1}\right] \sum_{\bm{a} \in \mathcal{A}(r-1)}T_r(\bm{a}) \nonumber\\
    &= \sum_{i \in G_\varepsilon^c} \sum_{r=R_{\max } + 1} ^{\infty} \mathbbm{1}\left[i \notin B_{r-1}\right] \sum_{\bm{a} \in \mathcal{A}(r-1)}T_r(\bm{a}) \nonumber\\
    &\leq \sum_{i \in G_\varepsilon^c} \sum_{r=1} ^{\infty} \mathbbm{1}\left[i \notin B_{r-1}\right] \sum_{\bm{a} \in \mathcal{A}(r-1)}T_r(\bm{a}) \nonumber\\
    &\leq \sum_{i \in G_\varepsilon^c} \sum_{r=1} ^{\infty} \mathbbm{1}\left[i \notin B_{r-1}\right] \left(\frac{2 d}{\varepsilon_{r}^2} \log \left(\frac{2K r(r+1)}{\delta}\right)+\frac{d(d+1)}{2}\right),
\end{align}
where the second line comes from the definition of $R_\textup{max}$ since $G_r = G_\varepsilon$ as round goes over $R_\textup{max}$. The third line comes from the fact that as long as $G_\varepsilon^c \backslash B_{r-1}$ is not the empty set, the corresponding indicator function is 1. The fourth line considers the indicator function for each arm in $G_\varepsilon$. The fifth and sixth lines consider the exchange of the double summation and enlarges the summation range of the round $r$. The last line comes from the inequality in Step 5.


\subsection{Step 7: Bound the expected total number of samples drawn by LinFACT}\label{algorithm2: Step 7}

Now we take expectations over the number of samples drawn regarding our actual bandit instance $\bm{\mu}$. These expectations are conditioned on the high probability event $\mathcal{E}_1$.
\begin{align}
    \mathbb{E}_{\bm{\mu}}\left[T_G \mid \mathcal{E}_1\right]
    &\leq \sum_{r=1}^{\infty} \mathbb{E}_{\bm{\mu}}\left[\mathbbm{1}\left[G_r \cup B_r \neq[K]\right] \mid \mathcal{E}_1\right] \sum_{\bm{a} \in \mathcal{A}(r-1)}T_r(\bm{a}) \nonumber\\
    &= \sum_{r=1}^{\infty} \mathbb{E}_{\bm{\mu}}\left[\mathbbm{1}\left[G_{r-1} \neq G_\varepsilon\right] \mathbbm{1}\left[G_{r-1} \cup B_{r-1} \neq[K]\right] \mid \mathcal{E}_1\right] \sum_{\bm{a} \in \mathcal{A}(r-1)}T_r(\bm{a}) \nonumber\\
    &+ \sum_{r=1}^{\infty} \mathbb{E}_{\bm{\mu}}\left[\mathbbm{1}\left[G_{r-1}=G_\varepsilon\right] \mathbbm{1}\left[G_{r-1} \cup B_{r-1} \neq[K]\right] \mid \mathcal{E}_1\right] \sum_{\bm{a} \in \mathcal{A}(r-1)}T_r(\bm{a}) \nonumber\\
    &\stackrel{\text {Step 5}}{\leq} c 2^{2R_{\max }+1} d \log \left(\frac{2K R_{\max }(R_{\max }+1)}{\delta}\right)+\frac{d(d+1)}{2}R_{\max } \nonumber\\
    &+ \sum_{r=1}^{\infty} \mathbb{E}_{\bm{\mu}}\left[\mathbbm{1}\left[G_{r-1}=G_\varepsilon\right] \mathbbm{1}\left[G_{r-1} \cup B_{r-1} \neq[K]\right] \mid \mathcal{E}_1\right] \sum_{\bm{a} \in \mathcal{A}(r-1)}T_r(\bm{a}) \nonumber\\
    &\stackrel{\text {Step 6}}{\leq} c 2^{2R_{\max }+1} d \log \left(\frac{2K R_{\max }(R_{\max }+1)}{\delta}\right)+\frac{d(d+1)}{2} R_{\max } \nonumber\\
    &+ \sum_{i \in G_\varepsilon^c} \sum_{r=1} ^{\infty} 
    \mathbb{E}_{\bm{\mu}}\left[ \mathbbm{1}\left[i \notin B_{r-1} \mid \mathcal{E}_1\right] \right] \left(\frac{2 d}{\varepsilon_{r}^2} \log \left(\frac{2K r(r+1)}{\delta}\right)+\frac{d(d+1)}{2}\right),\label{equation: decomposition of T}
\end{align}
where the first line comes from the stopping condition of LinFACT-G and the additivity of the expectation. The second line is the decomposition shown in Step 2. Subsequent lines follow the results in previous Step 5 and Step 6.


The result in equation~\eqref{equation: decomposition of T} still contains an implicit term within the summation across all rounds and the following Step 8 focus on bounding this term.

\subsection{Step 8: For $i \in G_\varepsilon^c$, bound $\sum_{r=1}^{\infty} \mathbb{E}_{{\mu}} \left[\mathbbm{1}\left[i \notin B_{r-1}\right] \mid \mathcal{E}_1\right]\left(\frac{2 d}{\varepsilon_{r}^2} \log \left(\frac{2K r(r+1)}{\delta}\right)+\frac{d(d+1)}{2}\right)$}\label{algorithm2: Step 8}

Next, we bound the expectation remaining in Step 7. First, for a given $i \in G_\varepsilon^c$ and a given $r$, we bound the probability that $i \notin B_r$. Note that by Borel-Cantelli, this implies that the probability that $i$ is never added to any $B_r$ is 0.

\begin{lemma}\label{claim8_1}
    For $i \in G_\varepsilon^c$ and $r \geq\left\lceil\log _2\left(\frac{4}{\Delta_i-\varepsilon}\right)\right\rceil$, we have $\mathbb{E}_{\boldsymbol{\mu}}\left[\mathbbm{1}\left[i \notin B_r\right] \mid \mathcal{E}_1\right] = 0$.
\end{lemma}


\proof\\{\textit{Proof.}}
First, we have that for any $i \in G_\varepsilon^c$
\begin{align}\label{eq_Ev_claim0}
    \mathbb{E}_{\bm{\mu}}\left[\mathbbm{1}\left[i \notin B_r\right] \mid \mathcal{E}_1\right] &= \mathbb{E}_{\boldsymbol{\mu}}\left[ \mathbbm{1}\left[\max_{j \in \mathcal{A}_I(r-1)} \hat{\mu}_j - \hat{\mu}_i \leq 2\varepsilon_r + \varepsilon\right] \mid \mathcal{E}_1, i \notin B_m (m = \{1, 2, \ldots, r-1\}) \right] \nonumber \\
    &\leq \mathbb{E}_{\boldsymbol{\mu}}\left[ \mathbbm{1}\left[\max_{j \in \mathcal{A}_I(r-1)} \hat{\mu}_j - \hat{\mu}_i \leq 2\varepsilon_r + \varepsilon\right] \mid \mathcal{E}_1\right],
\end{align}
where the first line comes from the fact that arm $i \in G_\varepsilon^c$ will never be added into $B_r$ from round 1 to round $r-1$ if $i \notin B_r$, and the second line considers the conditional expectation.




If $i \in B_{r-1}$, then $i \in B_r$ by definition. Otherwise, if $i \notin B_{r-1}$, then under event $\mathcal{E}_1$, for $i \in G_\varepsilon^c$ and $r \geq\left\lceil\log _2\left(\frac{4}{\Delta_i-\varepsilon}\right)\right\rceil$, we have
\begin{equation}
    \max_{j \in \mathcal{A}_I(r-1)} \hat{\mu}_j - \hat{\mu}_i \geq \Delta_i-2^{-r+1} \geq \varepsilon+2\varepsilon_r,
\end{equation}
which implies that $i \in B_r$ by line \ref{elimination: good 1} of the Algorithm \ref{alg_LinFACT_stopping}. In other words, we have proved the correctness of the algorithm if line \ref{elimination: good 1} happens in Step 1, and here we introduce when exactly line \ref{elimination: good 1} will happen. In particular, under event $\mathcal{E}_1$, if $i \notin B_{r-1}$, for all $i \in G_\varepsilon^c$ and $r \geq\left\lceil\log _2\left(\frac{4}{\Delta_i-\varepsilon}\right)\right\rceil$, we have
\begin{equation}\label{expectation about contradictory}
    \mathbb{E}_{\boldsymbol{\mu}}\left[ \mathbbm{1}\left[\max_{j \in \mathcal{A}_I(r-1)} \hat{\mu}_j - \hat{\mu}_i > 2\varepsilon_r + \varepsilon\right] \mid i \notin B_{r-1}, \mathcal{E}_1 \right] = 1.
\end{equation}

Therefore, for all $i \in G_\varepsilon^c$ and $r \geq\left\lceil\log _2\left(\frac{4}{\Delta_i-\varepsilon}\right)\right\rceil$, we have
\begin{align}
    &\mathbb{E}_{\boldsymbol{\mu}}\left[ \mathbbm{1}\left[\max_{j \in \mathcal{A}_I(r-1)} \hat{\mu}_j - \hat{\mu}_i \leq 2\varepsilon_r + \varepsilon\right] \mid \mathcal{E}_1 \right] \nonumber\\
    &= \mathbb{E}_{\boldsymbol{\mu}}\left[ \mathbbm{1}\left[\max_{j \in \mathcal{A}_I(r-1)} \hat{\mu}_j - \hat{\mu}_i \leq 2\varepsilon_r + \varepsilon\right] \mathbbm{1}\left[i \notin B_{r-1}\right] \mid \mathcal{E}_1 \right] \nonumber\\
    &+ \mathbb{E}_{\boldsymbol{\mu}}\left[ \mathbbm{1}\left[\max_{j \in \mathcal{A}_I(r-1)} \hat{\mu}_j - \hat{\mu}_i \leq 2\varepsilon_r + \varepsilon\right] \mathbbm{1}\left[i \in B_{r-1}\right] \mid \mathcal{E}_1 \right]\nonumber\\
    &= \mathbb{E}_{\boldsymbol{\mu}}\left[ \mathbbm{1}\left[\max_{j \in \mathcal{A}_I(r-1)} \hat{\mu}_j - \hat{\mu}_i \leq 2\varepsilon_r + \varepsilon\right] \mathbbm{1}\left[i \notin B_{r-1}\right] \mid \mathcal{E}_1 \right] \nonumber\\
    &=\mathbb{E}_{\boldsymbol{\mu}}\left[\mathbbm{1}\left[\max_{j \in \mathcal{A}_I(r-1)} \hat{\mu}_j - \hat{\mu}_i \leq 2\varepsilon_r + \varepsilon\right] \mathbbm{1}\left[i \notin B_{r-1}\right] \mid i \notin B_{r-1}, \mathcal{E}_1\right] \mathbb{P}\left(i \notin B_{r-1} \mid \mathcal{E}_1\right) \nonumber\\
    &+ \mathbb{E}_{\boldsymbol{\mu}}\left[\mathbbm{1}\left[\max_{j \in \mathcal{A}_I(r-1)} \hat{\mu}_j - \hat{\mu}_i \leq 2\varepsilon_r + \varepsilon\right] \mathbbm{1}\left[i \notin B_{r-1}\right] \mid i \in B_{r-1}, \mathcal{E}_1\right] \mathbb{P}\left(i \in B_{r-1} \mid \mathcal{E}_1\right) \nonumber\\
    &= \mathbb{E}_{\boldsymbol{\mu}}\left[\mathbbm{1}\left[\max_{j \in \mathcal{A}_I(r-1)} \hat{\mu}_j - \hat{\mu}_i \leq 2\varepsilon_r + \varepsilon\right] \mathbbm{1}\left[i \notin B_{r-1}\right] \mid i \notin B_{r-1}, \mathcal{E}_1\right] \mathbb{P}\left(i \notin B_{r-1} \mid \mathcal{E}_1\right) \nonumber\\
    &= \mathbb{E}_{\boldsymbol{\mu}}\left[\mathbbm{1}\left[\max_{j \in \mathcal{A}_I(r-1)} \hat{\mu}_j - \hat{\mu}_i \leq 2\varepsilon_r + \varepsilon\right] \mid i \notin B_{r-1}, \mathcal{E}_1\right] \mathbb{E}_{\bm{\mu}}\left[\mathbbm{1}\left[i \notin B_{r-1}\right] \mid \mathcal{E}_1\right] \nonumber\\
    &= 0,
\end{align}
where the second line comes from the additivity of expectation. The fourth line follows the deterministic result that $\mathbbm{1}\left[i \notin B_r\right] \mathbbm{1}\left[i \in B_{r-1}\right]=0$. The fifth line is the decomposition based on the conditional expectation. The eighth line comes from the fact that the expectation of the indicator function is simply the probability. The last line follows the result in equation~\eqref{expectation about contradictory}.


The lemma can thus be concluded together with equation \eqref{eq_Ev_claim0}. 
\hfill\Halmos
\endproof

\begin{lemma}\label{claim8_2}
    For $i \in G_\varepsilon^c$, based on the result in previous Lemma \ref{claim8_1}, we have
\begin{align}
    &\sum_{r=1} ^{\infty} \mathbb{E}_{\boldsymbol{\mu}}\left[\mathbbm{1}\left[i \notin B_{r-1}\right] \mid \mathcal{E}_1\right] \left(\frac{2 d}{\varepsilon_{r}^2} \log \left(\frac{2K r(r+1)}{\delta}\right)+\frac{d(d+1)}{2}\right) \nonumber\\ 
    &\leq \frac{d(d+1)}{2} \log _2\left(\frac{8}{\Delta_i-\varepsilon}\right) + \xi \frac{256d}{(\Delta_i - \varepsilon)^2} \log \left(\frac{2K}{\delta} \log_2 \frac{16}{\Delta_i - \varepsilon} \right).
\end{align}
\end{lemma}
    
\proof\\{\textit{Proof.}}
\begin{align}
& \sum_{r=1}^{\infty} \mathbb{E}_{\boldsymbol{\mu}}\left[\mathbbm{1}\left[i \notin B_{r-1}\right] \mid \mathcal{E}_1\right]\left(\frac{2 d}{\varepsilon_{r}^2} \log \left(\frac{2K r(r+1)}{\delta}\right)+\frac{d(d+1)}{2}\right) \nonumber\\
&= \sum_{r=1}^{\left\lceil\log _2\left(\frac{4}{\Delta_i-\varepsilon}\right)\right\rceil} \mathbb{E}_{\boldsymbol{\mu}}\left[\mathbbm{1}\left[i \notin B_{r-1}\right] \mid \mathcal{E}_1\right]\left(\frac{2 d}{\varepsilon_{r}^2} \log \left(\frac{2K r(r+1)}{\delta}\right)+\frac{d(d+1)}{2}\right) \nonumber\\
&+ \sum_{r=\left\lceil\log _2\left(\frac{4}{\Delta_i-\varepsilon}\right)\right\rceil+1}^{\infty} \mathbb{E}_{\boldsymbol{\mu}}\left[\mathbbm{1}\left[i \notin B_{r-1}\right] \mid \mathcal{E}_1\right]\left(\frac{2 d}{\varepsilon_{r}^2} \log \left(\frac{2K r(r+1)}{\delta}\right)+\frac{d(d+1)}{2}\right) \nonumber\\
&= \sum_{r=1}^{\left\lceil\log _2\left(\frac{4}{\Delta_i-\varepsilon}\right)\right\rceil} \mathbb{E}_{\boldsymbol{\mu}}\left[\mathbbm{1}\left[i \notin B_{r-1}\right] \mid \mathcal{E}_1\right]\left(\frac{2 d}{\varepsilon_{r}^2} \log \left(\frac{2K r(r+1)}{\delta}\right)+\frac{d(d+1)}{2}\right) + 0 \nonumber\\
&\leq \sum_{r=1}^{\left\lceil\log _2\left(\frac{4}{\Delta_i-\varepsilon}\right)\right\rceil} \left(d2^{2r+1} \log \left(\frac{2K r(r+1)}{\delta}\right)+\frac{d(d+1)}{2}\right) \nonumber\\
&\leq \frac{d(d+1)}{2} \log _2\left(\frac{8}{\Delta_i-\varepsilon}\right) + 2d\log \left(\frac{2K}{\delta}\right) \sum_{r=1}^{{\left\lceil\log _2\left(\frac{4}{\Delta_i-\varepsilon}\right)\right\rceil}} 2^{2r} + 4d \sum_{r=1}^{{\left\lceil\log _2\left(\frac{4}{\Delta_i-\varepsilon}\right)\right\rceil}} 2^{2r} \log (r+1) \nonumber\\
&\leq \frac{d(d+1)}{2} \log _2\left(\frac{8}{\Delta_i-\varepsilon}\right) + \xi \frac{256d}{(\Delta_i - \varepsilon)^2} \log \left(\frac{2K}{\delta} \log_2 \frac{16}{\Delta_i - \varepsilon} \right),
\end{align}
where $\xi$ is a sufficiently large universal constant. The second line comes from the decomposition of the summation across all rounds. The fourth line comes from the result in Lemma \ref{claim8_1} of Step 8. The fifth line enlarges the expectation of the indicator function to a maximum value of 1. The sixth and seventh lines enlarge round $r$ to its maximum value of $\left\lceil\log _2\left(\frac{4}{\Delta_i-\varepsilon}\right)\right\rceil$. 
\hfill\Halmos
\endproof

\subsection{Step 9: Apply the result of Step 8 to the result of Step 7}\label{algorithm2: Step 9}
\ 
Summarizing the aforementioned results, we have 
\begin{align}
    \mathbb{E}_{\boldsymbol{\mu}}\left[T_G \mid \mathcal{E}_1\right]
    &\leq c 2^{2R_{\max }+1} d \log \left(\frac{2K R_{\max }(R_{\max }+1)}{\delta}\right)+\frac{d(d+1)}{2}R_{\max } \nonumber\\
    &+ \sum_{i \in G_\varepsilon^c} \sum_{r=1} ^{\infty} 
    \mathbb{E}_{\boldsymbol{\mu}}\left[ \mathbbm{1}\left[i \notin B_{r-1} 
    \mid \mathcal{E}_1 \right] \right] \left(\frac{2 d}{\varepsilon_{r}^2} \log \left(\frac{2K r(r+1)}{\delta}\right)+\frac{d(d+1)}{2}\right)
\end{align}
from Step 7, where we have
\begin{equation}
    R_{\max } \leq \lceil h\left(0.25 \alpha_\varepsilon\right) \rceil = \log_2 \frac{8}{\alpha_\varepsilon}.
\end{equation}

From Step 8, we have
\begin{align}
& \sum_{r=1}^{\infty} \mathbb{E}_{\boldsymbol{\mu}}\left[\mathbbm{1}\left[i \notin B_{r-1}\right] \mid \mathcal{E}_1\right]\left(\frac{2 d}{\varepsilon_{r}^2} \log \left(\frac{2K r(r+1)}{\delta}\right)+\frac{d(d+1)}{2}\right) \nonumber\\
&\leq \frac{d(d+1)}{2} \log _2\left(\frac{8}{\Delta_i-\varepsilon}\right) + \xi \frac{256d}{(\Delta_i - \varepsilon)^2} \log \left(\frac{2K}{\delta} \log_2 \frac{16}{\Delta_i - \varepsilon} \right).
\end{align}

We will repeat the result of Step 8 for every $i \in G_\varepsilon^c$ and plug this inequality into the result of Step 7. Then we can return the final result as follows.
\begin{align}
    \mathbb{E}_{\boldsymbol{\mu}}\left[T_G \mid \mathcal{E}_1\right]
    &\leq c 2^{2R_{\max }+1} d \log \left(\frac{2K R_{\max }(R_{\max }+1)}{\delta}\right)+\frac{d(d+1)}{2}R_{\max } \nonumber\\
    &+ \sum_{i \in G_\varepsilon^c} \sum_{r=1} ^{\infty} 
    \mathbb{E}_{\boldsymbol{\mu}}\left[ \mathbbm{1}\left[i \notin B_{r-1} \mid \mathcal{E}_1 \right] \right] \left(\frac{2 d}{\varepsilon_{r}^2} \log \left(\frac{2K r(r+1)}{\delta}\right)+\frac{d(d+1)}{2}\right) \nonumber\\
    &\leq c \frac{256d}{\alpha_\varepsilon^2} \log \left( \frac{2K}{\delta} \log_2 \left( \frac{16}{\alpha_\varepsilon} \right) \right) + \frac{d(d+1)}{2} \log_2 \frac{8}{\alpha_\varepsilon} \nonumber\\
    &+ \sum_{i \in G_\varepsilon^c}\left( \frac{d(d+1)}{2} \log _2\left(\frac{8}{\Delta_i-\varepsilon}\right) + \xi \frac{256d}{(\Delta_i - \varepsilon)^2} \log \left(\frac{2K}{\delta} \log_2 \frac{16}{\Delta_i - \varepsilon} \right) \right).
\end{align}

\subsection{A refined version removing the summation}\label{refined version}
However, the final result derived from the former steps contains the form of summation over the set $G_\varepsilon$, which can be improved by removing the summation. Instead of merely focusing on the round $R_\textup{max}$, which is defined in Lemma \ref{claim8_1} in Step 8, when all the arms in $G_\varepsilon$ are classified into set $G_r$, we can define the round when all the classifications have been finished, \emph{i.e.}, $G_r \cup B_r = [K]$.

\begin{lemma}\label{claim_refine_1}
    For $i \in G_\varepsilon$ and $r \geq\left\lceil\log _2\left(\frac{4}{\varepsilon-\Delta_i}\right)\right\rceil$, we have $ \mathbb{E}_{\boldsymbol{\mu}}\left[\mathbbm{1}\left[i \notin G_r\right] \mid \mathcal{E}_1\right] = 0$.
\end{lemma}

\proof\\{\textit{Proof.}}
First, we have that for any $i \in G_\varepsilon$
\begin{align}\label{eq_EvG_claim0}
    \mathbb{E}_{\boldsymbol{\mu}}\left[\mathbbm{1}\left[i \notin G_r\right] \mid \mathcal{E}_1\right] &= \mathbb{E}_{\boldsymbol{\mu}}\left[ \mathbbm{1}\left[\max_{j \in \mathcal{A}_I(r-1)} \hat{\mu}_j - \hat{\mu}_i \geq -2\varepsilon_r + \varepsilon\right] \mid \mathcal{E}_1, i \notin G_m (m = \{1, 2, \ldots, r-1\}) \right] \nonumber \\
    &\leq \mathbb{E}_{\boldsymbol{\mu}}\left[ \mathbbm{1}\left[\max_{j \in \mathcal{A}_I(r-1)} \hat{\mu}_j - \hat{\mu}_i \geq -2\varepsilon_r + \varepsilon\right] \mid \mathcal{E}_1\right].
\end{align}

If $i \in G_{r-1}$, then $i \in G_r$ by definition. Otherwise, if $i \notin G_{r-1}$, then under event $\mathcal{E}_1$, for $i \in G_\varepsilon$ and $r \geq\left\lceil\log _2\left(\frac{4}{\varepsilon-\Delta_i}\right)\right\rceil$, we have
\begin{equation}
    \max_{j \in \mathcal{A}_I(r-1)} \hat{\mu}_j - \hat{\mu}_i \leq \mu_{\arg\max_{j \in \mathcal{A}_I(r-1)}\hat{\mu}_j} - \mu_i +2^{-r+1}\leq \Delta_i+2^{-r+1} \leq \varepsilon-2\varepsilon_r,
\end{equation}
which implies that $i \in G_r$ by line \ref{add to Gr} of the Algorithm \ref{alg_LinFACT_stopping}. In other words, here we introduce when exactly line \ref{add to Gr} will happen. In particular, under event $\mathcal{E}_1$, if $i \notin G_{r-1}$, for all $i \in G_\varepsilon$ and $r \geq\left\lceil\log _2\left(\frac{4}{\varepsilon-\Delta_i}\right)\right\rceil$, we have
\begin{equation}\label{expectation about contradictory_xy}
    \mathbb{E}_{\boldsymbol{\mu}}\left[ \mathbbm{1}\left[\max_{j \in \mathcal{A}_I(r-1)} \hat{\mu}_j - \hat{\mu}_i \leq -2\varepsilon_r + \varepsilon\right] \mid i \notin G_{r-1}, \mathcal{E}_1 \right] = 1.
\end{equation}

Deterministically, $\mathbbm{1}\left[i \notin G_r\right] \mathbbm{1}\left[i \in G_{r-1}\right]=0$. Therefore,
\begin{align}
    &\mathbb{E}_{\boldsymbol{\mu}}\left[ \mathbbm{1}\left[\max_{j \in \mathcal{A}_I(r-1)} \hat{\mu}_j - \hat{\mu}_i \geq -2\varepsilon_r + \varepsilon\right] \mid \mathcal{E}_1 \right] \nonumber\\
    &= \mathbb{E}_{\boldsymbol{\mu}}\left[ \mathbbm{1}\left[\max_{j \in \mathcal{A}_I(r-1)} \hat{\mu}_j - \hat{\mu}_i \geq -2\varepsilon_r + \varepsilon\right] \mathbbm{1}\left[i \notin G_{r-1}\right] \mid \mathcal{E}_1 \right] \nonumber\\
    &+ \mathbb{E}_{\boldsymbol{\mu}}\left[ \mathbbm{1}\left[\max_{j \in \mathcal{A}_I(r-1)} \hat{\mu}_j - \hat{\mu}_i \geq -2\varepsilon_r + \varepsilon\right] \mathbbm{1}\left[i \in G_{r-1}\right] \mid \mathcal{E}_1 \right]\nonumber\\
    &= \mathbb{E}_{\boldsymbol{\mu}}\left[ \mathbbm{1}\left[\max_{j \in \mathcal{A}_I(r-1)} \hat{\mu}_j - \hat{\mu}_i \geq -2\varepsilon_r + \varepsilon\right] \mathbbm{1}\left[i \notin G_{r-1}\right] \mid \mathcal{E}_1 \right] \nonumber\\
    &=\mathbb{E}_{\boldsymbol{\mu}}\left[\mathbbm{1}\left[\max_{j \in \mathcal{A}_I(r-1)} \hat{\mu}_j - \hat{\mu}_i \geq -2\varepsilon_r + \varepsilon\right] \mathbbm{1}\left[i \notin G_{r-1}\right] \mid i \notin G_{r-1}, \mathcal{E}_1\right] \mathbb{P}\left(i \notin G_{r-1} \mid \mathcal{E}_1\right) \nonumber\\
    &+ \mathbb{E}_{\boldsymbol{\mu}}\left[\mathbbm{1}\left[\max_{j \in \mathcal{A}_I(r-1)} \hat{\mu}_j - \hat{\mu}_i \geq -2\varepsilon_r + \varepsilon\right] \mathbbm{1}\left[i \notin G_{r-1}\right] \mid i \in G_{r-1}, \mathcal{E}_1\right] \mathbb{P}\left(i \in G_{r-1} \mid \mathcal{E}_1\right) \nonumber\\
    &= \mathbb{E}_{\boldsymbol{\mu}}\left[\mathbbm{1}\left[\max_{j \in \mathcal{A}_I(r-1)} \hat{\mu}_j - \hat{\mu}_i \geq -2\varepsilon_r + \varepsilon\right] \mathbbm{1}\left[i \notin G_{r-1}\right] \mid i \notin G_{r-1}, \mathcal{E}_1\right] \mathbb{P}\left(i \notin G_{r-1} \mid \mathcal{E}_1\right) \nonumber\\
    &= \mathbb{E}_{\boldsymbol{\mu}}\left[\mathbbm{1}\left[\max_{j \in \mathcal{A}_I(r-1)} \hat{\mu}_j - \hat{\mu}_i \geq -2\varepsilon_r + \varepsilon\right] \mid i \notin G_{r-1}, \mathcal{E}_1\right] \mathbb{E}_{\boldsymbol{\mu}}\left[\mathbbm{1}\left[i \notin G_{r-1}\right] \mid \mathcal{E}_1\right] \nonumber\\
    &= 0,
\end{align}
where the second line comes from the additivity of expectation. The fourth line follows the deterministic result that $\mathbbm{1}\left[i \notin G_r\right] \mathbbm{1}\left[i \in G_{r-1}\right]=0$. The fifth line is the decomposition based on the conditional expectation. The eighth line comes from the fact that the expectation of the indicator function is simply the probability. The last line follows the result in equation~\eqref{expectation about contradictory_xy}.

The lemma can thus be concluded together with equation \eqref{eq_EvG_claim0}.
\hfill\Halmos
\endproof

\begin{lemma}\label{claim_refine_2}
    $R_{\textup{upper}} = \max \left\{ \left\lceil \log_2 \frac{4}{\alpha_\varepsilon} \right\rceil, \left\lceil \log_2 \frac{4}{\beta_\varepsilon} \right\rceil \right\}$ is the round where all the classifications have been finished and the answer is returned.
\end{lemma}
\proof\\{\textit{Proof.}}
Combining the result in Lemma \ref{claim_refine_1} of this section and the result in Lemma \ref{claim8_1} of Step 8, instead of focusing on $R_{\max}$, we can define a new unknown round $R_{\textup{upper}} = \max \left\{ \left\lceil \log_2 \frac{4}{\alpha_\varepsilon} \right\rceil, \left\lceil \log_2 \frac{4}{\beta_\varepsilon} \right\rceil \right\}$ to assist the derivation of our upper bound. Specifically, for $i \in G_\varepsilon^c$ and $r \geq\left\lceil\log _2\left(\frac{4}{\Delta_i-\varepsilon}\right)\right\rceil$, we have $\mathbb{E}_{\boldsymbol{\mu}}\left[\mathbbm{1}\left[i \notin B_r\right] \mid \mathcal{E}_1\right] = 0$; for $i \in G_\varepsilon$ and $r \geq\left\lceil\log _2\left(\frac{4}{\varepsilon-\Delta_i}\right)\right\rceil$, we have $\mathbb{E}_{\boldsymbol{\mu}}\left[\mathbbm{1}\left[i \notin G_r\right] \mid \mathcal{E}_1\right] = 0$.

Considering that $\alpha_\varepsilon = \min_{i \in G_\varepsilon}(\varepsilon - \Delta_i)$ and $\beta_\varepsilon = \min_{i \in G_\varepsilon^c}(\Delta_i - \varepsilon)$, for the round $r \geq R_{\textup{upper}}$, all the arms have been added into set $G_r$ or $B_r$, representing the termination of the algorithm.
\hfill\Halmos
\endproof

\begin{lemma}\label{claim_refine_3}
    For the expected sample complexity with high probability event $\mathcal{E}_1$, we have
\begin{align}
    \mathbb{E}_{\boldsymbol{\mu}}\left[T_{\mathcal{XY}} \mid \mathcal{E}_1\right]
    &\leq \zeta \max\left\{ \frac{256d}{\alpha_\varepsilon^2} \log \left( \frac{2K}{\delta} \log_2 \frac{16}{\alpha_\varepsilon} \right), \frac{256d}{\beta_\varepsilon^2} \log \left( \frac{2K}{\delta} \log_2 \frac{16}{\beta_\varepsilon} \right) \right\} + \frac{d(d+1)}{2} R_{\textup{upper}},
\end{align}
where $\zeta$ is a universal constant and $R_{\textup{upper}} = \max \left\{ \left\lceil \log_2 \frac{4}{\alpha_\varepsilon} \right\rceil, \left\lceil \log_2 \frac{4}{\beta_\varepsilon} \right\rceil \right\}$ is the round where all the classifications have been finished and the answer is returned.
\end{lemma}

\proof\\{\textit{Proof.}}
Based on the analysis of this step, we can have another decomposition of $T$ in equation (\ref{equation: decomposition of T}). These expectations are conditioned on the high probability event $\mathcal{E}_1$, given by
\begin{align}
    \mathbb{E}_{\boldsymbol{\mu}}\left[T_{\mathcal{XY}} \mid \mathcal{E}_1\right]
    &\leq \sum_{r=1}^{\infty} \mathbb{E}_{\boldsymbol{\mu}}\left[\mathbbm{1}\left[G_r \cup B_r \neq[K]\right] \mid \mathcal{E}_1\right] \sum_{\bm{a} \in \mathcal{A}(r-1)}T_r(\bm{a}) \nonumber\\
    &\leq \sum_{r=1}^{R_{\textup{upper}}} \left( d 2^{2r+1} \log \left(\frac{2K r(r+1)}{\delta}\right)+\frac{d(d+1)}{2}\right) \nonumber\\
    &\leq \frac{d(d+1)}{2} R_{\textup{upper}} + 2d\log \left(\frac{2K}{\delta}\right) \sum_{r=1}^{{R_{\textup{upper}}}} 2^{2r} + 4d \sum_{r=1}^{{R_{\textup{upper}}}} 2^{2r} \log (r+1) \nonumber\\
    &\leq 4\log \left[ \frac{2K}{\delta}\left( R_{\textup{upper}} + 1 \right) \right] \sum_{r=1}^{{R_{\textup{upper}}}} d 2^{2r} + \frac{d(d+1)}{2} R_{\textup{upper}} \label{middle result_refined} \\
    &\leq \zeta \max\left\{ \frac{256d}{\alpha_\varepsilon^2} \log \left( \frac{2K}{\delta} \log_2 \frac{16}{\alpha_\varepsilon} \right), \frac{256d}{\beta_\varepsilon^2} \log \left( \frac{2K}{\delta} \log_2 \frac{16}{\beta_\varepsilon} \right) \right\} + \frac{d(d+1)}{2} R_{\textup{upper}},\label{another angle: decomposition of T - 1}
\end{align}
where $\zeta$ is a universal constant. The second line comes from Lemma \ref{claim_refine_2}.
\hfill\Halmos
\endproof

\section{Additional insights into the algorithm optimality}\label{additional insights}

From another perspective, we consider the relationship between the lower bound and the upper bound in the following section and give some additional insights into the algorithm optimality. This relationship serves as the basis for the derivation of a near-optimal upper bound in Theorem \ref{upper bound: Algorithm 1 XY}.

For $\forall i \in \left( \mathcal{A}_I(r-1) \cap G_\varepsilon\right)$ and $\forall j \in \mathcal{A}_I(r-1)$, $j \neq i$, in round $r$, we have
\begin{equation}
    \boldsymbol{y}_{j, i}^\top(\hat{\boldsymbol{\theta}}_r-\boldsymbol{\theta}^{}) \leq 2\varepsilon_r
\end{equation}

\noindent and
\begin{equation}\label{Gr prime}
    \boldsymbol{y}_{j, i}^\top\hat{\boldsymbol{\theta}}_r - \varepsilon \leq \boldsymbol{y}_{j, i}^\top \boldsymbol{\theta}^{} + 2\varepsilon_r - \varepsilon.
\end{equation}

\begin{lemma}\label{claim_add_1}
    Define $G_r^\prime$ as $G_r^\prime \coloneqq \left\{ \exists j \in \mathcal{A}_I(r-1), j \neq i, i: \boldsymbol{y}_{j,i}^\top\boldsymbol{\theta}^{} - \varepsilon > -4\varepsilon_r \right\}$. We can prove that for $i \in G_\varepsilon$, we always have $\left( \mathcal{A}_I(r-1) \cap G_\varepsilon \cap G_r^c \right) \subset G_r^\prime$.
\end{lemma}

\proof\\{\textit{Proof.}}
When $r = 1$, the lemma can be easily proved through the assumption in the Theorem \ref{upper bound: Algorithm 1 XY} that $\max_{i \in [K]}\vert \mu_1 - \varepsilon - \mu_i \vert \leq 2$. When $r \geq 2$, with the idea of contradiction, if $i \in \left( \mathcal{A}_I(r-1) \cap G_\varepsilon \cap G_r^c\right) \cap (G_r^\prime)^c$, then for $\forall j \in \mathcal{A}_I(r-1)$ and $j \neq i$, we have 
\begin{equation}
    \boldsymbol{y}_{j, i}^\top\boldsymbol{\theta}^{} \leq -4\varepsilon_r +\varepsilon.
\end{equation}

Hence, with equation (\ref{Gr prime}), for $\forall j \in \mathcal{A}_I(r-1)$ and $j \neq i$, we have
\begin{equation}
    \boldsymbol{y}_{j, i}^\top\hat{\boldsymbol{\theta}}_r - \varepsilon \leq-2\varepsilon_r,
\end{equation}
which is exactly the condition for the algorithm to add arm $i$ into $G_r$ by line \ref{add to Gr} of the algorithm. This contradiction leads to the result. Besides, note that when $r \geq \left\lceil \log_2 \frac{4}{\alpha_\varepsilon} \right\rceil$, we have $G_r^\prime \cap G_\varepsilon = \emptyset$. Furthermore, considering that $i \in G_\varepsilon$, we have $\bm{y}_{i, j}^\top\boldsymbol{\theta}^{} + \varepsilon > 0$.
\hfill\Halmos
\endproof

However, there is one exceptional case that would render the above derivation invalid. In the proof, when $i$ happens to be the index of the arm with the largest mean value, \emph{i.e.}, $i = \arg \max_{i \in \mathcal{A}_I(r-1)}$, the proof does not hold. To let this happen, $\arg \max_{i \in \mathcal{A}_I(r-1)} \in G_r^c$ must hold. This is equivalent to $\varepsilon \leq 2\varepsilon_r$, which will not last long and can thus be ignored.

On the other hand, for $i \in \left( \mathcal{A}_I(r-1) \cap G_\varepsilon^c \right)$, in any round $r$, we have
\begin{equation}
    \boldsymbol{y}_{1,i}^\top(\boldsymbol{\theta}^{}-\hat{\boldsymbol{\theta}}_r) \leq 2\varepsilon_r
\end{equation}
and
\begin{equation}\label{Br prime}
    \boldsymbol{y}_{1, i}^\top\hat{\boldsymbol{\theta}}_r - \varepsilon \geq \boldsymbol{y}_{1, i}^\top \boldsymbol{\theta}^{} - 2\varepsilon_r - \varepsilon.
\end{equation}

\begin{lemma}\label{claim_add_2}
    Define $B_r^\prime$ as $B_r^\prime \coloneqq \left\{ i: \boldsymbol{y}_{1,i}^\top\boldsymbol{\theta}^{} - \varepsilon < 4\varepsilon_r \right\}$. Then considering that 1 always belongs to $\mathcal{A}_I(r)$, we can prove that for $i \in G_\varepsilon^c$, we always have $\left( \mathcal{A}_I(r-1) \cap G_\varepsilon^c\right) \subset B_r^\prime$.
\end{lemma}
    
\proof\\{\textit{Proof.}}
To prove this, when $r = 1$, the lemma can be easily proved through the assumption in the Theorem \ref{upper bound: Algorithm 1 XY} that $\max_{i \in [K]}\vert \mu_1 - \varepsilon - \mu_i \vert \leq 2$. When $r \geq 2$, with the same idea of contradiction, if $i \in \left( \mathcal{A}_I(r-1) \cap G_\varepsilon^c\right) \cap (B_r^\prime)^c$, then we have
\begin{equation}
    \boldsymbol{y}_{1, i}^\top\boldsymbol{\theta}^{} \geq 4\varepsilon_r +\varepsilon.
\end{equation}

Hence, with equation (\ref{Br prime}), we have
\begin{equation}
    \boldsymbol{y}_{1, i}^\top\hat{\boldsymbol{\theta}}_r - \varepsilon \geq 2\varepsilon_r,
\end{equation}
which is exactly the condition for the algorithm to add arm $i$ into $B_r$ and eliminate it from $\mathcal{A}_I(r-1)$ by line \ref{elimination: good 1} of the algorithm. This contradiction leads to the result. Besides, note that when $r \geq \left\lceil \log_2 \frac{4}{\beta_\varepsilon} \right\rceil$, we have $B_r^\prime \cap G_\varepsilon^c = \emptyset$. Furthermore, for $i \in G_\varepsilon^c$, we can conclude that $\boldsymbol{y}_{1,i}^\top\boldsymbol{\theta}^{} - \varepsilon > 0$.
\hfill\Halmos
\endproof

Then we provide a critical lemma that represents the relationship between the newly proposed lower bound in Theorem \ref{lower bound: All-epsilon in the linear setting} and our upper bound.

\begin{lemma}\label{claim_add_3}
    For the stochastic linear bandit, considering the value of lower bound $(\Gamma^\ast)^{-1}$ in Theorem \ref{lower bound: All-epsilon in the linear setting}, we have
\begin{equation}
     (\Gamma^\ast)^{-1} \geq \frac{{\mathcal{G}_{\mathcal{Y}}}^2 L_2}{4R_{\textup{upper}} d L_1}  \sum_{r=1}^{R_{\textup{upper}}} 2^{2r-3} \min_{\bm{p} \in S_K} \max_{i, j \in \mathcal{A}_I(r-1)} \Vert \boldsymbol{y}_{i, j} \Vert_{\bm{V}_{\bm{p}}^{-1}}^2,
\end{equation}
where $R_{\textup{upper}} = \max \left\{ \left\lceil \log_2 \frac{4}{\alpha_\varepsilon} \right\rceil, \left\lceil \log_2 \frac{4}{\beta_\varepsilon} \right\rceil \right\}$ is the round where all the classifications have been finished and the answer is returned; ${\mathcal{G}_{\mathcal{Y}}}$, introduced in Lemma \ref{lemma: inequality iv}, is the gauge of $\mathcal{Y}(\mathcal{A})$, where $\mathcal{A}$ is the initial set of all arm vectors; $L_1$ and $L_2$ are constraint parameters for the bandit instance, introduced in the Section \ref{sec_Problem Formulation} and the Theorem \ref{upper bound: Algorithm 1 XY} respectively.
\end{lemma}

\proof\\{\textit{Proof.}}
Let $R_{\textup{upper}} = \max \left\{ \left\lceil \log_2 \frac{4}{\alpha_\varepsilon} \right\rceil, \left\lceil \log_2 \frac{4}{\beta_\varepsilon} \right\rceil \right\}$. When round $r$ is larger than $R_{\textup{upper}}$, we have $G_r^\prime \cap G_\varepsilon = \emptyset$ and $B_r^\prime \cap G_\varepsilon^c = \emptyset$, meaning $G_r \cup B_r = [K]$ and the algorithm is terminated. From Theorem \ref{lower bound: All-epsilon in the linear setting}, we have
\begin{align}
    (\Gamma^\ast)^{-1} &= \min_{\bm{p} \in S_K}\max_{(i, j, m) \in \mathcal{X}} \max \left\{ \frac{2 \Vert \boldsymbol{y}_{i, j} \Vert_{\bm{V}_{\bm{p}}^{-1}}^2}{\left ( \boldsymbol{y}_{i, j}^\top \boldsymbol{\theta}^{} + \varepsilon \right )^2}, \frac{2 \Vert \boldsymbol{y}_{1, m} \Vert_{\bm{V}_{\bm{p}}^{-1}}^2}{\left ( \boldsymbol{y}_{1, m}^\top \boldsymbol{\theta}^{} - \varepsilon \right )^2} \right\} \nonumber\\
    &= \min_{\bm{p} \in S_K} \max_{r \leq R_{\textup{upper}}} \max_{i \in G_r^\prime \cap G_\varepsilon} \max_{\substack{j \in \mathcal{A}_I(r-1) \\ j \neq i}} \max_{m \in B_r^\prime \cap G_\varepsilon^c} \max \left\{ \frac{2 \Vert \boldsymbol{y}_{i, j} \Vert_{\bm{V}_{\bm{p}}^{-1}}^2}{\left ( \boldsymbol{y}_{i, j}^\top \boldsymbol{\theta}^{} + \varepsilon \right )^2}, \frac{2 \Vert \boldsymbol{y}_{1, m} \Vert_{\bm{V}_{\bm{p}}^{-1}}^2}{\left ( \boldsymbol{y}_{1, m}^\top \boldsymbol{\theta}^{} - \varepsilon \right )^2} \right\} \nonumber\\
    &\geq \min_{\bm{p} \in S_K} \max_{r \leq R_{\textup{upper}}} \max_{i \in G_r^\prime \cap G_\varepsilon} \max_{\substack{j \in \mathcal{A}_I(r-1) \\ j \neq i}} \max_{m \in B_r^\prime \cap G_\varepsilon^c} \max \left\{ \frac{2 \Vert \boldsymbol{y}_{i, j} \Vert_{\bm{V}_{\bm{p}}^{-1}}^2}{\left (4\varepsilon_r \right )^2}, \frac{2 \Vert \boldsymbol{y}_{1, m} \Vert_{\bm{V}_{\bm{p}}^{-1}}^2}{\left ( 4\varepsilon_r \right )^2} \right\} \nonumber\\
    &\stackrel{\text{(i)}}{\geq} \frac{1}{R_{\textup{upper}}} \min_{\bm{p} \in S_K} \sum_{r=1}^{R_{\textup{upper}}} \max_{i \in G_r^\prime \cap G_\varepsilon} \max_{\substack{j \in \mathcal{A}_I(r-1) \\ j \neq i}} \max_{m \in B_r^\prime \cap G_\varepsilon^c} \max \left\{ \frac{2 \Vert \boldsymbol{y}_{i, j} \Vert_{\bm{V}_{\bm{p}}^{-1}}^2}{\left (4\varepsilon_r \right )^2}, \frac{2 \Vert \boldsymbol{y}_{1, m} \Vert_{\bm{V}_{\bm{p}}^{-1}}^2}{\left ( 4\varepsilon_r \right )^2} \right\} \nonumber\\
    &\stackrel{\text{(ii)}}{\geq} \frac{1}{R_{\textup{upper}}} \sum_{r=1}^{R_{\textup{upper}}} 2^{2r-3} \min_{\bm{p} \in S_K} \max_{i \in G_r^\prime \cap G_\varepsilon} \max_{\substack{j \in \mathcal{A}_I(r-1) \\ j \neq i}} \max_{m \in B_r^\prime \cap G_\varepsilon^c} \max \left\{ \Vert \boldsymbol{y}_{i, j} \Vert_{\bm{V}_{\bm{p}}^{-1}}^2, \Vert \boldsymbol{y}_{1, m} \Vert_{\bm{V}_{\bm{p}}^{-1}}^2 \right\} \nonumber\\
    &\stackrel{\text{(iii)}}{\geq} \frac{1}{R_{\textup{upper}}} \sum_{r=1}^{R_{\textup{upper}}} 2^{2r-3} \min_{\bm{p} \in S_K} \max_{i \in \mathcal{A}_I(r-1) \cap G_\varepsilon \cap G_r^c} \max_{\substack{j \in \mathcal{A}_I(r-1) \\ j \neq i}} \max_{m \in \mathcal{A}_I(r-1) \cap G_\varepsilon^c} \max \left\{ \Vert \boldsymbol{y}_{i, j} \Vert_{\bm{V}_{\bm{p}}^{-1}}^2, \Vert \boldsymbol{y}_{1, m} \Vert_{\bm{V}_{\bm{p}}^{-1}}^2 \right\} \nonumber\\
    &\stackrel{\text{(iv)}}{\geq} \frac{{\mathcal{G}_{\mathcal{Y}}}^2 L_2}{d L_1} \frac{1}{R_{\textup{upper}}} \sum_{r=1}^{R_{\textup{upper}}} 2^{2r-3} \min_{\bm{p} \in S_K} \max_{i \in \mathcal{A}_I(r-1) \cap G_\varepsilon \backslash \{ 1 \}} \max_{m \in \mathcal{A}_I(r-1) \cap G_\varepsilon^c} \max \left\{ \Vert \boldsymbol{y}_{1, i} \Vert_{\bm{V}_{\bm{p}}^{-1}}^2, \Vert \boldsymbol{y}_{1, m} \Vert_{\bm{V}_{\bm{p}}^{-1}}^2 \right\} \nonumber\\
    &\stackrel{\text{(v)}}{\geq} \frac{{\mathcal{G}_{\mathcal{Y}}}^2 L_2}{4R_{\textup{upper}} d L_1}  \sum_{r=1}^{R_{\textup{upper}}} 2^{2r-3} \min_{\bm{p} \in S_K} \max_{\substack{i, j \in \mathcal{A}_I(r-1) \\ i \neq j}} \Vert \boldsymbol{y}_{i, j} \Vert_{\bm{V}_{\bm{p}}^{-1}}^2 \nonumber\\
    &= \frac{{\mathcal{G}_{\mathcal{Y}}}^2 L_2}{32R_{\textup{upper}} d L_1}  \sum_{r=1}^{R_{\textup{upper}}} 2^{2r} g_{\mathcal{XY}}(\mathcal{Y}(\mathcal{A}(r-1))), \label{equation: relationship between the lower bound and the upper bound}
\end{align}
where ($i$) follows from the fact that the maximum of positive numbers is always less than the average, and ($ii$) by the fact that the minimum of a sum is greater than the sum of minimums. ($iii$) comes from the inclusion relationship between sets in Lemma \ref{claim_add_1} and Lemma \ref{claim_add_2}. Take $j = 1$ and ($iv$) is a conclusion of Lemma \ref{lemma: inequality iv}. To see ($v$), note that for $i,j \in \mathcal{A}_I(r-1)$, we have $\max_{i, j \in \mathcal{A}_I(r-1), i \neq j} \Vert \boldsymbol{y}_{i, j} \Vert_{\bm{V}_{\bm{p}}^{-1}}^2 \leq 4 \max_{i \in \mathcal{A}_I(r-1)\backslash \{ 1 \}} \Vert \boldsymbol{y}_{1, i} \Vert_{\bm{V}_{\bm{p}}^{-1}}^2$.
\hfill\Halmos
\endproof

Moreover, to give some additional insights for the G-optimal design, from equation \eqref{middle result_refined} in Section \ref{refined version}, we have
\begin{align}
    \mathbb{E}_{\boldsymbol{\mu}}\left[T \mid \mathcal{E}_1\right] &\leq 4 \log \left[ \frac{2K}{\delta}\left( R_{\textup{upper}} + 1 \right) \right] \sum_{r=1}^{{R_{\textup{upper}}}} d 2^{2r} + \frac{d(d+1)}{2} R_{\textup{upper}}.\label{upper bound: additional insights}
\end{align}

From inequalities (\ref{equation: relationship between the lower bound and the upper bound}) and (\ref{upper bound: additional insights}), it can be seen that to match the upper bound and the lower bound, one must establish that $d \leq c \min_{\bm{p} \in S_K} \max_{i, j \in \mathcal{A}_I(r-1), i \neq j} \Vert \boldsymbol{y}_{i, j} \Vert_{\bm{V}_{\bm{p}}^{-1}}^2$ for some universal constant. However, applying the Kiefer–Wolfowitz Theorem from Lemma \ref{theorem_KW} yields only the reverse inequality
\begin{equation}
    \min_{\bm{p} \in S_K} \max_{i, j \in \mathcal{A}_I(r-1), i \neq j} \Vert \boldsymbol{y}_{i, j} \Vert_{\bm{V}_{\bm{p}}^{-1}}^2 \leq 4\min_{\bm{p} \in S_K} \max_{i \in \mathcal{A}_I(r-1)} \Vert \boldsymbol{a}_{i} \Vert_{\bm{V}_{\bm{p}}^{-1}}^2 = 4d.
\end{equation}

This reversed inequality does not help in bridging the gap between the lower and upper bounds, thereby motivating our refinement of the algorithm's sampling policy. 
\hfill\Halmos
\endproof

\section{Proof of Theorem \ref{upper bound: Algorithm 1 XY}}\label{proof of theorem: upper bound of LinFACT XY}

From the former derivation and results in Theorem \ref{upper bound: Algorithm 1 G}, it can be seen that the upper bound of the proposed algorithm cannot match the lower bound in any form. Thus, to give a matching upper bound of the expected sample complexity, we provide a detailed proof, which proceeds as follows.

The key point of this proof is to show the direct relative size relationship between the lower bound and the key minimax summation terms in the
upper bound, providing a way to constrain the sample complexity with the lower bound term $(\Gamma^\ast)^{-1}$ directly. We begin by showing that the good event $\mathcal{E}_3$ holds with a probability of at least $1-\delta_r$ in each round $r$. We then show that the probability of this good event holding in every round is at least $1-\delta$. As a result, we can simply sum over the bound on the sample complexity in each round given in the good event to obtain the stated bound on the sample complexity.

Considering the difference between G-optimal allocation and $\mathcal{XY}$-optimal allocation, the upper bound analysis based on G-optimal design cannot match the lower bound. Here we rearrange the clean event and give the proof process for the $\mathcal{XY}$-optimal design in Algorithm \ref{alg_LinFACT_sampling_XY}. 

The initial point of the algorithm design and the proof is the definition of the events $\mathcal{E}_3$ below.
\begin{equation}
    \mathcal{E}_3 = \bigcap_{i \in \mathcal{A}_I(r-1)} \bigcap_{\substack{
    j \in \mathcal{A}_I(r-1) \\ j \neq i}} \bigcap_{r \in \mathbb{N}} \left| (\hat{\mu}_j(r) - \hat{\mu}_i(r)) - (\mu_j - \mu_i) \right| \leq 2\varepsilon_r. \label{clean event XY}
\end{equation}

Considering that the arm is sampled based on the preset allocation, \emph{i.e.}, the fixed design, here we introduce the following lemma to give a simple confidence region of the estimated parameter $\boldsymbol{\theta}$.

\begin{lemma}\label{claim_xy_1}
    Let $\delta \in (0,1)$, it holds that 
    $
      P(\mathcal{E}_3) \geq 1 - \delta.
    $
\end{lemma}
 
\proof\\{\textit{Proof.}}
Since $\hat{\boldsymbol{\theta}}_r$ is a ordinary least squares estimator of $\boldsymbol{\theta}^{}$ and the noise is i.i.d., we know that $\bm{y}^{\top}\left(\boldsymbol{\theta}^{}-\hat{\boldsymbol{\theta}}_r\right) \text { is }\| \bm{y} \|_{\bm{V}_r^{-1}}^2$-sub-Gaussian for all $\bm{y} \in \mathcal{Y}(\mathcal{A}(r-1))$. Furthermore, due to the guarantees of the rounding procedure, we have
\begin{equation}
    \| \bm{y} \|_{\bm{V}_r^{-1}}^2 \leq \left( 1+\epsilon \right) g_{\mathcal{XY}}(\mathcal{Y}(\mathcal{A}(r-1)))/T_r \leq \frac{2^{-2r-1}}{\log \frac{2K(K-1)r(r+1)}{\delta}}
\end{equation}
for all $\bm{y} \in \mathcal{Y}(\mathcal{A}(r-1))$ by our choice of $T_r$ in equation \eqref{equation_phase budget XY}. Since the right-hand side is deterministic, independent of the random reward of arms, for any $\rho > 0$, we have that
\begin{equation}
    \mathbb{P}\left\{ \vert \bm{y}^{\top}\left(\boldsymbol{\theta}^{}-\hat{\boldsymbol{\theta}}_r\right) \vert > \sqrt{2^{-2r} \frac{\log(2/\rho)}{\log \frac{2K(K-1)r(r+1)}{\delta}}} \right\} \leq \rho
\end{equation}
for all $\bm{y} \in \mathcal{Y}(\mathcal{A}(r-1))$. Taking $\rho = \frac{\delta}{K(K-1)r(r+1)}$ and and union bounding over all the possible $\bm{y} \in \mathcal{Y}(\mathcal{A}(r-1))$, where we have $\vert \mathcal{Y}(\mathcal{A}(r-1)) \vert \leq \vert \mathcal{Y}(\mathcal{A}(0)) \vert \leq K(K-1)$. Thus, to give the probability guarantee, we have 
\begin{align}
    \mathbb{P}(\mathcal{E}_3^c) &= \mathbb{P} \left\{ \bigcup_{i \in \mathcal{A}_I(r-1)} \bigcup_{\substack{
    j \in \mathcal{A}_I(r-1) \\ j \neq i}} \bigcup_{r \in \mathbb{N}} \left| (\hat{\mu}_j(r) - \hat{\mu}_i(r)) - (\mu_j - \mu_i) \right| > \varepsilon_r \right\}
    \nonumber\\ &\leq \sum_{r=1}^{\infty} \mathbb{P} \left\{ \bigcup_{i \in \mathcal{A}_I(r-1)} \bigcup_{\substack{
    j \in \mathcal{A}_I(r-1) \\ j \neq i}} \left| (\hat{\mu}_j(r) - \hat{\mu}_i(r)) - (\mu_j - \mu_i) \right| > \varepsilon_r \right\}
    \nonumber\\ &\leq \sum_{r=1}^{\infty} \sum_{i=1}^{K} \sum_{\substack{
    j = 1 \\ j \neq i}}^{K} \frac{\delta}{K(K-1)r(r+1)}\nonumber\\
    &= \delta.
\end{align}

Taking the union bound over all rounds $r \in \mathbb{N}$ completes the proof of the lemma. 
\hfill\Halmos
\endproof

Thus with the standard result of the $\mathcal{XY}$-optimal design, we have
\begin{equation}
    C_{\delta/K}(r) \coloneqq \varepsilon_r,
\end{equation}
which is the same as in the equation \eqref{eq_C_G} of G-optimal design.

\begin{lemma}\label{claim_xy_2}
    On the newly designed clean event $\mathcal{E}_3$, best arm index $1 \in \mathcal{A}_I(r)$ for all $r \in \mathbb{N}$.
\end{lemma}

\proof\\{\textit{Proof.}}
Firstly we show $1 \in \mathcal{A}_I(r)$ for all $r \in \mathbb{N}$, that is, the best arm is never eliminated from $\mathcal{A}(r-1)$ in any round. If the event $\mathcal{E}_3$ holds, note for any arm $i \in \mathcal{A}_I(r-1)$, we have
\begin{equation}
    \hat{\mu}_i(r) - \hat{\mu}_1(r) \leq \mu_i - \mu_1 + 2\varepsilon_r \leq 2\varepsilon_r < 2\varepsilon_r + \varepsilon,
\end{equation}

\noindent which particularly shows that $\hat{\mu}_1(r) + \varepsilon_r > \max_{i \in \mathcal{A}_I(r-1)} \hat{\mu}_i - \varepsilon_r - \varepsilon = L_r$ and $\hat{\mu}_1(r) + \varepsilon_r \geq \max_{i \in \mathcal{A}_I(r-1)} \hat{\mu}_i (r) - C_{\delta/K}$ showing that arm $1$ will never exit $\mathcal{A}_I(r)$ in line \ref{elimination: good 1} or line \ref{elimination: good 2} of LinFACT.
\hfill\Halmos
\endproof




\begin{lemma}\label{claim_xy_3}
    With probability greater than $1-\delta$, using an $\epsilon$-efficient rounding procedure. The proposed LinFACT algorithm correctly identifies all $\varepsilon$-best arms and is instance optimal up to logarithmic factors, given by
\begin{equation}
    T \leq c \left[d R_{\textup{upper}} \log \left( \frac{2K(R_{\textup{upper}}+1)}{\delta} \right) \right] (\Gamma^\ast)^{-1} + q\left( \epsilon \right) R_{\textup{upper}},
\end{equation}
where $c$ is a universal constant, $R_{\textup{upper}} = \max \left\{ \left\lceil \log_2 \frac{4}{\alpha_\varepsilon} \right\rceil, \left\lceil \log_2 \frac{4}{\beta_\varepsilon} \right\rceil \right\}$, and $(\Gamma^\ast)^{-1}$, proposed in Theorem \ref{lower bound: All-epsilon in the linear setting}, is the critical term in the lower bound of the linear bandit with all $\varepsilon$-best pure exploration task. 
\end{lemma}

\proof\\{\textit{Proof.}}
Combining the result of Lemma \ref{claim_xy_1} in this section, with a probability as least $1-\delta$, we can introduce the term $(\Gamma^\ast)^{-1}$, the lower bound in Theorem \ref{lower bound: All-epsilon in the linear setting} to assist our derivation.
\begin{align}
    T &\leq \sum_{r=1}^{R_{\textup{upper}}} \max \left\{ \left\lceil \frac{2 g_{\mathcal{XY}}\left( \mathcal{Y}(\mathcal{A}(r-1)) \right) (1+\epsilon)}{\varepsilon_r^2} \log \left( \frac{2K(K-1)r(r+1)}{\delta} \right) \right\rceil, q\left( \epsilon \right) \right\} \nonumber\\
    &\leq \sum_{r=1}^{R_{\textup{upper}}} 2 \cdot 2^{2r} g_{\mathcal{XY}}\left( \mathcal{Y}(\mathcal{A}(r-1)) \right) (1+\epsilon) \log \left( \frac{2K(K-1)r(r+1)}{\delta} \right) + (1 + q\left( \epsilon \right))R_{\textup{upper}} \nonumber\\
    &\leq \left[ 64\left( 1+\epsilon \right) \log \left( \frac{2K(K-1)R_{\textup{upper}}(R_{\textup{upper}}+1)}{\delta} \right) \frac{R_{\textup{upper}}d L_1}{{\mathcal{G}_{\mathcal{Y}}}^2 L_2} \right] (\Gamma^\ast)^{-1} + (1 + q\left( \epsilon \right))R_{\textup{upper}} \nonumber\\
    &\leq \left[ 128\left( 1+\epsilon \right) \log \left( \frac{2K(R_{\textup{upper}}+1)}{\delta} \right) \frac{R_{\textup{upper}}d L_1}{{\mathcal{G}_{\mathcal{Y}}}^2 L_2} \right] (\Gamma^\ast)^{-1} + (1 + q\left( \epsilon \right))R_{\textup{upper}} \nonumber\\
    &\leq c \left[ d R_{\textup{upper}} \log \left( \frac{2K(R_{\textup{upper}}+1)}{\delta} \right) \right] (\Gamma^\ast)^{-1} + q\left( \epsilon \right) R_{\textup{upper}},
\end{align}
where $c$ is a universal constant, $R_{\textup{upper}} = \max \left\{ \left\lceil \log_2 \frac{4}{\alpha_\varepsilon} \right\rceil, \left\lceil \log_2 \frac{4}{\beta_\varepsilon} \right\rceil \right\}$, and the third inequality comes from equation (\ref{equation: relationship between the lower bound and the upper bound}).
\hfill\Halmos
\endproof

\section{Proof of Theorem \ref{upper bound: Algorithm 1 G_mis_1}}\label{sec_Proof of upper bound: Algorithm 1 G_mis_1}
\subsection{Step 1: Rearrange the Clean Event With Model Misspecification}\label{subsec_Rearrange the Clean Event With Misspecification}
Following the derivation method in Section \ref{refined version}, we can provide the upper bound of LinFACT with the G-optimal design when the misspecification of the linear model is considered. The core of the proof is to similarly define the round when all the classifications have been done with the misspecified model. To achieve this, we have to reconstruct the anytime confidence radius for arms in each round $r$ and the high probability event during the whole process of our algorithm.

The initial point of this proof is the redesigned definition of the clean event $\mathcal{E}_{1m}$ below.
\begin{equation}\label{clean event mis_1}
    \mathcal{E}_{1m} = \left\{ \bigcap_{i \in \mathcal{A}_I(r-1)} \bigcap_{r \in \mathbb{N}} \left| \hat{\mu}_i(r) - \mu_i \right| \leq \varepsilon_r + L_m \sqrt{d} \right\}. 
\end{equation}

Similarly, considering that the arm is sampled based on the preset allocation, \emph{i.e.}, the fixed design, here we use previous Lemma \ref{lemma: confidence region for any single fixed arm} to give an adjusted confidence region of estimated parameter $\hat{\boldsymbol{\theta}}_t$. If we follow the G-optimal sampling rule as stated in line \ref{G_sampling_1} and line \ref{G_sampling_2} in the pseudocode, for each round $r$ we still have
\begin{equation}
    \bm{V}_r = \sum_{\bm{a} \in \text{Supp}(\pi_r)} T_r(\bm{a}) \bm{a} \bm{a}^\top \succeq \frac{2d}{\varepsilon_r^2} \log \left( \frac{2Kr(r+1)}{\delta} \right) \bm{V}(\pi).
\end{equation}

To give the confidence radius, we have the following decomposition.
\begin{align}\label{equation: confidence region_mis}
    \left|\left\langle \hat{\boldsymbol{\theta}}_r - \boldsymbol{\theta}, \bm{a} \right\rangle \right| &= \left| \boldsymbol{a}^\top \boldsymbol{V}_r^{-1} \sum_{s=1}^{T_r} \Delta_m(\boldsymbol{a}_{A_s})\boldsymbol{a}_{A_s} + \boldsymbol{a}^\top \boldsymbol{V}_r^{-1} \sum_{s=1}^{T_r} \eta_s \boldsymbol{a}_{A_s} \right| \nonumber \\
    &\leq \left| \boldsymbol{a}^\top \boldsymbol{V}_r^{-1} \sum_{s=1}^{T_r} \Delta_m(\boldsymbol{a}_{A_s})\boldsymbol{a}_{A_s} \right| + \left| \boldsymbol{a}^\top \boldsymbol{V}_r^{-1} \sum_{s=1}^{T_r} \eta_s \boldsymbol{a}_{A_s} \right|,
\end{align}
where the first term is bounded using Hölder's Inequality and Jensen’s Inequality, given by
\begin{align}\label{eq_mis_1}
    \left| \boldsymbol{a}^\top \boldsymbol{V}_r^{-1} \sum_{s=1}^{T_r} \Delta_m(\boldsymbol{a}_{A_s})\boldsymbol{a}_{A_s} \right| &= \left| \boldsymbol{a}^\top \boldsymbol{V}_r^{-1} \sum_{\bm{b} \in \mathcal{A}(r-1)} T_r(\bm{b}) \Delta_m(\bm{b}) \bm{b} \right| \nonumber\\
    &\leq L_m \sum_{\bm{b} \in \mathcal{A}(r-1)} T_r(\bm{b}) \left| \boldsymbol{a}^\top \boldsymbol{V}_r^{-1} \bm{b} \right| \nonumber\\
    &\leq L_m \sqrt{\left( \sum_{\bm{b} \in \mathcal{A}(r-1)} T_r(\bm{b}) \right) \boldsymbol{a}^\top \left( \sum_{\bm{b} \in \mathcal{A}(r-1)} T_r(\bm{b}) \boldsymbol{V}_r^{-1} \bm{b} \bm{b}^\top \boldsymbol{V}_r^{-1} \boldsymbol{a} \right)} \nonumber\\
    &= L_m \sqrt{\sum_{\bm{b} \in \mathcal{A}(r-1)} T_r(\bm{b}) \| \boldsymbol{a} \|_{\boldsymbol{V}_r^{-1}}^2} \nonumber\\
    &\leq L_m \sqrt{d},
\end{align}
where the first inequality follows Hölder's Inequality, the second is Jensen’s Inequality, and the last follows from the result of the G-optimal exploration policy that guarantees $\| \boldsymbol{a} \|_{\boldsymbol{V}_r^{-1}}^2 \leq {d}/{T_r}$.

The second term is still bounded with the inequality in Lemma \ref{lemma: confidence region for any single fixed arm} and the result in Lemma \ref{lemma: inversion reverses loewner orders}. For any arm $\bm{a} \in \mathcal{A}(r-1)$, with a probability of at least $1-\delta/{Kr(r+1)}$, we have
\begin{align}\label{eq_mis_2}
    \left| \boldsymbol{a}^\top \boldsymbol{V}_r^{-1} \sum_{s=1}^{T_r} \eta_s \boldsymbol{a}_{A_s} \right| &\leq \sqrt{2\left \lVert \bm{a} \right\rVert^2_{\bm{V}_r^{-1}} \log \left( \frac{2Kr(r+1)}{\delta} \right)} \nonumber\\
    &= \sqrt{2\bm{a}^\top\bm{V}_r^{-1}\bm{a} \log \left( \frac{2Kr(r+1)}{\delta} \right)} \nonumber\\
    &\leq \sqrt{2\bm{a}^\top \left( \frac{\varepsilon_r^2}{2d}\frac{1}{\log \left( \frac{2Kr(r+1)}{\delta} \right)}\bm{V}(\pi)^{-1} \right) \bm{a} \log \left( \frac{2Kr(r+1)}{\delta} \right)} \nonumber\\
    &\leq \varepsilon_r.
\end{align}

Thus with the standard result of the G-optimal design, we still have
\begin{equation}\label{eq_C_G_3}
    C_{\delta/K}(r) \coloneqq \varepsilon_r.
\end{equation}

To give the probability guarantee of newly defined event $\mathcal{E}_{1m}$, combining the result in equation \eqref{eq_mis_1} and equation \eqref{eq_mis_2}, we have
\begin{align}
    \mathbb{P}(\mathcal{E}_{1m}^c) &= \mathbb{P} \left\{ \bigcup_{i \in \mathcal{A}_I(r-1)} \bigcup_{r \in \mathbb{N}} \left| \hat{\mu}_i(r) - \mu_i \right| > \varepsilon_r + L_m \sqrt{d} \right\}
    \nonumber\\
    &\leq \sum_{r=1}^{\infty} \mathbb{P} \left\{ \bigcup_{i \in \mathcal{A}_I(r-1)} \left| \hat{\mu}_i(r) - \mu_i \right| > \varepsilon_r + L_m \sqrt{d} \right\} \nonumber\\
    &\leq \sum_{r=1}^{\infty} \sum_{i=1}^{K} \frac{\delta}{Kr(r+1)}\nonumber\\    &= \delta.
\end{align}

Therefore, considering the union bounds over the rounds $r \in \mathbb{N}$, we have
\begin{equation}
    P(\mathcal{E}_{1m}) \geq 1 - \delta.
\end{equation}

Considering another event describing the gaps between different arms, given by
\begin{equation}
    \mathcal{E}_{2m} = \bigcap_{i \in G_{\varepsilon}} \bigcap_{j \in \mathcal{A}_I(r-1)} \bigcap_{r \in \mathbb{N}} \left| (\hat{\mu}_j(r) - \hat{\mu}_i(r)) - (\mu_j - \mu_i) \right| \leq 2\varepsilon_r + 2L_m \sqrt{d}.
\end{equation}

By equation (\ref{equation: confidence region_mis}), for $i, j \in \mathcal{A}_I(r-1)$, we have
\begin{align}\label{confidence: arm filter}
    \mathbb{P}\left\{\left|(\hat{\mu}_j - \hat{\mu}_i) - (\mu_j - \mu_i)\right| > 2\varepsilon_r + 2L_m \sqrt{d} \mid \mathcal{E}_{1m} \right\} &\leq \mathbb{P}\left\{\left|\hat{\mu}_j - \mu_j \right| + \left|\hat{\mu}_i - \mu_i\right| > 2\varepsilon_r + 2L_m \sqrt{d} \mid \mathcal{E}_{1m} \right\} \nonumber\\
    &\leq \mathbb{P}\left\{\left|\hat{\mu}_j - \mu_j \right| > \varepsilon_r + L_m \sqrt{d} \mid \mathcal{E}_{1m} \right\} \nonumber\\
    &+ \mathbb{P}\left\{\left|\hat{\mu}_i - \mu_i\right| > \varepsilon_r + L_m \sqrt{d} \mid \mathcal{E}_{1m} \right\} \nonumber\\
    &= 0,
\end{align}
\noindent which means
\begin{equation}
    \mathbb{P}(\mathcal{E}_{2m} \mid \mathcal{E}_{1m}) = 1.
\end{equation}

\subsection{Step 2: Bound the Expected Sample Complexity With the Model Misspecification}\label{subsec_Bound the Expected Sample Complexity When the Model Misspecification is Considered}
To bound the expected sample complexity when the model misspecification is considered, we are trying to find the round where all the arms $i \in G_\varepsilon$ have been added into $G_r$ and the round where all the arms $i \in G_\varepsilon^c$ have been added into $B_r$.

\begin{lemma}\label{claim_mis_2_1}
    For $i \in G_\varepsilon$ and $L_m < \frac{\alpha_\varepsilon}{2 \sqrt{d}}$, if $r \geq\left\lceil\log _2\left(\frac{4}{\varepsilon-\Delta_i-2L_m\sqrt{d}}\right)\right\rceil$, then we have $ \mathbb{E}_{\boldsymbol{\mu}}\left[\mathbbm{1}\left[i \notin G_r\right] \mid \mathcal{E}_{1m}\right] = 0$.
\end{lemma}

\proof\\{\textit{Proof.}}
First, we have that for any $i \in G_\varepsilon$
\begin{align}\label{eq_EvG_claim0_mis}
    \mathbb{E}_{\boldsymbol{\mu}}\left[\mathbbm{1}\left[i \notin G_r\right] \mid \mathcal{E}_{1m}\right] &= \mathbb{E}_{\boldsymbol{\mu}}\left[ \mathbbm{1}\left[\max_{j \in \mathcal{A}_I(r-1)} \hat{\mu}_j - \hat{\mu}_i \geq -2\varepsilon_r + \varepsilon\right] \mid \mathcal{E}_{1m}, i \notin G_m (m = \{1, 2, \ldots, r-1\}) \right] \nonumber \\
    &\leq \mathbb{E}_{\boldsymbol{\mu}}\left[ \mathbbm{1}\left[\max_{j \in \mathcal{A}_I(r-1)} \hat{\mu}_j - \hat{\mu}_i \geq -2\varepsilon_r + \varepsilon\right] \mid \mathcal{E}_{1m}\right].
\end{align}

If $i \in G_{r-1}$, then $i \in G_r$ by definition. Otherwise, if $i \notin G_{r-1}$, then under event $\mathcal{E}_{1m}$, for $i \in G_\varepsilon$ and $r \geq\left\lceil\log _2\left(\frac{4}{\varepsilon-\Delta_i-2L_m\sqrt{d}}\right)\right\rceil$, we have
\begin{equation}
    \max_{j \in \mathcal{A}_I(r-1)} \hat{\mu}_j - \hat{\mu}_i \leq \mu_{\arg\max_{j \in \mathcal{A}_I(r-1)}\hat{\mu}_j} - \mu_i +2^{-r+1} + 2L_m \sqrt{d} \leq \Delta_i+2^{-r+1} + 2L_m \sqrt{d} \leq \varepsilon-2\varepsilon_r,
\end{equation}
which implies that $i \in G_r$ by line \ref{add to Gr} of the algorithm. In particular, under event $\mathcal{E}_{1m}$, if $i \notin G_{r-1}$, for all $i \in G_\varepsilon$ and $r \geq\left\lceil\log _2\left(\frac{4}{\varepsilon-\Delta_i-2L_m\sqrt{d}}\right)\right\rceil$, we have
\begin{equation}\label{expectation_mis}
    \mathbb{E}_{\boldsymbol{\mu}}\left[ \mathbbm{1}\left[\max_{j \in \mathcal{A}_I(r-1)} \hat{\mu}_j - \hat{\mu}_i \leq -2\varepsilon_r + \varepsilon\right] \mid i \notin G_{r-1}, \mathcal{E}_{1m} \right] = 1.
\end{equation}

Deterministically, $\mathbbm{1}\left[i \notin G_r\right] \mathbbm{1}\left[i \in G_{r-1}\right]=0$. Therefore,
\begin{align}
    &\mathbb{E}_{\boldsymbol{\mu}}\left[ \mathbbm{1}\left[\max_{j \in \mathcal{A}_I(r-1)} \hat{\mu}_j - \hat{\mu}_i \geq -2\varepsilon_r + \varepsilon\right] \mid \mathcal{E}_{1m} \right] \nonumber\\
    &= \mathbb{E}_{\boldsymbol{\mu}}\left[ \mathbbm{1}\left[\max_{j \in \mathcal{A}_I(r-1)} \hat{\mu}_j - \hat{\mu}_i \geq -2\varepsilon_r + \varepsilon\right] \mathbbm{1}\left[i \notin G_{r-1}\right] \mid \mathcal{E}_{1m} \right] \nonumber\\
    &+ \mathbb{E}_{\boldsymbol{\mu}}\left[ \mathbbm{1}\left[\max_{j \in \mathcal{A}_I(r-1)} \hat{\mu}_j - \hat{\mu}_i \geq -2\varepsilon_r + \varepsilon\right] \mathbbm{1}\left[i \in G_{r-1}\right] \mid \mathcal{E}_{1m} \right]\nonumber\\
    &= \mathbb{E}_{\boldsymbol{\mu}}\left[ \mathbbm{1}\left[\max_{j \in \mathcal{A}_I(r-1)} \hat{\mu}_j - \hat{\mu}_i \geq -2\varepsilon_r + \varepsilon\right] \mathbbm{1}\left[i \notin G_{r-1}\right] \mid \mathcal{E}_{1m} \right] \nonumber\\
    &=\mathbb{E}_{\boldsymbol{\mu}}\left[\mathbbm{1}\left[\max_{j \in \mathcal{A}_I(r-1)} \hat{\mu}_j - \hat{\mu}_i \geq -2\varepsilon_r + \varepsilon\right] \mathbbm{1}\left[i \notin G_{r-1}\right] \mid i \notin G_{r-1}, \mathcal{E}_{1m}\right] \mathbb{P}\left(i \notin G_{r-1} \mid \mathcal{E}_{1m}\right) \nonumber\\
    &+ \mathbb{E}_{\boldsymbol{\mu}}\left[\mathbbm{1}\left[\max_{j \in \mathcal{A}_I(r-1)} \hat{\mu}_j - \hat{\mu}_i \geq -2\varepsilon_r + \varepsilon\right] \mathbbm{1}\left[i \notin G_{r-1}\right] \mid i \in G_{r-1}, \mathcal{E}_{1m}\right] \mathbb{P}\left(i \in G_{r-1} \mid \mathcal{E}_{1m}\right) \nonumber\\
    &= \mathbb{E}_{\boldsymbol{\mu}}\left[\mathbbm{1}\left[\max_{j \in \mathcal{A}_I(r-1)} \hat{\mu}_j - \hat{\mu}_i \geq -2\varepsilon_r + \varepsilon\right] \mathbbm{1}\left[i \notin G_{r-1}\right] \mid i \notin G_{r-1}, \mathcal{E}_{1m}\right] \mathbb{P}\left(i \notin G_{r-1} \mid \mathcal{E}_{1m}\right) \nonumber\\
    &= \mathbb{E}_{\boldsymbol{\mu}}\left[\mathbbm{1}\left[\max_{j \in \mathcal{A}_I(r-1)} \hat{\mu}_j - \hat{\mu}_i \geq -2\varepsilon_r + \varepsilon\right] \mid i \notin G_{r-1}, \mathcal{E}_{1m}\right] \mathbb{E}_{\boldsymbol{\mu}}\left[\mathbbm{1}\left[i \notin G_{r-1}\right] \mid \mathcal{E}_{1m}\right] \nonumber\\
    &= 0,
\end{align}
where the second line comes from the additivity of expectation. The fourth line follows the deterministic result that $\mathbbm{1}\left[i \notin G_r\right] \mathbbm{1}\left[i \in G_{r-1}\right]=0$. The fifth line is the decomposition based on the conditional expectation. The eighth line comes from the fact that the expectation of the indicator function is simply the probability. The last line follows the result in equation~\eqref{expectation_mis}.

The lemma can thus be concluded together with equation \eqref{eq_EvG_claim0_mis}.

Besides, in the model assumed with perfect linearity, if $i \in G_\varepsilon$, then $\varepsilon - \Delta_i > 0$ always holds. However, when the misspecification is considered, the sign of this term inside the logarithm needs to be tested. To make it positive for every $i \in G_\varepsilon$, we have to make sure that $\alpha_\varepsilon > 2L_m \sqrt{d}$. As the misspecification $L_m \rightarrow \frac{\alpha_\varepsilon}{2 \sqrt{d}}$, the upper bound of the expected sample complexity will increase rapidly, since the deviation brought by misspecification will greatly influence the identification of $\varepsilon$-best arms. Furthermore, when $L_m \geq \frac{\alpha_\varepsilon}{2 \sqrt{d}}$, the sample complexity can no longer be bounded in this way, which is quite intuitive and aligns with previous general insights in Section \ref{subsec_General Insights Related to the Misspecified Linear Bandits in the Pure Exploration Setting}.
\hfill\Halmos
\endproof

\begin{lemma}\label{claim_mis_2_2}
    For $i \in G_\varepsilon^c$ and $L_m < \frac{\beta_\varepsilon}{2 \sqrt{d}}$, if $r \geq\left\lceil\log _2\left(\frac{4}{\Delta_i-\varepsilon - 2L_m\sqrt{d}}\right)\right\rceil$, then we have $ \mathbb{E}_{\boldsymbol{\mu}}\left[\mathbbm{1}\left[i \notin B_r\right] \mid \mathcal{E}_{1m}\right] = 0$.
\end{lemma}


\proof\\{\textit{Proof.}}
First, we have that for any $i \in G_\varepsilon^c$
\begin{align}\label{eq_Ev_claim0_2},
    \mathbb{E}_{\boldsymbol{\mu}}\left[\mathbbm{1}\left[i \notin B_r\right] \mid \mathcal{E}_{1m}\right] &= \mathbb{E}_{\boldsymbol{\mu}}\left[ \mathbbm{1}\left[\max_{j \in \mathcal{A}_I(r-1)} \hat{\mu}_j - \hat{\mu}_i \leq 2\varepsilon_r + \varepsilon\right] \mid \mathcal{E}_{1m}, i \notin B_m (m = \{1, 2, \ldots, r-1\}) \right] \nonumber \\
    &\leq \mathbb{E}_{\boldsymbol{\mu}}\left[ \mathbbm{1}\left[\max_{j \in \mathcal{A}_I(r-1)} \hat{\mu}_j - \hat{\mu}_i \leq 2\varepsilon_r + \varepsilon\right] \mid \mathcal{E}_{1m}\right].
\end{align}




If $i \in B_{r-1}$, then $i \in B_r$ by definition. Otherwise, if $i \notin B_{r-1}$, then under event $\mathcal{E}_{1m}$, for $i \in G_\varepsilon^c$ and $r \geq\left\lceil\log _2\left(\frac{4}{\Delta_i-\varepsilon - 2L_m\sqrt{d}}\right)\right\rceil$, we have
\begin{equation}
    \max_{j \in \mathcal{A}_I(r-1)} \hat{\mu}_j - \hat{\mu}_i \geq \Delta_i-2^{-r+1} - 2L_m\sqrt{d} \geq \varepsilon+2\varepsilon_r,
\end{equation}
which implies that $i \in B_r$ by line \ref{elimination: good 1} of the algorithm. In particular, under event $\mathcal{E}_{1m}$, if $i \notin B_{r-1}$, for all $i \in G_\varepsilon^c$ and $r \geq\left\lceil\log _2\left(\frac{4}{\Delta_i-\varepsilon - 2L_m\sqrt{d}}\right)\right\rceil$, we have
\begin{equation}\label{expectation_mis_op}
    \mathbb{E}_{\boldsymbol{\mu}}\left[ \mathbbm{1}\left[\max_{j \in \mathcal{A}_I(r-1)} \hat{\mu}_j - \hat{\mu}_i > 2\varepsilon_r + \varepsilon\right] \mid i \notin B_{r-1}, \mathcal{E}_{1m} \right] = 1.
\end{equation}

Deterministically, $\mathbbm{1}\left[i \notin B_r\right] \mathbbm{1}\left[i \in B_{r-1}\right]=0$. Therefore,
\begin{align}
    &\mathbb{E}_{\boldsymbol{\mu}}\left[ \mathbbm{1}\left[\max_{j \in \mathcal{A}_I(r-1)} \hat{\mu}_j - \hat{\mu}_i \leq 2\varepsilon_r + \varepsilon\right] \mid \mathcal{E}_{1m} \right] \nonumber\\
    &= \mathbb{E}_{\boldsymbol{\mu}}\left[ \mathbbm{1}\left[\max_{j \in \mathcal{A}_I(r-1)} \hat{\mu}_j - \hat{\mu}_i \leq 2\varepsilon_r + \varepsilon\right] \mathbbm{1}\left[i \notin B_{r-1}\right] \mid \mathcal{E}_{1m} \right] \nonumber\\
    &+ \mathbb{E}_{\boldsymbol{\mu}}\left[ \mathbbm{1}\left[\max_{j \in \mathcal{A}_I(r-1)} \hat{\mu}_j - \hat{\mu}_i \leq 2\varepsilon_r + \varepsilon\right] \mathbbm{1}\left[i \in B_{r-1}\right] \mid \mathcal{E}_{1m} \right]\nonumber\\
    &= \mathbb{E}_{\boldsymbol{\mu}}\left[ \mathbbm{1}\left[\max_{j \in \mathcal{A}_I(r-1)} \hat{\mu}_j - \hat{\mu}_i \leq 2\varepsilon_r + \varepsilon\right] \mathbbm{1}\left[i \notin B_{r-1}\right] \mid \mathcal{E}_{1m} \right] \nonumber\\
    &=\mathbb{E}_{\boldsymbol{\mu}}\left[\mathbbm{1}\left[\max_{j \in \mathcal{A}_I(r-1)} \hat{\mu}_j - \hat{\mu}_i \leq 2\varepsilon_r + \varepsilon\right] \mathbbm{1}\left[i \notin B_{r-1}\right] \mid i \notin B_{r-1}, \mathcal{E}_{1m}\right] \mathbb{P}\left(i \notin B_{r-1} \mid \mathcal{E}_{1m}\right) \nonumber\\
    &+ \mathbb{E}_{\boldsymbol{\mu}}\left[\mathbbm{1}\left[\max_{j \in \mathcal{A}_I(r-1)} \hat{\mu}_j - \hat{\mu}_i \leq 2\varepsilon_r + \varepsilon\right] \mathbbm{1}\left[i \notin B_{r-1}\right] \mid i \in B_{r-1}, \mathcal{E}_{1m}\right] \mathbb{P}\left(i \in B_{r-1} \mid \mathcal{E}_{1m}\right) \nonumber\\
    &= \mathbb{E}_{\boldsymbol{\mu}}\left[\mathbbm{1}\left[\max_{j \in \mathcal{A}_I(r-1)} \hat{\mu}_j - \hat{\mu}_i \leq 2\varepsilon_r + \varepsilon\right] \mathbbm{1}\left[i \notin B_{r-1}\right] \mid i \notin B_{r-1}, \mathcal{E}_{1m}\right] \mathbb{P}\left(i \notin B_{r-1} \mid \mathcal{E}_{1m}\right) \nonumber\\
    &= \mathbb{E}_{\boldsymbol{\mu}}\left[\mathbbm{1}\left[\max_{j \in \mathcal{A}_I(r-1)} \hat{\mu}_j - \hat{\mu}_i \leq 2\varepsilon_r + \varepsilon\right] \mid i \notin B_{r-1}, \mathcal{E}_{1m}\right] \mathbb{E}_{\boldsymbol{\mu}}\left[\mathbbm{1}\left[i \notin B_{r-1}\right] \mid \mathcal{E}_{1m}\right] \nonumber\\
    &= 0,
\end{align}
where the second line comes from the additivity of expectation. The fourth line follows the deterministic result that $\mathbbm{1}\left[i \notin B_r\right] \mathbbm{1}\left[i \in B_{r-1}\right]=0$. The fifth line is the decomposition based on the conditional expectation. The eighth line comes from the fact that the expectation of the indicator function is simply the probability. The last line follows the result in equation~\eqref{expectation_mis_op}.

The lemma can thus be concluded together with equation \eqref{eq_Ev_claim0_2}.

Similarly, we need to consider the sign of the term inside the logarithm. To make $\Delta_i-\varepsilon - 2L_m\sqrt{d}$ positive for every $i \in G_\varepsilon^c$, we have to make sure that $\beta_\varepsilon > 2L_m \sqrt{d}$.
\hfill\Halmos
\endproof

\begin{lemma}\label{claim_mis_2_3}
    For $L_m < \min \left\{\frac{\alpha_\varepsilon}{2 \sqrt{d}}, \frac{\beta_\varepsilon}{2 \sqrt{d}} \right\}$, $R_{\text{upper}}^\prime = \max \left\{ \left\lceil \log_2 \frac{4}{\alpha_\varepsilon - 2L_m \sqrt{d}} \right\rceil, \left\lceil \log_2 \frac{4}{\beta_\varepsilon - 2L_m \sqrt{d}} \right\rceil \right\}$ is the round where all the classifications have been finished and the answer is returned when the model misspecification is considered.
\end{lemma}

\proof\\{\textit{Proof.}}
Combining the result in Lemma \ref{claim_mis_2_1} and Lemma \ref{claim_mis_2_2} of this section, we can define an unknown round $R_{m}$ to assist the derivation of our upper bound with the model misspecification. Specifically, for $i \in G_\varepsilon^c$ and $r \geq\left\lceil\log _2\left(\frac{4}{\Delta_i-\varepsilon - 2L_m \sqrt{d}}\right)\right\rceil$, we have $\mathbb{E}_{\boldsymbol{\mu}}\left[\mathbbm{1}\left[i \notin B_r\right] \mid \mathcal{E}_{1m}\right] = 0$; for $i \in G_\varepsilon$ and $r \geq\left\lceil\log _2\left(\frac{4}{\varepsilon-\Delta_i - 2L_m \sqrt{d}}\right)\right\rceil$, we have $\mathbb{E}_{\boldsymbol{\mu}}\left[\mathbbm{1}\left[i \notin G_r\right] \mid \mathcal{E}_{1m}\right] = 0$.

Considering that $\alpha_\varepsilon = \min_{i \in G_\varepsilon}(\varepsilon - \Delta_i)$ and $\beta_\varepsilon = \min_{i \in G_\varepsilon^c}(\Delta_i - \varepsilon)$, for the round $r \geq R_{\text{upper}}^\prime$, all the arms have been added into set $G_r$ or $B_r$, representing the termination of the algorithm.
\hfill\Halmos
\endproof

\begin{lemma}\label{claim_mis_2_4}
    For the expected sample complexity with high probability event $\mathcal{E}_{1m}$, we have
\begin{align}
    \mathbb{E}_{\boldsymbol{\mu}}\left[T_{G_{\textup{mis}}} \mid \mathcal{E}_{1m}\right] &\leq c \max\left\{ \frac{256d}{(\alpha_\varepsilon - 2L_m \sqrt{d})^2} \log \left( \frac{2K}{\delta} \log_2 \frac{16}{\alpha_\varepsilon - 2L_m \sqrt{d}} \right), \right. \nonumber\\
    &\left. \quad \frac{256d}{(\beta_\varepsilon - 2L_m \sqrt{d})^2} \log \left( \frac{2K}{\delta} \log_2 \frac{16}{\beta_\varepsilon - 2L_m \sqrt{d}} \right) \right\} + \frac{d(d+1)}{2} R_{\text{upper}}^\prime, \label{eq_mis_sample_complexity}
\end{align}
where $c$ is a universal constant and $R_{\text{upper}}^\prime = \max \left\{ \left\lceil \log_2 \frac{4}{\alpha_\varepsilon - 2L_m \sqrt{d}} \right\rceil, \left\lceil \log_2 \frac{4}{\beta_\varepsilon - 2L_m \sqrt{d}} \right\rceil \right\}$ is the round where all the classifications have been finished and the answer is returned with the model misspecification.
\end{lemma}

\proof\\{\textit{Proof.}}
We can have another decomposition of $T$ in equation (\ref{equation: decomposition of T}). These expectations are conditioned on the high probability event $\mathcal{E}_{1m}$, given by
\begin{align}
    \mathbb{E}_{\boldsymbol{\mu}}\left[T_{G_{\textup{mis}}} \mid \mathcal{E}_{1m}\right]
    &\leq \sum_{r=1}^{\infty} \mathbb{E}_{\boldsymbol{\mu}}\left[\mathbbm{1}\left[G_r \cup B_r \neq[K]\right] \mid \mathcal{E}_{1m}\right] \sum_{\bm{a} \in \mathcal{A}(r-1)}T_r(\bm{a}) \nonumber\\
    &\leq \sum_{r=1}^{R_{\text{upper}}^\prime} \left( d 2^{2r+1} \log \left(\frac{2K r(r+1)}{\delta}\right)+\frac{d(d+1)}{2}\right) \nonumber\\
    &\leq \frac{d(d+1)}{2} R_{\text{upper}}^\prime + 2d\log \left(\frac{2K}{\delta}\right) \sum_{r=1}^{{R_{\text{upper}}^\prime}} 2^{2r} + 4d \sum_{r=1}^{{R_{\text{upper}}^\prime}} 2^{2r} \log (r+1) \nonumber\\
    &\leq 4\log \left[ \frac{2K}{\delta}\left( R_{\text{upper}}^\prime + 1 \right) \right] \sum_{r=1}^{{R_{\text{upper}}^\prime}} d 2^{2r} + \frac{d(d+1)}{2} R_{\text{upper}}^\prime \label{middle result_refined_mis} \nonumber\\
    &\leq c \max\left\{ \frac{256d}{(\alpha_\varepsilon - 2L_m \sqrt{d})^2} \log \left( \frac{2K}{\delta} \log_2 \frac{16}{\alpha_\varepsilon - 2L_m \sqrt{d}} \right), \right. \nonumber\\
    &\left. \quad \frac{256d}{(\beta_\varepsilon - 2L_m \sqrt{d})^2} \log \left( \frac{2K}{\delta} \log_2 \frac{16}{\beta_\varepsilon - 2L_m \sqrt{d}} \right) \right\} + \frac{d(d+1)}{2} R_{\text{upper}}^\prime,
    \label{another angle: decomposition of T - 1_mis_1}
\end{align}
where $c$ is a universal constant.
\hfill\Halmos
\endproof

\section{Proof of Theorem \ref{upper bound: Algorithm 1 G_mis_2}}\label{sec_Proof of upper bound: Algorithm 1 G_mis_2}

\subsection{Step 1: Orthogonal Parameterization}\label{subsec_Orthogonal Parameterization}
If we ignore the difference in confidence radius caused by the misspecification and make no improvements to our algorithm, then the supplemented analyses in Section \ref{subsec_Rearrange the Clean Event With Misspecification} show that the algorithm will suffer the expansion of sample complexity in the form of equation~\eqref{eq_mis_sample_complexity}. Whereas, in this section, we provide an adjusted version of LinFACT based on the idea of orthogonal parameterization to enhance its theoretical performance. 

\subsection{Step 2: Confidence Radius With Misspecification}\label{subsec_Confidence Radius With Misspecification}
In equation \eqref{equation: confidence region_mis}, we can find that the first term, which is brought by the unknown misspecification, is unavoidable without any prior knowledge. However, instead of focusing on the distance between the true vector $\bm{\theta}$ and the estimator $\hat{\bm{\theta}}_t$, we are now comparing the orthogonal vector $\bm{\theta}_t$ and the estimator $\hat{\bm{\theta}}_t$ in the direction $\bm{x} \in \mathbb{R}^d$.



Then based on the introduction of the empirically optimal vector $\hat{\bm{\mu}}_{o}(r)$ with $(\hat{\bm{\theta}}_o(r), \hat{\bm{\Delta}}_{mo}(r))$ being the solution, and the orthogonal parameterization $(\bm{\theta}_r, \bm{\Delta}_m(r))$, we could provide the confidence radius for all arms by the following decomposition. For any $i \in \mathcal{A}_I(r-1)$, let $\hat{\mu}_i(r) = \hat{\mu}_{oi}(r)$ be the value of optimal estimator vector on one of $K$ arms in round $r$, we have 
\begin{align}
    \left \lvert \hat{\mu}_i(r) - \mu_i \right\rvert &= \left \lvert\left\langle \hat{\bm{\theta}}_o(r) - \bm{\theta}, \bm{a}_i \right\rangle + \hat{\bm{\Delta}}_{mo}(r) - \Delta_{mi} \right\rvert \nonumber\\
    &\leq \left \lvert\left\langle \hat{\bm{\theta}}_o(r) - \bm{\theta}_r, \bm{a}_i \right\rangle \right\rvert + \left \lvert\left\langle \bm{\theta}_r - \bm{\theta}, \bm{a}_i \right\rangle \right\rvert + \left \lvert \hat{\bm{\Delta}}_{mo}(r) - \Delta_{mi} \right\rvert,
\end{align}
where the third term is simply bounded by the definition as $\left \lvert \hat{\bm{\Delta}}_{mo}(r) - \Delta_{mi} \right\rvert \leq 2L_m$ and the first two terms are bounded with the help of the following auxiliary lemmas.

\begin{lemma}\label{lemma_para_distance}
Let $r$ be any round such that $\bm{V}_r$ is invertible. Consider the orthogonal parameterization $(\bm{\theta}_r, \bm{\Delta}_m(r)$ for $\bm{\mu} = \bm{\Psi}\bm{\theta} + \bm{\Delta}_m$ with $\|\bm{\Delta_m}\|_\infty \leq L_m$. Then
\begin{equation}
    \|\bm{\theta}_r - \bm{\theta}\|_{\bm{V}_r} \leq L_m \sqrt{T_r},
\end{equation}
where $T_r$ is defined in equation \eqref{equation_phase budget G_mis_ortho}.
\end{lemma}
\proof\\{\textit{Proof.}}
We use the expression $\bm{\theta}_r = \bm{\theta} + \bm{V}_r^{-1} \bm{\Psi}^\top \bm{D}_{\bm{N}_r} \bm{\Delta}_m$ derived in the last paragraph, let $\bm{P}_{\bm{N}_r} = \bm{\Psi}_{\bm{N}_r} \left( \bm{\Psi}_{\bm{N}_r}^\top \bm{\Psi}_{\bm{N}_r} \right)^\dagger \bm{\Psi}_{\bm{N}_r}^\top$ be a projection, we have
\begin{align}
    \|\bm{\theta}_r - \bm{\theta}\|_{\bm{V}_r} &= \|\bm{V}_r^{-1} \bm{\Psi}^\top \bm{D}_{\bm{N}_r} \bm{\Delta_m}\|_{\bm{V}_r} \nonumber\\ 
    &= \sqrt{\bm{\Delta_m}^\top \bm{D}_{\bm{N}_r} \bm{\Psi} \bm{V}_r^{-1} \bm{\Psi}^\top \bm{D}_{\bm{N}_r} \bm{\Delta_m}} \nonumber\\
    &= \left\|\bm{D}_{\bm{N}_r}^{1/2} \bm{\Delta_m}\right\|_{\bm{P}_{\bm{N}_r}} \nonumber\\
    &\leq \left\|\bm{D}_{\bm{N}_r}^{1/2} \bm{\Delta_m}\right\| \nonumber\\
    &= \left\|\bm{\Delta_m}\right\|_{D_{\bm{N}_r}} \nonumber\\
    &\leq L_m \sqrt{T_r} .
\end{align}
\hfill\Halmos
\endproof

\begin{lemma}\label{lemma_optimi_distance}
\citep{reda2021dealing} (Lemma 10). Let $\hat{\bm{\mu}}_o(r) = \bm{\Psi}\hat{\bm{\theta}}_o(r) + \hat{\bm{\Delta}}_{mo}(r)$ in round $r$, where $(\hat{\bm{\theta}}_o(r), \hat{\bm{\Delta}}_{mo}(r))$ are the solution of \eqref{eq_optimization_esti}.Then the following relationship holds.
\begin{equation}
    \left\| \hat{\bm{\theta}}_o(r) - \bm{\theta}_r \right\|_{\bm{V}_r}^2 \leq \left\| \hat{\bm{\theta}}_r - \bm{\theta}_r \right\|_{\bm{V}_r}^2.
\end{equation}
\end{lemma}

\begin{lemma}\label{lemma_confidence_C_epsilon}
\citep{lattimore2020bandit} (Section 20). Let $\delta \in (0, 1)$. Then, with a probability of at least $1-\delta$, it holds that for all $t \in \mathbb{N}$,
\begin{equation}\label{log6_1}
     \left\| \bm{\hat{\theta}}_t - \bm{\theta} \right\|_{V_t} < 2 \sqrt{2 \left( d \log(6) + \log \left( \frac{1}{\delta} \right) \right)}.
\end{equation}

In our setting, when the misspecification is considered, we can show that in any round $r \in \mathbb{N}$, with a probability of at least $1-\delta$, it holds that
\begin{equation}\label{log6_2}
     \left\| \bm{\hat{\theta}}_r - \bm{\theta}_r \right\|_{V_t} < 2 \sqrt{2 \left( d \log(6) + \log \left( \frac{1}{\delta} \right) \right)}.
\end{equation}
\end{lemma}

The equation \eqref{log6_1} cannot be directly applied in our setting because of the deviation term brought by the model misspecification, which is shown in the equation \eqref{equation: confidence region_mis}. Whereas by utilizing the orthogonal parameterization, we can have the following critical observation
\begin{equation}\label{eq_self-normalized}
    \Vert \hat{\bm{\theta}}_r - \bm{\theta}_r \Vert_{\bm{V}_r}^2 = \Vert \bm{V}_r^{-1} \Psi^\top S_r \Vert_{\bm{V}_r}^2 = \Vert \Psi^\top S_r \Vert_{\bm{V}_r^{-1}}^2,
\end{equation}
where $S_r = \sum_{s=1}^{T_r} X_{A_s} - \mu_{A_s}$ is the standard self-normalized quantity from the linear bandit literature, meaning that all the ideas, including different concentration inequalities, can be applied in the misspecified setting directly. This observation leads to the result in equation \eqref{log6_2}, which serves as the basis for the design of the sampling budget and the theoretical analyses.

Hence, together with Lemma \ref{lemma_para_distance}, Lemma \ref{lemma_optimi_distance}, and Lemma \ref{lemma_confidence_C_epsilon}, the distance $\left \lvert \tilde{\mu}_i(r) - \mu_i \right\rvert$ can be bounded as the following statement with a probability of as least $1-\delta/{Kr(r+1)}$, we have
\begin{align}
    \left \lvert \tilde{\mu}_i(r) - \mu_i \right\rvert &\leq \left \lvert\left\langle \hat{\bm{\theta}}_o(r) - \bm{\theta}_r, \bm{a}_i \right\rangle \right\rvert + \left \lvert\left\langle \bm{\theta}_r - \bm{\theta}, \bm{a}_i \right\rangle \right\rvert + \left \lvert \hat{\bm{\Delta}}_{mo}(r) - \Delta_{mi} \right\rvert \nonumber\\
    &\leq \left \lVert \hat{\bm{\theta}}_o(r) - \bm{\theta}_r \right\rVert_{\bm{V}_r} \left \lVert \bm{a}_i \right\rVert_{\bm{V}_r^{-1}} + \left \lVert \bm{\theta}_r - \bm{\theta} \right\rVert_{\bm{V}_r} \left \lVert \bm{a}_i \right\rVert_{\bm{V}_r^{-1}} + 2L_m \nonumber\\
    &\leq \left \lVert \hat{\bm{\theta}}_r - \bm{\theta}_r \right\rVert_{\bm{V}_r} \left \lVert \bm{a}_i \right\rVert_{\bm{V}_r^{-1}} + \left \lVert \bm{\theta}_r - \bm{\theta} \right\rVert_{\bm{V}_r} \left \lVert \bm{a}_i \right\rVert_{\bm{V}_r^{-1}} + 2L_m \nonumber\\
    &\leq \left( 2 \sqrt{2 \left( d \log(6) + \log \left( \frac{Kr(r+1)}{\delta} \right) \right)} + L_m \sqrt{T_r}  \right) \sqrt{\frac{d}{T_r}} + 2L_m \nonumber\\
    &\leq \varepsilon_r + (\sqrt{d} + 2) L_m.
\end{align}

Then, we can be a new clean event $\mathcal{E}_{1m}^\prime$ as
\begin{equation}\label{clean event mis_2}
    \mathcal{E}_{1m}^\prime = \left\{ \bigcap_{i \in \mathcal{A}_I(r-1)} \bigcap_{r \in \mathbb{N}} \left| \hat{\mu}_i(r) - \mu_i \right| \leq \varepsilon_r + (\sqrt{d} + 2) L_m \right\},
\end{equation}
which is quite similar to the event defined in \eqref{clean event mis_1} and we can obtain all the results in Section \ref{sec_Proof of upper bound: Algorithm 1 G_mis_1}.

\subsection{Step 3: Bound the Expected Sample Complexity When the Model Misspecification is Considered and the Orthogonal Parameterization is Applied}\label{subsec_Bound the Expected Sample Complexity When the Model Misspecification is Considered and the Orthogonal Parameterization is Applied}

\begin{lemma}\label{claim_or_3_1}
   For $i \in G_\varepsilon$ and $L_m < \frac{\alpha_\varepsilon}{2 \left( \sqrt{d} + 2 \right)}$, if $r \geq\left\lceil\log _2\left(\frac{4}{\varepsilon-\Delta_i-2L_m\left( \sqrt{d} + 2 \right)}\right)\right\rceil$, then we have $ \mathbb{E}_{\boldsymbol{\mu}}\left[\mathbbm{1}\left[i \notin G_r\right] \mid \mathcal{E}_{1m}^\prime\right] = 0$. 
\end{lemma}

\begin{lemma}\label{claim_or_3_2}
   For $i \in G_\varepsilon^c$ and $L_m < \frac{\beta_\varepsilon}{2 \left( \sqrt{d} + 2 \right)}$, if $r \geq\left\lceil\log _2\left(\frac{4}{\Delta_i-\varepsilon - 2L_m\left( \sqrt{d} + 2 \right)}\right)\right\rceil$, then we have $ \mathbb{E}_{\boldsymbol{\mu}}\left[\mathbbm{1}\left[i \notin B_r\right] \mid \mathcal{E}_{1m}^\prime\right] = 0$. 
\end{lemma}

\begin{lemma}\label{claim_or_3_3}
   For the misspecification magnitude $L_m < \min \left\{\frac{\alpha_\varepsilon}{2 \left( \sqrt{d} + 2 \right)}, \frac{\beta_\varepsilon}{2 \left( \sqrt{d} + 2 \right)} \right\}$, $R_{\text{upper}}^{\prime\prime} = \max \left\{ \left\lceil \log_2 \frac{4}{\alpha_\varepsilon - 2L_m \left( \sqrt{d} + 2 \right)} \right\rceil, \left\lceil \log_2 \frac{4}{\beta_\varepsilon - 2L_m \left( \sqrt{d} + 2 \right)} \right\rceil \right\}$ is the round where all the classifications have been finished and the answer is returned when the model misspecification is considered with an estimation method based on the orthogonal parameterization.
\end{lemma}

The proofs for Lemma \ref{claim_or_3_1}, Lemma \ref{claim_or_3_2}, and Lemma \ref{claim_or_3_3} are quite similar to those in Section \ref{sec_Proof of upper bound: Algorithm 1 G_mis_1}, so they will not be elaborated upon further.

\begin{lemma}\label{claim_or_3_4}
   For the expected sample complexity with high probability event $\mathcal{E}_{1m}^\prime$, we have
\begin{align}\label{eq_mis_sample_complexity_orthogonal}
    \mathbb{E}_{\boldsymbol{\mu}}\left[T_{G_{\textup{op}}} \mid \mathcal{E}_{1m}^\prime\right] &\leq c \max\bigg\{ \frac{256d}{(\alpha_\varepsilon - 2L_m ( \sqrt{d} + 2 ))^2} \log \left( \frac{K 6^d}{\delta} \log_2 \frac{8}{\alpha_\varepsilon - 2L_m ( \sqrt{d} + 2 )} \right),\nonumber\\
    &\frac{256d}{(\beta_\varepsilon - 2L_m ( \sqrt{d} + 2 ))^2} \log \left( \frac{K 6^d}{\delta} \log_2 \frac{8}{\beta_\varepsilon - 2L_m ( \sqrt{d} + 2 )} \right) \bigg\} + \frac{d(d+1)}{2} R_{\text{upper}}^{\prime\prime},
\end{align}
where $c$ is a universal constant and $R_{\text{upper}}^{\prime\prime} = \max \left\{ \left\lceil \log_2 \frac{4}{\alpha_\varepsilon - 2L_m \left( \sqrt{d} + 2 \right)} \right\rceil, \left\lceil \log_2 \frac{4}{\beta_\varepsilon - 2L_m \left( \sqrt{d} + 2 \right)} \right\rceil \right\}$ is the round where all the classifications have been finished and the answer is returned with the model misspecification and the orthogonal parameterization.
\end{lemma}

\proof\\{\textit{Proof.}}
We can have another decomposition of $T$ in equation (\ref{equation: decomposition of T}). These expectations are conditioned on the high probability event $\mathcal{E}_{1m}^\prime$, given by
\begin{align}
    \mathbb{E}_{\boldsymbol{\mu}}\left[T_{G_{\textup{op}}} \mid \mathcal{E}_{1m}^\prime \right]
    &\leq \sum_{r=1}^{\infty} \mathbb{E}_{\boldsymbol{\mu}}\left[\mathbbm{1}\left[G_r \cup B_r \neq[K]\right] \mid \mathcal{E}_{1m}^\prime \right] \sum_{\bm{a} \in \mathcal{A}(r-1)}T_r(\bm{a}) \nonumber\\
    &\leq \sum_{r=1}^{R_{\text{upper}}^{\prime\prime}} \left( d2^{2r+3} \left( d\log(6) + \log\left(\dfrac{Kr(r + 1)}{\delta}\right) \right) + \frac{d(d+1)}{2}\right) \nonumber\\
    &\leq \frac{d(d+1)}{2} R_{\text{upper}}^{\prime\prime} + 8d\log \left(\frac{K 6^d}{\delta}\right) \sum_{r=1}^{{R_{\text{upper}}^{\prime\prime}}} 2^{2r} + 16d \sum_{r=1}^{{R_{\text{upper}}^{\prime\prime}}} 2^{2r} \log (r+1) \nonumber\\
    &\leq 16\log \left[ \frac{K 6^d}{\delta}\left( R_{\text{upper}}^{\prime\prime} + 1 \right) \right] \sum_{r=1}^{{R_{\text{upper}}^{\prime\prime}}} d 2^{2r} + \frac{d(d+1)}{2} R_{\text{upper}}^{\prime\prime} \label{middle result_refined_mis_ortho} \\
    &\leq c \max\bigg\{ \frac{256d}{(\alpha_\varepsilon - 2L_m ( \sqrt{d} + 2 ))^2} \log \left( \frac{K 6^d}{\delta} \log_2 \frac{16}{\alpha_\varepsilon - 2L_m ( \sqrt{d} + 2 )} \right),\nonumber\\
    &\frac{256d}{(\beta_\varepsilon - 2L_m ( \sqrt{d} + 2 ))^2} \log \left( \frac{K 6^d}{\delta} \log_2 \frac{16}{\beta_\varepsilon - 2L_m ( \sqrt{d} + 2 )} \right) \bigg\} + \frac{d(d+1)}{2} R_{\text{upper}}^{\prime\prime},\label{another angle: decomposition of T - 1_mis_ortho}
\end{align}
where $c$ is a universal constant.
\hfill\Halmos
\endproof


\section{Proof of Theorem \ref{upper bound: Algorithm 1 GLM}}\label{proof of theorem: upper bound of LinFACT GLM}
\subsection{Step 1: Rearrange the Clean Event With GLM}\label{subsubsec_Step 1: Rearrange the Clean Event With GLM}
Following the derivation method in Section \ref{refined version}, we can provide the upper bound of LinFACT with the G-optimal design when the linear model is extended to the GLM. The core of the proof is to similarly define the round when all the classifications have been down with GLM. To achieve this, we will reconstruct the anytime confidence radius for arms in each round $r$ and the high probability event during the whole process of our algorithm.

Let $\check{\bm{V}}_r = \sum_{s=1}^{T_r} \dot{\mu}_{\text{link}}(\bm{a}^\top\check{\bm{\theta}}_r)\bm{a}_s\bm{a}_s^\top$, where $\check{\bm{\theta}}_r$ is some convex combination of true parameter $\bm{\theta}$ and parameter $\hat{\bm{\theta}}_r$ based on maximum likelihood estimation (MLE). It can be checked that the unweighted matrix $\bm{V}_r = \sum_{s=1}^{T_r} \bm{a}^\top\check{\bm{\theta}}_r)\bm{a}_s\bm{a}_s^\top$ in the standard linear model is the special case of this newly defined matrix $\check{\bm{V}}_r$ when the inverse link function $\mu_{\text{link}}(x) = x$.

For each arm $i$, we define the following auxiliary vector.
\begin{equation}
    W_i = \left( W_{i,1},W_{i,2},\ldots,W_{i,T_r} \right) = \bm{a}_i^\top \check{\bm{V}}_r^{-1} \left( \bm{a}_{A_1}, \bm{a}_{A_2},\ldots,\bm{a}_{A_{T_r}} \right) \in \mathbb{R}^{T_r},
\end{equation}
and thus we have
\begin{equation}
    \Vert W_i \Vert_2^2 = W_i W_i^\top = \bm{a}_i^\top \check{\bm{V}}_r^{-1} \bm{V}_r \check{\bm{V}}_r^{-1}  \bm{a}_i.
\end{equation}

To give the confidence radius when the GLM is considered, we have the following statement for any arm $i \in \mathcal{A}_I(r-1)$ in round $r$.
\begin{align}
    \left \lvert \hat{\mu}_i(r) - \mu_i \right\rvert &= \left \lvert \bm{a}_i^\top (\hat{\bm{\theta}}_r - \bm{\theta}) \right\rvert \nonumber\\
    &= \left \lvert \bm{a}_i^\top \check{\bm{V}}_r^{-1} \sum_{s=1}^{T_r} \bm{a}_{A_s} \eta_s \right\rvert \nonumber\\
    &= \left \lvert \sum_{s=1}^{T_r} W_{i,s}\eta_{A_s} \right\rvert,
\end{align}
where the second equality is established with Lemma 1 proposed by \cite{kveton2023randomized} and $T_r$, which is defined in equation \eqref{equation_phase budget G_GLM}, is the adjusted sampling budget in each round $r$ for the GLM. Since $(\eta_{A_s})_{s \in T_r}$ are independent, mean zero, 1-sub-Gaussian random variables, then $\sum_{s=1}^{T_r} W_{i,s}\eta_{A_s}$ is a $\Vert W_i \Vert_2$-sub-Gaussian variable for each arm $i$, then we have
\begin{align}\label{eq_GLM_prob_1}
    \mathbb{P}\left( \left \lvert \hat{\mu}_i(r) - \mu_i \right\rvert > \varepsilon_r \right) &\leq 2\text{exp} \left( -\frac{\varepsilon_r^2}{2\Vert W_i \Vert_2^2} \right).
\end{align}

Since $\check{\bm{\theta}}_r$ is not known in the process, we need to find another way to represent this term. By assumption, we know $\dot{\mu}_{\text{link}} \geq c_{\min}$ for some $c_{\min} \in \mathbb{R}^+$ and for all $i \in \mathcal{A}_I(r-1)$. Therefore $c_{\min}^{-1} \bm{V}_r^{-1} \succeq \check{\bm{V}}_r^{-1}$ by definition of $\check{\bm{V}}_r$, and then we have
\begin{align}
    \Vert W_i \Vert_2^2 &= \bm{a}_i^\top \check{\bm{V}}_r^{-1} \bm{V}_r \check{\bm{V}}_r^{-1} \bm{a}_i \nonumber\\
    &\leq \bm{a}_i^\top c_{\min}^{-1} \bm{V}_r^{-1} \bm{V}_r c_{\min}^{-1} \bm{V}_r^{-1} \bm{a}_i \nonumber\\
    &=c_{\min}^{-2} \Vert \bm{a}_i \Vert_{\bm{V}_r^{-1}}^2.
\end{align}

Furthermore, if G-optimal design is considered, we have $\Vert \bm{a}_i \Vert_{\bm{V}_r^{-1}}^2 \leq \frac{d}{T_r}$. Together with equation \eqref{eq_GLM_prob_1}, we have
\begin{align}\label{eq_GLM_prob_2}
    \mathbb{P}\left( \left \lvert \hat{\mu}_i(r) - \mu_i \right\rvert > \varepsilon_r \right) &\leq 2\text{exp} \left( -\frac{\varepsilon_r^2}{2c_{\min}^{-2} \Vert \bm{a}_i \Vert_{\bm{V}_r^{-1}}^2} \right) \nonumber\\
    &\leq 2\text{exp} \left( -\frac{\varepsilon_r^2 c_{\min}^{2}}{2 d} T_r \right).
\end{align}

Finally, considering the definition of $T_r$ in equation \eqref{equation_phase budget G_GLM}, with a probability of as least $1-\delta/{Kr(r+1)}$, we have
\begin{equation}
    \left \lvert \hat{\mu}_i(r) - \mu_i \right\rvert > \varepsilon_r.
\end{equation}

Thus with the standard result of the G-optimal design, we still have
\begin{equation}\label{eq_C_G_2}
    C_{\delta/K}(r) \coloneqq \varepsilon_r,
\end{equation}
with which the events $\mathcal{E}_1$ and $\mathcal{E}_2$ in Section \ref{algorithm2: preliminary} hold with a probability of at least $1-\delta$. 

\subsection{Step 2: Bound the Expected Sample Complexity When the GLM is Assumed}\label{subsubsec_Bound the Expected Sample Complexity When the GLM is Assumed}

\begin{lemma}\label{claim_glm_2_1}
   For $i \in G_\varepsilon$, if $r \geq\left\lceil\log _2\left(\frac{4}{\varepsilon-\Delta_i}\right)\right\rceil$, then we have $ \mathbb{E}_{\boldsymbol{\mu}}\left[\mathbbm{1}\left[i \notin G_r\right] \mid \mathcal{E}_{1}\right] = 0$. 
\end{lemma}

\begin{lemma}\label{claim_glm_2_2}
   For $i \in G_\varepsilon^c$, if $r \geq\left\lceil\log _2\left(\frac{4}{\Delta_i-\varepsilon}\right)\right\rceil$, then we have $ \mathbb{E}_{\boldsymbol{\mu}}\left[\mathbbm{1}\left[i \notin B_r\right] \mid \mathcal{E}_{1}\right] = 0$. 
\end{lemma}

\begin{lemma}\label{claim_glm_2_3}
   $R_{\textup{GLM}} = R_{\textup{upper}} = \max \left\{ \left\lceil \log_2 \frac{4}{\alpha_\varepsilon} \right\rceil, \left\lceil \log_2 \frac{4}{\beta_\varepsilon} \right\rceil \right\}$ is the round where all the classifications have been finished and the answer is returned when the GLM is considered. 
\end{lemma}

The proofs for Lemma \ref{claim_glm_2_1}, Lemma \ref{claim_glm_2_2}, and Lemma \ref{claim_glm_2_3} are quite similar to those in Section \ref{refined version}, so they will not be elaborated upon further.

\begin{lemma}\label{claim_glm_2_4}
   For the expected sample complexity with high probability event $\mathcal{E}_{1}$, we have
\begin{align}\label{eq_GLM_sample_complexity}
    \mathbb{E}_{\boldsymbol{\mu}}\left[T_{\textup{GLM}} \mid \mathcal{E}_{1}\right] &\leq c \max\left\{ \frac{256d}{\alpha_\varepsilon^2c_{\min}^{2}} \log \left( \frac{2K}{\delta} \log_2 \frac{16}{\alpha_\varepsilon} \right), \frac{256d}{\beta_\varepsilon^2c_{\min}^{2}} \log \left( \frac{2K}{\delta} \log_2 \frac{16}{\beta_\varepsilon} \right) \right\} + \frac{d(d+1)}{2} R_{\textup{GLM}},
\end{align}
where $c$ is a universal constant, $c_{\min}$ is the known constant controlling the first-order derivative of the inverse link function, and $R_{\textup{GLM}} = R_{\textup{upper}} = \max \left\{ \left\lceil \log_2 \frac{4}{\alpha_\varepsilon} \right\rceil, \left\lceil \log_2 \frac{4}{\beta_\varepsilon} \right\rceil \right\}$ is the round where all the classifications have been finished and the answer is returned with the GLM assumption. 
\end{lemma}

\proof\\{\textit{Proof.}}
We can have another decomposition of $T$ in equation (\ref{equation: decomposition of T}). These expectations are conditioned on the high probability event $\mathcal{E}_{1}$, given by
\begin{align}
    \mathbb{E}_{\boldsymbol{\mu}}\left[T_{\textup{GLM}} \mid \mathcal{E}_{1} \right]
    &\leq \sum_{r=1}^{\infty} \mathbb{E}_{\boldsymbol{\mu}}\left[\mathbbm{1}\left[G_r \cup B_r \neq[K]\right] \mid \mathcal{E}_{1} \right] \sum_{\bm{a} \in \mathcal{A}(r-1)}T_r(\bm{a}) \nonumber\\
    &\leq \sum_{r=1}^{R_{\textup{GLM}}} \left( \dfrac{d2^{2r+1}}{c_{\min}^2} \log\left(\dfrac{2Kr(r + 1)}{\delta}\right) + \frac{d(d+1)}{2} \right) \nonumber\\
    &\leq \frac{d(d+1)}{2} R_{\textup{GLM}} + 2c_{\min}^{-2}d\log \left(\frac{2K}{\delta}\right) \sum_{r=1}^{{R_{\textup{GLM}}}} 2^{2r} + 4c_{\min}^{-2}d \sum_{r=1}^{{R_{\textup{GLM}}}} 2^{2r} \log (r+1) \nonumber\\
    &\leq 4c_{\min}^{-2}\log \left[ \frac{2K}{\delta}\left( R_{\textup{GLM}} + 1 \right) \right] \sum_{r=1}^{{R_{\textup{GLM}}}} d 2^{2r} + \frac{d(d+1)}{2} R_{\textup{GLM}} \label{middle result_refined_GLM} \\
    &\leq c \max\left\{ \frac{256d}{\alpha_\varepsilon^2c_{\min}^{2}} \log \left( \frac{2K}{\delta} \log_2 \frac{16}{\alpha_\varepsilon} \right), \frac{256d}{\beta_\varepsilon^2c_{\min}^{2}} \log \left( \frac{2K}{\delta} \log_2 \frac{16}{\beta_\varepsilon} \right) \right\} + \frac{d(d+1)}{2} R_{\textup{GLM}},\label{another angle: decomposition of T - 1_GLM}
\end{align}
where $c$ is a universal constant.
\hfill\Halmos
\endproof

\section{Modification to Fixed-Batch Sampling}
As previously noted, drug development is a key application where identifying all $\varepsilon$-best arms is a common objective. In practice, drug experiments are typically conducted in fixed-size batches. However, most existing algorithms, such as BayesGap, KGCB, and LinGIFA, employ a sequential sampling strategy, selecting one arm at a time. At each step, these algorithms identify the arm that is expected to be most informative for discovering, for example, the best arm, an $\varepsilon$-best arm, the top-$m$ arms, or arms above a threshold, and then sample that arm. The result is observed, and the process is repeated.

This one-by-one sampling approach does not align with the batch-based nature of real-world drug experiments. In contrast, our LinFACT algorithm can be readily adapted to batch sampling, making it more suitable for practical deployment in drug discovery.

The main idea is to slightly modify the sampling rule in algorithm \ref{alg_LinFACT_sampling} or algorithm \ref{alg_LinFACT_sampling_XY}. Instead of directly computing sample budget $T_r(a)$ for arm $a$ at round $r$ according to the requirement for confidence increment at round $r$, we fix the total budget $\sum_{a}T_r(a)$ for round $r$ as a constant $\chi$, where $m$ is the batch size in real drug development setting. Then, we round the continuous allocation $m\pi_r(a)$ to a discrete allocation $T_r(a)$ for each arm $a\in\mathcal{A}_I(r-1)$. 

To be specific, we replace equation~\eqref{equation_phase budget G} with 
\begin{equation}
    T_r(a) = \text{ROUND}(\pi_r, m\chi),
\end{equation}
where ROUND can be the rounding algorithm given by \cite{allenzhu2017nearoptimal} with a $(1+\epsilon)$ approximation, or an efficient rounding given by \cite{pukelsheim2006optimal} in Chapter 12.

In real-world applications, observed data can be utilized incrementally, allowing regression updates after each experimental batch. This practical approach intuitively leads to increasing confidence in the parameter estimates as more data accumulate. However, such incremental confidence gain is difficult to capture theoretically, as previously discussed. For theoretical guarantees, our algorithm instead performs a fresh linear regression at the end of each round $r = 1, 2, \ldots$, using only the data collected during that round. This design ensures compatibility with the theory of phase-based optimal design. 

\section{Detailed Settings for Synthetic Experiments}\label{sec_Detailed Settings for Synthetic Experiments}
\begin{figure}[htbp]
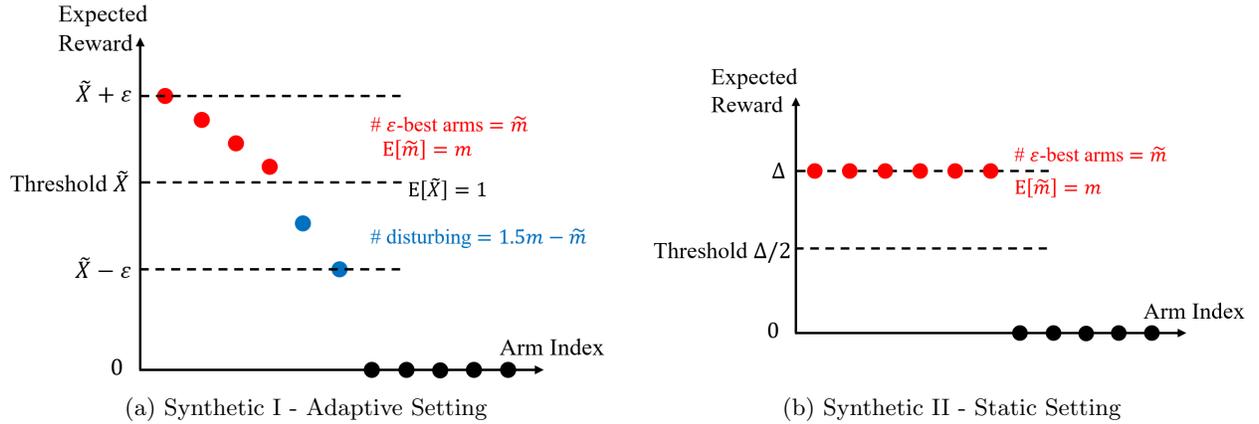

    \centering
    \begin{subfigure}{0.48\textwidth}
        \centering
        \includegraphics[width=\textwidth]{img/Synthetic_Adaptive.png}  
        \caption{Synthetic I - Adaptive Setting}
        \label{fig:synthetic-adaptive}
    \end{subfigure}
    \hfill
    \begin{subfigure}{0.48\textwidth}
        \centering
        \includegraphics[width=\textwidth]{img/Synthetic_Static.png}  
        \caption{Synthetic II - Static Setting}
        \label{fig:synthetic-static}
    \end{subfigure}
    
    \caption{Illustration on Synthetic Settings}
    \label{fig:synthetic}
\end{figure}


\subsection{Synthetic I - Adaptive Setting.}

First, we randomly sample $\tilde{m}$, representing the number of $\varepsilon$-best arms, from a distribution with an expected value of $m$ (used as input for a top $m$ algorithm). Next, we randomly sample $\tilde{X}$, representing the best arm reward minus $\varepsilon$, from a distribution with an expected value of $X$ (used as input for a threshold bandit algorithm). We then assign $\tilde{m}$ $\varepsilon$-best arms with expected rewards uniformly distributed between $\tilde{X} + \varepsilon$ and $\tilde{X}$. Additionally, we assign $(1.5m - \tilde{m})$ arms that are not $\varepsilon$-best with expected rewards uniformly distributed between $\tilde{X}$ and $\tilde{X} - \varepsilon$, as illustrated in Figure~\ref{fig:synthetic}. 

Based on these designed arm rewards, we define the linear model parameter as:
$$\bm{\theta}=(\tilde{X}+\varepsilon, \tilde{X}+(\tilde{m}-1)\varepsilon/\tilde{m}, \ldots,\tilde{X}+\varepsilon/\tilde{m}, 0,\ldots,0)^\top.$$

Arms are $d$-dimensional canonical basis $e_1, e_2, \ldots,e_d$ and $(1.5m-\tilde{m})$ additional disturbing arms 
$$\bm{x_i}=\left(\frac{\tilde{X}-(1.5m-\tilde{m}-i)\varepsilon/(1.5m-\tilde{m})}{\tilde{X}+\varepsilon}, 0,\cdots,0, \sqrt{1-\left(\frac{\tilde{X}-(1.5m-\tilde{m}-i)\varepsilon/(1.5m-\tilde{m})}{\tilde{X}+\varepsilon}\right)^2}\right)^\top$$ with $i\in[1.5m-\tilde{m}]$.


In the adaptive setting, pulling one arm can provide information about the distributions of other arms. The optimal policy in this setting should adaptively refine its sampling and stopping strategy based on historical data. This allows the algorithm to focus more on the disturbing arms, making adaptive strategies particularly effective as the algorithm progresses. In our experiments, we set $m=4$, $X=1$, with $d=10$ and $\varepsilon\in\{0.1, 0.2, 0.3\}$. A total of six different problem instances are evaluated to compare the performance of the algorithms. 

\subsection{Synthetic II - Static Setting.}

We consider a static synthetic setting, similar to the one proposed by \citet{xu2018fully}, where arms are represented as $d$-dimensional canonical basis vectors $e_1, e_2, \ldots, e_d$. We set the parameter vector $\bm{\theta} = (\Delta, \ldots, \Delta, 0, \ldots, 0)^\top$, where $\tilde{m}$ elements are $\Delta$ and $d - \tilde{m}$ elements are 0. In this setting, $\mathbb{E}[\tilde{m}] = m$, and only the value $m$ is provided as input to top $m$ algorithms. Consequently, the true mean values consist of some $\Delta$'s and some $0$'s.

If we set $\varepsilon = \Delta / 2$, as $\Delta$ approaches 0, it becomes difficult to distinguish between the $\varepsilon$-best arms and the arms that are not $\varepsilon$-best. In the static setting, knowledge of the rewards does not alter the sampling strategy, as all arms must be estimated with equal accuracy to effectively differentiate between them. Therefore, a static policy is optimal in this case, and the goal of this setting is to assess the ability of our algorithm to adapt to such static conditions. In our experiment, we set $m = 4$ with $d \in \{8, 12, 16\}$ and $\Delta = 1$. A total of three different problem instances are evaluated to compare the algorithms. 

\section{Auxiliary Results}\label{sec_Auxiliary Results}


The following lemma shows that matrix inversion reverses the order relation.

\begin{lemma}\label{lemma: inversion reverses loewner orders}
    \text{(Inversion Reverses Loewner orders)} Let $\bm{A}, \bm{B} \in \mathbb{R}^{d \times d}$. Suppose that $\bm{A} \succeq \bm{B}$ and $\bm{B}$ is invertible, we have
\begin{equation}
    \bm{A}^{-1} \preceq \bm{B}^{-1}.
\end{equation}
\end{lemma}

\proof\\{\textit{Proof.}}
Be definition, to show $\boldsymbol{B}^{-1}-\boldsymbol{A}^{-1}$ is a positive semi-definite matrix, it suffices to show that $\|\bm{x}\|_{\bm{B}^{-1}}^2-\|\bm{x}\|_{\bm{A}^{-1}}^2=\|\bm{x}\|_{\bm{B}^{-1}-\bm{A}^{-1}}^2 \geq 0$ for any $\bm{x} \in \mathbb{R}^d$. Then, by the Cauchy-Schwarz inequality,
\begin{equation}
    \|\bm{x}\|_{\bm{A}^{-1}}^2=\left\langle \bm{x}, \bm{A}^{-1} \bm{x}\right\rangle \leq\|\bm{x}\|_{\bm{B}^{-1}}\left\|\bm{A}^{-1} \bm{x}\right\|_{\bm{B}} \leq\|\bm{x}\|_{\bm{B}^{-1}}\left\|\bm{A}^{-1} \bm{x}\right\|_{\bm{A}}=\|\bm{x}\|_{\bm{B}^{-1}}\|\bm{x}\|_{\bm{A}^{-1}} .
\end{equation}

Hence $\|\bm{x}\|_{\bm{A}^{-1}} \leq\|\bm{x}\|_{\bm{B}^{-1}}$ for all $\bm{x}$, which completes the lemma.
\hfill\Halmos
\endproof

The following lemma provides an upper bound for the ratio of two optimization problems involving some bandit instance-related information.

\begin{lemma}\label{lemma: inequality iv}
    In this lemma, $\mathcal{A}_I = [K]$ always holds, \emph{i.e.}, we are considering the whole arm set. For any arm $i \in \mathcal{A}_I \cap G_\varepsilon \backslash \{1\}$, we have
\begin{equation}
    \frac{\min_{\bm{p}\in S_K}\max_{i \in \mathcal{A}_I \cap G_\varepsilon \backslash \{1\}} \Vert \boldsymbol{y}_{1, i} \Vert_{\bm{V}_{\bm{p}}^{-1}}^2 }{\min_{\bm{p}\in S_K}\min_{i \in \mathcal{A}_I \cap G_\varepsilon \backslash \{1\}} \Vert \boldsymbol{y}_{1, i} \Vert_{\bm{V}_{\bm{p}}^{-1}}^2} \leq \frac{d L_1}{{\mathcal{G}_{\mathcal{Y}}}^2 L_2}.
\end{equation}

\end{lemma}

\proof\\{\textit{Proof.}}
For any arm $i \in [K]$, from a perspective of geometry quantity, let conv($\mathcal{A}\cup-\mathcal{A}$) denote the convex hull of symmetric $\mathcal{A}\cup-\mathcal{A}$. Then for any set $\mathcal{Y}\subset \mathbb{R}^d$ define the gauge of $\mathcal{Y}$ as
\begin{equation}
    \mathcal{G}_\mathcal{Y} = \max \left\{ c>0: c\mathcal{Y}\subset \text{conv(}\mathcal{A}\cup-\mathcal{A}\text{)} \right\}.
\end{equation}

We can then provide a natural upper bound for ${\min_{\bm{p}\in S_K}\max_{i \in \mathcal{A}_I \cap G_\varepsilon \backslash \{1\}} \Vert \boldsymbol{y}_{1, i} \Vert_{\bm{V}_{\bm{p}}^{-1}}^2}$, given by
\begin{align}\label{numerator}
    {\min_{\bm{p}\in S_K}\max_{i \in \mathcal{A}_I \cap G_\varepsilon \backslash \{1\}} \Vert \boldsymbol{y}_{1, i} \Vert_{\bm{V}_{\bm{p}}^{-1}}^2} &\leq {\min_{\bm{p}\in S_K}\max_{y\in \mathcal{Y}(\mathcal{A}_I)} \Vert \boldsymbol{y} \Vert_{\bm{V}_{\bm{p}}^{-1}}^2} \nonumber\\ 
    &= \frac{1}{{\mathcal{G}_{\mathcal{Y}}}^2} {\min_{\bm{p}\in S_K}\max_{y\in \mathcal{Y}(\mathcal{A}_I)} \Vert \boldsymbol{y}{\mathcal{G}_\mathcal{Y}} \Vert_{\bm{V}_{\bm{p}}^{-1}}^2} \nonumber\\
    &\leq \frac{1}{{\mathcal{G}_{\mathcal{Y}}}^2} {\min_{\bm{p}\in S_K}\max_{\bm{a} \in \text{conv(}\mathcal{A}\cup-\mathcal{A}\text{)}} \Vert \bm{a} \Vert_{\bm{V}_{\bm{p}}^{-1}}^2} \nonumber\\
    &= \frac{1}{{\mathcal{G}_{\mathcal{Y}}}^2} {\min_{\bm{p}\in S_K}\max_{i \in \mathcal{A}_I} \Vert \bm{a}_i \Vert_{\bm{V}_{\bm{p}}^{-1}}^2} \nonumber\\
    &\leq \frac{d}{{\mathcal{G}_{\mathcal{Y}}}^2},
\end{align}
where the third line follows from the fact that the maximum value of a convex function on a convex set must occur at a vertex. With the Kiefer-Wolfowitz Theorem for the G-optimal design, the last inequality is achieved. 

Furthermore, for any arm $i \in \mathcal{A}_I \cap G_\varepsilon \backslash \{1\}$, we have 
\begin{align}
    \min_{\bm{p}\in S_K}\min_{i \in \mathcal{A}_I \cap G_\varepsilon \backslash \{1\}} \Vert \boldsymbol{y}_{1, i} \Vert_{\bm{V}_{\bm{p}}^{-1}}^2 &\geq \min_{\bm{p}\in S_K}\min_{i \in \mathcal{A}_I \cap G_\varepsilon \backslash \{1\}} \text{eig}_{\min }(\bm{V}_{\bm{p}}^{-1}) \Vert \boldsymbol{y}_{1, i} \Vert_2^2 \nonumber\\
    &= \min_{\bm{p}\in S_K}\min_{i \in \mathcal{A}_I \cap G_\varepsilon \backslash \{1\}} \frac{1}{\text{eig}_{\max }(\bm{V}_{\bm{p}})} \Vert \boldsymbol{y}_{1, i} \Vert_2^2 \nonumber\\
    &\geq \frac{1}{\max_{i \in \mathcal{A}_I} \Vert \bm{a}_i \Vert_2} \min_{\bm{p}\in S_K}\min_{i \in \mathcal{A}_I \cap G_\varepsilon \backslash \{1\}} \Vert \boldsymbol{y}_{1, i} \Vert_2^2,
\end{align}
where $\text{eig}_{\max }(\cdot)$ and $\text{eig}_{\min }(\cdot)$ are respectively the largest and smallest eigenvalue operators of a matrix. The first line follows from the Rayleigh Quotient and Rayleigh Theorem. The last line is derived by the relationship $\text{eig}_{\max }(\bm{V}_{\bm{p}}) \leq \max_{i \in \mathcal{A}_I} \Vert \bm{a}_i \Vert_2$. Recall the assumption in Theorem \ref{upper bound: Algorithm 1 XY} that $\min_{i \in G_\varepsilon \backslash \{ 1 \}} \Vert \bm{a}_1 - \bm{a}_i \Vert^2 \geq L_2$ and the assumption in Section \ref{sec_Problem Formulation} that $\Vert \boldsymbol{a}_i \Vert_2 \leq L_1 \text{ for } \forall i \in [K]$, we have
\begin{equation}\label{denominator}
     \min_{\bm{p}\in S_K}\min_{i \in \mathcal{A}_I \cap G_\varepsilon \backslash \{1\}} \Vert \boldsymbol{y}_{1, i} \Vert_{\bm{V}_{\bm{p}}^{-1}}^2 \geq \frac{L_2}{L_1}.
\end{equation}

Finally, combining inequalities (\ref{numerator}) and (\ref{denominator}) just completes the lemma.
\hfill\Halmos
\endproof


\begin{lemma}\label{theorem_KW}
    \citep{kiefer1960equivalence}. If the arm vectors $\boldsymbol{a} \in \mathcal{A}$ span $\mathbb{R}^d$, then for any probability distribution $\pi \in \mathcal{P}(\mathcal{A})$, the following statements are equivalent:
    \begin{enumerate}
        \item $\pi^\ast$ minimizes the function $g(\pi) = \max_{\boldsymbol{a} \in \mathcal{A}} \|\boldsymbol{a}\|_{\boldsymbol{V}(\pi)^{-1}}^2$.
\item $\pi^\ast$ maximizes the function $f(\pi) = \log \det \boldsymbol{V}(\pi)$.
\item $g(\pi^\ast) = d$.
    \end{enumerate}
     Additionally, there exists a $\pi^\ast$ of $g(\pi)$ such that the size of its support, $\vert \text{Supp}(\pi^\ast) \vert$, is at most $d(d+1)/2$.
\end{lemma}

\end{appendices}

\bibliographystyle{informs2014} 
\bibliography{reference}